%% file: master.tex
\documentclass[11pt,a4paper]{article}
\PassOptionsToPackage{breaklinks,final}{hyperref}
\usepackage{naaclhlt2019}
\usepackage{times}
\usepackage{latexsym}
\usepackage[T1]{fontenc}
\usepackage[utf8]{inputenc}
\usepackage{url}

\usepackage{array}
\usepackage{booktabs}
\usepackage{enumitem}
\newlist{enuminline}{enumerate*}{1}
\setlist*[enuminline,1]{%
  label=(\roman*),
}
\usepackage{linguex}
\usepackage{subcaption}
\usepackage{tikz}
\usetikzlibrary{arrows,positioning}
\usepackage{todonotes}

\newcommand{\ceri}{CER\textsubscript{I}}
\newcommand{\lingchar}[1]{$\langle$\textit{#1}$\rangle$}
\newcommand{\normpair}[2]{\textit{#1}~-- \textit{#2}}

\aclfinalcopy 
\def\aclpaperid{1229} 

\title{A Large-Scale Comparison of Historical Text Normalization Systems}

\author{Marcel Bollmann\\
  Department of Computer Science\\
  University of Copenhagen, Denmark\\
  {\tt marcel@di.ku.dk} \\}

\date{}

\begin{document}
\maketitle
\begin{abstract}
There is no consensus on the state-of-the-art approach to historical text normalization. 
Many techniques have been proposed, including rule-based methods, distance metrics, character-based statistical machine translation, and neural encoder--decoder models, but studies have used different datasets, different evaluation methods, and have come to different conclusions.
This paper presents the largest study of historical text normalization done so far.
We critically survey the existing literature and report experiments on eight languages, comparing systems spanning all categories of proposed normalization techniques, analysing the effect of training data quantity, and using different evaluation methods. 
The datasets and scripts are made publicly available.
\end{abstract}

\section{Introduction\protect\footnote{This work largely builds upon the author's doctoral thesis \citep{BollmannPhD}, the research for which was carried out at Ruhr-Universität Bochum, Germany.}}
\label{sec:intro}

Spelling variation is one of the key challenges for NLP on historical texts, affecting the performance of tools such as part-of-speech taggers or parsers and complicating users' search queries on a corpus.  \emph{Normalization} is often proposed as a solution; it is commonly defined as the mapping of historical variant spellings to a single, contemporary ``normal form'' as exemplified in Figure~\ref{fig:example}.

Automatic normalization of historical texts has a long history, going back to at least \citet{Fix1980}.  
Earlier approaches often rely on hand-crafted algorithms tailored to one specific language, while more recent approaches have focused on supervised machine learning, particularly character-based statistical machine translation~(SMT) and its neural equivalent~(NMT).  However, no clear consensus has emerged about the state of the art for this task, with papers either reporting an advantage for NMT \citep{Hamalainen-etal2018}, SMT \citep{Domingo-Casacuberta2018}, or language-specific algorithms \citep{Schneider-etal2017}.
Moreover, the quantity of annotated training data varies considerably between studies, making it difficult to obtain practical recommendations for new projects seeking to use normalization techniques.

\input{fig_normexample.tex}

\paragraph{Contributions}
This paper aims to provide the most comprehensive evaluation and analysis of historical text normalization systems so far.  Motivated by a systematic review of previous work on this topic~(Sec.\,\ref{sec:survey}), only publicly available normalization systems covering a wide range of proposed techniques are selected~(Sec.\,\ref{sec:setup}) and evaluated across a diverse collection of historical datasets covering eight languages~(Sec.\,\ref{sec:eval}).  This is followed by a detailed analysis of the effect of training data quantity and a critical discussion of evaluation methods for assessing normalization quality~(Sec.\,\ref{sec:analysis}).

The datasets and code are made freely available whenever possible,\footnote{\url{https://github.com/coastalcph/histnorm}; one dataset could not be included due to licensing restrictions.} along with detailed instructions on how to reproduce the experiments.

\section{A Brief Survey of Automatic Historical Text Normalization}
\label{sec:survey}

The following overview is broadly organized by categories that each represent a conceptually or methodically different approach.

\subsection{Substitution Lists}

The conceptually simplest form of normalization is to look up each
historical variant in a pre-compiled list that maps it to its intended normalization.  This approach can go by many names, such as \emph{lexical substitution}, \emph{dictionary lookup}, \emph{wordlist mapping}, or \emph{memorization}.  While it does not generalize in any way to variants that are not covered by the list, it has proven highly effective as a component in several normalization systems, such as the semi-automatic VARD tool~\citep{Rayson-etal2005,Baron-Rayson2008} or the fully automatic Norma tool~\citep{Bollmann2012}.

\subsection{Rule-based Methods}

Rule-based approaches try to encode regularities in spelling variants---e.g., historical~\lingchar{v} often representing modern~\lingchar{u}---in the form of replacement rules, typically including context
information to discriminate between different usages of a character.  Some of the earliest approaches to normalization are rule-based, with rules being created manually for one particular language, such as Old Icelandic~\citep{Fix1980} or Old German~\citep{Koller1983}.

VARD~2 uses ``letter replacement rules'' to construct normalization candidates, but is not necessarily concerned with precision due to its interactive nature~\citep{Baron-Rayson2008}.
\citet{Bollmann-etal2011} describe a supervised learning algorithm to automatically derive context-aware replacement rules from training data, including ``identity rules'' that leave a character unchanged, then apply one rule to each character of a historical word form to produce a normalization.
\citet{Porta-etal2013} model phonological sound change rules for Old Spanish using finite-state transducers; \citet{Etxeberria-etal2016} describe a similarly motivated model that can be trained in a supervised manner. 

Rule-based methods are also commonly found when the goal is not to produce a single best normalization, but to cluster a group of spelling variants~\citep{Giusti-etal2007} or to retrieve occurrences of variant spellings given a modern form in an information retrieval~(IR) scenario~\citep{ErnstGerlach-Fuhr2006,Koolen-etal2006}.

\subsection{Distance-based Methods}

Approaches using edit distance measures~\citep[such as Levenshtein distance;][]{Levenshtein1966} are most commonly found in an IR~context, since measures that compare two word forms are a natural fit for matching a search term with relevant word forms in a historical document~\citep[e.g.,][]{Robertson-Willett1993}.  Weighted variants of distance measures can be used to assign lower costs to more likely edit operations~\citep{Kempken-etal2006,Hauser-Schulz2007}.

In a normalization context, distance measures can be used to compare historical variants to entries in a contemporary full-form lexicon~\citep{Kestemont-etal2010,Jurish2010b}.
Norma includes a distance-based component whose edit weights can be learned from a training set of normalizations~\citep{Bollmann2012}. \citet{Pettersson-Megyesi-Nivre2013} find a similar approach to be more effective than hand-crafted rules on Swedish.
Sometimes, the line between distance-based and rule-based methods get blurred; \citet{Adesam-etal2012} use the Levenshtein algorithm to derive ``substitution rules'' from training data, which are then used to link up historical Swedish forms with lexicon entries; \citet{vanHalteren-Rem2013} describe a comparable approach for Dutch.

Furthermore, distance measures also lend themselves to unsupervised approaches for clustering historical variants of the same modern form, where identifying the precise modern form is not necessarily required~\citep{Amoia-Martinez2013,Barteld-etal2015}.

\subsection{Statistical Models}

In a probabilistic view of the normalization task, 
the goal is to optimize the probability~$p(t|s)$ that a contemporary word
form~$t$ is the normalization of a historical word form~$s$.  This can be
seen as a \emph{noisy channel model}, which has been used for normalization by, e.g., \citet{Oravecz-etal2010} and \citet{Etxeberria-etal2016}.

More commonly, \emph{character-based statistical machine translation~(CSMT)} has been applied to the normalization task.  Instead of translating a sentence as a sequence of tokens, these approaches ``translate'' a historical word form as a sequence of characters.  This has been found to be very effective for a variety of historical languages, such as Spanish~\citep{SanchezMartinez-etal2013}, Icelandic and Swedish~\citep{Pettersson-etal2013}, Slovene~\citep{Scherrer-Erjavec2013,Scherrer-Erjavec2016,Ljubesic-etal2016}, as well as Hungarian, German, and English~\citep{Pettersson2016}, where it is usually found to outperform previous approaches.

\citet{Pettersson-etal2014} find that a CSMT~system often performs best in a comparison with a filtering method and a distance-based approach on five different languages.  \citet{Schneider-etal2017} compare VARD~2 to CSMT on English and find that VARD~2 performs slightly better.  \citet{Domingo-Casacuberta2018} evaluate both word-based and character-based models and find that SMT outperforms a neural network model.

\subsection{Neural Models}

Neural network architectures have become popular for a variety of NLP~tasks, and historical normalization is no exception. 
\emph{Character-based neural machine translation~(CNMT)} is the logical neural equivalent to the CSMT~approach, and has first been used for normalization of historical German~\citep{Bollmann-etal2017,Korchagina2017} using encoder--decoder models with long short-term memory~(LSTM) units.
\citet{Robertson-Goldwater2018} present a more detailed evaluation of this architecture on five different languages.  \citet{Hamalainen-etal2018} evaluate SMT, NMT, an edit-distance approach, and a rule-based finite-state transducer, and advocate for a combination of these approaches to make use of their individual strengths; however, they restrict their evaluation to English.

Other neural architectures have rarely been used for normalization so far.  \citet{AlAzawi-etal2013} and \citet{Bollmann-Sogaard2016} frame the normalization task as a sequence labelling problem, labelling each character in the historical word form with its normalized equivalent.  \citet{Kestemont-etal2016} use convolutional networks for lemmatization of historical Dutch.  Overall, though, the encoder--decoder model with recurrent layers is the dominant approach.

\subsection{Beyond Token-Level Normalization}
\label{sec:survey-beyond}

The presented literature almost exclusively focuses on models where the input is a single token.  In theory, it would be desirable to include context from the surrounding tokens, as some historical spellings can have more than one modern equivalent depending on the context in which they are used (e.g., historical \textit{ther} could represent \textit{their} or \textit{there}).  Remarkably few studies have attempted this so far: \citet{Jurish2010} uses hidden Markov models to select between normalization candidates; \citet{Mitankin-etal2014} use a language model in a similar vein; \citet{Ljubesic-etal2016} experiment with ``segment-level'' input, i.e., a string of several historical tokens as input to a normalizer.  Since this area is currently very underexplored, it warrants a deeper investigation that goes beyond the scope of this paper.

\input{tab_datasets}

\section{Experimental Setup}
\label{sec:setup}

\paragraph{Systems}
The selection of normalization systems follows two goals:
\begin{enuminline}
    \item to include at least one system for each major category as identified in Sec.\,\ref{sec:survey}; and
    \item to use only freely available tools in order to facilitate reproduction and application of the described methods.
\end{enuminline}
To that effect, this study compares the following approaches:

\begin{itemize}
    \item \textbf{Norma}\footnote{\url{https://github.com/comphist/norma}}~\citep{Bollmann2012}, which combines substitution lists, a rule-based normalizer, and a distance-based algorithm, with the option of running them separately or combined.  Importantly, it implements supervised learning algorithms for all of these components and is not restricted to a particular language.
    \item \textbf{cSMTiser}\footnote{\url{https://github.com/clarinsi/csmtiser}}~\citep{Ljubesic-etal2016,Scherrer-Ljubesic2016}, which implements a normalization pipeline using character-based statistical machine translation~(CSMT) using the Moses toolkit~\citep{Moses}.
    \item \textbf{Neural machine translation~(NMT)}, in the form of two publicly available implementations:
    \begin{enuminline}
        \item the model by~\citet{BollmannPhD}, also used in \citet{Bollmann-etal2018};\footnote{I reimplemented the model here using the XNMT~toolkit \citep{Neubig-etal2018}.} and
        \item the model by \citet{Tang-etal2018}.\footnote{\url{https://github.com/tanggongbo/normalization-NMT}; their model uses the deep transition architecture of \citet[Sec.\,2.3.1]{Sennrich-etal2017} as implemented by Marian~\citep{mariannmt}.}
    \end{enuminline}
\end{itemize}
Two systems were chosen for the NMT approach as they use very different hyperparameters, despite both using comparable neural encoder--decoder models: \citet{BollmannPhD} uses a single LSTM~layer with dimensionality~300 in the encoder and decoder, while \citet{Tang-etal2018} use six vanilla RNN cells with dimensionality~1024.

\paragraph{Datasets}
Table~\ref{tab:datasets} gives an overview of the historical datasets.  They are taken from \citet{BollmannPhD} and represent the largest and most varied collection of datasets used for historical text normalization so far, covering eight languages from different language families---English, German, Hungarian, Icelandic, Spanish, Portuguese, Slovene, and Swedish---as well as different text genres and time periods.  Furthermore, most of these have also been used in previous work, such as the English, Hungarian, Icelandic, and Swedish datasets \citep[e.g.,][]{Pettersson-etal2014,Pettersson2016,Robertson-Goldwater2018,Tang-etal2018} and the Slovene datasets \citep[e.g.,][]{Ljubesic-etal2016,Scherrer-Erjavec2016,Etxeberria-etal2016,Domingo-Casacuberta2018}.  

Additionally, contemporary datasets are required for the rule-based and distance-based components of Norma, as they expect a list of valid target word forms to function properly.  For this, we want to choose resources that are readily available for many languages and are reliable, i.e., consist of carefully edited text.  Here, I choose a combination of three sources:\footnote{Detailed descriptions of the data extraction procedure can be found in the Supplementary Material.}
\begin{enuminline}
    \item the normalizations in the training sets,
    \item the Europarl corpus \citep{Koehn2005}, and
    \item the parallel Bible corpus by \citet{Christo-Steedman2015}.
\end{enuminline}
The only exception is Icelandic, which is not covered by Europarl; here, we can follow \citet{Pettersson2016} instead by using data from two specialized resources, the B\'IN~database \citep{Bjarnadottir2012} and the M\'IM~corpus \citep{Helgadottir-etal2012}.  This way, we obtain full-form lexica of 12k--64k~word types from the Bible corpus, 55k--268k types from Europarl, and 2.8M~types from the Icelandic resources.

\paragraph{Preprocessing}
The most important preprocessing decisions\footnote{The full preprocessing steps can be found in the Supplementary Material.} are
\begin{enuminline}
    \item to lowercase all characters and
    \item to remove all punctuation-only tokens.
\end{enuminline}
Both capitalization and punctuation often cannot be handled correctly without considering token context, which all current normalization models do not do.  Furthermore, their usage can be very erratic in historical texts, potentially distorting the evaluation; e.g., when a text uses punctuation marks according to modern conventions, their normalization is usually trivial, resulting in artificial gains in normalization accuracy that other texts do not get.
At the same time, most previous work has not followed these same preprocessing guidelines, making a direct comparison more difficult.  This work tries to make up for this by evaluating many different systems, effectively reproducing some of these previous results instead.

\input{tab_eval_test}

\section{Evaluation}
\label{sec:eval}

All models are trained and evaluated separately for each dataset by calculating word accuracy over all tokens.  In particular, there is no step to discriminate between tokens that require normalization and those that do not; all word forms in the datasets are treated equally.

For Norma, all components are evaluated both separately and combined, as the former gives us insight into the performance of each individual component, while the latter is reported to produce the best results~\citep{Bollmann2012}.
For cSMTiser, the authors suggest using additional monolingual data to improve the language model; the contemporary datasets are used for this purpose and the model is trained both without and with this additional data; the latter is denoted cSMTiser\textsubscript{+LM}.
For~NMT, the model by \citet{BollmannPhD} is evaluated using an ensemble of five models; the model by \citet{Tang-etal2018} is trained on character-level input using the default settings provided by their implementation.\footnote{This is the ``Att-RNN'' setting reported in their paper; due to the high computational demands of the model, it was not feasible to run experiments with multiple configurations.}

To illustrate how challenging the normalization task is on different datasets, we can additionally look at the \textit{identity baseline}---i.e., the percentage of tokens that do \emph{not} need to be normalized---as well as the \textit{maximum accuracy} obtainable if each word type was mapped to its most frequently occurring normalization.  The latter gives an indication of the extent of ambiguity in the datasets and the disadvantage of not considering token context (cf.\ Sec.\,\ref{sec:survey-beyond}).

\input{tab_eval_ceri}

\paragraph{Results}
Table~\ref{tab:eval-test} shows the results of this evaluation.  The extent of spelling variation varies greatly between datasets, with less than 15\% of tokens requiring normalization (SL\textsubscript{G}) to more than 80\%~(HU).  The maximum accuracy is above 97\% for most datasets, suggesting that we can obtain high normalization accuracy in principle even without considering token context.

For the normalization systems,
we observe significantly better word accuracy with SMT than NMT on four of the datasets, and non-significant differences on five others.  There is only one dataset (DE\textsubscript{A}) where the NMT system by \citet{Tang-etal2018} gets significantly better word accuracy than other systems.  This somewhat contradicts the results from \citet{Tang-etal2018}, who find NMT to usually outperform the SMT baseline by~\citet{Pettersson-etal2014}.  However, note that the results for the cSMTiser system are often significantly better than reported in previous work: e.g., on Hungarian, cSMTiser obtains 91.7\% accuracy, but only 80.1\% with the SMT system from~\citet{Pettersson-etal2014}.

Overall, the deep NMT model by \citet{Tang-etal2018} consistently outperforms the shallow one by \citet{BollmannPhD}.
cSMTiser seems to benefit from the added contemporary data for language modelling, though the effect is not significant on any individual dataset.  Finally, while Norma does produce competitive results on several datasets (particularly in the ``combined'' setting), it is generally significantly behind the SMT and NMT methods.

\section{Analysis}
\label{sec:analysis}

\subsection{Measuring Normalization Quality}
\label{sec:analysis-quality}

While word accuracy is easily interpretable, it is also a very crude measure, as it classifies predictions as correct/incorrect without considering the type of error(s) made by the model.  \emph{Character error rate~(CER)} has sometimes been suggested as a complement to address this issue, but I~believe this is not very insightful:  For any normalization system that achieves a reasonably high word accuracy, CER~will highly correlate with accuracy simply because CER equals zero for any word that is accurately normalized.\footnote{When comparing word accuracy scores in Table~\ref{tab:eval-test} with the same configurations evaluated using CER, they correlate with Pearson's $r \approx -0.96$.}
At the same time, there is a need for a more fine-grained way to assess the normalization quality.  Consider the following example from the Hungarian dataset with its predicted normalization from the NMT~system by \citet{BollmannPhD}:

\alignSubExtrue
\setlength{\Exlabelsep}{2.5em}%
\setlength{\alignSubExnegindent}{\Exlabelsep}%
\ifalignSubEx\addtolength{\Exlabelsep}{1em}%
\addtolength{\alignSubExnegindent}{1em}\fi%
\ex. \label{ex:hungarian}
  \a.[\sc Orig] yduewzewlendewk
  \b.[\sc Gold] {üdvözülend\H{o}}ek
  \b.[\sc Pred] {üdvözülend\H{o}}k

Here, the prediction matches the correct target form almost perfectly, but would be counted as incorrect since it misses an insertion of the letter~\lingchar{e} towards the end.  In this vein, it will be treated the same by the word accuracy measure as a prediction that, e.g., had left the original form unchanged.

\paragraph{\ceri}
One alternative is to consider character error rate \emph{on the subset of incorrect normalizations only.}  This way, CER becomes a true complement to word accuracy by assessing the \emph{magnitude of error} that a normalization model makes when it is not perfectly accurate.  The results of this measure, denoted \ceri{}, are shown in Table~\ref{tab:eval-ceri}.  The lowest \ceri{} score is often achieved by Norma's lookup module, which leaves historical word forms unchanged if they are not in its lookup wordlist learned during training.  This suggests that the incorrect predictions made by other systems are often \emph{worse} than just leaving the historical spelling unchanged.

\paragraph{Stemming}
Another problem of CER is that all types of errors are treated the same: a one-letter difference in inflection, such as \normpair{king}{kings} or \normpair{came}{come}, would be treated identically to an error that changes the meaning of the word (\normpair{bids}{beds}) or results in a non-word (\normpair{creature}{cryature}).  
I propose an approach that, to the best of my knowledge, has not been used in normalization evaluation before: measure accuracy on \emph{word stems,} i.e., process both the reference normalization and the prediction with an automatic stemming algorithm and check if both stems match.
For this evaluation, I choose the Snowball stemmer~\citep{Snowball} as it contains stemming algorithms for many languages (including the ones represented here except for Icelandic and Slovene) and is publicly available.\footnote{\url{http://snowballstem.org/}}

Table~\ref{tab:eval-stem} shows the accuracy on word stems, again only evaluated on the subset of incorrect normalizations, as this better highlights the differences between settings.  This evaluation reveals some notable differences between \emph{datasets}: For example, while the English and Spanish datasets have very comparable accuracy scores overall (cf.\ Tab.\,\ref{tab:eval-test}), they show very different characteristics in the stemming evaluation; for English, only up to 9.86\% of incorrect predictions show the correct word stem, while for Spanish the number is up to 43.82\%.  Examining predictions on the dev set, many of the incorrectly predicted cases in Spanish result from mistakes in placement of diacritics, such as \normpair{\'esta}{est\'a} or \normpair{env\'ie}{envi\'e}; the stemming algorithm removes diacritics and can therefore match these instances.  Overall, this gives an indication that the errors made on the Spanish dataset are less severe than those on English, despite comparable word accuracy scores and a usually higher \ceri{} for Spanish.

This case study shows that stemming can be a useful tool for error analysis in normalization models and reveal characteristics that neither word accuracy nor CER alone can show.

\subsection{Effect of Training Data Quantity}

Supervised methods for historical text normalization have been evaluated with highly varying amounts of training data: e.g., \citet{Domingo-Casacuberta2018} train a normalizer for 17\textsuperscript{th}~century Spanish on 436k~tokens; \citet{Etxeberria-etal2016} use only 8k~tokens to train a normalizer for Basque.  Even in the evaluation in Sec.\,\ref{sec:eval}, training set sizes varied between 24k and 234k~tokens, depending on the dataset.  Furthermore, many research projects seeking to use automatic normalization techniques cannot afford to produce training data in high quantity.  All of this raises the question how different normalization systems perform with varying amounts of training data, and whether reasonable normalization results can be achieved in a low-resource scenario.

\begin{figure*}
     \centering
     \begin{subfigure}[t]{0.48\linewidth}
     \hspace{-1em}
 	  \input{curves_english-icamet.pgf}
       \subcaption{English}\label{fig:curve-en}
     \end{subfigure}
     \begin{subfigure}[t]{0.48\linewidth}
     \hspace{-1em}
 	  \input{curves_hungarian-hgds.pgf}
       \subcaption{Hungarian}\label{fig:curve-hu}
     \end{subfigure}

     \caption{Word accuracy on the development sets for different amounts of training data (note that the $x$-axis is log-scaled); NMT-1 is the model by \citet{BollmannPhD}, NMT-2 is the model by \citet{Tang-etal2018}.}
     \label{fig:curves}
\end{figure*}

\paragraph{Methodology}
All models are retrained on varying subsets of the training data, with sizes ranging from 100~tokens to 50,000~tokens.  However, the lower the training set size is, the higher the potential variance when training on it, since random factors such as the covered spelling variants or vocabulary are more likely to impact the results.  Therefore, I choose the following approach:
For each dataset and training size, up to ten different training splits are extracted,\footnote{The ten training splits consist of chunks of $n$~tokens that are spaced equidistantly across the full training set; for larger~$n$, the number of chunks is reduced so that no splits overlap to more than 50\%.} and a separate model is trained on each one.  Each model is then evaluated on the respective development dataset, and only the average accuracy across all splits is considered.

\input{tab_eval_oov}

\paragraph{Results}
Figure~\ref{fig:curves} shows two learning curves that are representative for most of the datasets.\footnote{Plots for all datasets can be found in the Appendix.}  They reveal that Norma (in the ``combined'' setting) performs best in extremely low-resource scenarios, but is overtaken by the SMT~approach as more training data becomes available; usually already around 500--1000~tokens.  The NMT~models have a steeper learning curve, needing more training data to become competitive.  Extrapolating this trend, it is conceivable that the NMT models would simply need more training data than our current datasets provide in order to consistently outperform the SMT approach.
On the other hand, there appears to be no correlation between the size of the training set (cf.\ Tab.\,\ref{tab:datasets}) and the relative performance of NMT vs.\ SMT (cf.\ Tab.\,\ref{tab:eval-test}) in the experiments.  Since I am not aware of larger datasets for the historical normalization task, this remains an open question for now.

A remarkable result is that very small amounts of training data can already be helpful for the normalization task.  The English dataset has comparatively little spelling variation to begin with: leaving all words unnormalized already results in an accuracy of~75.5\%.   Still, with as little as 100~tokens for training, applying the Norma tool raises the accuracy above~83\%.  For Hungarian, the same amount of training data raises the accuracy from 17.8\%~(unnormalized) to around~50\%.  It would be interesting to further compare these results with fully unsupervised methods.

\subsection{Out-of-Vocabulary Words}
\label{sec:analysis-oov}

\citet{Robertson-Goldwater2018} highlight the importance of evaluating separately on seen vs.\ unseen tokens, i.e., tokens that have also been in the training set (in-vocabulary) and those that have not (out-of-vocabulary), as well as comparing to a naive memorization baseline.  These numbers are presented in Table~\ref{tab:eval-iv-oov}.  For unseen tokens (Tab.\,\ref{tab:eval-oov}), the accuracy scores follow generally the same trend as in the full evaluation of Tab.\,\ref{tab:eval-test}; i.e., SMT performs best in most cases.  For seen tokens (Tab.\,\ref{tab:eval-iv}), however, Norma's lookup component---which implements naive memorization---obtains the highest score on nine datasets.

These observations suggest a new normalization strategy: apply the naive lookup on the subset of in-vocabulary tokens and the SMT/NMT models on the subset of out-of-vocabulary tokens only.
Table~\ref{tab:eval-combined} shows the results of this strategy.\footnote{The non-lookup components of Norma are not included in this evaluation since ``Norma (Combined)'' effectively implements such a strategy already.}  On nine datasets, it performs better than always using the learned models (as in Tab.\,\ref{tab:eval-test}), and this difference is statistically significant on five of them.  These results support the claim from \citet{Robertson-Goldwater2018} that ``learned models should typically only be applied to \emph{unseen} tokens.''

\section{Conclusion}

This paper presented a large study of historical text normalization.  Starting with a systematic survey of the existing literature, four different systems (based on supervised learning) were evaluated and compared on datasets from eight different languages.
On the basis of these results, we can extract some practical recommendations for projects seeking to employ normalization techniques:
\begin{enumerate}
    \item to use the Norma tool when only little training data ($<$500~tokens) is available;
    \item to use cSMTiser otherwise, ideally with additional data for language modelling; and
    \item to make use of the naive memorization/lookup technique for in-vocabulary tokens when possible.
\end{enumerate}
Furthermore, the qualitative analysis (in Sec.\,\ref{sec:analysis-quality}) should encourage authors evaluating normalization systems to use task-motivated approaches, such as evaluation on word stems, to provide deeper insight into the properties of their models and datasets.

Detailed information on how to train and apply all of the evaluated techniques is also made available online at {\small\url{https://github.com/coastalcph/histnorm}}.

\section*{Acknowledgments}

I would like to thank my PhD supervisor, Stefanie Dipper, for her continuous support over many years that culminated in the doctoral thesis on which this paper is based; the acknowledgments in that thesis largely extend to this paper as well.  Many thanks to Anders S{\o}gaard for many helpful discussions and for supporting the follow-up experiments conducted for this paper.  Further thanks go to the anonymous reviewers whose helpful suggestions have largely been incorporated here.

I gratefully acknowledge the donation of a Titan Xp GPU by the NVIDIA Corporation that was used for a substantial part of this research.

\newpage
\bibliography{master}
\bibliographystyle{acl_natbib}

\newpage
\appendix

\begin{figure*}[!hb]
     \centering
     \begin{subfigure}[t]{0.48\linewidth}
     \hspace{-1em}
 	  \input{curves_german-anselm.pgf}
       \caption{German (Anselm)}
     \end{subfigure}
     \begin{subfigure}[t]{0.48\linewidth}
     \hspace{-1em}
 	  \input{curves_german-ridges.pgf}
       \caption{German (RIDGES)}
     \end{subfigure}

     \begin{subfigure}[t]{0.48\linewidth}
     \hspace{-1em}
 	  \input{curves_english-icamet.pgf}
       \caption{English}
     \end{subfigure}
     \begin{subfigure}[t]{0.48\linewidth}
     \hspace{-1em}
 	  \input{curves_hungarian-hgds.pgf}
       \caption{Hungarian}
     \end{subfigure}
 
     \begin{subfigure}[t]{0.48\linewidth}
     \hspace{-1em}
 	  \input{curves_spanish-ps.pgf}
       \caption{Spanish}
     \end{subfigure}
     \begin{subfigure}[t]{0.48\linewidth}
     \hspace{-1em}
 	  \input{curves_portuguese-ps.pgf}
       \caption{Portuguese}
     \end{subfigure}

     \caption{Word accuracy on the development sets for different amounts of training data (note that the $x$-axis is log-scaled); NMT-1 is the model by \citet{BollmannPhD}, NMT-2 is the model by \citet{Tang-etal2018}.}
     \label{fig:curves-all1}
\end{figure*}
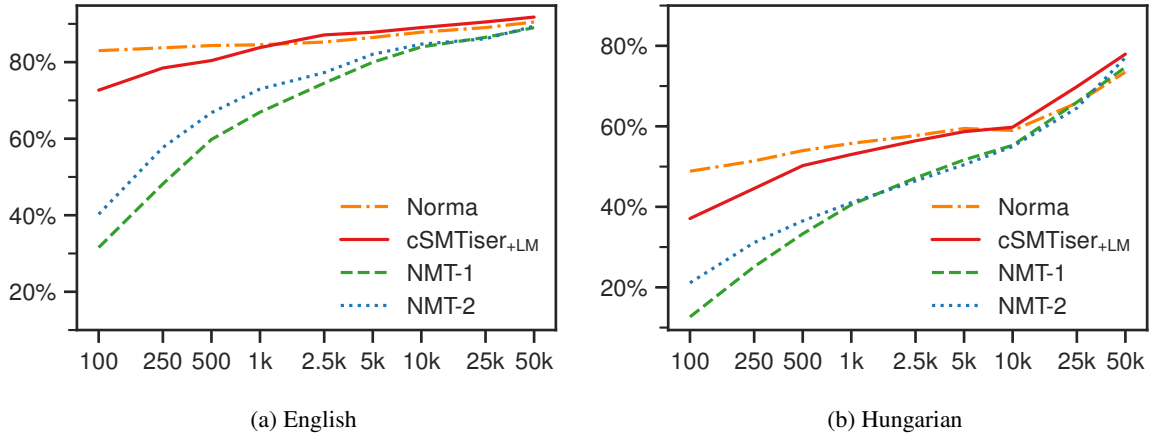

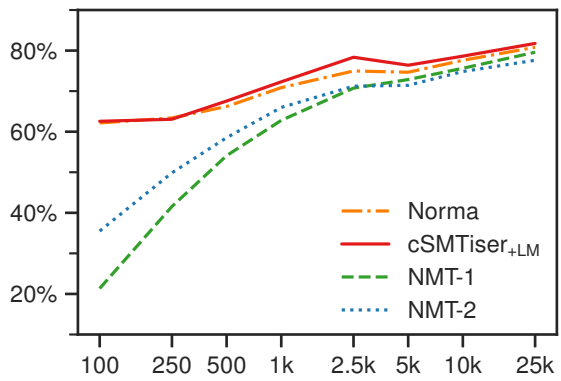
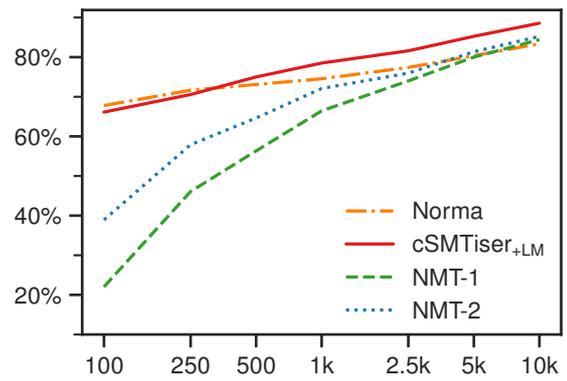
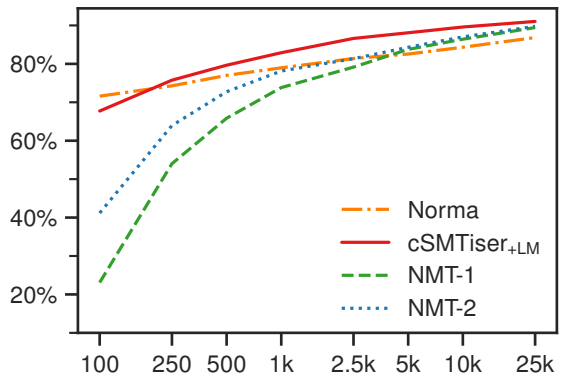
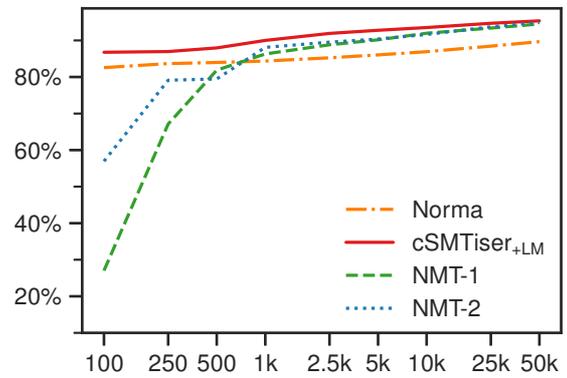
\begin{figure*}
\centering
     \begin{subfigure}[t]{0.48\linewidth}
     \hspace{-1em}
 	  \input{curves_icelandic-icepahc.pgf}
       \caption{Icelandic}
     \end{subfigure}
     \begin{subfigure}[t]{0.48\linewidth}
     \hspace{-1em}
 	  \input{curves_swedish-gaw.pgf}
       \caption{Swedish}
     \end{subfigure}

     \begin{subfigure}[t]{0.48\linewidth}
     \hspace{-1em}
 	  \input{curves_slovene-goo300k-bohoric.pgf}
       \caption{Slovene (Bohori\v{c})}
     \end{subfigure}
     \begin{subfigure}[t]{0.48\linewidth}
     \hspace{-1em}
 	  \input{curves_slovene-goo300k-gaj.pgf}
       \caption{Slovene (Gaj)}
     \end{subfigure}
     
     \caption{Word accuracy on the development sets (continued); NMT-1 is the model by \citet{BollmannPhD}, NMT-2 is the model by \citet{Tang-etal2018}.}
     \label{fig:curves-all2}
\end{figure*}

\section{Appendix}
\label{sec:appendix}
Figures~\ref{fig:curves-all1} and~\ref{fig:curves-all2} show plots of the learning curves for all of the datasets.

\subsection{Preprocessing}

The full preprocessing steps of all datasets (both historical and contemporary) comprise of:
\begin{enumerate}
    \item lowercasing all tokens;
    \item filtering out pairs where either the historical token or the reference normalization is empty;
    \item filtering out pairs where either the historical token or the reference normalization consists \emph{only} of punctuation marks, defined as characters that belong to one of the Unicode ``Punctuation'' categories;
    \item replacing all digits with zeroes \emph{iff} the digits in the historical token and the reference normalization match;
    \item replacing actual space characters in either the historical token or the reference normalization with a special symbol that does not otherwise occur in the dataset; and
    \item performing Unicode normalization according to the NFC standard.
\end{enumerate}
Additionally, the preprocessing script can also be found in the Supplementary Material or at \url{https://github.com/coastalcph/histnorm}.

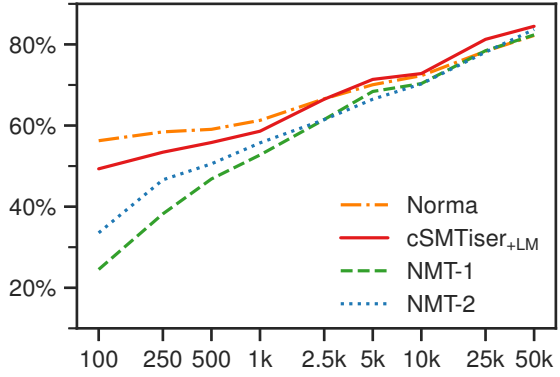
\end{document}

%% file: fig_normexample.tex
\begin{figure}
    \centering
    \begin{tikzpicture}[transform shape,>=stealth',auto]
    
    \tikzset{token/.style={text height=0.6em,text depth=0.2em,font=\itshape},
             centertoken/.style={draw,circle,fill=lightgray!40,%
                                 text height=1.4em,text depth=0.6em,font=\Large\bfseries}
    }
    
    \node[centertoken]    (C) at (0, 0)  {their};
    
    \foreach \a/\t in {1/thair,2.15/thaire,3.1/thear,4/thayr,4.9/theaire,5.85/theiare,7/theire}{
        \node[token] (La) at ({270 + \a*180/8}: 2cm) {\t};
        \draw[->] (La) to (C);
    }
    \foreach \a/\t in {1/{\th}er,2.15/{\th}ere,3.1/{\th}air,4/theyr,4.9/thir,5.85/ther,7/thar}{
        \node[token] (Ra) at ({90 + \a*180/8}: 2cm) {\t};
        \draw[->] (Ra) to (C);
    }
    
    \end{tikzpicture}
    \caption{Historical text normalization exemplified: mapping variant spellings from historical English texts to their normalization \textit{`their'}}
    \label{fig:example}
\end{figure}

%% file: tab_datasets.tex
\begin{table*}
  \centering
  \begin{tabular}{llccrrr}
\toprule
\multicolumn{2}{l}{\textbf{Dataset/Language}} & \textbf{Time Period} & \textbf{Genre} & \multicolumn{3}{c}{\textbf{Tokens}} \\
\cmidrule(lr){5-7}
& & & & \textsc{Train} & \textsc{Dev} & \textsc{Test} \\
\midrule
DE\textsubscript{A} & German (Anselm) & 14\textsuperscript{th}--16\textsuperscript{th} c. & Religious  & 233,947 & 45,996 & 45,999 \\
DE\textsubscript{R} &  German (RIDGES) & 1482--1652 & Science & 41,857 & 9,712 & 9,587 \\
EN & English & 1386--1698 & Letters & 147,826 & 16,334 & 17,644 \\
ES & Spanish & 15\textsuperscript{th}--19\textsuperscript{th} c. & Letters & 97,320 & 11,650 & 12,479 \\
HU & Hungarian & 1440--1541 & Religious & 134,028 & 16,707 & 16,779 \\
IS & Icelandic & 15\textsuperscript{th} c. & Religious & 49,633 & 6,109 & 6,037  \\
PT & Portuguese & 15\textsuperscript{th}--19\textsuperscript{th} c. & Letters & 222,525 & 26,749 & 27,078\\  
SL\textsubscript{B} & Slovene (Bohori{\v c}) & 1750--1840s & Diverse &  50,023 & 5,841 & 5,969 \\ 
SL\textsubscript{G} & Slovene (Gaj) & 1840s--1899 & Diverse & 161,211 & 20,878 & 21,493 \\
SV & Swedish & 1527--1812 & Diverse & 24,458 & 2,245 & 29,184 \\
 \bottomrule
\end{tabular}
\caption{Historical datasets used in the experiments}
  \label{tab:datasets}
\end{table*}

%% file: tab_eval_test.tex
\makeatletter
\newlength\mysuperscriptlen
\DeclareRobustCommand*\Textsuperscript[1]{%
\@Textsuperscript{\selectfont#1}}
\def\@Textsuperscript#1{%
\settoheight\mysuperscriptlen{\fontsize\f@size\z@ A}%
{\m@th\ensuremath{\raise.3\mysuperscriptlen\hbox{\fontsize\sf@size\z@#1}}}}
\makeatother
\newcommand{\notsig}[1]{\Textsuperscript{\footnotesize{}*}#1}

\begin{table*}
\resizebox{\textwidth}{!}{%

  \centering
\setlength{\tabcolsep}{3pt}
\begin{tabular}{lrrrrrrrrrr}
\toprule
\textbf{Method} & \multicolumn{10}{c}{\textbf{Dataset}} \\
\cmidrule(lr){2-11}
 & \multicolumn{1}{c}{DE\textsubscript{A}} & \multicolumn{1}{c}{DE\textsubscript{R}} & \multicolumn{1}{c}{EN} & \multicolumn{1}{c}{ES} & \multicolumn{1}{c}{HU} & \multicolumn{1}{c}{IS} & \multicolumn{1}{c}{PT} & \multicolumn{1}{c}{SL\textsubscript{B}} & \multicolumn{1}{c}{SL\textsubscript{G}} & \multicolumn{1}{c}{SV}\\
\midrule
\textit{Identity} & \itshape 30.63 & \itshape 44.36 & \itshape 75.29 & \itshape 73.40 & \itshape 17.53 & \itshape 47.62 & \itshape 65.19 & \itshape 40.74 & \itshape 85.38 & \itshape 58.59 \\
\textit{Maximum} & \itshape 94.64 & \itshape 96.46 & \itshape 98.57 & \itshape 97.40 & \itshape 98.70 & \itshape 93.46 & \itshape 97.65 & \itshape 98.71 & \itshape 98.96 & \itshape 98.97 \\
\midrule
Norma, Lookup & 83.86 & 82.15 & 92.45 & 92.51 & 74.58 & 82.84 & 91.67 & 81.76 & 93.90 & 83.80\\
Norma, Rule-based & 76.48 & 82.52 & 90.85 & 88.59 & 78.73 & 83.72 & 86.33 & 86.09 & 91.63 & 85.23 \\
Norma, Distance-based & 58.92 & 73.30 & 83.92 & 84.41 & 62.38 & 69.95 & 77.28 & 71.02 & 88.20 & 76.03 \\
Norma (Combined) & 88.02 & 86.55 & 94.60 & 94.41 & 86.83 & \notsig{86.85} & 94.19 & 89.45 & 91.44 & 87.12\\
\midrule
cSMTiser                    & 88.82 & \notsig{88.06} & \notsig{95.21} & \notsig{95.01} & \notsig{91.63} & \notsig{87.10} & \notsig{95.09} & \notsig{93.18} & \notsig{95.99} & \textbf{91.13}\\
cSMTiser\textsubscript{+LM} & 86.69 & \notsig{88.19} & \textbf{95.24} & \textbf{95.02} & \textbf{91.70} & \notsig{86.83} & \textbf{95.18} & \textbf{93.30} & \textbf{96.01} & \notsig{91.11}\\
\midrule
NMT {\small\citep{BollmannPhD}} & 89.16 & \notsig{88.07} & 94.80 & \notsig{94.83} & 91.17 & 86.45 & 94.64 & 91.61 & 95.19 & 90.27\\
NMT {\small\citep{Tang-etal2018}} & \textbf{89.64} & \textbf{88.22} & 94.95 & \notsig{94.84} & \notsig{91.65} & \textbf{87.31} & 94.51 & 92.60 & \notsig{95.85} & 90.39\\
\midrule
$\dagger$SMT {\small\citep{Pettersson-etal2014}} & -- & -- & 94.3-- & -- & 80.1-- & 71.8-- & -- & -- & -- & 92.9-- \\
$\dagger$NMT {\small\citep{Tang-etal2018}} & -- & -- & 94.69 & -- & 91.69 & 87.59 & -- & -- & -- & 91.56 \\
\bottomrule
\end{tabular}

}

\caption{Word accuracy of different normalization methods on the test sets of the historical datasets, in percent; best result for each dataset in \textbf{bold}; results marked with an asterisk~(*) are not significantly different from the best result using McNemar's test at $p < 0.05$.  $\dagger$~indicates scores that were not (re)produced here, but reported in previous work; they might not be strictly comparable due to differences in data preprocessing (cf.\ Sec.\,\ref{sec:setup}).
Additionally, \textit{Identity} shows the accuracy when leaving all word forms unchanged, while \textit{Maximum} gives the theoretical maximum accuracy with purely token-level methods.}
  \label{tab:eval-test}
\end{table*}

%% file: tab_eval_ceri.tex
\newsavebox\cellbox
\newlength{\myl}
\newcolumntype{W}
{>{\begin{lrbox}\cellbox}%
l%
<{\end{lrbox}%
\let\hss\hfil
\settowidth{\myl}{00.00}%
\makebox[\myl][r]{\unhbox\cellbox}}}

\begin{table*}[t]

\newcommand{\na}{--}

\begin{subtable}[t]{1\textwidth}
  \centering
\resizebox{\textwidth}{!}{%
\setlength{\tabcolsep}{5pt}
\begin{tabular}{lWWWWWWWWWW}
\toprule
\textbf{Method} & \multicolumn{10}{c}{\textbf{Dataset}} \\
\cmidrule(lr){2-11}
 & \multicolumn{1}{c}{DE\textsubscript{A}} & \multicolumn{1}{c}{DE\textsubscript{R}} & \multicolumn{1}{c}{EN} & \multicolumn{1}{c}{ES} & \multicolumn{1}{c}{HU} & \multicolumn{1}{c}{IS} & \multicolumn{1}{c}{PT} & \multicolumn{1}{c}{SL\textsubscript{B}} & \multicolumn{1}{c}{SL\textsubscript{G}} & \multicolumn{1}{c}{SV}\\
\midrule
Norma, Lookup & 0.41 & 0.31 & \textbf{0.38} & \textbf{0.35} & 0.43 & \textbf{0.38} & \textbf{0.39} & \textbf{0.44} & 0.44 & 0.29\\
Norma, Rule-based & 0.40 & 0.33 & 0.43 & 0.39 & 0.38 & 0.40 & 0.45 & 0.47 & 0.46 & 0.32\\
Norma, Distance-based & 0.42 & 0.34 & 0.46 & 0.44 & 0.41 & 0.44 & 0.50 & 0.52 & \textbf{0.39} & 0.38\\
Norma (Combined) & {0.41} & {0.33} & {0.45} & {0.42} & {0.34} & {0.42} & {0.51} & {0.51} & {0.42} & {0.31}\\
\midrule
cSMTiser                    & \textbf{0.37} & \textbf{0.26} & 0.39 & 0.41 & \textbf{0.26} & 0.40 & 0.50 & 0.53 & 0.56 & 0.24\\
cSMTiser\textsubscript{+LM} & 0.39 & 0.27 & 0.39 & 0.42 & 0.27 & 0.41 & 0.50 & 0.53 & 0.56 & 0.24\\
\midrule
NMT {\small\citep{BollmannPhD}} & 0.38 & \textbf{0.26} & 0.39 & 0.43 & 0.27 & 0.40 & 0.48 & 0.47 & 0.51 & \textbf{0.23}\\
NMT {\small\citep{Tang-etal2018}} & 0.38 & 0.27 & \textbf{0.38} & 0.42 & \textbf{0.26} & 0.41 & 0.46 & 0.50 & 0.56 & 0.24\\
\bottomrule
\end{tabular}
}
\subcaption{\ceri{}: character error rate on the subset of incorrect normalizations (lower is better)\label{tab:eval-ceri}}
\end{subtable}

\vspace{.5em}
\begin{subtable}[t]{1\textwidth}
  \centering
\resizebox{\textwidth}{!}{%
\setlength{\tabcolsep}{5pt}
\begin{tabular}{lWWWWWWWWWW}
\toprule
Norma, Lookup & \hphantom{0}8.47 & 16.72 & \hphantom{0}8.33 & 27.81 & \hphantom{0}2.86 & \na & \hphantom{0}4.57 & \na & \na & \hphantom{0}7.42\\
Norma, Rule-based & 12.82 & 26.73 & \hphantom{0}8.48 & 26.33 & 12.44 & \na & \hphantom{0}5.65 & \na & \na & 15.38\\
Norma, Distance-based & \hphantom{0}7.71 & 19.10 & \hphantom{0}7.26 & 28.62 & 10.93 & \na & \hphantom{0}5.58 & \na & \na & 12.18\\
Norma (Combined) & 16.97 & 28.94 & \hphantom{0}\textbf{9.86} & 43.55 & 20.00 & \na & \textbf{11.13} & \na & \na & \textbf{20.75}\\
\midrule
cSMTiser                    & \textbf{17.92} & \textbf{34.67} & \hphantom{0}8.27 & \textbf{43.82} & \textbf{20.44} & \na & \hphantom{0}6.09 & \na & \na & 17.73 \\
cSMTiser\textsubscript{+LM} & 12.23 & 33.83 & \hphantom{0}8.46 & 42.93 & 20.40 & \na & \hphantom{0}6.60 & \na & \na & 17.58 \\
\midrule
NMT {\small\citep{BollmannPhD}} & 17.24 & 33.65 & \hphantom{0}7.96 & 39.22 & 18.30 & \na & \hphantom{0}5.92 & \na & \na & 16.65\\
NMT {\small\citep{Tang-etal2018}} & 16.34 & 34.19 & \hphantom{0}9.43 & 40.99 & 19.84 & \na & \hphantom{0}6.46 & \na & \na & 18.93\\
\bottomrule
\end{tabular}
}
\subcaption{Stemming accuracy: percentage of incorrect normalizations with correct word stems (higher is better)\label{tab:eval-stem}}
\end{subtable}


\caption{Evaluations on the subset of incorrect normalizations only; best results for each dataset in \textbf{bold}.  Note that this subset is different for each system, so for comparisons between systems, these numbers should be considered in conjunction with word accuracy scores from Table~\protect\ref{tab:eval-test}.}
  \label{tab:eval-incorrect}
\end{table*}

%% file: curves_english-icamet.pgf
\begingroup%
\makeatletter%
\begin{pgfpicture}%
\pgfpathrectangle{\pgfpointorigin}{\pgfqpoint{3.078809in}{2.168389in}}%
\pgfusepath{use as bounding box, clip}%
\begin{pgfscope}%
\pgfsetbuttcap%
\pgfsetmiterjoin%
\definecolor{currentfill}{rgb}{1.000000,1.000000,1.000000}%
\pgfsetfillcolor{currentfill}%
\pgfsetlinewidth{0.000000pt}%
\definecolor{currentstroke}{rgb}{1.000000,1.000000,1.000000}%
\pgfsetstrokecolor{currentstroke}%
\pgfsetdash{}{0pt}%
\pgfpathmoveto{\pgfqpoint{0.000000in}{0.000000in}}%
\pgfpathlineto{\pgfqpoint{3.078809in}{0.000000in}}%
\pgfpathlineto{\pgfqpoint{3.078809in}{2.168389in}}%
\pgfpathlineto{\pgfqpoint{0.000000in}{2.168389in}}%
\pgfpathclose%
\pgfusepath{fill}%
\end{pgfscope}%
\begin{pgfscope}%
\pgfsetbuttcap%
\pgfsetmiterjoin%
\definecolor{currentfill}{rgb}{1.000000,1.000000,1.000000}%
\pgfsetfillcolor{currentfill}%
\pgfsetlinewidth{0.000000pt}%
\definecolor{currentstroke}{rgb}{0.000000,0.000000,0.000000}%
\pgfsetstrokecolor{currentstroke}%
\pgfsetstrokeopacity{0.000000}%
\pgfsetdash{}{0pt}%
\pgfpathmoveto{\pgfqpoint{0.450809in}{0.326389in}}%
\pgfpathlineto{\pgfqpoint{2.930809in}{0.326389in}}%
\pgfpathlineto{\pgfqpoint{2.930809in}{2.020389in}}%
\pgfpathlineto{\pgfqpoint{0.450809in}{2.020389in}}%
\pgfpathclose%
\pgfusepath{fill}%
\end{pgfscope}%
\begin{pgfscope}%
\pgfsetbuttcap%
\pgfsetroundjoin%
\definecolor{currentfill}{rgb}{0.150000,0.150000,0.150000}%
\pgfsetfillcolor{currentfill}%
\pgfsetlinewidth{1.003750pt}%
\definecolor{currentstroke}{rgb}{0.150000,0.150000,0.150000}%
\pgfsetstrokecolor{currentstroke}%
\pgfsetdash{}{0pt}%
\pgfsys@defobject{currentmarker}{\pgfqpoint{0.000000in}{-0.066667in}}{\pgfqpoint{0.000000in}{0.000000in}}{%
\pgfpathmoveto{\pgfqpoint{0.000000in}{0.000000in}}%
\pgfpathlineto{\pgfqpoint{0.000000in}{-0.066667in}}%
\pgfusepath{stroke,fill}%
}%
\begin{pgfscope}%
\pgfsys@transformshift{0.563536in}{0.326389in}%
\pgfsys@useobject{currentmarker}{}%
\end{pgfscope}%
\end{pgfscope}%
\begin{pgfscope}%
\definecolor{textcolor}{rgb}{0.150000,0.150000,0.150000}%
\pgfsetstrokecolor{textcolor}%
\pgfsetfillcolor{textcolor}%
\pgftext[x=0.563536in,y=0.211111in,,top]{\color{textcolor}\sffamily\fontsize{8.800000}{10.560000}\selectfont 100}%
\end{pgfscope}%
\begin{pgfscope}%
\pgfsetbuttcap%
\pgfsetroundjoin%
\definecolor{currentfill}{rgb}{0.150000,0.150000,0.150000}%
\pgfsetfillcolor{currentfill}%
\pgfsetlinewidth{1.003750pt}%
\definecolor{currentstroke}{rgb}{0.150000,0.150000,0.150000}%
\pgfsetstrokecolor{currentstroke}%
\pgfsetdash{}{0pt}%
\pgfsys@defobject{currentmarker}{\pgfqpoint{0.000000in}{-0.066667in}}{\pgfqpoint{0.000000in}{0.000000in}}{%
\pgfpathmoveto{\pgfqpoint{0.000000in}{0.000000in}}%
\pgfpathlineto{\pgfqpoint{0.000000in}{-0.066667in}}%
\pgfusepath{stroke,fill}%
}%
\begin{pgfscope}%
\pgfsys@transformshift{0.895949in}{0.326389in}%
\pgfsys@useobject{currentmarker}{}%
\end{pgfscope}%
\end{pgfscope}%
\begin{pgfscope}%
\definecolor{textcolor}{rgb}{0.150000,0.150000,0.150000}%
\pgfsetstrokecolor{textcolor}%
\pgfsetfillcolor{textcolor}%
\pgftext[x=0.895949in,y=0.211111in,,top]{\color{textcolor}\sffamily\fontsize{8.800000}{10.560000}\selectfont 250}%
\end{pgfscope}%
\begin{pgfscope}%
\pgfsetbuttcap%
\pgfsetroundjoin%
\definecolor{currentfill}{rgb}{0.150000,0.150000,0.150000}%
\pgfsetfillcolor{currentfill}%
\pgfsetlinewidth{1.003750pt}%
\definecolor{currentstroke}{rgb}{0.150000,0.150000,0.150000}%
\pgfsetstrokecolor{currentstroke}%
\pgfsetdash{}{0pt}%
\pgfsys@defobject{currentmarker}{\pgfqpoint{0.000000in}{-0.066667in}}{\pgfqpoint{0.000000in}{0.000000in}}{%
\pgfpathmoveto{\pgfqpoint{0.000000in}{0.000000in}}%
\pgfpathlineto{\pgfqpoint{0.000000in}{-0.066667in}}%
\pgfusepath{stroke,fill}%
}%
\begin{pgfscope}%
\pgfsys@transformshift{1.147410in}{0.326389in}%
\pgfsys@useobject{currentmarker}{}%
\end{pgfscope}%
\end{pgfscope}%
\begin{pgfscope}%
\definecolor{textcolor}{rgb}{0.150000,0.150000,0.150000}%
\pgfsetstrokecolor{textcolor}%
\pgfsetfillcolor{textcolor}%
\pgftext[x=1.147410in,y=0.211111in,,top]{\color{textcolor}\sffamily\fontsize{8.800000}{10.560000}\selectfont 500}%
\end{pgfscope}%
\begin{pgfscope}%
\pgfsetbuttcap%
\pgfsetroundjoin%
\definecolor{currentfill}{rgb}{0.150000,0.150000,0.150000}%
\pgfsetfillcolor{currentfill}%
\pgfsetlinewidth{1.003750pt}%
\definecolor{currentstroke}{rgb}{0.150000,0.150000,0.150000}%
\pgfsetstrokecolor{currentstroke}%
\pgfsetdash{}{0pt}%
\pgfsys@defobject{currentmarker}{\pgfqpoint{0.000000in}{-0.066667in}}{\pgfqpoint{0.000000in}{0.000000in}}{%
\pgfpathmoveto{\pgfqpoint{0.000000in}{0.000000in}}%
\pgfpathlineto{\pgfqpoint{0.000000in}{-0.066667in}}%
\pgfusepath{stroke,fill}%
}%
\begin{pgfscope}%
\pgfsys@transformshift{1.398871in}{0.326389in}%
\pgfsys@useobject{currentmarker}{}%
\end{pgfscope}%
\end{pgfscope}%
\begin{pgfscope}%
\definecolor{textcolor}{rgb}{0.150000,0.150000,0.150000}%
\pgfsetstrokecolor{textcolor}%
\pgfsetfillcolor{textcolor}%
\pgftext[x=1.398871in,y=0.211111in,,top]{\color{textcolor}\sffamily\fontsize{8.800000}{10.560000}\selectfont 1k}%
\end{pgfscope}%
\begin{pgfscope}%
\pgfsetbuttcap%
\pgfsetroundjoin%
\definecolor{currentfill}{rgb}{0.150000,0.150000,0.150000}%
\pgfsetfillcolor{currentfill}%
\pgfsetlinewidth{1.003750pt}%
\definecolor{currentstroke}{rgb}{0.150000,0.150000,0.150000}%
\pgfsetstrokecolor{currentstroke}%
\pgfsetdash{}{0pt}%
\pgfsys@defobject{currentmarker}{\pgfqpoint{0.000000in}{-0.066667in}}{\pgfqpoint{0.000000in}{0.000000in}}{%
\pgfpathmoveto{\pgfqpoint{0.000000in}{0.000000in}}%
\pgfpathlineto{\pgfqpoint{0.000000in}{-0.066667in}}%
\pgfusepath{stroke,fill}%
}%
\begin{pgfscope}%
\pgfsys@transformshift{1.731285in}{0.326389in}%
\pgfsys@useobject{currentmarker}{}%
\end{pgfscope}%
\end{pgfscope}%
\begin{pgfscope}%
\definecolor{textcolor}{rgb}{0.150000,0.150000,0.150000}%
\pgfsetstrokecolor{textcolor}%
\pgfsetfillcolor{textcolor}%
\pgftext[x=1.731285in,y=0.211111in,,top]{\color{textcolor}\sffamily\fontsize{8.800000}{10.560000}\selectfont 2.5k}%
\end{pgfscope}%
\begin{pgfscope}%
\pgfsetbuttcap%
\pgfsetroundjoin%
\definecolor{currentfill}{rgb}{0.150000,0.150000,0.150000}%
\pgfsetfillcolor{currentfill}%
\pgfsetlinewidth{1.003750pt}%
\definecolor{currentstroke}{rgb}{0.150000,0.150000,0.150000}%
\pgfsetstrokecolor{currentstroke}%
\pgfsetdash{}{0pt}%
\pgfsys@defobject{currentmarker}{\pgfqpoint{0.000000in}{-0.066667in}}{\pgfqpoint{0.000000in}{0.000000in}}{%
\pgfpathmoveto{\pgfqpoint{0.000000in}{0.000000in}}%
\pgfpathlineto{\pgfqpoint{0.000000in}{-0.066667in}}%
\pgfusepath{stroke,fill}%
}%
\begin{pgfscope}%
\pgfsys@transformshift{1.982746in}{0.326389in}%
\pgfsys@useobject{currentmarker}{}%
\end{pgfscope}%
\end{pgfscope}%
\begin{pgfscope}%
\definecolor{textcolor}{rgb}{0.150000,0.150000,0.150000}%
\pgfsetstrokecolor{textcolor}%
\pgfsetfillcolor{textcolor}%
\pgftext[x=1.982746in,y=0.211111in,,top]{\color{textcolor}\sffamily\fontsize{8.800000}{10.560000}\selectfont 5k}%
\end{pgfscope}%
\begin{pgfscope}%
\pgfsetbuttcap%
\pgfsetroundjoin%
\definecolor{currentfill}{rgb}{0.150000,0.150000,0.150000}%
\pgfsetfillcolor{currentfill}%
\pgfsetlinewidth{1.003750pt}%
\definecolor{currentstroke}{rgb}{0.150000,0.150000,0.150000}%
\pgfsetstrokecolor{currentstroke}%
\pgfsetdash{}{0pt}%
\pgfsys@defobject{currentmarker}{\pgfqpoint{0.000000in}{-0.066667in}}{\pgfqpoint{0.000000in}{0.000000in}}{%
\pgfpathmoveto{\pgfqpoint{0.000000in}{0.000000in}}%
\pgfpathlineto{\pgfqpoint{0.000000in}{-0.066667in}}%
\pgfusepath{stroke,fill}%
}%
\begin{pgfscope}%
\pgfsys@transformshift{2.234207in}{0.326389in}%
\pgfsys@useobject{currentmarker}{}%
\end{pgfscope}%
\end{pgfscope}%
\begin{pgfscope}%
\definecolor{textcolor}{rgb}{0.150000,0.150000,0.150000}%
\pgfsetstrokecolor{textcolor}%
\pgfsetfillcolor{textcolor}%
\pgftext[x=2.234207in,y=0.211111in,,top]{\color{textcolor}\sffamily\fontsize{8.800000}{10.560000}\selectfont 10k}%
\end{pgfscope}%
\begin{pgfscope}%
\pgfsetbuttcap%
\pgfsetroundjoin%
\definecolor{currentfill}{rgb}{0.150000,0.150000,0.150000}%
\pgfsetfillcolor{currentfill}%
\pgfsetlinewidth{1.003750pt}%
\definecolor{currentstroke}{rgb}{0.150000,0.150000,0.150000}%
\pgfsetstrokecolor{currentstroke}%
\pgfsetdash{}{0pt}%
\pgfsys@defobject{currentmarker}{\pgfqpoint{0.000000in}{-0.066667in}}{\pgfqpoint{0.000000in}{0.000000in}}{%
\pgfpathmoveto{\pgfqpoint{0.000000in}{0.000000in}}%
\pgfpathlineto{\pgfqpoint{0.000000in}{-0.066667in}}%
\pgfusepath{stroke,fill}%
}%
\begin{pgfscope}%
\pgfsys@transformshift{2.566620in}{0.326389in}%
\pgfsys@useobject{currentmarker}{}%
\end{pgfscope}%
\end{pgfscope}%
\begin{pgfscope}%
\definecolor{textcolor}{rgb}{0.150000,0.150000,0.150000}%
\pgfsetstrokecolor{textcolor}%
\pgfsetfillcolor{textcolor}%
\pgftext[x=2.566620in,y=0.211111in,,top]{\color{textcolor}\sffamily\fontsize{8.800000}{10.560000}\selectfont 25k}%
\end{pgfscope}%
\begin{pgfscope}%
\pgfsetbuttcap%
\pgfsetroundjoin%
\definecolor{currentfill}{rgb}{0.150000,0.150000,0.150000}%
\pgfsetfillcolor{currentfill}%
\pgfsetlinewidth{1.003750pt}%
\definecolor{currentstroke}{rgb}{0.150000,0.150000,0.150000}%
\pgfsetstrokecolor{currentstroke}%
\pgfsetdash{}{0pt}%
\pgfsys@defobject{currentmarker}{\pgfqpoint{0.000000in}{-0.066667in}}{\pgfqpoint{0.000000in}{0.000000in}}{%
\pgfpathmoveto{\pgfqpoint{0.000000in}{0.000000in}}%
\pgfpathlineto{\pgfqpoint{0.000000in}{-0.066667in}}%
\pgfusepath{stroke,fill}%
}%
\begin{pgfscope}%
\pgfsys@transformshift{2.818081in}{0.326389in}%
\pgfsys@useobject{currentmarker}{}%
\end{pgfscope}%
\end{pgfscope}%
\begin{pgfscope}%
\definecolor{textcolor}{rgb}{0.150000,0.150000,0.150000}%
\pgfsetstrokecolor{textcolor}%
\pgfsetfillcolor{textcolor}%
\pgftext[x=2.818081in,y=0.211111in,,top]{\color{textcolor}\sffamily\fontsize{8.800000}{10.560000}\selectfont 50k}%
\end{pgfscope}%
\begin{pgfscope}%
\pgfsetbuttcap%
\pgfsetroundjoin%
\definecolor{currentfill}{rgb}{0.150000,0.150000,0.150000}%
\pgfsetfillcolor{currentfill}%
\pgfsetlinewidth{1.003750pt}%
\definecolor{currentstroke}{rgb}{0.150000,0.150000,0.150000}%
\pgfsetstrokecolor{currentstroke}%
\pgfsetdash{}{0pt}%
\pgfsys@defobject{currentmarker}{\pgfqpoint{-0.066667in}{0.000000in}}{\pgfqpoint{0.000000in}{0.000000in}}{%
\pgfpathmoveto{\pgfqpoint{0.000000in}{0.000000in}}%
\pgfpathlineto{\pgfqpoint{-0.066667in}{0.000000in}}%
\pgfusepath{stroke,fill}%
}%
\begin{pgfscope}%
\pgfsys@transformshift{0.450809in}{0.526191in}%
\pgfsys@useobject{currentmarker}{}%
\end{pgfscope}%
\end{pgfscope}%
\begin{pgfscope}%
\definecolor{textcolor}{rgb}{0.150000,0.150000,0.150000}%
\pgfsetstrokecolor{textcolor}%
\pgfsetfillcolor{textcolor}%
\pgftext[x=0.100000in,y=0.482788in,left,base]{\color{textcolor}\sffamily\fontsize{8.800000}{10.560000}\selectfont 20\%}%
\end{pgfscope}%
\begin{pgfscope}%
\pgfsetbuttcap%
\pgfsetroundjoin%
\definecolor{currentfill}{rgb}{0.150000,0.150000,0.150000}%
\pgfsetfillcolor{currentfill}%
\pgfsetlinewidth{1.003750pt}%
\definecolor{currentstroke}{rgb}{0.150000,0.150000,0.150000}%
\pgfsetstrokecolor{currentstroke}%
\pgfsetdash{}{0pt}%
\pgfsys@defobject{currentmarker}{\pgfqpoint{-0.066667in}{0.000000in}}{\pgfqpoint{0.000000in}{0.000000in}}{%
\pgfpathmoveto{\pgfqpoint{0.000000in}{0.000000in}}%
\pgfpathlineto{\pgfqpoint{-0.066667in}{0.000000in}}%
\pgfusepath{stroke,fill}%
}%
\begin{pgfscope}%
\pgfsys@transformshift{0.450809in}{0.925794in}%
\pgfsys@useobject{currentmarker}{}%
\end{pgfscope}%
\end{pgfscope}%
\begin{pgfscope}%
\definecolor{textcolor}{rgb}{0.150000,0.150000,0.150000}%
\pgfsetstrokecolor{textcolor}%
\pgfsetfillcolor{textcolor}%
\pgftext[x=0.100000in,y=0.882391in,left,base]{\color{textcolor}\sffamily\fontsize{8.800000}{10.560000}\selectfont 40\%}%
\end{pgfscope}%
\begin{pgfscope}%
\pgfsetbuttcap%
\pgfsetroundjoin%
\definecolor{currentfill}{rgb}{0.150000,0.150000,0.150000}%
\pgfsetfillcolor{currentfill}%
\pgfsetlinewidth{1.003750pt}%
\definecolor{currentstroke}{rgb}{0.150000,0.150000,0.150000}%
\pgfsetstrokecolor{currentstroke}%
\pgfsetdash{}{0pt}%
\pgfsys@defobject{currentmarker}{\pgfqpoint{-0.066667in}{0.000000in}}{\pgfqpoint{0.000000in}{0.000000in}}{%
\pgfpathmoveto{\pgfqpoint{0.000000in}{0.000000in}}%
\pgfpathlineto{\pgfqpoint{-0.066667in}{0.000000in}}%
\pgfusepath{stroke,fill}%
}%
\begin{pgfscope}%
\pgfsys@transformshift{0.450809in}{1.325398in}%
\pgfsys@useobject{currentmarker}{}%
\end{pgfscope}%
\end{pgfscope}%
\begin{pgfscope}%
\definecolor{textcolor}{rgb}{0.150000,0.150000,0.150000}%
\pgfsetstrokecolor{textcolor}%
\pgfsetfillcolor{textcolor}%
\pgftext[x=0.100000in,y=1.281995in,left,base]{\color{textcolor}\sffamily\fontsize{8.800000}{10.560000}\selectfont 60\%}%
\end{pgfscope}%
\begin{pgfscope}%
\pgfsetbuttcap%
\pgfsetroundjoin%
\definecolor{currentfill}{rgb}{0.150000,0.150000,0.150000}%
\pgfsetfillcolor{currentfill}%
\pgfsetlinewidth{1.003750pt}%
\definecolor{currentstroke}{rgb}{0.150000,0.150000,0.150000}%
\pgfsetstrokecolor{currentstroke}%
\pgfsetdash{}{0pt}%
\pgfsys@defobject{currentmarker}{\pgfqpoint{-0.066667in}{0.000000in}}{\pgfqpoint{0.000000in}{0.000000in}}{%
\pgfpathmoveto{\pgfqpoint{0.000000in}{0.000000in}}%
\pgfpathlineto{\pgfqpoint{-0.066667in}{0.000000in}}%
\pgfusepath{stroke,fill}%
}%
\begin{pgfscope}%
\pgfsys@transformshift{0.450809in}{1.725001in}%
\pgfsys@useobject{currentmarker}{}%
\end{pgfscope}%
\end{pgfscope}%
\begin{pgfscope}%
\definecolor{textcolor}{rgb}{0.150000,0.150000,0.150000}%
\pgfsetstrokecolor{textcolor}%
\pgfsetfillcolor{textcolor}%
\pgftext[x=0.100000in,y=1.681599in,left,base]{\color{textcolor}\sffamily\fontsize{8.800000}{10.560000}\selectfont 80\%}%
\end{pgfscope}%
\begin{pgfscope}%
\pgfsetbuttcap%
\pgfsetroundjoin%
\definecolor{currentfill}{rgb}{0.150000,0.150000,0.150000}%
\pgfsetfillcolor{currentfill}%
\pgfsetlinewidth{0.803000pt}%
\definecolor{currentstroke}{rgb}{0.150000,0.150000,0.150000}%
\pgfsetstrokecolor{currentstroke}%
\pgfsetdash{}{0pt}%
\pgfsys@defobject{currentmarker}{\pgfqpoint{-0.044444in}{0.000000in}}{\pgfqpoint{0.000000in}{0.000000in}}{%
\pgfpathmoveto{\pgfqpoint{0.000000in}{0.000000in}}%
\pgfpathlineto{\pgfqpoint{-0.044444in}{0.000000in}}%
\pgfusepath{stroke,fill}%
}%
\begin{pgfscope}%
\pgfsys@transformshift{0.450809in}{0.326389in}%
\pgfsys@useobject{currentmarker}{}%
\end{pgfscope}%
\end{pgfscope}%
\begin{pgfscope}%
\pgfsetbuttcap%
\pgfsetroundjoin%
\definecolor{currentfill}{rgb}{0.150000,0.150000,0.150000}%
\pgfsetfillcolor{currentfill}%
\pgfsetlinewidth{0.803000pt}%
\definecolor{currentstroke}{rgb}{0.150000,0.150000,0.150000}%
\pgfsetstrokecolor{currentstroke}%
\pgfsetdash{}{0pt}%
\pgfsys@defobject{currentmarker}{\pgfqpoint{-0.044444in}{0.000000in}}{\pgfqpoint{0.000000in}{0.000000in}}{%
\pgfpathmoveto{\pgfqpoint{0.000000in}{0.000000in}}%
\pgfpathlineto{\pgfqpoint{-0.044444in}{0.000000in}}%
\pgfusepath{stroke,fill}%
}%
\begin{pgfscope}%
\pgfsys@transformshift{0.450809in}{0.725992in}%
\pgfsys@useobject{currentmarker}{}%
\end{pgfscope}%
\end{pgfscope}%
\begin{pgfscope}%
\pgfsetbuttcap%
\pgfsetroundjoin%
\definecolor{currentfill}{rgb}{0.150000,0.150000,0.150000}%
\pgfsetfillcolor{currentfill}%
\pgfsetlinewidth{0.803000pt}%
\definecolor{currentstroke}{rgb}{0.150000,0.150000,0.150000}%
\pgfsetstrokecolor{currentstroke}%
\pgfsetdash{}{0pt}%
\pgfsys@defobject{currentmarker}{\pgfqpoint{-0.044444in}{0.000000in}}{\pgfqpoint{0.000000in}{0.000000in}}{%
\pgfpathmoveto{\pgfqpoint{0.000000in}{0.000000in}}%
\pgfpathlineto{\pgfqpoint{-0.044444in}{0.000000in}}%
\pgfusepath{stroke,fill}%
}%
\begin{pgfscope}%
\pgfsys@transformshift{0.450809in}{1.125596in}%
\pgfsys@useobject{currentmarker}{}%
\end{pgfscope}%
\end{pgfscope}%
\begin{pgfscope}%
\pgfsetbuttcap%
\pgfsetroundjoin%
\definecolor{currentfill}{rgb}{0.150000,0.150000,0.150000}%
\pgfsetfillcolor{currentfill}%
\pgfsetlinewidth{0.803000pt}%
\definecolor{currentstroke}{rgb}{0.150000,0.150000,0.150000}%
\pgfsetstrokecolor{currentstroke}%
\pgfsetdash{}{0pt}%
\pgfsys@defobject{currentmarker}{\pgfqpoint{-0.044444in}{0.000000in}}{\pgfqpoint{0.000000in}{0.000000in}}{%
\pgfpathmoveto{\pgfqpoint{0.000000in}{0.000000in}}%
\pgfpathlineto{\pgfqpoint{-0.044444in}{0.000000in}}%
\pgfusepath{stroke,fill}%
}%
\begin{pgfscope}%
\pgfsys@transformshift{0.450809in}{1.525200in}%
\pgfsys@useobject{currentmarker}{}%
\end{pgfscope}%
\end{pgfscope}%
\begin{pgfscope}%
\pgfsetbuttcap%
\pgfsetroundjoin%
\definecolor{currentfill}{rgb}{0.150000,0.150000,0.150000}%
\pgfsetfillcolor{currentfill}%
\pgfsetlinewidth{0.803000pt}%
\definecolor{currentstroke}{rgb}{0.150000,0.150000,0.150000}%
\pgfsetstrokecolor{currentstroke}%
\pgfsetdash{}{0pt}%
\pgfsys@defobject{currentmarker}{\pgfqpoint{-0.044444in}{0.000000in}}{\pgfqpoint{0.000000in}{0.000000in}}{%
\pgfpathmoveto{\pgfqpoint{0.000000in}{0.000000in}}%
\pgfpathlineto{\pgfqpoint{-0.044444in}{0.000000in}}%
\pgfusepath{stroke,fill}%
}%
\begin{pgfscope}%
\pgfsys@transformshift{0.450809in}{1.924803in}%
\pgfsys@useobject{currentmarker}{}%
\end{pgfscope}%
\end{pgfscope}%
\begin{pgfscope}%
\pgfpathrectangle{\pgfqpoint{0.450809in}{0.326389in}}{\pgfqpoint{2.480000in}{1.694000in}}%
\pgfusepath{clip}%
\pgfsetbuttcap%
\pgfsetroundjoin%
\pgfsetlinewidth{1.204500pt}%
\definecolor{currentstroke}{rgb}{1.000000,0.498039,0.000000}%
\pgfsetstrokecolor{currentstroke}%
\pgfsetdash{{7.680000pt}{1.920000pt}{1.200000pt}{1.920000pt}}{0.000000pt}%
\pgfpathmoveto{\pgfqpoint{0.563536in}{1.784781in}}%
\pgfpathlineto{\pgfqpoint{0.895949in}{1.799777in}}%
\pgfpathlineto{\pgfqpoint{1.147410in}{1.811655in}}%
\pgfpathlineto{\pgfqpoint{1.398871in}{1.815594in}}%
\pgfpathlineto{\pgfqpoint{1.731285in}{1.829930in}}%
\pgfpathlineto{\pgfqpoint{1.982746in}{1.853624in}}%
\pgfpathlineto{\pgfqpoint{2.234207in}{1.881391in}}%
\pgfpathlineto{\pgfqpoint{2.566620in}{1.905427in}}%
\pgfpathlineto{\pgfqpoint{2.818081in}{1.934607in}}%
\pgfusepath{stroke}%
\end{pgfscope}%
\begin{pgfscope}%
\pgfpathrectangle{\pgfqpoint{0.450809in}{0.326389in}}{\pgfqpoint{2.480000in}{1.694000in}}%
\pgfusepath{clip}%
\pgfsetroundcap%
\pgfsetroundjoin%
\pgfsetlinewidth{1.204500pt}%
\definecolor{currentstroke}{rgb}{0.890196,0.101961,0.109804}%
\pgfsetstrokecolor{currentstroke}%
\pgfsetdash{}{0pt}%
\pgfpathmoveto{\pgfqpoint{0.563536in}{1.578655in}}%
\pgfpathlineto{\pgfqpoint{0.895949in}{1.694494in}}%
\pgfpathlineto{\pgfqpoint{1.147410in}{1.733222in}}%
\pgfpathlineto{\pgfqpoint{1.398871in}{1.801160in}}%
\pgfpathlineto{\pgfqpoint{1.731285in}{1.867091in}}%
\pgfpathlineto{\pgfqpoint{1.982746in}{1.881195in}}%
\pgfpathlineto{\pgfqpoint{2.234207in}{1.905574in}}%
\pgfpathlineto{\pgfqpoint{2.566620in}{1.934883in}}%
\pgfpathlineto{\pgfqpoint{2.818081in}{1.960234in}}%
\pgfusepath{stroke}%
\end{pgfscope}%
\begin{pgfscope}%
\pgfpathrectangle{\pgfqpoint{0.450809in}{0.326389in}}{\pgfqpoint{2.480000in}{1.694000in}}%
\pgfusepath{clip}%
\pgfsetbuttcap%
\pgfsetroundjoin%
\pgfsetlinewidth{1.204500pt}%
\definecolor{currentstroke}{rgb}{0.200000,0.627451,0.172549}%
\pgfsetstrokecolor{currentstroke}%
\pgfsetdash{{4.440000pt}{1.920000pt}}{0.000000pt}%
\pgfpathmoveto{\pgfqpoint{0.563536in}{0.757136in}}%
\pgfpathlineto{\pgfqpoint{0.895949in}{1.089474in}}%
\pgfpathlineto{\pgfqpoint{1.147410in}{1.321337in}}%
\pgfpathlineto{\pgfqpoint{1.398871in}{1.463806in}}%
\pgfpathlineto{\pgfqpoint{1.731285in}{1.615119in}}%
\pgfpathlineto{\pgfqpoint{1.982746in}{1.724634in}}%
\pgfpathlineto{\pgfqpoint{2.234207in}{1.804462in}}%
\pgfpathlineto{\pgfqpoint{2.566620in}{1.854455in}}%
\pgfpathlineto{\pgfqpoint{2.818081in}{1.905128in}}%
\pgfusepath{stroke}%
\end{pgfscope}%
\begin{pgfscope}%
\pgfpathrectangle{\pgfqpoint{0.450809in}{0.326389in}}{\pgfqpoint{2.480000in}{1.694000in}}%
\pgfusepath{clip}%
\pgfsetbuttcap%
\pgfsetroundjoin%
\pgfsetlinewidth{1.204500pt}%
\definecolor{currentstroke}{rgb}{0.121569,0.470588,0.705882}%
\pgfsetstrokecolor{currentstroke}%
\pgfsetdash{{1.200000pt}{1.980000pt}}{0.000000pt}%
\pgfpathmoveto{\pgfqpoint{0.563536in}{0.931030in}}%
\pgfpathlineto{\pgfqpoint{0.895949in}{1.279661in}}%
\pgfpathlineto{\pgfqpoint{1.147410in}{1.461164in}}%
\pgfpathlineto{\pgfqpoint{1.398871in}{1.584600in}}%
\pgfpathlineto{\pgfqpoint{1.731285in}{1.670372in}}%
\pgfpathlineto{\pgfqpoint{1.982746in}{1.766285in}}%
\pgfpathlineto{\pgfqpoint{2.234207in}{1.818945in}}%
\pgfpathlineto{\pgfqpoint{2.566620in}{1.847043in}}%
\pgfpathlineto{\pgfqpoint{2.818081in}{1.912956in}}%
\pgfusepath{stroke}%
\end{pgfscope}%
\begin{pgfscope}%
\pgfsetrectcap%
\pgfsetmiterjoin%
\pgfsetlinewidth{1.003750pt}%
\definecolor{currentstroke}{rgb}{0.150000,0.150000,0.150000}%
\pgfsetstrokecolor{currentstroke}%
\pgfsetdash{}{0pt}%
\pgfpathmoveto{\pgfqpoint{0.450809in}{0.326389in}}%
\pgfpathlineto{\pgfqpoint{0.450809in}{2.020389in}}%
\pgfusepath{stroke}%
\end{pgfscope}%
\begin{pgfscope}%
\pgfsetrectcap%
\pgfsetmiterjoin%
\pgfsetlinewidth{1.003750pt}%
\definecolor{currentstroke}{rgb}{0.150000,0.150000,0.150000}%
\pgfsetstrokecolor{currentstroke}%
\pgfsetdash{}{0pt}%
\pgfpathmoveto{\pgfqpoint{2.930809in}{0.326389in}}%
\pgfpathlineto{\pgfqpoint{2.930809in}{2.020389in}}%
\pgfusepath{stroke}%
\end{pgfscope}%
\begin{pgfscope}%
\pgfsetrectcap%
\pgfsetmiterjoin%
\pgfsetlinewidth{1.003750pt}%
\definecolor{currentstroke}{rgb}{0.150000,0.150000,0.150000}%
\pgfsetstrokecolor{currentstroke}%
\pgfsetdash{}{0pt}%
\pgfpathmoveto{\pgfqpoint{0.450809in}{0.326389in}}%
\pgfpathlineto{\pgfqpoint{2.930809in}{0.326389in}}%
\pgfusepath{stroke}%
\end{pgfscope}%
\begin{pgfscope}%
\pgfsetrectcap%
\pgfsetmiterjoin%
\pgfsetlinewidth{1.003750pt}%
\definecolor{currentstroke}{rgb}{0.150000,0.150000,0.150000}%
\pgfsetstrokecolor{currentstroke}%
\pgfsetdash{}{0pt}%
\pgfpathmoveto{\pgfqpoint{0.450809in}{2.020389in}}%
\pgfpathlineto{\pgfqpoint{2.930809in}{2.020389in}}%
\pgfusepath{stroke}%
\end{pgfscope}%
\begin{pgfscope}%
\pgfsetbuttcap%
\pgfsetroundjoin%
\pgfsetlinewidth{1.204500pt}%
\definecolor{currentstroke}{rgb}{1.000000,0.498039,0.000000}%
\pgfsetstrokecolor{currentstroke}%
\pgfsetdash{{7.680000pt}{1.920000pt}{1.200000pt}{1.920000pt}}{0.000000pt}%
\pgfpathmoveto{\pgfqpoint{1.816819in}{0.971250in}}%
\pgfpathlineto{\pgfqpoint{2.061263in}{0.971250in}}%
\pgfusepath{stroke}%
\end{pgfscope}%
\begin{pgfscope}%
\definecolor{textcolor}{rgb}{0.150000,0.150000,0.150000}%
\pgfsetstrokecolor{textcolor}%
\pgfsetfillcolor{textcolor}%
\pgftext[x=2.159041in,y=0.928472in,left,base]{\color{textcolor}\sffamily\fontsize{8.800000}{10.560000}\selectfont Norma}%
\end{pgfscope}%
\begin{pgfscope}%
\pgfsetroundcap%
\pgfsetroundjoin%
\pgfsetlinewidth{1.204500pt}%
\definecolor{currentstroke}{rgb}{0.890196,0.101961,0.109804}%
\pgfsetstrokecolor{currentstroke}%
\pgfsetdash{}{0pt}%
\pgfpathmoveto{\pgfqpoint{1.816819in}{0.799028in}}%
\pgfpathlineto{\pgfqpoint{2.061263in}{0.799028in}}%
\pgfusepath{stroke}%
\end{pgfscope}%
\begin{pgfscope}%
\definecolor{textcolor}{rgb}{0.150000,0.150000,0.150000}%
\pgfsetstrokecolor{textcolor}%
\pgfsetfillcolor{textcolor}%
\pgftext[x=2.159041in,y=0.756250in,left,base]{\color{textcolor}\sffamily\fontsize{8.800000}{10.560000}\selectfont cSMTiser\textsubscript{+LM}}%
\end{pgfscope}%
\begin{pgfscope}%
\pgfsetbuttcap%
\pgfsetroundjoin%
\pgfsetlinewidth{1.204500pt}%
\definecolor{currentstroke}{rgb}{0.200000,0.627451,0.172549}%
\pgfsetstrokecolor{currentstroke}%
\pgfsetdash{{4.440000pt}{1.920000pt}}{0.000000pt}%
\pgfpathmoveto{\pgfqpoint{1.816819in}{0.626806in}}%
\pgfpathlineto{\pgfqpoint{2.061263in}{0.626806in}}%
\pgfusepath{stroke}%
\end{pgfscope}%
\begin{pgfscope}%
\definecolor{textcolor}{rgb}{0.150000,0.150000,0.150000}%
\pgfsetstrokecolor{textcolor}%
\pgfsetfillcolor{textcolor}%
\pgftext[x=2.159041in,y=0.584028in,left,base]{\color{textcolor}\sffamily\fontsize{8.800000}{10.560000}\selectfont NMT-1}%
\end{pgfscope}%
\begin{pgfscope}%
\pgfsetbuttcap%
\pgfsetroundjoin%
\pgfsetlinewidth{1.204500pt}%
\definecolor{currentstroke}{rgb}{0.121569,0.470588,0.705882}%
\pgfsetstrokecolor{currentstroke}%
\pgfsetdash{{1.200000pt}{1.980000pt}}{0.000000pt}%
\pgfpathmoveto{\pgfqpoint{1.816819in}{0.454583in}}%
\pgfpathlineto{\pgfqpoint{2.061263in}{0.454583in}}%
\pgfusepath{stroke}%
\end{pgfscope}%
\begin{pgfscope}%
\definecolor{textcolor}{rgb}{0.150000,0.150000,0.150000}%
\pgfsetstrokecolor{textcolor}%
\pgfsetfillcolor{textcolor}%
\pgftext[x=2.159041in,y=0.411806in,left,base]{\color{textcolor}\sffamily\fontsize{8.800000}{10.560000}\selectfont NMT-2}%
\end{pgfscope}%
\end{pgfpicture}%
\makeatother%
\endgroup%

%% file: curves_hungarian-hgds.pgf
\begingroup%
\makeatletter%
\begin{pgfpicture}%
\pgfpathrectangle{\pgfpointorigin}{\pgfqpoint{3.078809in}{2.168389in}}%
\pgfusepath{use as bounding box, clip}%
\begin{pgfscope}%
\pgfsetbuttcap%
\pgfsetmiterjoin%
\definecolor{currentfill}{rgb}{1.000000,1.000000,1.000000}%
\pgfsetfillcolor{currentfill}%
\pgfsetlinewidth{0.000000pt}%
\definecolor{currentstroke}{rgb}{1.000000,1.000000,1.000000}%
\pgfsetstrokecolor{currentstroke}%
\pgfsetdash{}{0pt}%
\pgfpathmoveto{\pgfqpoint{0.000000in}{0.000000in}}%
\pgfpathlineto{\pgfqpoint{3.078809in}{0.000000in}}%
\pgfpathlineto{\pgfqpoint{3.078809in}{2.168389in}}%
\pgfpathlineto{\pgfqpoint{0.000000in}{2.168389in}}%
\pgfpathclose%
\pgfusepath{fill}%
\end{pgfscope}%
\begin{pgfscope}%
\pgfsetbuttcap%
\pgfsetmiterjoin%
\definecolor{currentfill}{rgb}{1.000000,1.000000,1.000000}%
\pgfsetfillcolor{currentfill}%
\pgfsetlinewidth{0.000000pt}%
\definecolor{currentstroke}{rgb}{0.000000,0.000000,0.000000}%
\pgfsetstrokecolor{currentstroke}%
\pgfsetstrokeopacity{0.000000}%
\pgfsetdash{}{0pt}%
\pgfpathmoveto{\pgfqpoint{0.450809in}{0.326389in}}%
\pgfpathlineto{\pgfqpoint{2.930809in}{0.326389in}}%
\pgfpathlineto{\pgfqpoint{2.930809in}{2.020389in}}%
\pgfpathlineto{\pgfqpoint{0.450809in}{2.020389in}}%
\pgfpathclose%
\pgfusepath{fill}%
\end{pgfscope}%
\begin{pgfscope}%
\pgfsetbuttcap%
\pgfsetroundjoin%
\definecolor{currentfill}{rgb}{0.150000,0.150000,0.150000}%
\pgfsetfillcolor{currentfill}%
\pgfsetlinewidth{1.003750pt}%
\definecolor{currentstroke}{rgb}{0.150000,0.150000,0.150000}%
\pgfsetstrokecolor{currentstroke}%
\pgfsetdash{}{0pt}%
\pgfsys@defobject{currentmarker}{\pgfqpoint{0.000000in}{-0.066667in}}{\pgfqpoint{0.000000in}{0.000000in}}{%
\pgfpathmoveto{\pgfqpoint{0.000000in}{0.000000in}}%
\pgfpathlineto{\pgfqpoint{0.000000in}{-0.066667in}}%
\pgfusepath{stroke,fill}%
}%
\begin{pgfscope}%
\pgfsys@transformshift{0.563536in}{0.326389in}%
\pgfsys@useobject{currentmarker}{}%
\end{pgfscope}%
\end{pgfscope}%
\begin{pgfscope}%
\definecolor{textcolor}{rgb}{0.150000,0.150000,0.150000}%
\pgfsetstrokecolor{textcolor}%
\pgfsetfillcolor{textcolor}%
\pgftext[x=0.563536in,y=0.211111in,,top]{\color{textcolor}\sffamily\fontsize{8.800000}{10.560000}\selectfont 100}%
\end{pgfscope}%
\begin{pgfscope}%
\pgfsetbuttcap%
\pgfsetroundjoin%
\definecolor{currentfill}{rgb}{0.150000,0.150000,0.150000}%
\pgfsetfillcolor{currentfill}%
\pgfsetlinewidth{1.003750pt}%
\definecolor{currentstroke}{rgb}{0.150000,0.150000,0.150000}%
\pgfsetstrokecolor{currentstroke}%
\pgfsetdash{}{0pt}%
\pgfsys@defobject{currentmarker}{\pgfqpoint{0.000000in}{-0.066667in}}{\pgfqpoint{0.000000in}{0.000000in}}{%
\pgfpathmoveto{\pgfqpoint{0.000000in}{0.000000in}}%
\pgfpathlineto{\pgfqpoint{0.000000in}{-0.066667in}}%
\pgfusepath{stroke,fill}%
}%
\begin{pgfscope}%
\pgfsys@transformshift{0.895949in}{0.326389in}%
\pgfsys@useobject{currentmarker}{}%
\end{pgfscope}%
\end{pgfscope}%
\begin{pgfscope}%
\definecolor{textcolor}{rgb}{0.150000,0.150000,0.150000}%
\pgfsetstrokecolor{textcolor}%
\pgfsetfillcolor{textcolor}%
\pgftext[x=0.895949in,y=0.211111in,,top]{\color{textcolor}\sffamily\fontsize{8.800000}{10.560000}\selectfont 250}%
\end{pgfscope}%
\begin{pgfscope}%
\pgfsetbuttcap%
\pgfsetroundjoin%
\definecolor{currentfill}{rgb}{0.150000,0.150000,0.150000}%
\pgfsetfillcolor{currentfill}%
\pgfsetlinewidth{1.003750pt}%
\definecolor{currentstroke}{rgb}{0.150000,0.150000,0.150000}%
\pgfsetstrokecolor{currentstroke}%
\pgfsetdash{}{0pt}%
\pgfsys@defobject{currentmarker}{\pgfqpoint{0.000000in}{-0.066667in}}{\pgfqpoint{0.000000in}{0.000000in}}{%
\pgfpathmoveto{\pgfqpoint{0.000000in}{0.000000in}}%
\pgfpathlineto{\pgfqpoint{0.000000in}{-0.066667in}}%
\pgfusepath{stroke,fill}%
}%
\begin{pgfscope}%
\pgfsys@transformshift{1.147410in}{0.326389in}%
\pgfsys@useobject{currentmarker}{}%
\end{pgfscope}%
\end{pgfscope}%
\begin{pgfscope}%
\definecolor{textcolor}{rgb}{0.150000,0.150000,0.150000}%
\pgfsetstrokecolor{textcolor}%
\pgfsetfillcolor{textcolor}%
\pgftext[x=1.147410in,y=0.211111in,,top]{\color{textcolor}\sffamily\fontsize{8.800000}{10.560000}\selectfont 500}%
\end{pgfscope}%
\begin{pgfscope}%
\pgfsetbuttcap%
\pgfsetroundjoin%
\definecolor{currentfill}{rgb}{0.150000,0.150000,0.150000}%
\pgfsetfillcolor{currentfill}%
\pgfsetlinewidth{1.003750pt}%
\definecolor{currentstroke}{rgb}{0.150000,0.150000,0.150000}%
\pgfsetstrokecolor{currentstroke}%
\pgfsetdash{}{0pt}%
\pgfsys@defobject{currentmarker}{\pgfqpoint{0.000000in}{-0.066667in}}{\pgfqpoint{0.000000in}{0.000000in}}{%
\pgfpathmoveto{\pgfqpoint{0.000000in}{0.000000in}}%
\pgfpathlineto{\pgfqpoint{0.000000in}{-0.066667in}}%
\pgfusepath{stroke,fill}%
}%
\begin{pgfscope}%
\pgfsys@transformshift{1.398871in}{0.326389in}%
\pgfsys@useobject{currentmarker}{}%
\end{pgfscope}%
\end{pgfscope}%
\begin{pgfscope}%
\definecolor{textcolor}{rgb}{0.150000,0.150000,0.150000}%
\pgfsetstrokecolor{textcolor}%
\pgfsetfillcolor{textcolor}%
\pgftext[x=1.398871in,y=0.211111in,,top]{\color{textcolor}\sffamily\fontsize{8.800000}{10.560000}\selectfont 1k}%
\end{pgfscope}%
\begin{pgfscope}%
\pgfsetbuttcap%
\pgfsetroundjoin%
\definecolor{currentfill}{rgb}{0.150000,0.150000,0.150000}%
\pgfsetfillcolor{currentfill}%
\pgfsetlinewidth{1.003750pt}%
\definecolor{currentstroke}{rgb}{0.150000,0.150000,0.150000}%
\pgfsetstrokecolor{currentstroke}%
\pgfsetdash{}{0pt}%
\pgfsys@defobject{currentmarker}{\pgfqpoint{0.000000in}{-0.066667in}}{\pgfqpoint{0.000000in}{0.000000in}}{%
\pgfpathmoveto{\pgfqpoint{0.000000in}{0.000000in}}%
\pgfpathlineto{\pgfqpoint{0.000000in}{-0.066667in}}%
\pgfusepath{stroke,fill}%
}%
\begin{pgfscope}%
\pgfsys@transformshift{1.731285in}{0.326389in}%
\pgfsys@useobject{currentmarker}{}%
\end{pgfscope}%
\end{pgfscope}%
\begin{pgfscope}%
\definecolor{textcolor}{rgb}{0.150000,0.150000,0.150000}%
\pgfsetstrokecolor{textcolor}%
\pgfsetfillcolor{textcolor}%
\pgftext[x=1.731285in,y=0.211111in,,top]{\color{textcolor}\sffamily\fontsize{8.800000}{10.560000}\selectfont 2.5k}%
\end{pgfscope}%
\begin{pgfscope}%
\pgfsetbuttcap%
\pgfsetroundjoin%
\definecolor{currentfill}{rgb}{0.150000,0.150000,0.150000}%
\pgfsetfillcolor{currentfill}%
\pgfsetlinewidth{1.003750pt}%
\definecolor{currentstroke}{rgb}{0.150000,0.150000,0.150000}%
\pgfsetstrokecolor{currentstroke}%
\pgfsetdash{}{0pt}%
\pgfsys@defobject{currentmarker}{\pgfqpoint{0.000000in}{-0.066667in}}{\pgfqpoint{0.000000in}{0.000000in}}{%
\pgfpathmoveto{\pgfqpoint{0.000000in}{0.000000in}}%
\pgfpathlineto{\pgfqpoint{0.000000in}{-0.066667in}}%
\pgfusepath{stroke,fill}%
}%
\begin{pgfscope}%
\pgfsys@transformshift{1.982746in}{0.326389in}%
\pgfsys@useobject{currentmarker}{}%
\end{pgfscope}%
\end{pgfscope}%
\begin{pgfscope}%
\definecolor{textcolor}{rgb}{0.150000,0.150000,0.150000}%
\pgfsetstrokecolor{textcolor}%
\pgfsetfillcolor{textcolor}%
\pgftext[x=1.982746in,y=0.211111in,,top]{\color{textcolor}\sffamily\fontsize{8.800000}{10.560000}\selectfont 5k}%
\end{pgfscope}%
\begin{pgfscope}%
\pgfsetbuttcap%
\pgfsetroundjoin%
\definecolor{currentfill}{rgb}{0.150000,0.150000,0.150000}%
\pgfsetfillcolor{currentfill}%
\pgfsetlinewidth{1.003750pt}%
\definecolor{currentstroke}{rgb}{0.150000,0.150000,0.150000}%
\pgfsetstrokecolor{currentstroke}%
\pgfsetdash{}{0pt}%
\pgfsys@defobject{currentmarker}{\pgfqpoint{0.000000in}{-0.066667in}}{\pgfqpoint{0.000000in}{0.000000in}}{%
\pgfpathmoveto{\pgfqpoint{0.000000in}{0.000000in}}%
\pgfpathlineto{\pgfqpoint{0.000000in}{-0.066667in}}%
\pgfusepath{stroke,fill}%
}%
\begin{pgfscope}%
\pgfsys@transformshift{2.234207in}{0.326389in}%
\pgfsys@useobject{currentmarker}{}%
\end{pgfscope}%
\end{pgfscope}%
\begin{pgfscope}%
\definecolor{textcolor}{rgb}{0.150000,0.150000,0.150000}%
\pgfsetstrokecolor{textcolor}%
\pgfsetfillcolor{textcolor}%
\pgftext[x=2.234207in,y=0.211111in,,top]{\color{textcolor}\sffamily\fontsize{8.800000}{10.560000}\selectfont 10k}%
\end{pgfscope}%
\begin{pgfscope}%
\pgfsetbuttcap%
\pgfsetroundjoin%
\definecolor{currentfill}{rgb}{0.150000,0.150000,0.150000}%
\pgfsetfillcolor{currentfill}%
\pgfsetlinewidth{1.003750pt}%
\definecolor{currentstroke}{rgb}{0.150000,0.150000,0.150000}%
\pgfsetstrokecolor{currentstroke}%
\pgfsetdash{}{0pt}%
\pgfsys@defobject{currentmarker}{\pgfqpoint{0.000000in}{-0.066667in}}{\pgfqpoint{0.000000in}{0.000000in}}{%
\pgfpathmoveto{\pgfqpoint{0.000000in}{0.000000in}}%
\pgfpathlineto{\pgfqpoint{0.000000in}{-0.066667in}}%
\pgfusepath{stroke,fill}%
}%
\begin{pgfscope}%
\pgfsys@transformshift{2.566620in}{0.326389in}%
\pgfsys@useobject{currentmarker}{}%
\end{pgfscope}%
\end{pgfscope}%
\begin{pgfscope}%
\definecolor{textcolor}{rgb}{0.150000,0.150000,0.150000}%
\pgfsetstrokecolor{textcolor}%
\pgfsetfillcolor{textcolor}%
\pgftext[x=2.566620in,y=0.211111in,,top]{\color{textcolor}\sffamily\fontsize{8.800000}{10.560000}\selectfont 25k}%
\end{pgfscope}%
\begin{pgfscope}%
\pgfsetbuttcap%
\pgfsetroundjoin%
\definecolor{currentfill}{rgb}{0.150000,0.150000,0.150000}%
\pgfsetfillcolor{currentfill}%
\pgfsetlinewidth{1.003750pt}%
\definecolor{currentstroke}{rgb}{0.150000,0.150000,0.150000}%
\pgfsetstrokecolor{currentstroke}%
\pgfsetdash{}{0pt}%
\pgfsys@defobject{currentmarker}{\pgfqpoint{0.000000in}{-0.066667in}}{\pgfqpoint{0.000000in}{0.000000in}}{%
\pgfpathmoveto{\pgfqpoint{0.000000in}{0.000000in}}%
\pgfpathlineto{\pgfqpoint{0.000000in}{-0.066667in}}%
\pgfusepath{stroke,fill}%
}%
\begin{pgfscope}%
\pgfsys@transformshift{2.818081in}{0.326389in}%
\pgfsys@useobject{currentmarker}{}%
\end{pgfscope}%
\end{pgfscope}%
\begin{pgfscope}%
\definecolor{textcolor}{rgb}{0.150000,0.150000,0.150000}%
\pgfsetstrokecolor{textcolor}%
\pgfsetfillcolor{textcolor}%
\pgftext[x=2.818081in,y=0.211111in,,top]{\color{textcolor}\sffamily\fontsize{8.800000}{10.560000}\selectfont 50k}%
\end{pgfscope}%
\begin{pgfscope}%
\pgfsetbuttcap%
\pgfsetroundjoin%
\definecolor{currentfill}{rgb}{0.150000,0.150000,0.150000}%
\pgfsetfillcolor{currentfill}%
\pgfsetlinewidth{1.003750pt}%
\definecolor{currentstroke}{rgb}{0.150000,0.150000,0.150000}%
\pgfsetstrokecolor{currentstroke}%
\pgfsetdash{}{0pt}%
\pgfsys@defobject{currentmarker}{\pgfqpoint{-0.066667in}{0.000000in}}{\pgfqpoint{0.000000in}{0.000000in}}{%
\pgfpathmoveto{\pgfqpoint{0.000000in}{0.000000in}}%
\pgfpathlineto{\pgfqpoint{-0.066667in}{0.000000in}}%
\pgfusepath{stroke,fill}%
}%
\begin{pgfscope}%
\pgfsys@transformshift{0.450809in}{0.549365in}%
\pgfsys@useobject{currentmarker}{}%
\end{pgfscope}%
\end{pgfscope}%
\begin{pgfscope}%
\definecolor{textcolor}{rgb}{0.150000,0.150000,0.150000}%
\pgfsetstrokecolor{textcolor}%
\pgfsetfillcolor{textcolor}%
\pgftext[x=0.100000in,y=0.505962in,left,base]{\color{textcolor}\sffamily\fontsize{8.800000}{10.560000}\selectfont 20\%}%
\end{pgfscope}%
\begin{pgfscope}%
\pgfsetbuttcap%
\pgfsetroundjoin%
\definecolor{currentfill}{rgb}{0.150000,0.150000,0.150000}%
\pgfsetfillcolor{currentfill}%
\pgfsetlinewidth{1.003750pt}%
\definecolor{currentstroke}{rgb}{0.150000,0.150000,0.150000}%
\pgfsetstrokecolor{currentstroke}%
\pgfsetdash{}{0pt}%
\pgfsys@defobject{currentmarker}{\pgfqpoint{-0.066667in}{0.000000in}}{\pgfqpoint{0.000000in}{0.000000in}}{%
\pgfpathmoveto{\pgfqpoint{0.000000in}{0.000000in}}%
\pgfpathlineto{\pgfqpoint{-0.066667in}{0.000000in}}%
\pgfusepath{stroke,fill}%
}%
\begin{pgfscope}%
\pgfsys@transformshift{0.450809in}{0.969657in}%
\pgfsys@useobject{currentmarker}{}%
\end{pgfscope}%
\end{pgfscope}%
\begin{pgfscope}%
\definecolor{textcolor}{rgb}{0.150000,0.150000,0.150000}%
\pgfsetstrokecolor{textcolor}%
\pgfsetfillcolor{textcolor}%
\pgftext[x=0.100000in,y=0.926254in,left,base]{\color{textcolor}\sffamily\fontsize{8.800000}{10.560000}\selectfont 40\%}%
\end{pgfscope}%
\begin{pgfscope}%
\pgfsetbuttcap%
\pgfsetroundjoin%
\definecolor{currentfill}{rgb}{0.150000,0.150000,0.150000}%
\pgfsetfillcolor{currentfill}%
\pgfsetlinewidth{1.003750pt}%
\definecolor{currentstroke}{rgb}{0.150000,0.150000,0.150000}%
\pgfsetstrokecolor{currentstroke}%
\pgfsetdash{}{0pt}%
\pgfsys@defobject{currentmarker}{\pgfqpoint{-0.066667in}{0.000000in}}{\pgfqpoint{0.000000in}{0.000000in}}{%
\pgfpathmoveto{\pgfqpoint{0.000000in}{0.000000in}}%
\pgfpathlineto{\pgfqpoint{-0.066667in}{0.000000in}}%
\pgfusepath{stroke,fill}%
}%
\begin{pgfscope}%
\pgfsys@transformshift{0.450809in}{1.389950in}%
\pgfsys@useobject{currentmarker}{}%
\end{pgfscope}%
\end{pgfscope}%
\begin{pgfscope}%
\definecolor{textcolor}{rgb}{0.150000,0.150000,0.150000}%
\pgfsetstrokecolor{textcolor}%
\pgfsetfillcolor{textcolor}%
\pgftext[x=0.100000in,y=1.346547in,left,base]{\color{textcolor}\sffamily\fontsize{8.800000}{10.560000}\selectfont 60\%}%
\end{pgfscope}%
\begin{pgfscope}%
\pgfsetbuttcap%
\pgfsetroundjoin%
\definecolor{currentfill}{rgb}{0.150000,0.150000,0.150000}%
\pgfsetfillcolor{currentfill}%
\pgfsetlinewidth{1.003750pt}%
\definecolor{currentstroke}{rgb}{0.150000,0.150000,0.150000}%
\pgfsetstrokecolor{currentstroke}%
\pgfsetdash{}{0pt}%
\pgfsys@defobject{currentmarker}{\pgfqpoint{-0.066667in}{0.000000in}}{\pgfqpoint{0.000000in}{0.000000in}}{%
\pgfpathmoveto{\pgfqpoint{0.000000in}{0.000000in}}%
\pgfpathlineto{\pgfqpoint{-0.066667in}{0.000000in}}%
\pgfusepath{stroke,fill}%
}%
\begin{pgfscope}%
\pgfsys@transformshift{0.450809in}{1.810243in}%
\pgfsys@useobject{currentmarker}{}%
\end{pgfscope}%
\end{pgfscope}%
\begin{pgfscope}%
\definecolor{textcolor}{rgb}{0.150000,0.150000,0.150000}%
\pgfsetstrokecolor{textcolor}%
\pgfsetfillcolor{textcolor}%
\pgftext[x=0.100000in,y=1.766840in,left,base]{\color{textcolor}\sffamily\fontsize{8.800000}{10.560000}\selectfont 80\%}%
\end{pgfscope}%
\begin{pgfscope}%
\pgfsetbuttcap%
\pgfsetroundjoin%
\definecolor{currentfill}{rgb}{0.150000,0.150000,0.150000}%
\pgfsetfillcolor{currentfill}%
\pgfsetlinewidth{0.803000pt}%
\definecolor{currentstroke}{rgb}{0.150000,0.150000,0.150000}%
\pgfsetstrokecolor{currentstroke}%
\pgfsetdash{}{0pt}%
\pgfsys@defobject{currentmarker}{\pgfqpoint{-0.044444in}{0.000000in}}{\pgfqpoint{0.000000in}{0.000000in}}{%
\pgfpathmoveto{\pgfqpoint{0.000000in}{0.000000in}}%
\pgfpathlineto{\pgfqpoint{-0.044444in}{0.000000in}}%
\pgfusepath{stroke,fill}%
}%
\begin{pgfscope}%
\pgfsys@transformshift{0.450809in}{0.339218in}%
\pgfsys@useobject{currentmarker}{}%
\end{pgfscope}%
\end{pgfscope}%
\begin{pgfscope}%
\pgfsetbuttcap%
\pgfsetroundjoin%
\definecolor{currentfill}{rgb}{0.150000,0.150000,0.150000}%
\pgfsetfillcolor{currentfill}%
\pgfsetlinewidth{0.803000pt}%
\definecolor{currentstroke}{rgb}{0.150000,0.150000,0.150000}%
\pgfsetstrokecolor{currentstroke}%
\pgfsetdash{}{0pt}%
\pgfsys@defobject{currentmarker}{\pgfqpoint{-0.044444in}{0.000000in}}{\pgfqpoint{0.000000in}{0.000000in}}{%
\pgfpathmoveto{\pgfqpoint{0.000000in}{0.000000in}}%
\pgfpathlineto{\pgfqpoint{-0.044444in}{0.000000in}}%
\pgfusepath{stroke,fill}%
}%
\begin{pgfscope}%
\pgfsys@transformshift{0.450809in}{0.759511in}%
\pgfsys@useobject{currentmarker}{}%
\end{pgfscope}%
\end{pgfscope}%
\begin{pgfscope}%
\pgfsetbuttcap%
\pgfsetroundjoin%
\definecolor{currentfill}{rgb}{0.150000,0.150000,0.150000}%
\pgfsetfillcolor{currentfill}%
\pgfsetlinewidth{0.803000pt}%
\definecolor{currentstroke}{rgb}{0.150000,0.150000,0.150000}%
\pgfsetstrokecolor{currentstroke}%
\pgfsetdash{}{0pt}%
\pgfsys@defobject{currentmarker}{\pgfqpoint{-0.044444in}{0.000000in}}{\pgfqpoint{0.000000in}{0.000000in}}{%
\pgfpathmoveto{\pgfqpoint{0.000000in}{0.000000in}}%
\pgfpathlineto{\pgfqpoint{-0.044444in}{0.000000in}}%
\pgfusepath{stroke,fill}%
}%
\begin{pgfscope}%
\pgfsys@transformshift{0.450809in}{1.179804in}%
\pgfsys@useobject{currentmarker}{}%
\end{pgfscope}%
\end{pgfscope}%
\begin{pgfscope}%
\pgfsetbuttcap%
\pgfsetroundjoin%
\definecolor{currentfill}{rgb}{0.150000,0.150000,0.150000}%
\pgfsetfillcolor{currentfill}%
\pgfsetlinewidth{0.803000pt}%
\definecolor{currentstroke}{rgb}{0.150000,0.150000,0.150000}%
\pgfsetstrokecolor{currentstroke}%
\pgfsetdash{}{0pt}%
\pgfsys@defobject{currentmarker}{\pgfqpoint{-0.044444in}{0.000000in}}{\pgfqpoint{0.000000in}{0.000000in}}{%
\pgfpathmoveto{\pgfqpoint{0.000000in}{0.000000in}}%
\pgfpathlineto{\pgfqpoint{-0.044444in}{0.000000in}}%
\pgfusepath{stroke,fill}%
}%
\begin{pgfscope}%
\pgfsys@transformshift{0.450809in}{1.600096in}%
\pgfsys@useobject{currentmarker}{}%
\end{pgfscope}%
\end{pgfscope}%
\begin{pgfscope}%
\pgfsetbuttcap%
\pgfsetroundjoin%
\definecolor{currentfill}{rgb}{0.150000,0.150000,0.150000}%
\pgfsetfillcolor{currentfill}%
\pgfsetlinewidth{0.803000pt}%
\definecolor{currentstroke}{rgb}{0.150000,0.150000,0.150000}%
\pgfsetstrokecolor{currentstroke}%
\pgfsetdash{}{0pt}%
\pgfsys@defobject{currentmarker}{\pgfqpoint{-0.044444in}{0.000000in}}{\pgfqpoint{0.000000in}{0.000000in}}{%
\pgfpathmoveto{\pgfqpoint{0.000000in}{0.000000in}}%
\pgfpathlineto{\pgfqpoint{-0.044444in}{0.000000in}}%
\pgfusepath{stroke,fill}%
}%
\begin{pgfscope}%
\pgfsys@transformshift{0.450809in}{2.020389in}%
\pgfsys@useobject{currentmarker}{}%
\end{pgfscope}%
\end{pgfscope}%
\begin{pgfscope}%
\pgfpathrectangle{\pgfqpoint{0.450809in}{0.326389in}}{\pgfqpoint{2.480000in}{1.694000in}}%
\pgfusepath{clip}%
\pgfsetbuttcap%
\pgfsetroundjoin%
\pgfsetlinewidth{1.204500pt}%
\definecolor{currentstroke}{rgb}{1.000000,0.498039,0.000000}%
\pgfsetstrokecolor{currentstroke}%
\pgfsetdash{{7.680000pt}{1.920000pt}{1.200000pt}{1.920000pt}}{0.000000pt}%
\pgfpathmoveto{\pgfqpoint{0.563536in}{1.155490in}}%
\pgfpathlineto{\pgfqpoint{0.895949in}{1.209375in}}%
\pgfpathlineto{\pgfqpoint{1.147410in}{1.263122in}}%
\pgfpathlineto{\pgfqpoint{1.398871in}{1.300694in}}%
\pgfpathlineto{\pgfqpoint{1.731285in}{1.341435in}}%
\pgfpathlineto{\pgfqpoint{1.982746in}{1.378642in}}%
\pgfpathlineto{\pgfqpoint{2.234207in}{1.368504in}}%
\pgfpathlineto{\pgfqpoint{2.566620in}{1.511306in}}%
\pgfpathlineto{\pgfqpoint{2.818081in}{1.673503in}}%
\pgfusepath{stroke}%
\end{pgfscope}%
\begin{pgfscope}%
\pgfpathrectangle{\pgfqpoint{0.450809in}{0.326389in}}{\pgfqpoint{2.480000in}{1.694000in}}%
\pgfusepath{clip}%
\pgfsetroundcap%
\pgfsetroundjoin%
\pgfsetlinewidth{1.204500pt}%
\definecolor{currentstroke}{rgb}{0.890196,0.101961,0.109804}%
\pgfsetstrokecolor{currentstroke}%
\pgfsetdash{}{0pt}%
\pgfpathmoveto{\pgfqpoint{0.563536in}{0.908514in}}%
\pgfpathlineto{\pgfqpoint{0.895949in}{1.065353in}}%
\pgfpathlineto{\pgfqpoint{1.147410in}{1.184747in}}%
\pgfpathlineto{\pgfqpoint{1.398871in}{1.242896in}}%
\pgfpathlineto{\pgfqpoint{1.731285in}{1.314165in}}%
\pgfpathlineto{\pgfqpoint{1.982746in}{1.362076in}}%
\pgfpathlineto{\pgfqpoint{2.234207in}{1.385271in}}%
\pgfpathlineto{\pgfqpoint{2.566620in}{1.596741in}}%
\pgfpathlineto{\pgfqpoint{2.818081in}{1.767023in}}%
\pgfusepath{stroke}%
\end{pgfscope}%
\begin{pgfscope}%
\pgfpathrectangle{\pgfqpoint{0.450809in}{0.326389in}}{\pgfqpoint{2.480000in}{1.694000in}}%
\pgfusepath{clip}%
\pgfsetbuttcap%
\pgfsetroundjoin%
\pgfsetlinewidth{1.204500pt}%
\definecolor{currentstroke}{rgb}{0.200000,0.627451,0.172549}%
\pgfsetstrokecolor{currentstroke}%
\pgfsetdash{{4.440000pt}{1.920000pt}}{0.000000pt}%
\pgfpathmoveto{\pgfqpoint{0.563536in}{0.394991in}}%
\pgfpathlineto{\pgfqpoint{0.895949in}{0.656997in}}%
\pgfpathlineto{\pgfqpoint{1.147410in}{0.828767in}}%
\pgfpathlineto{\pgfqpoint{1.398871in}{0.980500in}}%
\pgfpathlineto{\pgfqpoint{1.731285in}{1.120497in}}%
\pgfpathlineto{\pgfqpoint{1.982746in}{1.214734in}}%
\pgfpathlineto{\pgfqpoint{2.234207in}{1.290770in}}%
\pgfpathlineto{\pgfqpoint{2.566620in}{1.515485in}}%
\pgfpathlineto{\pgfqpoint{2.818081in}{1.698283in}}%
\pgfusepath{stroke}%
\end{pgfscope}%
\begin{pgfscope}%
\pgfpathrectangle{\pgfqpoint{0.450809in}{0.326389in}}{\pgfqpoint{2.480000in}{1.694000in}}%
\pgfusepath{clip}%
\pgfsetbuttcap%
\pgfsetroundjoin%
\pgfsetlinewidth{1.204500pt}%
\definecolor{currentstroke}{rgb}{0.121569,0.470588,0.705882}%
\pgfsetstrokecolor{currentstroke}%
\pgfsetdash{{1.200000pt}{1.980000pt}}{0.000000pt}%
\pgfpathmoveto{\pgfqpoint{0.563536in}{0.572936in}}%
\pgfpathlineto{\pgfqpoint{0.895949in}{0.782944in}}%
\pgfpathlineto{\pgfqpoint{1.147410in}{0.896061in}}%
\pgfpathlineto{\pgfqpoint{1.398871in}{0.991707in}}%
\pgfpathlineto{\pgfqpoint{1.731285in}{1.105038in}}%
\pgfpathlineto{\pgfqpoint{1.982746in}{1.189401in}}%
\pgfpathlineto{\pgfqpoint{2.234207in}{1.284304in}}%
\pgfpathlineto{\pgfqpoint{2.566620in}{1.485185in}}%
\pgfpathlineto{\pgfqpoint{2.818081in}{1.749256in}}%
\pgfusepath{stroke}%
\end{pgfscope}%
\begin{pgfscope}%
\pgfsetrectcap%
\pgfsetmiterjoin%
\pgfsetlinewidth{1.003750pt}%
\definecolor{currentstroke}{rgb}{0.150000,0.150000,0.150000}%
\pgfsetstrokecolor{currentstroke}%
\pgfsetdash{}{0pt}%
\pgfpathmoveto{\pgfqpoint{0.450809in}{0.326389in}}%
\pgfpathlineto{\pgfqpoint{0.450809in}{2.020389in}}%
\pgfusepath{stroke}%
\end{pgfscope}%
\begin{pgfscope}%
\pgfsetrectcap%
\pgfsetmiterjoin%
\pgfsetlinewidth{1.003750pt}%
\definecolor{currentstroke}{rgb}{0.150000,0.150000,0.150000}%
\pgfsetstrokecolor{currentstroke}%
\pgfsetdash{}{0pt}%
\pgfpathmoveto{\pgfqpoint{2.930809in}{0.326389in}}%
\pgfpathlineto{\pgfqpoint{2.930809in}{2.020389in}}%
\pgfusepath{stroke}%
\end{pgfscope}%
\begin{pgfscope}%
\pgfsetrectcap%
\pgfsetmiterjoin%
\pgfsetlinewidth{1.003750pt}%
\definecolor{currentstroke}{rgb}{0.150000,0.150000,0.150000}%
\pgfsetstrokecolor{currentstroke}%
\pgfsetdash{}{0pt}%
\pgfpathmoveto{\pgfqpoint{0.450809in}{0.326389in}}%
\pgfpathlineto{\pgfqpoint{2.930809in}{0.326389in}}%
\pgfusepath{stroke}%
\end{pgfscope}%
\begin{pgfscope}%
\pgfsetrectcap%
\pgfsetmiterjoin%
\pgfsetlinewidth{1.003750pt}%
\definecolor{currentstroke}{rgb}{0.150000,0.150000,0.150000}%
\pgfsetstrokecolor{currentstroke}%
\pgfsetdash{}{0pt}%
\pgfpathmoveto{\pgfqpoint{0.450809in}{2.020389in}}%
\pgfpathlineto{\pgfqpoint{2.930809in}{2.020389in}}%
\pgfusepath{stroke}%
\end{pgfscope}%
\begin{pgfscope}%
\pgfsetbuttcap%
\pgfsetroundjoin%
\pgfsetlinewidth{1.204500pt}%
\definecolor{currentstroke}{rgb}{1.000000,0.498039,0.000000}%
\pgfsetstrokecolor{currentstroke}%
\pgfsetdash{{7.680000pt}{1.920000pt}{1.200000pt}{1.920000pt}}{0.000000pt}%
\pgfpathmoveto{\pgfqpoint{1.816819in}{0.971250in}}%
\pgfpathlineto{\pgfqpoint{2.061263in}{0.971250in}}%
\pgfusepath{stroke}%
\end{pgfscope}%
\begin{pgfscope}%
\definecolor{textcolor}{rgb}{0.150000,0.150000,0.150000}%
\pgfsetstrokecolor{textcolor}%
\pgfsetfillcolor{textcolor}%
\pgftext[x=2.159041in,y=0.928472in,left,base]{\color{textcolor}\sffamily\fontsize{8.800000}{10.560000}\selectfont Norma}%
\end{pgfscope}%
\begin{pgfscope}%
\pgfsetroundcap%
\pgfsetroundjoin%
\pgfsetlinewidth{1.204500pt}%
\definecolor{currentstroke}{rgb}{0.890196,0.101961,0.109804}%
\pgfsetstrokecolor{currentstroke}%
\pgfsetdash{}{0pt}%
\pgfpathmoveto{\pgfqpoint{1.816819in}{0.799028in}}%
\pgfpathlineto{\pgfqpoint{2.061263in}{0.799028in}}%
\pgfusepath{stroke}%
\end{pgfscope}%
\begin{pgfscope}%
\definecolor{textcolor}{rgb}{0.150000,0.150000,0.150000}%
\pgfsetstrokecolor{textcolor}%
\pgfsetfillcolor{textcolor}%
\pgftext[x=2.159041in,y=0.756250in,left,base]{\color{textcolor}\sffamily\fontsize{8.800000}{10.560000}\selectfont cSMTiser\textsubscript{+LM}}%
\end{pgfscope}%
\begin{pgfscope}%
\pgfsetbuttcap%
\pgfsetroundjoin%
\pgfsetlinewidth{1.204500pt}%
\definecolor{currentstroke}{rgb}{0.200000,0.627451,0.172549}%
\pgfsetstrokecolor{currentstroke}%
\pgfsetdash{{4.440000pt}{1.920000pt}}{0.000000pt}%
\pgfpathmoveto{\pgfqpoint{1.816819in}{0.626806in}}%
\pgfpathlineto{\pgfqpoint{2.061263in}{0.626806in}}%
\pgfusepath{stroke}%
\end{pgfscope}%
\begin{pgfscope}%
\definecolor{textcolor}{rgb}{0.150000,0.150000,0.150000}%
\pgfsetstrokecolor{textcolor}%
\pgfsetfillcolor{textcolor}%
\pgftext[x=2.159041in,y=0.584028in,left,base]{\color{textcolor}\sffamily\fontsize{8.800000}{10.560000}\selectfont NMT-1}%
\end{pgfscope}%
\begin{pgfscope}%
\pgfsetbuttcap%
\pgfsetroundjoin%
\pgfsetlinewidth{1.204500pt}%
\definecolor{currentstroke}{rgb}{0.121569,0.470588,0.705882}%
\pgfsetstrokecolor{currentstroke}%
\pgfsetdash{{1.200000pt}{1.980000pt}}{0.000000pt}%
\pgfpathmoveto{\pgfqpoint{1.816819in}{0.454583in}}%
\pgfpathlineto{\pgfqpoint{2.061263in}{0.454583in}}%
\pgfusepath{stroke}%
\end{pgfscope}%
\begin{pgfscope}%
\definecolor{textcolor}{rgb}{0.150000,0.150000,0.150000}%
\pgfsetstrokecolor{textcolor}%
\pgfsetfillcolor{textcolor}%
\pgftext[x=2.159041in,y=0.411806in,left,base]{\color{textcolor}\sffamily\fontsize{8.800000}{10.560000}\selectfont NMT-2}%
\end{pgfscope}%
\end{pgfpicture}%
\makeatother%
\endgroup%

%% file: tab_eval_oov.tex
\begin{table*}[t!]

\begin{subtable}[t]{1\textwidth}
  \centering
\resizebox{\textwidth}{!}{%
\setlength{\tabcolsep}{5pt}
\begin{tabular}{lWWWWWWWWWW}
\toprule
\textbf{Method} & \multicolumn{10}{c}{\textbf{Dataset}} \\
\cmidrule(lr){2-11}
 & \multicolumn{1}{c}{DE\textsubscript{A}} & \multicolumn{1}{c}{DE\textsubscript{R}} & \multicolumn{1}{c}{EN} & \multicolumn{1}{c}{ES} & \multicolumn{1}{c}{HU} & \multicolumn{1}{c}{IS} & \multicolumn{1}{c}{PT} & \multicolumn{1}{c}{SL\textsubscript{B}} & \multicolumn{1}{c}{SL\textsubscript{G}} & \multicolumn{1}{c}{SV}\\
\midrule
Norma, Lookup/Combined & \textbf{92.36} & \textbf{93.66} & \textbf{97.46} & \textbf{96.59} & \textbf{96.81} & \textbf{89.51} & \textbf{97.04} & \textbf{97.15} & \textbf{98.17} & 97.61\\
Norma, Rule-based & 80.34 & 89.26 & 93.17 & 89.98 & 88.42 & 86.21 & 88.77 & 93.61 & 96.10 & 92.10\\
Norma, Distance-based & 60.79 & 78.85 & 86.77 & 85.89 & 67.04 & 70.99 & 78.93 & 77.49 & 94.54 & 84.17\\
\midrule
cSMTiser                    & 92.18 & 93.25 & 97.10 & 96.33 & 96.33 & 89.27 & 96.80 & 96.73 & 98.09 & \textbf{97.66}\\
cSMTiser\textsubscript{+LM} & 90.52 & 93.45 & 97.15 & 96.42 & 96.33 & 88.99 & 96.82 & 96.90 & 98.07 & \textbf{97.66}\\
\midrule
NMT {\small\citep{BollmannPhD}} & 91.91 & 93.29 & 97.19 & 96.37 & 96.18 & 88.67 & 96.72 & 95.92 & 97.56 & 97.01\\
NMT {\small\citep{Tang-etal2018}} & 92.25 & 93.41 & 97.19 & 96.27 & 96.43 & 89.34 & 96.60 & 96.84 & 97.89 & 96.90\\
\bottomrule
\end{tabular}
}
\subcaption{In-vocabulary/seen tokens\label{tab:eval-iv}}
\end{subtable}

\vspace{.5em}
\begin{subtable}[t]{1\textwidth}
  \centering
\resizebox{\textwidth}{!}{%
\setlength{\tabcolsep}{5pt}
\begin{tabular}{lWWWWWWWWWW}
\toprule
Norma, Lookup & 3.91 & 19.92 & 30.42 & 46.72 & 3.28 & 29.40 & 28.25 & 18.07 & 68.07 & 39.85\\
Norma, Rule-based & 40.27 & 46.06 & 62.06 & 72.97 & 47.66 & 63.73 & 57.50 & 54.99 & 64.59 & 63.37 \\
Norma, Distance-based & 41.33 & 43.25 & 48.63 & 67.78 & 47.43 & 61.64 & 57.79 & 44.23 & 49.82 & 50.15 \\
Norma (Combined) & 47.25 & 48.13 & 59.18 & 69.93 & 54.83 & 65.52 & 60.62 & 57.57 & 50.71 & 53.73\\
\midrule
cSMTiser                    & 57.25 & 59.96 & \textbf{71.78} & \textbf{80.22} & 76.59 & 69.70 & 74.69 & \textbf{78.49} & 83.30 & \textbf{70.35} \\
cSMTiser\textsubscript{+LM} & 50.69 & 59.76 & 71.70 & 79.24 & \textbf{76.86} & 69.55 & \textbf{75.91} & 78.40 & \textbf{83.56} & 70.26 \\
\midrule
NMT {\small\citep{BollmannPhD}} & 63.24 & 59.83 & 65.17 & 77.57 & 75.13 & 68.66 & 70.00 & 73.75 & 80.87 & 68.78 \\
NMT {\small\citep{Tang-etal2018}} & \textbf{65.09} & \textbf{60.16} & 67.15 & 78.75 & 76.33 & \textbf{71.04} & 69.90 & 75.04 & 83.46 & 69.66\\
\bottomrule
\end{tabular}
}
\subcaption{Out-of-vocabulary/unseen tokens\label{tab:eval-oov}}
\end{subtable}


\caption{Word accuracy for seen/unseen tokens separately (cf.\ Sec.\,\ref{sec:analysis-oov}); best results for each dataset in \textbf{bold}.}
  \label{tab:eval-iv-oov}

\vspace{1.5em}
 \resizebox{\textwidth}{!}{%

  \centering
\setlength{\tabcolsep}{3pt}
\begin{tabular}{lrrrrrrrrrr}
\toprule
\textbf{Method} & \multicolumn{10}{c}{\textbf{Dataset}} \\
\cmidrule(lr){2-11}
 & \multicolumn{1}{c}{DE\textsubscript{A}} & \multicolumn{1}{c}{DE\textsubscript{R}} & \multicolumn{1}{c}{EN} & \multicolumn{1}{c}{ES} & \multicolumn{1}{c}{HU} & \multicolumn{1}{c}{IS} & \multicolumn{1}{c}{PT} & \multicolumn{1}{c}{SL\textsubscript{B}} & \multicolumn{1}{c}{SL\textsubscript{G}} & \multicolumn{1}{c}{SV}\\
\midrule
\textit{Best without lookup} & \itshape \notsig{89.64} & \itshape \notsig{88.22} & \itshape 95.24 & \itshape 95.02 & \itshape 91.70 & \itshape \notsig{87.31} & \itshape 95.18 & \itshape 93.30 & \itshape \notsig{96.01} & \itshape \textbf{91.13} \\
\midrule
cSMTiser                    & 88.98 & \notsig{88.41} & \textbf{95.54} & \textbf{95.25} & \notsig{92.00} & \notsig{87.31} & \notsig{95.31} & \textbf{93.52} & \notsig{96.06} & \notsig{91.09} \\
cSMTiser\textsubscript{+LM} & 88.35 & \notsig{88.37} & \notsig{95.53} & \notsig{95.17} & \textbf{92.07} & \notsig{87.30} & \textbf{95.39} & \notsig{93.50} & \textbf{96.10} & \notsig{91.07} \\
\midrule
NMT {\small\citep{BollmannPhD}} & 89.56 & \notsig{88.38} & 95.05 & 95.03 & 91.66 & \notsig{87.20} & 94.93 & 92.60 & 95.71 & 90.72\\
NMT {\small\citep{Tang-etal2018}} & \textbf{89.74} & \textbf{88.45} & 95.19 & \notsig{95.13} & \notsig{91.94} & \textbf{87.46} & 94.92 & 92.85 & \notsig{96.08} & \notsig{90.93}  \\
\bottomrule
\end{tabular}
}

\caption{Word accuracy for the ``lookup on \textit{seen} tokens, learned models on \textit{unseen} tokens'' strategy, following \citet{Robertson-Goldwater2018} (cf.\ Sec.\,\ref{sec:analysis-oov}), compared to the best result without this strategy (according to Table~\ref{tab:eval-test}).  Best result for each dataset in \textbf{bold}; results marked with an asterisk~(*) are not significantly different from the best result using McNemar's test at $p < 0.05$.}
  \label{tab:eval-combined}
\end{table*}

%% file: curves_german-anselm.pgf
\begingroup%
\makeatletter%
\begin{pgfpicture}%
\pgfpathrectangle{\pgfpointorigin}{\pgfqpoint{3.078809in}{2.168389in}}%
\pgfusepath{use as bounding box, clip}%
\begin{pgfscope}%
\pgfsetbuttcap%
\pgfsetmiterjoin%
\definecolor{currentfill}{rgb}{1.000000,1.000000,1.000000}%
\pgfsetfillcolor{currentfill}%
\pgfsetlinewidth{0.000000pt}%
\definecolor{currentstroke}{rgb}{1.000000,1.000000,1.000000}%
\pgfsetstrokecolor{currentstroke}%
\pgfsetdash{}{0pt}%
\pgfpathmoveto{\pgfqpoint{0.000000in}{0.000000in}}%
\pgfpathlineto{\pgfqpoint{3.078809in}{0.000000in}}%
\pgfpathlineto{\pgfqpoint{3.078809in}{2.168389in}}%
\pgfpathlineto{\pgfqpoint{0.000000in}{2.168389in}}%
\pgfpathclose%
\pgfusepath{fill}%
\end{pgfscope}%
\begin{pgfscope}%
\pgfsetbuttcap%
\pgfsetmiterjoin%
\definecolor{currentfill}{rgb}{1.000000,1.000000,1.000000}%
\pgfsetfillcolor{currentfill}%
\pgfsetlinewidth{0.000000pt}%
\definecolor{currentstroke}{rgb}{0.000000,0.000000,0.000000}%
\pgfsetstrokecolor{currentstroke}%
\pgfsetstrokeopacity{0.000000}%
\pgfsetdash{}{0pt}%
\pgfpathmoveto{\pgfqpoint{0.450809in}{0.326389in}}%
\pgfpathlineto{\pgfqpoint{2.930809in}{0.326389in}}%
\pgfpathlineto{\pgfqpoint{2.930809in}{2.020389in}}%
\pgfpathlineto{\pgfqpoint{0.450809in}{2.020389in}}%
\pgfpathclose%
\pgfusepath{fill}%
\end{pgfscope}%
\begin{pgfscope}%
\pgfsetbuttcap%
\pgfsetroundjoin%
\definecolor{currentfill}{rgb}{0.150000,0.150000,0.150000}%
\pgfsetfillcolor{currentfill}%
\pgfsetlinewidth{1.003750pt}%
\definecolor{currentstroke}{rgb}{0.150000,0.150000,0.150000}%
\pgfsetstrokecolor{currentstroke}%
\pgfsetdash{}{0pt}%
\pgfsys@defobject{currentmarker}{\pgfqpoint{0.000000in}{-0.066667in}}{\pgfqpoint{0.000000in}{0.000000in}}{%
\pgfpathmoveto{\pgfqpoint{0.000000in}{0.000000in}}%
\pgfpathlineto{\pgfqpoint{0.000000in}{-0.066667in}}%
\pgfusepath{stroke,fill}%
}%
\begin{pgfscope}%
\pgfsys@transformshift{0.563536in}{0.326389in}%
\pgfsys@useobject{currentmarker}{}%
\end{pgfscope}%
\end{pgfscope}%
\begin{pgfscope}%
\definecolor{textcolor}{rgb}{0.150000,0.150000,0.150000}%
\pgfsetstrokecolor{textcolor}%
\pgfsetfillcolor{textcolor}%
\pgftext[x=0.563536in,y=0.211111in,,top]{\color{textcolor}\sffamily\fontsize{8.800000}{10.560000}\selectfont 100}%
\end{pgfscope}%
\begin{pgfscope}%
\pgfsetbuttcap%
\pgfsetroundjoin%
\definecolor{currentfill}{rgb}{0.150000,0.150000,0.150000}%
\pgfsetfillcolor{currentfill}%
\pgfsetlinewidth{1.003750pt}%
\definecolor{currentstroke}{rgb}{0.150000,0.150000,0.150000}%
\pgfsetstrokecolor{currentstroke}%
\pgfsetdash{}{0pt}%
\pgfsys@defobject{currentmarker}{\pgfqpoint{0.000000in}{-0.066667in}}{\pgfqpoint{0.000000in}{0.000000in}}{%
\pgfpathmoveto{\pgfqpoint{0.000000in}{0.000000in}}%
\pgfpathlineto{\pgfqpoint{0.000000in}{-0.066667in}}%
\pgfusepath{stroke,fill}%
}%
\begin{pgfscope}%
\pgfsys@transformshift{0.895949in}{0.326389in}%
\pgfsys@useobject{currentmarker}{}%
\end{pgfscope}%
\end{pgfscope}%
\begin{pgfscope}%
\definecolor{textcolor}{rgb}{0.150000,0.150000,0.150000}%
\pgfsetstrokecolor{textcolor}%
\pgfsetfillcolor{textcolor}%
\pgftext[x=0.895949in,y=0.211111in,,top]{\color{textcolor}\sffamily\fontsize{8.800000}{10.560000}\selectfont 250}%
\end{pgfscope}%
\begin{pgfscope}%
\pgfsetbuttcap%
\pgfsetroundjoin%
\definecolor{currentfill}{rgb}{0.150000,0.150000,0.150000}%
\pgfsetfillcolor{currentfill}%
\pgfsetlinewidth{1.003750pt}%
\definecolor{currentstroke}{rgb}{0.150000,0.150000,0.150000}%
\pgfsetstrokecolor{currentstroke}%
\pgfsetdash{}{0pt}%
\pgfsys@defobject{currentmarker}{\pgfqpoint{0.000000in}{-0.066667in}}{\pgfqpoint{0.000000in}{0.000000in}}{%
\pgfpathmoveto{\pgfqpoint{0.000000in}{0.000000in}}%
\pgfpathlineto{\pgfqpoint{0.000000in}{-0.066667in}}%
\pgfusepath{stroke,fill}%
}%
\begin{pgfscope}%
\pgfsys@transformshift{1.147410in}{0.326389in}%
\pgfsys@useobject{currentmarker}{}%
\end{pgfscope}%
\end{pgfscope}%
\begin{pgfscope}%
\definecolor{textcolor}{rgb}{0.150000,0.150000,0.150000}%
\pgfsetstrokecolor{textcolor}%
\pgfsetfillcolor{textcolor}%
\pgftext[x=1.147410in,y=0.211111in,,top]{\color{textcolor}\sffamily\fontsize{8.800000}{10.560000}\selectfont 500}%
\end{pgfscope}%
\begin{pgfscope}%
\pgfsetbuttcap%
\pgfsetroundjoin%
\definecolor{currentfill}{rgb}{0.150000,0.150000,0.150000}%
\pgfsetfillcolor{currentfill}%
\pgfsetlinewidth{1.003750pt}%
\definecolor{currentstroke}{rgb}{0.150000,0.150000,0.150000}%
\pgfsetstrokecolor{currentstroke}%
\pgfsetdash{}{0pt}%
\pgfsys@defobject{currentmarker}{\pgfqpoint{0.000000in}{-0.066667in}}{\pgfqpoint{0.000000in}{0.000000in}}{%
\pgfpathmoveto{\pgfqpoint{0.000000in}{0.000000in}}%
\pgfpathlineto{\pgfqpoint{0.000000in}{-0.066667in}}%
\pgfusepath{stroke,fill}%
}%
\begin{pgfscope}%
\pgfsys@transformshift{1.398871in}{0.326389in}%
\pgfsys@useobject{currentmarker}{}%
\end{pgfscope}%
\end{pgfscope}%
\begin{pgfscope}%
\definecolor{textcolor}{rgb}{0.150000,0.150000,0.150000}%
\pgfsetstrokecolor{textcolor}%
\pgfsetfillcolor{textcolor}%
\pgftext[x=1.398871in,y=0.211111in,,top]{\color{textcolor}\sffamily\fontsize{8.800000}{10.560000}\selectfont 1k}%
\end{pgfscope}%
\begin{pgfscope}%
\pgfsetbuttcap%
\pgfsetroundjoin%
\definecolor{currentfill}{rgb}{0.150000,0.150000,0.150000}%
\pgfsetfillcolor{currentfill}%
\pgfsetlinewidth{1.003750pt}%
\definecolor{currentstroke}{rgb}{0.150000,0.150000,0.150000}%
\pgfsetstrokecolor{currentstroke}%
\pgfsetdash{}{0pt}%
\pgfsys@defobject{currentmarker}{\pgfqpoint{0.000000in}{-0.066667in}}{\pgfqpoint{0.000000in}{0.000000in}}{%
\pgfpathmoveto{\pgfqpoint{0.000000in}{0.000000in}}%
\pgfpathlineto{\pgfqpoint{0.000000in}{-0.066667in}}%
\pgfusepath{stroke,fill}%
}%
\begin{pgfscope}%
\pgfsys@transformshift{1.731285in}{0.326389in}%
\pgfsys@useobject{currentmarker}{}%
\end{pgfscope}%
\end{pgfscope}%
\begin{pgfscope}%
\definecolor{textcolor}{rgb}{0.150000,0.150000,0.150000}%
\pgfsetstrokecolor{textcolor}%
\pgfsetfillcolor{textcolor}%
\pgftext[x=1.731285in,y=0.211111in,,top]{\color{textcolor}\sffamily\fontsize{8.800000}{10.560000}\selectfont 2.5k}%
\end{pgfscope}%
\begin{pgfscope}%
\pgfsetbuttcap%
\pgfsetroundjoin%
\definecolor{currentfill}{rgb}{0.150000,0.150000,0.150000}%
\pgfsetfillcolor{currentfill}%
\pgfsetlinewidth{1.003750pt}%
\definecolor{currentstroke}{rgb}{0.150000,0.150000,0.150000}%
\pgfsetstrokecolor{currentstroke}%
\pgfsetdash{}{0pt}%
\pgfsys@defobject{currentmarker}{\pgfqpoint{0.000000in}{-0.066667in}}{\pgfqpoint{0.000000in}{0.000000in}}{%
\pgfpathmoveto{\pgfqpoint{0.000000in}{0.000000in}}%
\pgfpathlineto{\pgfqpoint{0.000000in}{-0.066667in}}%
\pgfusepath{stroke,fill}%
}%
\begin{pgfscope}%
\pgfsys@transformshift{1.982746in}{0.326389in}%
\pgfsys@useobject{currentmarker}{}%
\end{pgfscope}%
\end{pgfscope}%
\begin{pgfscope}%
\definecolor{textcolor}{rgb}{0.150000,0.150000,0.150000}%
\pgfsetstrokecolor{textcolor}%
\pgfsetfillcolor{textcolor}%
\pgftext[x=1.982746in,y=0.211111in,,top]{\color{textcolor}\sffamily\fontsize{8.800000}{10.560000}\selectfont 5k}%
\end{pgfscope}%
\begin{pgfscope}%
\pgfsetbuttcap%
\pgfsetroundjoin%
\definecolor{currentfill}{rgb}{0.150000,0.150000,0.150000}%
\pgfsetfillcolor{currentfill}%
\pgfsetlinewidth{1.003750pt}%
\definecolor{currentstroke}{rgb}{0.150000,0.150000,0.150000}%
\pgfsetstrokecolor{currentstroke}%
\pgfsetdash{}{0pt}%
\pgfsys@defobject{currentmarker}{\pgfqpoint{0.000000in}{-0.066667in}}{\pgfqpoint{0.000000in}{0.000000in}}{%
\pgfpathmoveto{\pgfqpoint{0.000000in}{0.000000in}}%
\pgfpathlineto{\pgfqpoint{0.000000in}{-0.066667in}}%
\pgfusepath{stroke,fill}%
}%
\begin{pgfscope}%
\pgfsys@transformshift{2.234207in}{0.326389in}%
\pgfsys@useobject{currentmarker}{}%
\end{pgfscope}%
\end{pgfscope}%
\begin{pgfscope}%
\definecolor{textcolor}{rgb}{0.150000,0.150000,0.150000}%
\pgfsetstrokecolor{textcolor}%
\pgfsetfillcolor{textcolor}%
\pgftext[x=2.234207in,y=0.211111in,,top]{\color{textcolor}\sffamily\fontsize{8.800000}{10.560000}\selectfont 10k}%
\end{pgfscope}%
\begin{pgfscope}%
\pgfsetbuttcap%
\pgfsetroundjoin%
\definecolor{currentfill}{rgb}{0.150000,0.150000,0.150000}%
\pgfsetfillcolor{currentfill}%
\pgfsetlinewidth{1.003750pt}%
\definecolor{currentstroke}{rgb}{0.150000,0.150000,0.150000}%
\pgfsetstrokecolor{currentstroke}%
\pgfsetdash{}{0pt}%
\pgfsys@defobject{currentmarker}{\pgfqpoint{0.000000in}{-0.066667in}}{\pgfqpoint{0.000000in}{0.000000in}}{%
\pgfpathmoveto{\pgfqpoint{0.000000in}{0.000000in}}%
\pgfpathlineto{\pgfqpoint{0.000000in}{-0.066667in}}%
\pgfusepath{stroke,fill}%
}%
\begin{pgfscope}%
\pgfsys@transformshift{2.566620in}{0.326389in}%
\pgfsys@useobject{currentmarker}{}%
\end{pgfscope}%
\end{pgfscope}%
\begin{pgfscope}%
\definecolor{textcolor}{rgb}{0.150000,0.150000,0.150000}%
\pgfsetstrokecolor{textcolor}%
\pgfsetfillcolor{textcolor}%
\pgftext[x=2.566620in,y=0.211111in,,top]{\color{textcolor}\sffamily\fontsize{8.800000}{10.560000}\selectfont 25k}%
\end{pgfscope}%
\begin{pgfscope}%
\pgfsetbuttcap%
\pgfsetroundjoin%
\definecolor{currentfill}{rgb}{0.150000,0.150000,0.150000}%
\pgfsetfillcolor{currentfill}%
\pgfsetlinewidth{1.003750pt}%
\definecolor{currentstroke}{rgb}{0.150000,0.150000,0.150000}%
\pgfsetstrokecolor{currentstroke}%
\pgfsetdash{}{0pt}%
\pgfsys@defobject{currentmarker}{\pgfqpoint{0.000000in}{-0.066667in}}{\pgfqpoint{0.000000in}{0.000000in}}{%
\pgfpathmoveto{\pgfqpoint{0.000000in}{0.000000in}}%
\pgfpathlineto{\pgfqpoint{0.000000in}{-0.066667in}}%
\pgfusepath{stroke,fill}%
}%
\begin{pgfscope}%
\pgfsys@transformshift{2.818081in}{0.326389in}%
\pgfsys@useobject{currentmarker}{}%
\end{pgfscope}%
\end{pgfscope}%
\begin{pgfscope}%
\definecolor{textcolor}{rgb}{0.150000,0.150000,0.150000}%
\pgfsetstrokecolor{textcolor}%
\pgfsetfillcolor{textcolor}%
\pgftext[x=2.818081in,y=0.211111in,,top]{\color{textcolor}\sffamily\fontsize{8.800000}{10.560000}\selectfont 50k}%
\end{pgfscope}%
\begin{pgfscope}%
\pgfsetbuttcap%
\pgfsetroundjoin%
\definecolor{currentfill}{rgb}{0.150000,0.150000,0.150000}%
\pgfsetfillcolor{currentfill}%
\pgfsetlinewidth{1.003750pt}%
\definecolor{currentstroke}{rgb}{0.150000,0.150000,0.150000}%
\pgfsetstrokecolor{currentstroke}%
\pgfsetdash{}{0pt}%
\pgfsys@defobject{currentmarker}{\pgfqpoint{-0.066667in}{0.000000in}}{\pgfqpoint{0.000000in}{0.000000in}}{%
\pgfpathmoveto{\pgfqpoint{0.000000in}{0.000000in}}%
\pgfpathlineto{\pgfqpoint{-0.066667in}{0.000000in}}%
\pgfusepath{stroke,fill}%
}%
\begin{pgfscope}%
\pgfsys@transformshift{0.450809in}{0.538139in}%
\pgfsys@useobject{currentmarker}{}%
\end{pgfscope}%
\end{pgfscope}%
\begin{pgfscope}%
\definecolor{textcolor}{rgb}{0.150000,0.150000,0.150000}%
\pgfsetstrokecolor{textcolor}%
\pgfsetfillcolor{textcolor}%
\pgftext[x=0.100000in,y=0.494736in,left,base]{\color{textcolor}\sffamily\fontsize{8.800000}{10.560000}\selectfont 20\%}%
\end{pgfscope}%
\begin{pgfscope}%
\pgfsetbuttcap%
\pgfsetroundjoin%
\definecolor{currentfill}{rgb}{0.150000,0.150000,0.150000}%
\pgfsetfillcolor{currentfill}%
\pgfsetlinewidth{1.003750pt}%
\definecolor{currentstroke}{rgb}{0.150000,0.150000,0.150000}%
\pgfsetstrokecolor{currentstroke}%
\pgfsetdash{}{0pt}%
\pgfsys@defobject{currentmarker}{\pgfqpoint{-0.066667in}{0.000000in}}{\pgfqpoint{0.000000in}{0.000000in}}{%
\pgfpathmoveto{\pgfqpoint{0.000000in}{0.000000in}}%
\pgfpathlineto{\pgfqpoint{-0.066667in}{0.000000in}}%
\pgfusepath{stroke,fill}%
}%
\begin{pgfscope}%
\pgfsys@transformshift{0.450809in}{0.961639in}%
\pgfsys@useobject{currentmarker}{}%
\end{pgfscope}%
\end{pgfscope}%
\begin{pgfscope}%
\definecolor{textcolor}{rgb}{0.150000,0.150000,0.150000}%
\pgfsetstrokecolor{textcolor}%
\pgfsetfillcolor{textcolor}%
\pgftext[x=0.100000in,y=0.918236in,left,base]{\color{textcolor}\sffamily\fontsize{8.800000}{10.560000}\selectfont 40\%}%
\end{pgfscope}%
\begin{pgfscope}%
\pgfsetbuttcap%
\pgfsetroundjoin%
\definecolor{currentfill}{rgb}{0.150000,0.150000,0.150000}%
\pgfsetfillcolor{currentfill}%
\pgfsetlinewidth{1.003750pt}%
\definecolor{currentstroke}{rgb}{0.150000,0.150000,0.150000}%
\pgfsetstrokecolor{currentstroke}%
\pgfsetdash{}{0pt}%
\pgfsys@defobject{currentmarker}{\pgfqpoint{-0.066667in}{0.000000in}}{\pgfqpoint{0.000000in}{0.000000in}}{%
\pgfpathmoveto{\pgfqpoint{0.000000in}{0.000000in}}%
\pgfpathlineto{\pgfqpoint{-0.066667in}{0.000000in}}%
\pgfusepath{stroke,fill}%
}%
\begin{pgfscope}%
\pgfsys@transformshift{0.450809in}{1.385139in}%
\pgfsys@useobject{currentmarker}{}%
\end{pgfscope}%
\end{pgfscope}%
\begin{pgfscope}%
\definecolor{textcolor}{rgb}{0.150000,0.150000,0.150000}%
\pgfsetstrokecolor{textcolor}%
\pgfsetfillcolor{textcolor}%
\pgftext[x=0.100000in,y=1.341736in,left,base]{\color{textcolor}\sffamily\fontsize{8.800000}{10.560000}\selectfont 60\%}%
\end{pgfscope}%
\begin{pgfscope}%
\pgfsetbuttcap%
\pgfsetroundjoin%
\definecolor{currentfill}{rgb}{0.150000,0.150000,0.150000}%
\pgfsetfillcolor{currentfill}%
\pgfsetlinewidth{1.003750pt}%
\definecolor{currentstroke}{rgb}{0.150000,0.150000,0.150000}%
\pgfsetstrokecolor{currentstroke}%
\pgfsetdash{}{0pt}%
\pgfsys@defobject{currentmarker}{\pgfqpoint{-0.066667in}{0.000000in}}{\pgfqpoint{0.000000in}{0.000000in}}{%
\pgfpathmoveto{\pgfqpoint{0.000000in}{0.000000in}}%
\pgfpathlineto{\pgfqpoint{-0.066667in}{0.000000in}}%
\pgfusepath{stroke,fill}%
}%
\begin{pgfscope}%
\pgfsys@transformshift{0.450809in}{1.808639in}%
\pgfsys@useobject{currentmarker}{}%
\end{pgfscope}%
\end{pgfscope}%
\begin{pgfscope}%
\definecolor{textcolor}{rgb}{0.150000,0.150000,0.150000}%
\pgfsetstrokecolor{textcolor}%
\pgfsetfillcolor{textcolor}%
\pgftext[x=0.100000in,y=1.765236in,left,base]{\color{textcolor}\sffamily\fontsize{8.800000}{10.560000}\selectfont 80\%}%
\end{pgfscope}%
\begin{pgfscope}%
\pgfsetbuttcap%
\pgfsetroundjoin%
\definecolor{currentfill}{rgb}{0.150000,0.150000,0.150000}%
\pgfsetfillcolor{currentfill}%
\pgfsetlinewidth{0.803000pt}%
\definecolor{currentstroke}{rgb}{0.150000,0.150000,0.150000}%
\pgfsetstrokecolor{currentstroke}%
\pgfsetdash{}{0pt}%
\pgfsys@defobject{currentmarker}{\pgfqpoint{-0.044444in}{0.000000in}}{\pgfqpoint{0.000000in}{0.000000in}}{%
\pgfpathmoveto{\pgfqpoint{0.000000in}{0.000000in}}%
\pgfpathlineto{\pgfqpoint{-0.044444in}{0.000000in}}%
\pgfusepath{stroke,fill}%
}%
\begin{pgfscope}%
\pgfsys@transformshift{0.450809in}{0.326389in}%
\pgfsys@useobject{currentmarker}{}%
\end{pgfscope}%
\end{pgfscope}%
\begin{pgfscope}%
\pgfsetbuttcap%
\pgfsetroundjoin%
\definecolor{currentfill}{rgb}{0.150000,0.150000,0.150000}%
\pgfsetfillcolor{currentfill}%
\pgfsetlinewidth{0.803000pt}%
\definecolor{currentstroke}{rgb}{0.150000,0.150000,0.150000}%
\pgfsetstrokecolor{currentstroke}%
\pgfsetdash{}{0pt}%
\pgfsys@defobject{currentmarker}{\pgfqpoint{-0.044444in}{0.000000in}}{\pgfqpoint{0.000000in}{0.000000in}}{%
\pgfpathmoveto{\pgfqpoint{0.000000in}{0.000000in}}%
\pgfpathlineto{\pgfqpoint{-0.044444in}{0.000000in}}%
\pgfusepath{stroke,fill}%
}%
\begin{pgfscope}%
\pgfsys@transformshift{0.450809in}{0.749889in}%
\pgfsys@useobject{currentmarker}{}%
\end{pgfscope}%
\end{pgfscope}%
\begin{pgfscope}%
\pgfsetbuttcap%
\pgfsetroundjoin%
\definecolor{currentfill}{rgb}{0.150000,0.150000,0.150000}%
\pgfsetfillcolor{currentfill}%
\pgfsetlinewidth{0.803000pt}%
\definecolor{currentstroke}{rgb}{0.150000,0.150000,0.150000}%
\pgfsetstrokecolor{currentstroke}%
\pgfsetdash{}{0pt}%
\pgfsys@defobject{currentmarker}{\pgfqpoint{-0.044444in}{0.000000in}}{\pgfqpoint{0.000000in}{0.000000in}}{%
\pgfpathmoveto{\pgfqpoint{0.000000in}{0.000000in}}%
\pgfpathlineto{\pgfqpoint{-0.044444in}{0.000000in}}%
\pgfusepath{stroke,fill}%
}%
\begin{pgfscope}%
\pgfsys@transformshift{0.450809in}{1.173389in}%
\pgfsys@useobject{currentmarker}{}%
\end{pgfscope}%
\end{pgfscope}%
\begin{pgfscope}%
\pgfsetbuttcap%
\pgfsetroundjoin%
\definecolor{currentfill}{rgb}{0.150000,0.150000,0.150000}%
\pgfsetfillcolor{currentfill}%
\pgfsetlinewidth{0.803000pt}%
\definecolor{currentstroke}{rgb}{0.150000,0.150000,0.150000}%
\pgfsetstrokecolor{currentstroke}%
\pgfsetdash{}{0pt}%
\pgfsys@defobject{currentmarker}{\pgfqpoint{-0.044444in}{0.000000in}}{\pgfqpoint{0.000000in}{0.000000in}}{%
\pgfpathmoveto{\pgfqpoint{0.000000in}{0.000000in}}%
\pgfpathlineto{\pgfqpoint{-0.044444in}{0.000000in}}%
\pgfusepath{stroke,fill}%
}%
\begin{pgfscope}%
\pgfsys@transformshift{0.450809in}{1.596889in}%
\pgfsys@useobject{currentmarker}{}%
\end{pgfscope}%
\end{pgfscope}%
\begin{pgfscope}%
\pgfsetbuttcap%
\pgfsetroundjoin%
\definecolor{currentfill}{rgb}{0.150000,0.150000,0.150000}%
\pgfsetfillcolor{currentfill}%
\pgfsetlinewidth{0.803000pt}%
\definecolor{currentstroke}{rgb}{0.150000,0.150000,0.150000}%
\pgfsetstrokecolor{currentstroke}%
\pgfsetdash{}{0pt}%
\pgfsys@defobject{currentmarker}{\pgfqpoint{-0.044444in}{0.000000in}}{\pgfqpoint{0.000000in}{0.000000in}}{%
\pgfpathmoveto{\pgfqpoint{0.000000in}{0.000000in}}%
\pgfpathlineto{\pgfqpoint{-0.044444in}{0.000000in}}%
\pgfusepath{stroke,fill}%
}%
\begin{pgfscope}%
\pgfsys@transformshift{0.450809in}{2.020389in}%
\pgfsys@useobject{currentmarker}{}%
\end{pgfscope}%
\end{pgfscope}%
\begin{pgfscope}%
\pgfpathrectangle{\pgfqpoint{0.450809in}{0.326389in}}{\pgfqpoint{2.480000in}{1.694000in}}%
\pgfusepath{clip}%
\pgfsetbuttcap%
\pgfsetroundjoin%
\pgfsetlinewidth{1.204500pt}%
\definecolor{currentstroke}{rgb}{1.000000,0.498039,0.000000}%
\pgfsetstrokecolor{currentstroke}%
\pgfsetdash{{7.680000pt}{1.920000pt}{1.200000pt}{1.920000pt}}{0.000000pt}%
\pgfpathmoveto{\pgfqpoint{0.563536in}{1.305827in}}%
\pgfpathlineto{\pgfqpoint{0.895949in}{1.352011in}}%
\pgfpathlineto{\pgfqpoint{1.147410in}{1.365283in}}%
\pgfpathlineto{\pgfqpoint{1.398871in}{1.412056in}}%
\pgfpathlineto{\pgfqpoint{1.731285in}{1.524657in}}%
\pgfpathlineto{\pgfqpoint{1.982746in}{1.598150in}}%
\pgfpathlineto{\pgfqpoint{2.234207in}{1.645481in}}%
\pgfpathlineto{\pgfqpoint{2.566620in}{1.772864in}}%
\pgfpathlineto{\pgfqpoint{2.818081in}{1.855416in}}%
\pgfusepath{stroke}%
\end{pgfscope}%
\begin{pgfscope}%
\pgfpathrectangle{\pgfqpoint{0.450809in}{0.326389in}}{\pgfqpoint{2.480000in}{1.694000in}}%
\pgfusepath{clip}%
\pgfsetroundcap%
\pgfsetroundjoin%
\pgfsetlinewidth{1.204500pt}%
\definecolor{currentstroke}{rgb}{0.890196,0.101961,0.109804}%
\pgfsetstrokecolor{currentstroke}%
\pgfsetdash{}{0pt}%
\pgfpathmoveto{\pgfqpoint{0.563536in}{1.159053in}}%
\pgfpathlineto{\pgfqpoint{0.895949in}{1.245883in}}%
\pgfpathlineto{\pgfqpoint{1.147410in}{1.296509in}}%
\pgfpathlineto{\pgfqpoint{1.398871in}{1.355431in}}%
\pgfpathlineto{\pgfqpoint{1.731285in}{1.522429in}}%
\pgfpathlineto{\pgfqpoint{1.982746in}{1.625915in}}%
\pgfpathlineto{\pgfqpoint{2.234207in}{1.656387in}}%
\pgfpathlineto{\pgfqpoint{2.566620in}{1.835124in}}%
\pgfpathlineto{\pgfqpoint{2.818081in}{1.903253in}}%
\pgfusepath{stroke}%
\end{pgfscope}%
\begin{pgfscope}%
\pgfpathrectangle{\pgfqpoint{0.450809in}{0.326389in}}{\pgfqpoint{2.480000in}{1.694000in}}%
\pgfusepath{clip}%
\pgfsetbuttcap%
\pgfsetroundjoin%
\pgfsetlinewidth{1.204500pt}%
\definecolor{currentstroke}{rgb}{0.200000,0.627451,0.172549}%
\pgfsetstrokecolor{currentstroke}%
\pgfsetdash{{4.440000pt}{1.920000pt}}{0.000000pt}%
\pgfpathmoveto{\pgfqpoint{0.563536in}{0.633665in}}%
\pgfpathlineto{\pgfqpoint{0.895949in}{0.923935in}}%
\pgfpathlineto{\pgfqpoint{1.147410in}{1.106171in}}%
\pgfpathlineto{\pgfqpoint{1.398871in}{1.231510in}}%
\pgfpathlineto{\pgfqpoint{1.731285in}{1.416361in}}%
\pgfpathlineto{\pgfqpoint{1.982746in}{1.562988in}}%
\pgfpathlineto{\pgfqpoint{2.234207in}{1.604061in}}%
\pgfpathlineto{\pgfqpoint{2.566620in}{1.775801in}}%
\pgfpathlineto{\pgfqpoint{2.818081in}{1.858460in}}%
\pgfusepath{stroke}%
\end{pgfscope}%
\begin{pgfscope}%
\pgfpathrectangle{\pgfqpoint{0.450809in}{0.326389in}}{\pgfqpoint{2.480000in}{1.694000in}}%
\pgfusepath{clip}%
\pgfsetbuttcap%
\pgfsetroundjoin%
\pgfsetlinewidth{1.204500pt}%
\definecolor{currentstroke}{rgb}{0.121569,0.470588,0.705882}%
\pgfsetstrokecolor{currentstroke}%
\pgfsetdash{{1.200000pt}{1.980000pt}}{0.000000pt}%
\pgfpathmoveto{\pgfqpoint{0.563536in}{0.824887in}}%
\pgfpathlineto{\pgfqpoint{0.895949in}{1.101650in}}%
\pgfpathlineto{\pgfqpoint{1.147410in}{1.185248in}}%
\pgfpathlineto{\pgfqpoint{1.398871in}{1.294594in}}%
\pgfpathlineto{\pgfqpoint{1.731285in}{1.415896in}}%
\pgfpathlineto{\pgfqpoint{1.982746in}{1.522475in}}%
\pgfpathlineto{\pgfqpoint{2.234207in}{1.602436in}}%
\pgfpathlineto{\pgfqpoint{2.566620in}{1.768182in}}%
\pgfpathlineto{\pgfqpoint{2.818081in}{1.886864in}}%
\pgfusepath{stroke}%
\end{pgfscope}%
\begin{pgfscope}%
\pgfsetrectcap%
\pgfsetmiterjoin%
\pgfsetlinewidth{1.003750pt}%
\definecolor{currentstroke}{rgb}{0.150000,0.150000,0.150000}%
\pgfsetstrokecolor{currentstroke}%
\pgfsetdash{}{0pt}%
\pgfpathmoveto{\pgfqpoint{0.450809in}{0.326389in}}%
\pgfpathlineto{\pgfqpoint{0.450809in}{2.020389in}}%
\pgfusepath{stroke}%
\end{pgfscope}%
\begin{pgfscope}%
\pgfsetrectcap%
\pgfsetmiterjoin%
\pgfsetlinewidth{1.003750pt}%
\definecolor{currentstroke}{rgb}{0.150000,0.150000,0.150000}%
\pgfsetstrokecolor{currentstroke}%
\pgfsetdash{}{0pt}%
\pgfpathmoveto{\pgfqpoint{2.930809in}{0.326389in}}%
\pgfpathlineto{\pgfqpoint{2.930809in}{2.020389in}}%
\pgfusepath{stroke}%
\end{pgfscope}%
\begin{pgfscope}%
\pgfsetrectcap%
\pgfsetmiterjoin%
\pgfsetlinewidth{1.003750pt}%
\definecolor{currentstroke}{rgb}{0.150000,0.150000,0.150000}%
\pgfsetstrokecolor{currentstroke}%
\pgfsetdash{}{0pt}%
\pgfpathmoveto{\pgfqpoint{0.450809in}{0.326389in}}%
\pgfpathlineto{\pgfqpoint{2.930809in}{0.326389in}}%
\pgfusepath{stroke}%
\end{pgfscope}%
\begin{pgfscope}%
\pgfsetrectcap%
\pgfsetmiterjoin%
\pgfsetlinewidth{1.003750pt}%
\definecolor{currentstroke}{rgb}{0.150000,0.150000,0.150000}%
\pgfsetstrokecolor{currentstroke}%
\pgfsetdash{}{0pt}%
\pgfpathmoveto{\pgfqpoint{0.450809in}{2.020389in}}%
\pgfpathlineto{\pgfqpoint{2.930809in}{2.020389in}}%
\pgfusepath{stroke}%
\end{pgfscope}%
\begin{pgfscope}%
\pgfsetbuttcap%
\pgfsetroundjoin%
\pgfsetlinewidth{1.204500pt}%
\definecolor{currentstroke}{rgb}{1.000000,0.498039,0.000000}%
\pgfsetstrokecolor{currentstroke}%
\pgfsetdash{{7.680000pt}{1.920000pt}{1.200000pt}{1.920000pt}}{0.000000pt}%
\pgfpathmoveto{\pgfqpoint{1.816819in}{0.971250in}}%
\pgfpathlineto{\pgfqpoint{2.061263in}{0.971250in}}%
\pgfusepath{stroke}%
\end{pgfscope}%
\begin{pgfscope}%
\definecolor{textcolor}{rgb}{0.150000,0.150000,0.150000}%
\pgfsetstrokecolor{textcolor}%
\pgfsetfillcolor{textcolor}%
\pgftext[x=2.159041in,y=0.928472in,left,base]{\color{textcolor}\sffamily\fontsize{8.800000}{10.560000}\selectfont Norma}%
\end{pgfscope}%
\begin{pgfscope}%
\pgfsetroundcap%
\pgfsetroundjoin%
\pgfsetlinewidth{1.204500pt}%
\definecolor{currentstroke}{rgb}{0.890196,0.101961,0.109804}%
\pgfsetstrokecolor{currentstroke}%
\pgfsetdash{}{0pt}%
\pgfpathmoveto{\pgfqpoint{1.816819in}{0.799028in}}%
\pgfpathlineto{\pgfqpoint{2.061263in}{0.799028in}}%
\pgfusepath{stroke}%
\end{pgfscope}%
\begin{pgfscope}%
\definecolor{textcolor}{rgb}{0.150000,0.150000,0.150000}%
\pgfsetstrokecolor{textcolor}%
\pgfsetfillcolor{textcolor}%
\pgftext[x=2.159041in,y=0.756250in,left,base]{\color{textcolor}\sffamily\fontsize{8.800000}{10.560000}\selectfont cSMTiser\textsubscript{+LM}}%
\end{pgfscope}%
\begin{pgfscope}%
\pgfsetbuttcap%
\pgfsetroundjoin%
\pgfsetlinewidth{1.204500pt}%
\definecolor{currentstroke}{rgb}{0.200000,0.627451,0.172549}%
\pgfsetstrokecolor{currentstroke}%
\pgfsetdash{{4.440000pt}{1.920000pt}}{0.000000pt}%
\pgfpathmoveto{\pgfqpoint{1.816819in}{0.626806in}}%
\pgfpathlineto{\pgfqpoint{2.061263in}{0.626806in}}%
\pgfusepath{stroke}%
\end{pgfscope}%
\begin{pgfscope}%
\definecolor{textcolor}{rgb}{0.150000,0.150000,0.150000}%
\pgfsetstrokecolor{textcolor}%
\pgfsetfillcolor{textcolor}%
\pgftext[x=2.159041in,y=0.584028in,left,base]{\color{textcolor}\sffamily\fontsize{8.800000}{10.560000}\selectfont NMT-1}%
\end{pgfscope}%
\begin{pgfscope}%
\pgfsetbuttcap%
\pgfsetroundjoin%
\pgfsetlinewidth{1.204500pt}%
\definecolor{currentstroke}{rgb}{0.121569,0.470588,0.705882}%
\pgfsetstrokecolor{currentstroke}%
\pgfsetdash{{1.200000pt}{1.980000pt}}{0.000000pt}%
\pgfpathmoveto{\pgfqpoint{1.816819in}{0.454583in}}%
\pgfpathlineto{\pgfqpoint{2.061263in}{0.454583in}}%
\pgfusepath{stroke}%
\end{pgfscope}%
\begin{pgfscope}%
\definecolor{textcolor}{rgb}{0.150000,0.150000,0.150000}%
\pgfsetstrokecolor{textcolor}%
\pgfsetfillcolor{textcolor}%
\pgftext[x=2.159041in,y=0.411806in,left,base]{\color{textcolor}\sffamily\fontsize{8.800000}{10.560000}\selectfont NMT-2}%
\end{pgfscope}%
\end{pgfpicture}%
\makeatother%
\endgroup%

%% file: curves_german-ridges.pgf
\begingroup%
\makeatletter%
\begin{pgfpicture}%
\pgfpathrectangle{\pgfpointorigin}{\pgfqpoint{3.078809in}{2.168389in}}%
\pgfusepath{use as bounding box, clip}%
\begin{pgfscope}%
\pgfsetbuttcap%
\pgfsetmiterjoin%
\definecolor{currentfill}{rgb}{1.000000,1.000000,1.000000}%
\pgfsetfillcolor{currentfill}%
\pgfsetlinewidth{0.000000pt}%
\definecolor{currentstroke}{rgb}{1.000000,1.000000,1.000000}%
\pgfsetstrokecolor{currentstroke}%
\pgfsetdash{}{0pt}%
\pgfpathmoveto{\pgfqpoint{0.000000in}{0.000000in}}%
\pgfpathlineto{\pgfqpoint{3.078809in}{0.000000in}}%
\pgfpathlineto{\pgfqpoint{3.078809in}{2.168389in}}%
\pgfpathlineto{\pgfqpoint{0.000000in}{2.168389in}}%
\pgfpathclose%
\pgfusepath{fill}%
\end{pgfscope}%
\begin{pgfscope}%
\pgfsetbuttcap%
\pgfsetmiterjoin%
\definecolor{currentfill}{rgb}{1.000000,1.000000,1.000000}%
\pgfsetfillcolor{currentfill}%
\pgfsetlinewidth{0.000000pt}%
\definecolor{currentstroke}{rgb}{0.000000,0.000000,0.000000}%
\pgfsetstrokecolor{currentstroke}%
\pgfsetstrokeopacity{0.000000}%
\pgfsetdash{}{0pt}%
\pgfpathmoveto{\pgfqpoint{0.450809in}{0.326389in}}%
\pgfpathlineto{\pgfqpoint{2.930809in}{0.326389in}}%
\pgfpathlineto{\pgfqpoint{2.930809in}{2.020389in}}%
\pgfpathlineto{\pgfqpoint{0.450809in}{2.020389in}}%
\pgfpathclose%
\pgfusepath{fill}%
\end{pgfscope}%
\begin{pgfscope}%
\pgfsetbuttcap%
\pgfsetroundjoin%
\definecolor{currentfill}{rgb}{0.150000,0.150000,0.150000}%
\pgfsetfillcolor{currentfill}%
\pgfsetlinewidth{1.003750pt}%
\definecolor{currentstroke}{rgb}{0.150000,0.150000,0.150000}%
\pgfsetstrokecolor{currentstroke}%
\pgfsetdash{}{0pt}%
\pgfsys@defobject{currentmarker}{\pgfqpoint{0.000000in}{-0.066667in}}{\pgfqpoint{0.000000in}{0.000000in}}{%
\pgfpathmoveto{\pgfqpoint{0.000000in}{0.000000in}}%
\pgfpathlineto{\pgfqpoint{0.000000in}{-0.066667in}}%
\pgfusepath{stroke,fill}%
}%
\begin{pgfscope}%
\pgfsys@transformshift{0.563536in}{0.326389in}%
\pgfsys@useobject{currentmarker}{}%
\end{pgfscope}%
\end{pgfscope}%
\begin{pgfscope}%
\definecolor{textcolor}{rgb}{0.150000,0.150000,0.150000}%
\pgfsetstrokecolor{textcolor}%
\pgfsetfillcolor{textcolor}%
\pgftext[x=0.563536in,y=0.211111in,,top]{\color{textcolor}\sffamily\fontsize{8.800000}{10.560000}\selectfont 100}%
\end{pgfscope}%
\begin{pgfscope}%
\pgfsetbuttcap%
\pgfsetroundjoin%
\definecolor{currentfill}{rgb}{0.150000,0.150000,0.150000}%
\pgfsetfillcolor{currentfill}%
\pgfsetlinewidth{1.003750pt}%
\definecolor{currentstroke}{rgb}{0.150000,0.150000,0.150000}%
\pgfsetstrokecolor{currentstroke}%
\pgfsetdash{}{0pt}%
\pgfsys@defobject{currentmarker}{\pgfqpoint{0.000000in}{-0.066667in}}{\pgfqpoint{0.000000in}{0.000000in}}{%
\pgfpathmoveto{\pgfqpoint{0.000000in}{0.000000in}}%
\pgfpathlineto{\pgfqpoint{0.000000in}{-0.066667in}}%
\pgfusepath{stroke,fill}%
}%
\begin{pgfscope}%
\pgfsys@transformshift{0.937679in}{0.326389in}%
\pgfsys@useobject{currentmarker}{}%
\end{pgfscope}%
\end{pgfscope}%
\begin{pgfscope}%
\definecolor{textcolor}{rgb}{0.150000,0.150000,0.150000}%
\pgfsetstrokecolor{textcolor}%
\pgfsetfillcolor{textcolor}%
\pgftext[x=0.937679in,y=0.211111in,,top]{\color{textcolor}\sffamily\fontsize{8.800000}{10.560000}\selectfont 250}%
\end{pgfscope}%
\begin{pgfscope}%
\pgfsetbuttcap%
\pgfsetroundjoin%
\definecolor{currentfill}{rgb}{0.150000,0.150000,0.150000}%
\pgfsetfillcolor{currentfill}%
\pgfsetlinewidth{1.003750pt}%
\definecolor{currentstroke}{rgb}{0.150000,0.150000,0.150000}%
\pgfsetstrokecolor{currentstroke}%
\pgfsetdash{}{0pt}%
\pgfsys@defobject{currentmarker}{\pgfqpoint{0.000000in}{-0.066667in}}{\pgfqpoint{0.000000in}{0.000000in}}{%
\pgfpathmoveto{\pgfqpoint{0.000000in}{0.000000in}}%
\pgfpathlineto{\pgfqpoint{0.000000in}{-0.066667in}}%
\pgfusepath{stroke,fill}%
}%
\begin{pgfscope}%
\pgfsys@transformshift{1.220708in}{0.326389in}%
\pgfsys@useobject{currentmarker}{}%
\end{pgfscope}%
\end{pgfscope}%
\begin{pgfscope}%
\definecolor{textcolor}{rgb}{0.150000,0.150000,0.150000}%
\pgfsetstrokecolor{textcolor}%
\pgfsetfillcolor{textcolor}%
\pgftext[x=1.220708in,y=0.211111in,,top]{\color{textcolor}\sffamily\fontsize{8.800000}{10.560000}\selectfont 500}%
\end{pgfscope}%
\begin{pgfscope}%
\pgfsetbuttcap%
\pgfsetroundjoin%
\definecolor{currentfill}{rgb}{0.150000,0.150000,0.150000}%
\pgfsetfillcolor{currentfill}%
\pgfsetlinewidth{1.003750pt}%
\definecolor{currentstroke}{rgb}{0.150000,0.150000,0.150000}%
\pgfsetstrokecolor{currentstroke}%
\pgfsetdash{}{0pt}%
\pgfsys@defobject{currentmarker}{\pgfqpoint{0.000000in}{-0.066667in}}{\pgfqpoint{0.000000in}{0.000000in}}{%
\pgfpathmoveto{\pgfqpoint{0.000000in}{0.000000in}}%
\pgfpathlineto{\pgfqpoint{0.000000in}{-0.066667in}}%
\pgfusepath{stroke,fill}%
}%
\begin{pgfscope}%
\pgfsys@transformshift{1.503737in}{0.326389in}%
\pgfsys@useobject{currentmarker}{}%
\end{pgfscope}%
\end{pgfscope}%
\begin{pgfscope}%
\definecolor{textcolor}{rgb}{0.150000,0.150000,0.150000}%
\pgfsetstrokecolor{textcolor}%
\pgfsetfillcolor{textcolor}%
\pgftext[x=1.503737in,y=0.211111in,,top]{\color{textcolor}\sffamily\fontsize{8.800000}{10.560000}\selectfont 1k}%
\end{pgfscope}%
\begin{pgfscope}%
\pgfsetbuttcap%
\pgfsetroundjoin%
\definecolor{currentfill}{rgb}{0.150000,0.150000,0.150000}%
\pgfsetfillcolor{currentfill}%
\pgfsetlinewidth{1.003750pt}%
\definecolor{currentstroke}{rgb}{0.150000,0.150000,0.150000}%
\pgfsetstrokecolor{currentstroke}%
\pgfsetdash{}{0pt}%
\pgfsys@defobject{currentmarker}{\pgfqpoint{0.000000in}{-0.066667in}}{\pgfqpoint{0.000000in}{0.000000in}}{%
\pgfpathmoveto{\pgfqpoint{0.000000in}{0.000000in}}%
\pgfpathlineto{\pgfqpoint{0.000000in}{-0.066667in}}%
\pgfusepath{stroke,fill}%
}%
\begin{pgfscope}%
\pgfsys@transformshift{1.877880in}{0.326389in}%
\pgfsys@useobject{currentmarker}{}%
\end{pgfscope}%
\end{pgfscope}%
\begin{pgfscope}%
\definecolor{textcolor}{rgb}{0.150000,0.150000,0.150000}%
\pgfsetstrokecolor{textcolor}%
\pgfsetfillcolor{textcolor}%
\pgftext[x=1.877880in,y=0.211111in,,top]{\color{textcolor}\sffamily\fontsize{8.800000}{10.560000}\selectfont 2.5k}%
\end{pgfscope}%
\begin{pgfscope}%
\pgfsetbuttcap%
\pgfsetroundjoin%
\definecolor{currentfill}{rgb}{0.150000,0.150000,0.150000}%
\pgfsetfillcolor{currentfill}%
\pgfsetlinewidth{1.003750pt}%
\definecolor{currentstroke}{rgb}{0.150000,0.150000,0.150000}%
\pgfsetstrokecolor{currentstroke}%
\pgfsetdash{}{0pt}%
\pgfsys@defobject{currentmarker}{\pgfqpoint{0.000000in}{-0.066667in}}{\pgfqpoint{0.000000in}{0.000000in}}{%
\pgfpathmoveto{\pgfqpoint{0.000000in}{0.000000in}}%
\pgfpathlineto{\pgfqpoint{0.000000in}{-0.066667in}}%
\pgfusepath{stroke,fill}%
}%
\begin{pgfscope}%
\pgfsys@transformshift{2.160909in}{0.326389in}%
\pgfsys@useobject{currentmarker}{}%
\end{pgfscope}%
\end{pgfscope}%
\begin{pgfscope}%
\definecolor{textcolor}{rgb}{0.150000,0.150000,0.150000}%
\pgfsetstrokecolor{textcolor}%
\pgfsetfillcolor{textcolor}%
\pgftext[x=2.160909in,y=0.211111in,,top]{\color{textcolor}\sffamily\fontsize{8.800000}{10.560000}\selectfont 5k}%
\end{pgfscope}%
\begin{pgfscope}%
\pgfsetbuttcap%
\pgfsetroundjoin%
\definecolor{currentfill}{rgb}{0.150000,0.150000,0.150000}%
\pgfsetfillcolor{currentfill}%
\pgfsetlinewidth{1.003750pt}%
\definecolor{currentstroke}{rgb}{0.150000,0.150000,0.150000}%
\pgfsetstrokecolor{currentstroke}%
\pgfsetdash{}{0pt}%
\pgfsys@defobject{currentmarker}{\pgfqpoint{0.000000in}{-0.066667in}}{\pgfqpoint{0.000000in}{0.000000in}}{%
\pgfpathmoveto{\pgfqpoint{0.000000in}{0.000000in}}%
\pgfpathlineto{\pgfqpoint{0.000000in}{-0.066667in}}%
\pgfusepath{stroke,fill}%
}%
\begin{pgfscope}%
\pgfsys@transformshift{2.443938in}{0.326389in}%
\pgfsys@useobject{currentmarker}{}%
\end{pgfscope}%
\end{pgfscope}%
\begin{pgfscope}%
\definecolor{textcolor}{rgb}{0.150000,0.150000,0.150000}%
\pgfsetstrokecolor{textcolor}%
\pgfsetfillcolor{textcolor}%
\pgftext[x=2.443938in,y=0.211111in,,top]{\color{textcolor}\sffamily\fontsize{8.800000}{10.560000}\selectfont 10k}%
\end{pgfscope}%
\begin{pgfscope}%
\pgfsetbuttcap%
\pgfsetroundjoin%
\definecolor{currentfill}{rgb}{0.150000,0.150000,0.150000}%
\pgfsetfillcolor{currentfill}%
\pgfsetlinewidth{1.003750pt}%
\definecolor{currentstroke}{rgb}{0.150000,0.150000,0.150000}%
\pgfsetstrokecolor{currentstroke}%
\pgfsetdash{}{0pt}%
\pgfsys@defobject{currentmarker}{\pgfqpoint{0.000000in}{-0.066667in}}{\pgfqpoint{0.000000in}{0.000000in}}{%
\pgfpathmoveto{\pgfqpoint{0.000000in}{0.000000in}}%
\pgfpathlineto{\pgfqpoint{0.000000in}{-0.066667in}}%
\pgfusepath{stroke,fill}%
}%
\begin{pgfscope}%
\pgfsys@transformshift{2.818081in}{0.326389in}%
\pgfsys@useobject{currentmarker}{}%
\end{pgfscope}%
\end{pgfscope}%
\begin{pgfscope}%
\definecolor{textcolor}{rgb}{0.150000,0.150000,0.150000}%
\pgfsetstrokecolor{textcolor}%
\pgfsetfillcolor{textcolor}%
\pgftext[x=2.818081in,y=0.211111in,,top]{\color{textcolor}\sffamily\fontsize{8.800000}{10.560000}\selectfont 25k}%
\end{pgfscope}%
\begin{pgfscope}%
\pgfsetbuttcap%
\pgfsetroundjoin%
\definecolor{currentfill}{rgb}{0.150000,0.150000,0.150000}%
\pgfsetfillcolor{currentfill}%
\pgfsetlinewidth{1.003750pt}%
\definecolor{currentstroke}{rgb}{0.150000,0.150000,0.150000}%
\pgfsetstrokecolor{currentstroke}%
\pgfsetdash{}{0pt}%
\pgfsys@defobject{currentmarker}{\pgfqpoint{-0.066667in}{0.000000in}}{\pgfqpoint{0.000000in}{0.000000in}}{%
\pgfpathmoveto{\pgfqpoint{0.000000in}{0.000000in}}%
\pgfpathlineto{\pgfqpoint{-0.066667in}{0.000000in}}%
\pgfusepath{stroke,fill}%
}%
\begin{pgfscope}%
\pgfsys@transformshift{0.450809in}{0.538139in}%
\pgfsys@useobject{currentmarker}{}%
\end{pgfscope}%
\end{pgfscope}%
\begin{pgfscope}%
\definecolor{textcolor}{rgb}{0.150000,0.150000,0.150000}%
\pgfsetstrokecolor{textcolor}%
\pgfsetfillcolor{textcolor}%
\pgftext[x=0.100000in,y=0.494736in,left,base]{\color{textcolor}\sffamily\fontsize{8.800000}{10.560000}\selectfont 20\%}%
\end{pgfscope}%
\begin{pgfscope}%
\pgfsetbuttcap%
\pgfsetroundjoin%
\definecolor{currentfill}{rgb}{0.150000,0.150000,0.150000}%
\pgfsetfillcolor{currentfill}%
\pgfsetlinewidth{1.003750pt}%
\definecolor{currentstroke}{rgb}{0.150000,0.150000,0.150000}%
\pgfsetstrokecolor{currentstroke}%
\pgfsetdash{}{0pt}%
\pgfsys@defobject{currentmarker}{\pgfqpoint{-0.066667in}{0.000000in}}{\pgfqpoint{0.000000in}{0.000000in}}{%
\pgfpathmoveto{\pgfqpoint{0.000000in}{0.000000in}}%
\pgfpathlineto{\pgfqpoint{-0.066667in}{0.000000in}}%
\pgfusepath{stroke,fill}%
}%
\begin{pgfscope}%
\pgfsys@transformshift{0.450809in}{0.961639in}%
\pgfsys@useobject{currentmarker}{}%
\end{pgfscope}%
\end{pgfscope}%
\begin{pgfscope}%
\definecolor{textcolor}{rgb}{0.150000,0.150000,0.150000}%
\pgfsetstrokecolor{textcolor}%
\pgfsetfillcolor{textcolor}%
\pgftext[x=0.100000in,y=0.918236in,left,base]{\color{textcolor}\sffamily\fontsize{8.800000}{10.560000}\selectfont 40\%}%
\end{pgfscope}%
\begin{pgfscope}%
\pgfsetbuttcap%
\pgfsetroundjoin%
\definecolor{currentfill}{rgb}{0.150000,0.150000,0.150000}%
\pgfsetfillcolor{currentfill}%
\pgfsetlinewidth{1.003750pt}%
\definecolor{currentstroke}{rgb}{0.150000,0.150000,0.150000}%
\pgfsetstrokecolor{currentstroke}%
\pgfsetdash{}{0pt}%
\pgfsys@defobject{currentmarker}{\pgfqpoint{-0.066667in}{0.000000in}}{\pgfqpoint{0.000000in}{0.000000in}}{%
\pgfpathmoveto{\pgfqpoint{0.000000in}{0.000000in}}%
\pgfpathlineto{\pgfqpoint{-0.066667in}{0.000000in}}%
\pgfusepath{stroke,fill}%
}%
\begin{pgfscope}%
\pgfsys@transformshift{0.450809in}{1.385139in}%
\pgfsys@useobject{currentmarker}{}%
\end{pgfscope}%
\end{pgfscope}%
\begin{pgfscope}%
\definecolor{textcolor}{rgb}{0.150000,0.150000,0.150000}%
\pgfsetstrokecolor{textcolor}%
\pgfsetfillcolor{textcolor}%
\pgftext[x=0.100000in,y=1.341736in,left,base]{\color{textcolor}\sffamily\fontsize{8.800000}{10.560000}\selectfont 60\%}%
\end{pgfscope}%
\begin{pgfscope}%
\pgfsetbuttcap%
\pgfsetroundjoin%
\definecolor{currentfill}{rgb}{0.150000,0.150000,0.150000}%
\pgfsetfillcolor{currentfill}%
\pgfsetlinewidth{1.003750pt}%
\definecolor{currentstroke}{rgb}{0.150000,0.150000,0.150000}%
\pgfsetstrokecolor{currentstroke}%
\pgfsetdash{}{0pt}%
\pgfsys@defobject{currentmarker}{\pgfqpoint{-0.066667in}{0.000000in}}{\pgfqpoint{0.000000in}{0.000000in}}{%
\pgfpathmoveto{\pgfqpoint{0.000000in}{0.000000in}}%
\pgfpathlineto{\pgfqpoint{-0.066667in}{0.000000in}}%
\pgfusepath{stroke,fill}%
}%
\begin{pgfscope}%
\pgfsys@transformshift{0.450809in}{1.808639in}%
\pgfsys@useobject{currentmarker}{}%
\end{pgfscope}%
\end{pgfscope}%
\begin{pgfscope}%
\definecolor{textcolor}{rgb}{0.150000,0.150000,0.150000}%
\pgfsetstrokecolor{textcolor}%
\pgfsetfillcolor{textcolor}%
\pgftext[x=0.100000in,y=1.765236in,left,base]{\color{textcolor}\sffamily\fontsize{8.800000}{10.560000}\selectfont 80\%}%
\end{pgfscope}%
\begin{pgfscope}%
\pgfsetbuttcap%
\pgfsetroundjoin%
\definecolor{currentfill}{rgb}{0.150000,0.150000,0.150000}%
\pgfsetfillcolor{currentfill}%
\pgfsetlinewidth{0.803000pt}%
\definecolor{currentstroke}{rgb}{0.150000,0.150000,0.150000}%
\pgfsetstrokecolor{currentstroke}%
\pgfsetdash{}{0pt}%
\pgfsys@defobject{currentmarker}{\pgfqpoint{-0.044444in}{0.000000in}}{\pgfqpoint{0.000000in}{0.000000in}}{%
\pgfpathmoveto{\pgfqpoint{0.000000in}{0.000000in}}%
\pgfpathlineto{\pgfqpoint{-0.044444in}{0.000000in}}%
\pgfusepath{stroke,fill}%
}%
\begin{pgfscope}%
\pgfsys@transformshift{0.450809in}{0.326389in}%
\pgfsys@useobject{currentmarker}{}%
\end{pgfscope}%
\end{pgfscope}%
\begin{pgfscope}%
\pgfsetbuttcap%
\pgfsetroundjoin%
\definecolor{currentfill}{rgb}{0.150000,0.150000,0.150000}%
\pgfsetfillcolor{currentfill}%
\pgfsetlinewidth{0.803000pt}%
\definecolor{currentstroke}{rgb}{0.150000,0.150000,0.150000}%
\pgfsetstrokecolor{currentstroke}%
\pgfsetdash{}{0pt}%
\pgfsys@defobject{currentmarker}{\pgfqpoint{-0.044444in}{0.000000in}}{\pgfqpoint{0.000000in}{0.000000in}}{%
\pgfpathmoveto{\pgfqpoint{0.000000in}{0.000000in}}%
\pgfpathlineto{\pgfqpoint{-0.044444in}{0.000000in}}%
\pgfusepath{stroke,fill}%
}%
\begin{pgfscope}%
\pgfsys@transformshift{0.450809in}{0.749889in}%
\pgfsys@useobject{currentmarker}{}%
\end{pgfscope}%
\end{pgfscope}%
\begin{pgfscope}%
\pgfsetbuttcap%
\pgfsetroundjoin%
\definecolor{currentfill}{rgb}{0.150000,0.150000,0.150000}%
\pgfsetfillcolor{currentfill}%
\pgfsetlinewidth{0.803000pt}%
\definecolor{currentstroke}{rgb}{0.150000,0.150000,0.150000}%
\pgfsetstrokecolor{currentstroke}%
\pgfsetdash{}{0pt}%
\pgfsys@defobject{currentmarker}{\pgfqpoint{-0.044444in}{0.000000in}}{\pgfqpoint{0.000000in}{0.000000in}}{%
\pgfpathmoveto{\pgfqpoint{0.000000in}{0.000000in}}%
\pgfpathlineto{\pgfqpoint{-0.044444in}{0.000000in}}%
\pgfusepath{stroke,fill}%
}%
\begin{pgfscope}%
\pgfsys@transformshift{0.450809in}{1.173389in}%
\pgfsys@useobject{currentmarker}{}%
\end{pgfscope}%
\end{pgfscope}%
\begin{pgfscope}%
\pgfsetbuttcap%
\pgfsetroundjoin%
\definecolor{currentfill}{rgb}{0.150000,0.150000,0.150000}%
\pgfsetfillcolor{currentfill}%
\pgfsetlinewidth{0.803000pt}%
\definecolor{currentstroke}{rgb}{0.150000,0.150000,0.150000}%
\pgfsetstrokecolor{currentstroke}%
\pgfsetdash{}{0pt}%
\pgfsys@defobject{currentmarker}{\pgfqpoint{-0.044444in}{0.000000in}}{\pgfqpoint{0.000000in}{0.000000in}}{%
\pgfpathmoveto{\pgfqpoint{0.000000in}{0.000000in}}%
\pgfpathlineto{\pgfqpoint{-0.044444in}{0.000000in}}%
\pgfusepath{stroke,fill}%
}%
\begin{pgfscope}%
\pgfsys@transformshift{0.450809in}{1.596889in}%
\pgfsys@useobject{currentmarker}{}%
\end{pgfscope}%
\end{pgfscope}%
\begin{pgfscope}%
\pgfsetbuttcap%
\pgfsetroundjoin%
\definecolor{currentfill}{rgb}{0.150000,0.150000,0.150000}%
\pgfsetfillcolor{currentfill}%
\pgfsetlinewidth{0.803000pt}%
\definecolor{currentstroke}{rgb}{0.150000,0.150000,0.150000}%
\pgfsetstrokecolor{currentstroke}%
\pgfsetdash{}{0pt}%
\pgfsys@defobject{currentmarker}{\pgfqpoint{-0.044444in}{0.000000in}}{\pgfqpoint{0.000000in}{0.000000in}}{%
\pgfpathmoveto{\pgfqpoint{0.000000in}{0.000000in}}%
\pgfpathlineto{\pgfqpoint{-0.044444in}{0.000000in}}%
\pgfusepath{stroke,fill}%
}%
\begin{pgfscope}%
\pgfsys@transformshift{0.450809in}{2.020389in}%
\pgfsys@useobject{currentmarker}{}%
\end{pgfscope}%
\end{pgfscope}%
\begin{pgfscope}%
\pgfpathrectangle{\pgfqpoint{0.450809in}{0.326389in}}{\pgfqpoint{2.480000in}{1.694000in}}%
\pgfusepath{clip}%
\pgfsetbuttcap%
\pgfsetroundjoin%
\pgfsetlinewidth{1.204500pt}%
\definecolor{currentstroke}{rgb}{1.000000,0.498039,0.000000}%
\pgfsetstrokecolor{currentstroke}%
\pgfsetdash{{7.680000pt}{1.920000pt}{1.200000pt}{1.920000pt}}{0.000000pt}%
\pgfpathmoveto{\pgfqpoint{0.563536in}{1.576896in}}%
\pgfpathlineto{\pgfqpoint{0.937679in}{1.624862in}}%
\pgfpathlineto{\pgfqpoint{1.220708in}{1.659965in}}%
\pgfpathlineto{\pgfqpoint{1.503737in}{1.673613in}}%
\pgfpathlineto{\pgfqpoint{1.877880in}{1.725199in}}%
\pgfpathlineto{\pgfqpoint{2.160909in}{1.747940in}}%
\pgfpathlineto{\pgfqpoint{2.443938in}{1.797856in}}%
\pgfpathlineto{\pgfqpoint{2.818081in}{1.895066in}}%
\pgfusepath{stroke}%
\end{pgfscope}%
\begin{pgfscope}%
\pgfpathrectangle{\pgfqpoint{0.450809in}{0.326389in}}{\pgfqpoint{2.480000in}{1.694000in}}%
\pgfusepath{clip}%
\pgfsetroundcap%
\pgfsetroundjoin%
\pgfsetlinewidth{1.204500pt}%
\definecolor{currentstroke}{rgb}{0.890196,0.101961,0.109804}%
\pgfsetstrokecolor{currentstroke}%
\pgfsetdash{}{0pt}%
\pgfpathmoveto{\pgfqpoint{0.563536in}{1.368045in}}%
\pgfpathlineto{\pgfqpoint{0.937679in}{1.551866in}}%
\pgfpathlineto{\pgfqpoint{1.220708in}{1.659681in}}%
\pgfpathlineto{\pgfqpoint{1.503737in}{1.703658in}}%
\pgfpathlineto{\pgfqpoint{1.877880in}{1.783653in}}%
\pgfpathlineto{\pgfqpoint{2.160909in}{1.814024in}}%
\pgfpathlineto{\pgfqpoint{2.443938in}{1.865258in}}%
\pgfpathlineto{\pgfqpoint{2.818081in}{1.932785in}}%
\pgfusepath{stroke}%
\end{pgfscope}%
\begin{pgfscope}%
\pgfpathrectangle{\pgfqpoint{0.450809in}{0.326389in}}{\pgfqpoint{2.480000in}{1.694000in}}%
\pgfusepath{clip}%
\pgfsetbuttcap%
\pgfsetroundjoin%
\pgfsetlinewidth{1.204500pt}%
\definecolor{currentstroke}{rgb}{0.200000,0.627451,0.172549}%
\pgfsetstrokecolor{currentstroke}%
\pgfsetdash{{4.440000pt}{1.920000pt}}{0.000000pt}%
\pgfpathmoveto{\pgfqpoint{0.563536in}{0.644668in}}%
\pgfpathlineto{\pgfqpoint{0.937679in}{0.935192in}}%
\pgfpathlineto{\pgfqpoint{1.220708in}{1.198680in}}%
\pgfpathlineto{\pgfqpoint{1.503737in}{1.397959in}}%
\pgfpathlineto{\pgfqpoint{1.877880in}{1.607376in}}%
\pgfpathlineto{\pgfqpoint{2.160909in}{1.687240in}}%
\pgfpathlineto{\pgfqpoint{2.443938in}{1.782563in}}%
\pgfpathlineto{\pgfqpoint{2.818081in}{1.887980in}}%
\pgfusepath{stroke}%
\end{pgfscope}%
\begin{pgfscope}%
\pgfpathrectangle{\pgfqpoint{0.450809in}{0.326389in}}{\pgfqpoint{2.480000in}{1.694000in}}%
\pgfusepath{clip}%
\pgfsetbuttcap%
\pgfsetroundjoin%
\pgfsetlinewidth{1.204500pt}%
\definecolor{currentstroke}{rgb}{0.121569,0.470588,0.705882}%
\pgfsetstrokecolor{currentstroke}%
\pgfsetdash{{1.200000pt}{1.980000pt}}{0.000000pt}%
\pgfpathmoveto{\pgfqpoint{0.563536in}{0.838888in}}%
\pgfpathlineto{\pgfqpoint{0.937679in}{1.142516in}}%
\pgfpathlineto{\pgfqpoint{1.220708in}{1.343866in}}%
\pgfpathlineto{\pgfqpoint{1.503737in}{1.496683in}}%
\pgfpathlineto{\pgfqpoint{1.877880in}{1.649543in}}%
\pgfpathlineto{\pgfqpoint{2.160909in}{1.719269in}}%
\pgfpathlineto{\pgfqpoint{2.443938in}{1.793339in}}%
\pgfpathlineto{\pgfqpoint{2.818081in}{1.898336in}}%
\pgfusepath{stroke}%
\end{pgfscope}%
\begin{pgfscope}%
\pgfsetrectcap%
\pgfsetmiterjoin%
\pgfsetlinewidth{1.003750pt}%
\definecolor{currentstroke}{rgb}{0.150000,0.150000,0.150000}%
\pgfsetstrokecolor{currentstroke}%
\pgfsetdash{}{0pt}%
\pgfpathmoveto{\pgfqpoint{0.450809in}{0.326389in}}%
\pgfpathlineto{\pgfqpoint{0.450809in}{2.020389in}}%
\pgfusepath{stroke}%
\end{pgfscope}%
\begin{pgfscope}%
\pgfsetrectcap%
\pgfsetmiterjoin%
\pgfsetlinewidth{1.003750pt}%
\definecolor{currentstroke}{rgb}{0.150000,0.150000,0.150000}%
\pgfsetstrokecolor{currentstroke}%
\pgfsetdash{}{0pt}%
\pgfpathmoveto{\pgfqpoint{2.930809in}{0.326389in}}%
\pgfpathlineto{\pgfqpoint{2.930809in}{2.020389in}}%
\pgfusepath{stroke}%
\end{pgfscope}%
\begin{pgfscope}%
\pgfsetrectcap%
\pgfsetmiterjoin%
\pgfsetlinewidth{1.003750pt}%
\definecolor{currentstroke}{rgb}{0.150000,0.150000,0.150000}%
\pgfsetstrokecolor{currentstroke}%
\pgfsetdash{}{0pt}%
\pgfpathmoveto{\pgfqpoint{0.450809in}{0.326389in}}%
\pgfpathlineto{\pgfqpoint{2.930809in}{0.326389in}}%
\pgfusepath{stroke}%
\end{pgfscope}%
\begin{pgfscope}%
\pgfsetrectcap%
\pgfsetmiterjoin%
\pgfsetlinewidth{1.003750pt}%
\definecolor{currentstroke}{rgb}{0.150000,0.150000,0.150000}%
\pgfsetstrokecolor{currentstroke}%
\pgfsetdash{}{0pt}%
\pgfpathmoveto{\pgfqpoint{0.450809in}{2.020389in}}%
\pgfpathlineto{\pgfqpoint{2.930809in}{2.020389in}}%
\pgfusepath{stroke}%
\end{pgfscope}%
\begin{pgfscope}%
\pgfsetbuttcap%
\pgfsetroundjoin%
\pgfsetlinewidth{1.204500pt}%
\definecolor{currentstroke}{rgb}{1.000000,0.498039,0.000000}%
\pgfsetstrokecolor{currentstroke}%
\pgfsetdash{{7.680000pt}{1.920000pt}{1.200000pt}{1.920000pt}}{0.000000pt}%
\pgfpathmoveto{\pgfqpoint{1.816819in}{0.971250in}}%
\pgfpathlineto{\pgfqpoint{2.061263in}{0.971250in}}%
\pgfusepath{stroke}%
\end{pgfscope}%
\begin{pgfscope}%
\definecolor{textcolor}{rgb}{0.150000,0.150000,0.150000}%
\pgfsetstrokecolor{textcolor}%
\pgfsetfillcolor{textcolor}%
\pgftext[x=2.159041in,y=0.928472in,left,base]{\color{textcolor}\sffamily\fontsize{8.800000}{10.560000}\selectfont Norma}%
\end{pgfscope}%
\begin{pgfscope}%
\pgfsetroundcap%
\pgfsetroundjoin%
\pgfsetlinewidth{1.204500pt}%
\definecolor{currentstroke}{rgb}{0.890196,0.101961,0.109804}%
\pgfsetstrokecolor{currentstroke}%
\pgfsetdash{}{0pt}%
\pgfpathmoveto{\pgfqpoint{1.816819in}{0.799028in}}%
\pgfpathlineto{\pgfqpoint{2.061263in}{0.799028in}}%
\pgfusepath{stroke}%
\end{pgfscope}%
\begin{pgfscope}%
\definecolor{textcolor}{rgb}{0.150000,0.150000,0.150000}%
\pgfsetstrokecolor{textcolor}%
\pgfsetfillcolor{textcolor}%
\pgftext[x=2.159041in,y=0.756250in,left,base]{\color{textcolor}\sffamily\fontsize{8.800000}{10.560000}\selectfont cSMTiser\textsubscript{+LM}}%
\end{pgfscope}%
\begin{pgfscope}%
\pgfsetbuttcap%
\pgfsetroundjoin%
\pgfsetlinewidth{1.204500pt}%
\definecolor{currentstroke}{rgb}{0.200000,0.627451,0.172549}%
\pgfsetstrokecolor{currentstroke}%
\pgfsetdash{{4.440000pt}{1.920000pt}}{0.000000pt}%
\pgfpathmoveto{\pgfqpoint{1.816819in}{0.626806in}}%
\pgfpathlineto{\pgfqpoint{2.061263in}{0.626806in}}%
\pgfusepath{stroke}%
\end{pgfscope}%
\begin{pgfscope}%
\definecolor{textcolor}{rgb}{0.150000,0.150000,0.150000}%
\pgfsetstrokecolor{textcolor}%
\pgfsetfillcolor{textcolor}%
\pgftext[x=2.159041in,y=0.584028in,left,base]{\color{textcolor}\sffamily\fontsize{8.800000}{10.560000}\selectfont NMT-1}%
\end{pgfscope}%
\begin{pgfscope}%
\pgfsetbuttcap%
\pgfsetroundjoin%
\pgfsetlinewidth{1.204500pt}%
\definecolor{currentstroke}{rgb}{0.121569,0.470588,0.705882}%
\pgfsetstrokecolor{currentstroke}%
\pgfsetdash{{1.200000pt}{1.980000pt}}{0.000000pt}%
\pgfpathmoveto{\pgfqpoint{1.816819in}{0.454583in}}%
\pgfpathlineto{\pgfqpoint{2.061263in}{0.454583in}}%
\pgfusepath{stroke}%
\end{pgfscope}%
\begin{pgfscope}%
\definecolor{textcolor}{rgb}{0.150000,0.150000,0.150000}%
\pgfsetstrokecolor{textcolor}%
\pgfsetfillcolor{textcolor}%
\pgftext[x=2.159041in,y=0.411806in,left,base]{\color{textcolor}\sffamily\fontsize{8.800000}{10.560000}\selectfont NMT-2}%
\end{pgfscope}%
\end{pgfpicture}%
\makeatother%
\endgroup%

%% file: curves_spanish-ps.pgf
\begingroup%
\makeatletter%
\begin{pgfpicture}%
\pgfpathrectangle{\pgfpointorigin}{\pgfqpoint{3.078809in}{2.168389in}}%
\pgfusepath{use as bounding box, clip}%
\begin{pgfscope}%
\pgfsetbuttcap%
\pgfsetmiterjoin%
\definecolor{currentfill}{rgb}{1.000000,1.000000,1.000000}%
\pgfsetfillcolor{currentfill}%
\pgfsetlinewidth{0.000000pt}%
\definecolor{currentstroke}{rgb}{1.000000,1.000000,1.000000}%
\pgfsetstrokecolor{currentstroke}%
\pgfsetdash{}{0pt}%
\pgfpathmoveto{\pgfqpoint{0.000000in}{0.000000in}}%
\pgfpathlineto{\pgfqpoint{3.078809in}{0.000000in}}%
\pgfpathlineto{\pgfqpoint{3.078809in}{2.168389in}}%
\pgfpathlineto{\pgfqpoint{0.000000in}{2.168389in}}%
\pgfpathclose%
\pgfusepath{fill}%
\end{pgfscope}%
\begin{pgfscope}%
\pgfsetbuttcap%
\pgfsetmiterjoin%
\definecolor{currentfill}{rgb}{1.000000,1.000000,1.000000}%
\pgfsetfillcolor{currentfill}%
\pgfsetlinewidth{0.000000pt}%
\definecolor{currentstroke}{rgb}{0.000000,0.000000,0.000000}%
\pgfsetstrokecolor{currentstroke}%
\pgfsetstrokeopacity{0.000000}%
\pgfsetdash{}{0pt}%
\pgfpathmoveto{\pgfqpoint{0.450809in}{0.326389in}}%
\pgfpathlineto{\pgfqpoint{2.930809in}{0.326389in}}%
\pgfpathlineto{\pgfqpoint{2.930809in}{2.020389in}}%
\pgfpathlineto{\pgfqpoint{0.450809in}{2.020389in}}%
\pgfpathclose%
\pgfusepath{fill}%
\end{pgfscope}%
\begin{pgfscope}%
\pgfsetbuttcap%
\pgfsetroundjoin%
\definecolor{currentfill}{rgb}{0.150000,0.150000,0.150000}%
\pgfsetfillcolor{currentfill}%
\pgfsetlinewidth{1.003750pt}%
\definecolor{currentstroke}{rgb}{0.150000,0.150000,0.150000}%
\pgfsetstrokecolor{currentstroke}%
\pgfsetdash{}{0pt}%
\pgfsys@defobject{currentmarker}{\pgfqpoint{0.000000in}{-0.066667in}}{\pgfqpoint{0.000000in}{0.000000in}}{%
\pgfpathmoveto{\pgfqpoint{0.000000in}{0.000000in}}%
\pgfpathlineto{\pgfqpoint{0.000000in}{-0.066667in}}%
\pgfusepath{stroke,fill}%
}%
\begin{pgfscope}%
\pgfsys@transformshift{0.563536in}{0.326389in}%
\pgfsys@useobject{currentmarker}{}%
\end{pgfscope}%
\end{pgfscope}%
\begin{pgfscope}%
\definecolor{textcolor}{rgb}{0.150000,0.150000,0.150000}%
\pgfsetstrokecolor{textcolor}%
\pgfsetfillcolor{textcolor}%
\pgftext[x=0.563536in,y=0.211111in,,top]{\color{textcolor}\sffamily\fontsize{8.800000}{10.560000}\selectfont 100}%
\end{pgfscope}%
\begin{pgfscope}%
\pgfsetbuttcap%
\pgfsetroundjoin%
\definecolor{currentfill}{rgb}{0.150000,0.150000,0.150000}%
\pgfsetfillcolor{currentfill}%
\pgfsetlinewidth{1.003750pt}%
\definecolor{currentstroke}{rgb}{0.150000,0.150000,0.150000}%
\pgfsetstrokecolor{currentstroke}%
\pgfsetdash{}{0pt}%
\pgfsys@defobject{currentmarker}{\pgfqpoint{0.000000in}{-0.066667in}}{\pgfqpoint{0.000000in}{0.000000in}}{%
\pgfpathmoveto{\pgfqpoint{0.000000in}{0.000000in}}%
\pgfpathlineto{\pgfqpoint{0.000000in}{-0.066667in}}%
\pgfusepath{stroke,fill}%
}%
\begin{pgfscope}%
\pgfsys@transformshift{0.895949in}{0.326389in}%
\pgfsys@useobject{currentmarker}{}%
\end{pgfscope}%
\end{pgfscope}%
\begin{pgfscope}%
\definecolor{textcolor}{rgb}{0.150000,0.150000,0.150000}%
\pgfsetstrokecolor{textcolor}%
\pgfsetfillcolor{textcolor}%
\pgftext[x=0.895949in,y=0.211111in,,top]{\color{textcolor}\sffamily\fontsize{8.800000}{10.560000}\selectfont 250}%
\end{pgfscope}%
\begin{pgfscope}%
\pgfsetbuttcap%
\pgfsetroundjoin%
\definecolor{currentfill}{rgb}{0.150000,0.150000,0.150000}%
\pgfsetfillcolor{currentfill}%
\pgfsetlinewidth{1.003750pt}%
\definecolor{currentstroke}{rgb}{0.150000,0.150000,0.150000}%
\pgfsetstrokecolor{currentstroke}%
\pgfsetdash{}{0pt}%
\pgfsys@defobject{currentmarker}{\pgfqpoint{0.000000in}{-0.066667in}}{\pgfqpoint{0.000000in}{0.000000in}}{%
\pgfpathmoveto{\pgfqpoint{0.000000in}{0.000000in}}%
\pgfpathlineto{\pgfqpoint{0.000000in}{-0.066667in}}%
\pgfusepath{stroke,fill}%
}%
\begin{pgfscope}%
\pgfsys@transformshift{1.147410in}{0.326389in}%
\pgfsys@useobject{currentmarker}{}%
\end{pgfscope}%
\end{pgfscope}%
\begin{pgfscope}%
\definecolor{textcolor}{rgb}{0.150000,0.150000,0.150000}%
\pgfsetstrokecolor{textcolor}%
\pgfsetfillcolor{textcolor}%
\pgftext[x=1.147410in,y=0.211111in,,top]{\color{textcolor}\sffamily\fontsize{8.800000}{10.560000}\selectfont 500}%
\end{pgfscope}%
\begin{pgfscope}%
\pgfsetbuttcap%
\pgfsetroundjoin%
\definecolor{currentfill}{rgb}{0.150000,0.150000,0.150000}%
\pgfsetfillcolor{currentfill}%
\pgfsetlinewidth{1.003750pt}%
\definecolor{currentstroke}{rgb}{0.150000,0.150000,0.150000}%
\pgfsetstrokecolor{currentstroke}%
\pgfsetdash{}{0pt}%
\pgfsys@defobject{currentmarker}{\pgfqpoint{0.000000in}{-0.066667in}}{\pgfqpoint{0.000000in}{0.000000in}}{%
\pgfpathmoveto{\pgfqpoint{0.000000in}{0.000000in}}%
\pgfpathlineto{\pgfqpoint{0.000000in}{-0.066667in}}%
\pgfusepath{stroke,fill}%
}%
\begin{pgfscope}%
\pgfsys@transformshift{1.398871in}{0.326389in}%
\pgfsys@useobject{currentmarker}{}%
\end{pgfscope}%
\end{pgfscope}%
\begin{pgfscope}%
\definecolor{textcolor}{rgb}{0.150000,0.150000,0.150000}%
\pgfsetstrokecolor{textcolor}%
\pgfsetfillcolor{textcolor}%
\pgftext[x=1.398871in,y=0.211111in,,top]{\color{textcolor}\sffamily\fontsize{8.800000}{10.560000}\selectfont 1k}%
\end{pgfscope}%
\begin{pgfscope}%
\pgfsetbuttcap%
\pgfsetroundjoin%
\definecolor{currentfill}{rgb}{0.150000,0.150000,0.150000}%
\pgfsetfillcolor{currentfill}%
\pgfsetlinewidth{1.003750pt}%
\definecolor{currentstroke}{rgb}{0.150000,0.150000,0.150000}%
\pgfsetstrokecolor{currentstroke}%
\pgfsetdash{}{0pt}%
\pgfsys@defobject{currentmarker}{\pgfqpoint{0.000000in}{-0.066667in}}{\pgfqpoint{0.000000in}{0.000000in}}{%
\pgfpathmoveto{\pgfqpoint{0.000000in}{0.000000in}}%
\pgfpathlineto{\pgfqpoint{0.000000in}{-0.066667in}}%
\pgfusepath{stroke,fill}%
}%
\begin{pgfscope}%
\pgfsys@transformshift{1.731285in}{0.326389in}%
\pgfsys@useobject{currentmarker}{}%
\end{pgfscope}%
\end{pgfscope}%
\begin{pgfscope}%
\definecolor{textcolor}{rgb}{0.150000,0.150000,0.150000}%
\pgfsetstrokecolor{textcolor}%
\pgfsetfillcolor{textcolor}%
\pgftext[x=1.731285in,y=0.211111in,,top]{\color{textcolor}\sffamily\fontsize{8.800000}{10.560000}\selectfont 2.5k}%
\end{pgfscope}%
\begin{pgfscope}%
\pgfsetbuttcap%
\pgfsetroundjoin%
\definecolor{currentfill}{rgb}{0.150000,0.150000,0.150000}%
\pgfsetfillcolor{currentfill}%
\pgfsetlinewidth{1.003750pt}%
\definecolor{currentstroke}{rgb}{0.150000,0.150000,0.150000}%
\pgfsetstrokecolor{currentstroke}%
\pgfsetdash{}{0pt}%
\pgfsys@defobject{currentmarker}{\pgfqpoint{0.000000in}{-0.066667in}}{\pgfqpoint{0.000000in}{0.000000in}}{%
\pgfpathmoveto{\pgfqpoint{0.000000in}{0.000000in}}%
\pgfpathlineto{\pgfqpoint{0.000000in}{-0.066667in}}%
\pgfusepath{stroke,fill}%
}%
\begin{pgfscope}%
\pgfsys@transformshift{1.982746in}{0.326389in}%
\pgfsys@useobject{currentmarker}{}%
\end{pgfscope}%
\end{pgfscope}%
\begin{pgfscope}%
\definecolor{textcolor}{rgb}{0.150000,0.150000,0.150000}%
\pgfsetstrokecolor{textcolor}%
\pgfsetfillcolor{textcolor}%
\pgftext[x=1.982746in,y=0.211111in,,top]{\color{textcolor}\sffamily\fontsize{8.800000}{10.560000}\selectfont 5k}%
\end{pgfscope}%
\begin{pgfscope}%
\pgfsetbuttcap%
\pgfsetroundjoin%
\definecolor{currentfill}{rgb}{0.150000,0.150000,0.150000}%
\pgfsetfillcolor{currentfill}%
\pgfsetlinewidth{1.003750pt}%
\definecolor{currentstroke}{rgb}{0.150000,0.150000,0.150000}%
\pgfsetstrokecolor{currentstroke}%
\pgfsetdash{}{0pt}%
\pgfsys@defobject{currentmarker}{\pgfqpoint{0.000000in}{-0.066667in}}{\pgfqpoint{0.000000in}{0.000000in}}{%
\pgfpathmoveto{\pgfqpoint{0.000000in}{0.000000in}}%
\pgfpathlineto{\pgfqpoint{0.000000in}{-0.066667in}}%
\pgfusepath{stroke,fill}%
}%
\begin{pgfscope}%
\pgfsys@transformshift{2.234207in}{0.326389in}%
\pgfsys@useobject{currentmarker}{}%
\end{pgfscope}%
\end{pgfscope}%
\begin{pgfscope}%
\definecolor{textcolor}{rgb}{0.150000,0.150000,0.150000}%
\pgfsetstrokecolor{textcolor}%
\pgfsetfillcolor{textcolor}%
\pgftext[x=2.234207in,y=0.211111in,,top]{\color{textcolor}\sffamily\fontsize{8.800000}{10.560000}\selectfont 10k}%
\end{pgfscope}%
\begin{pgfscope}%
\pgfsetbuttcap%
\pgfsetroundjoin%
\definecolor{currentfill}{rgb}{0.150000,0.150000,0.150000}%
\pgfsetfillcolor{currentfill}%
\pgfsetlinewidth{1.003750pt}%
\definecolor{currentstroke}{rgb}{0.150000,0.150000,0.150000}%
\pgfsetstrokecolor{currentstroke}%
\pgfsetdash{}{0pt}%
\pgfsys@defobject{currentmarker}{\pgfqpoint{0.000000in}{-0.066667in}}{\pgfqpoint{0.000000in}{0.000000in}}{%
\pgfpathmoveto{\pgfqpoint{0.000000in}{0.000000in}}%
\pgfpathlineto{\pgfqpoint{0.000000in}{-0.066667in}}%
\pgfusepath{stroke,fill}%
}%
\begin{pgfscope}%
\pgfsys@transformshift{2.566620in}{0.326389in}%
\pgfsys@useobject{currentmarker}{}%
\end{pgfscope}%
\end{pgfscope}%
\begin{pgfscope}%
\definecolor{textcolor}{rgb}{0.150000,0.150000,0.150000}%
\pgfsetstrokecolor{textcolor}%
\pgfsetfillcolor{textcolor}%
\pgftext[x=2.566620in,y=0.211111in,,top]{\color{textcolor}\sffamily\fontsize{8.800000}{10.560000}\selectfont 25k}%
\end{pgfscope}%
\begin{pgfscope}%
\pgfsetbuttcap%
\pgfsetroundjoin%
\definecolor{currentfill}{rgb}{0.150000,0.150000,0.150000}%
\pgfsetfillcolor{currentfill}%
\pgfsetlinewidth{1.003750pt}%
\definecolor{currentstroke}{rgb}{0.150000,0.150000,0.150000}%
\pgfsetstrokecolor{currentstroke}%
\pgfsetdash{}{0pt}%
\pgfsys@defobject{currentmarker}{\pgfqpoint{0.000000in}{-0.066667in}}{\pgfqpoint{0.000000in}{0.000000in}}{%
\pgfpathmoveto{\pgfqpoint{0.000000in}{0.000000in}}%
\pgfpathlineto{\pgfqpoint{0.000000in}{-0.066667in}}%
\pgfusepath{stroke,fill}%
}%
\begin{pgfscope}%
\pgfsys@transformshift{2.818081in}{0.326389in}%
\pgfsys@useobject{currentmarker}{}%
\end{pgfscope}%
\end{pgfscope}%
\begin{pgfscope}%
\definecolor{textcolor}{rgb}{0.150000,0.150000,0.150000}%
\pgfsetstrokecolor{textcolor}%
\pgfsetfillcolor{textcolor}%
\pgftext[x=2.818081in,y=0.211111in,,top]{\color{textcolor}\sffamily\fontsize{8.800000}{10.560000}\selectfont 50k}%
\end{pgfscope}%
\begin{pgfscope}%
\pgfsetbuttcap%
\pgfsetroundjoin%
\definecolor{currentfill}{rgb}{0.150000,0.150000,0.150000}%
\pgfsetfillcolor{currentfill}%
\pgfsetlinewidth{1.003750pt}%
\definecolor{currentstroke}{rgb}{0.150000,0.150000,0.150000}%
\pgfsetstrokecolor{currentstroke}%
\pgfsetdash{}{0pt}%
\pgfsys@defobject{currentmarker}{\pgfqpoint{-0.066667in}{0.000000in}}{\pgfqpoint{0.000000in}{0.000000in}}{%
\pgfpathmoveto{\pgfqpoint{0.000000in}{0.000000in}}%
\pgfpathlineto{\pgfqpoint{-0.066667in}{0.000000in}}%
\pgfusepath{stroke,fill}%
}%
\begin{pgfscope}%
\pgfsys@transformshift{0.450809in}{0.520870in}%
\pgfsys@useobject{currentmarker}{}%
\end{pgfscope}%
\end{pgfscope}%
\begin{pgfscope}%
\definecolor{textcolor}{rgb}{0.150000,0.150000,0.150000}%
\pgfsetstrokecolor{textcolor}%
\pgfsetfillcolor{textcolor}%
\pgftext[x=0.100000in,y=0.477467in,left,base]{\color{textcolor}\sffamily\fontsize{8.800000}{10.560000}\selectfont 20\%}%
\end{pgfscope}%
\begin{pgfscope}%
\pgfsetbuttcap%
\pgfsetroundjoin%
\definecolor{currentfill}{rgb}{0.150000,0.150000,0.150000}%
\pgfsetfillcolor{currentfill}%
\pgfsetlinewidth{1.003750pt}%
\definecolor{currentstroke}{rgb}{0.150000,0.150000,0.150000}%
\pgfsetstrokecolor{currentstroke}%
\pgfsetdash{}{0pt}%
\pgfsys@defobject{currentmarker}{\pgfqpoint{-0.066667in}{0.000000in}}{\pgfqpoint{0.000000in}{0.000000in}}{%
\pgfpathmoveto{\pgfqpoint{0.000000in}{0.000000in}}%
\pgfpathlineto{\pgfqpoint{-0.066667in}{0.000000in}}%
\pgfusepath{stroke,fill}%
}%
\begin{pgfscope}%
\pgfsys@transformshift{0.450809in}{0.909831in}%
\pgfsys@useobject{currentmarker}{}%
\end{pgfscope}%
\end{pgfscope}%
\begin{pgfscope}%
\definecolor{textcolor}{rgb}{0.150000,0.150000,0.150000}%
\pgfsetstrokecolor{textcolor}%
\pgfsetfillcolor{textcolor}%
\pgftext[x=0.100000in,y=0.866429in,left,base]{\color{textcolor}\sffamily\fontsize{8.800000}{10.560000}\selectfont 40\%}%
\end{pgfscope}%
\begin{pgfscope}%
\pgfsetbuttcap%
\pgfsetroundjoin%
\definecolor{currentfill}{rgb}{0.150000,0.150000,0.150000}%
\pgfsetfillcolor{currentfill}%
\pgfsetlinewidth{1.003750pt}%
\definecolor{currentstroke}{rgb}{0.150000,0.150000,0.150000}%
\pgfsetstrokecolor{currentstroke}%
\pgfsetdash{}{0pt}%
\pgfsys@defobject{currentmarker}{\pgfqpoint{-0.066667in}{0.000000in}}{\pgfqpoint{0.000000in}{0.000000in}}{%
\pgfpathmoveto{\pgfqpoint{0.000000in}{0.000000in}}%
\pgfpathlineto{\pgfqpoint{-0.066667in}{0.000000in}}%
\pgfusepath{stroke,fill}%
}%
\begin{pgfscope}%
\pgfsys@transformshift{0.450809in}{1.298793in}%
\pgfsys@useobject{currentmarker}{}%
\end{pgfscope}%
\end{pgfscope}%
\begin{pgfscope}%
\definecolor{textcolor}{rgb}{0.150000,0.150000,0.150000}%
\pgfsetstrokecolor{textcolor}%
\pgfsetfillcolor{textcolor}%
\pgftext[x=0.100000in,y=1.255390in,left,base]{\color{textcolor}\sffamily\fontsize{8.800000}{10.560000}\selectfont 60\%}%
\end{pgfscope}%
\begin{pgfscope}%
\pgfsetbuttcap%
\pgfsetroundjoin%
\definecolor{currentfill}{rgb}{0.150000,0.150000,0.150000}%
\pgfsetfillcolor{currentfill}%
\pgfsetlinewidth{1.003750pt}%
\definecolor{currentstroke}{rgb}{0.150000,0.150000,0.150000}%
\pgfsetstrokecolor{currentstroke}%
\pgfsetdash{}{0pt}%
\pgfsys@defobject{currentmarker}{\pgfqpoint{-0.066667in}{0.000000in}}{\pgfqpoint{0.000000in}{0.000000in}}{%
\pgfpathmoveto{\pgfqpoint{0.000000in}{0.000000in}}%
\pgfpathlineto{\pgfqpoint{-0.066667in}{0.000000in}}%
\pgfusepath{stroke,fill}%
}%
\begin{pgfscope}%
\pgfsys@transformshift{0.450809in}{1.687755in}%
\pgfsys@useobject{currentmarker}{}%
\end{pgfscope}%
\end{pgfscope}%
\begin{pgfscope}%
\definecolor{textcolor}{rgb}{0.150000,0.150000,0.150000}%
\pgfsetstrokecolor{textcolor}%
\pgfsetfillcolor{textcolor}%
\pgftext[x=0.100000in,y=1.644352in,left,base]{\color{textcolor}\sffamily\fontsize{8.800000}{10.560000}\selectfont 80\%}%
\end{pgfscope}%
\begin{pgfscope}%
\pgfsetbuttcap%
\pgfsetroundjoin%
\definecolor{currentfill}{rgb}{0.150000,0.150000,0.150000}%
\pgfsetfillcolor{currentfill}%
\pgfsetlinewidth{0.803000pt}%
\definecolor{currentstroke}{rgb}{0.150000,0.150000,0.150000}%
\pgfsetstrokecolor{currentstroke}%
\pgfsetdash{}{0pt}%
\pgfsys@defobject{currentmarker}{\pgfqpoint{-0.044444in}{0.000000in}}{\pgfqpoint{0.000000in}{0.000000in}}{%
\pgfpathmoveto{\pgfqpoint{0.000000in}{0.000000in}}%
\pgfpathlineto{\pgfqpoint{-0.044444in}{0.000000in}}%
\pgfusepath{stroke,fill}%
}%
\begin{pgfscope}%
\pgfsys@transformshift{0.450809in}{0.326389in}%
\pgfsys@useobject{currentmarker}{}%
\end{pgfscope}%
\end{pgfscope}%
\begin{pgfscope}%
\pgfsetbuttcap%
\pgfsetroundjoin%
\definecolor{currentfill}{rgb}{0.150000,0.150000,0.150000}%
\pgfsetfillcolor{currentfill}%
\pgfsetlinewidth{0.803000pt}%
\definecolor{currentstroke}{rgb}{0.150000,0.150000,0.150000}%
\pgfsetstrokecolor{currentstroke}%
\pgfsetdash{}{0pt}%
\pgfsys@defobject{currentmarker}{\pgfqpoint{-0.044444in}{0.000000in}}{\pgfqpoint{0.000000in}{0.000000in}}{%
\pgfpathmoveto{\pgfqpoint{0.000000in}{0.000000in}}%
\pgfpathlineto{\pgfqpoint{-0.044444in}{0.000000in}}%
\pgfusepath{stroke,fill}%
}%
\begin{pgfscope}%
\pgfsys@transformshift{0.450809in}{0.715351in}%
\pgfsys@useobject{currentmarker}{}%
\end{pgfscope}%
\end{pgfscope}%
\begin{pgfscope}%
\pgfsetbuttcap%
\pgfsetroundjoin%
\definecolor{currentfill}{rgb}{0.150000,0.150000,0.150000}%
\pgfsetfillcolor{currentfill}%
\pgfsetlinewidth{0.803000pt}%
\definecolor{currentstroke}{rgb}{0.150000,0.150000,0.150000}%
\pgfsetstrokecolor{currentstroke}%
\pgfsetdash{}{0pt}%
\pgfsys@defobject{currentmarker}{\pgfqpoint{-0.044444in}{0.000000in}}{\pgfqpoint{0.000000in}{0.000000in}}{%
\pgfpathmoveto{\pgfqpoint{0.000000in}{0.000000in}}%
\pgfpathlineto{\pgfqpoint{-0.044444in}{0.000000in}}%
\pgfusepath{stroke,fill}%
}%
\begin{pgfscope}%
\pgfsys@transformshift{0.450809in}{1.104312in}%
\pgfsys@useobject{currentmarker}{}%
\end{pgfscope}%
\end{pgfscope}%
\begin{pgfscope}%
\pgfsetbuttcap%
\pgfsetroundjoin%
\definecolor{currentfill}{rgb}{0.150000,0.150000,0.150000}%
\pgfsetfillcolor{currentfill}%
\pgfsetlinewidth{0.803000pt}%
\definecolor{currentstroke}{rgb}{0.150000,0.150000,0.150000}%
\pgfsetstrokecolor{currentstroke}%
\pgfsetdash{}{0pt}%
\pgfsys@defobject{currentmarker}{\pgfqpoint{-0.044444in}{0.000000in}}{\pgfqpoint{0.000000in}{0.000000in}}{%
\pgfpathmoveto{\pgfqpoint{0.000000in}{0.000000in}}%
\pgfpathlineto{\pgfqpoint{-0.044444in}{0.000000in}}%
\pgfusepath{stroke,fill}%
}%
\begin{pgfscope}%
\pgfsys@transformshift{0.450809in}{1.493274in}%
\pgfsys@useobject{currentmarker}{}%
\end{pgfscope}%
\end{pgfscope}%
\begin{pgfscope}%
\pgfsetbuttcap%
\pgfsetroundjoin%
\definecolor{currentfill}{rgb}{0.150000,0.150000,0.150000}%
\pgfsetfillcolor{currentfill}%
\pgfsetlinewidth{0.803000pt}%
\definecolor{currentstroke}{rgb}{0.150000,0.150000,0.150000}%
\pgfsetstrokecolor{currentstroke}%
\pgfsetdash{}{0pt}%
\pgfsys@defobject{currentmarker}{\pgfqpoint{-0.044444in}{0.000000in}}{\pgfqpoint{0.000000in}{0.000000in}}{%
\pgfpathmoveto{\pgfqpoint{0.000000in}{0.000000in}}%
\pgfpathlineto{\pgfqpoint{-0.044444in}{0.000000in}}%
\pgfusepath{stroke,fill}%
}%
\begin{pgfscope}%
\pgfsys@transformshift{0.450809in}{1.882236in}%
\pgfsys@useobject{currentmarker}{}%
\end{pgfscope}%
\end{pgfscope}%
\begin{pgfscope}%
\pgfpathrectangle{\pgfqpoint{0.450809in}{0.326389in}}{\pgfqpoint{2.480000in}{1.694000in}}%
\pgfusepath{clip}%
\pgfsetbuttcap%
\pgfsetroundjoin%
\pgfsetlinewidth{1.204500pt}%
\definecolor{currentstroke}{rgb}{1.000000,0.498039,0.000000}%
\pgfsetstrokecolor{currentstroke}%
\pgfsetdash{{7.680000pt}{1.920000pt}{1.200000pt}{1.920000pt}}{0.000000pt}%
\pgfpathmoveto{\pgfqpoint{0.563536in}{1.678156in}}%
\pgfpathlineto{\pgfqpoint{0.895949in}{1.733729in}}%
\pgfpathlineto{\pgfqpoint{1.147410in}{1.775196in}}%
\pgfpathlineto{\pgfqpoint{1.398871in}{1.808533in}}%
\pgfpathlineto{\pgfqpoint{1.731285in}{1.852170in}}%
\pgfpathlineto{\pgfqpoint{1.982746in}{1.881601in}}%
\pgfpathlineto{\pgfqpoint{2.234207in}{1.902452in}}%
\pgfpathlineto{\pgfqpoint{2.566620in}{1.928004in}}%
\pgfpathlineto{\pgfqpoint{2.818081in}{1.949261in}}%
\pgfusepath{stroke}%
\end{pgfscope}%
\begin{pgfscope}%
\pgfpathrectangle{\pgfqpoint{0.450809in}{0.326389in}}{\pgfqpoint{2.480000in}{1.694000in}}%
\pgfusepath{clip}%
\pgfsetroundcap%
\pgfsetroundjoin%
\pgfsetlinewidth{1.204500pt}%
\definecolor{currentstroke}{rgb}{0.890196,0.101961,0.109804}%
\pgfsetstrokecolor{currentstroke}%
\pgfsetdash{}{0pt}%
\pgfpathmoveto{\pgfqpoint{0.563536in}{1.562369in}}%
\pgfpathlineto{\pgfqpoint{0.895949in}{1.692346in}}%
\pgfpathlineto{\pgfqpoint{1.147410in}{1.724898in}}%
\pgfpathlineto{\pgfqpoint{1.398871in}{1.802223in}}%
\pgfpathlineto{\pgfqpoint{1.731285in}{1.861703in}}%
\pgfpathlineto{\pgfqpoint{1.982746in}{1.900782in}}%
\pgfpathlineto{\pgfqpoint{2.234207in}{1.919963in}}%
\pgfpathlineto{\pgfqpoint{2.566620in}{1.942027in}}%
\pgfpathlineto{\pgfqpoint{2.818081in}{1.963033in}}%
\pgfusepath{stroke}%
\end{pgfscope}%
\begin{pgfscope}%
\pgfpathrectangle{\pgfqpoint{0.450809in}{0.326389in}}{\pgfqpoint{2.480000in}{1.694000in}}%
\pgfusepath{clip}%
\pgfsetbuttcap%
\pgfsetroundjoin%
\pgfsetlinewidth{1.204500pt}%
\definecolor{currentstroke}{rgb}{0.200000,0.627451,0.172549}%
\pgfsetstrokecolor{currentstroke}%
\pgfsetdash{{4.440000pt}{1.920000pt}}{0.000000pt}%
\pgfpathmoveto{\pgfqpoint{0.563536in}{0.815913in}}%
\pgfpathlineto{\pgfqpoint{0.895949in}{1.247844in}}%
\pgfpathlineto{\pgfqpoint{1.147410in}{1.478634in}}%
\pgfpathlineto{\pgfqpoint{1.398871in}{1.616106in}}%
\pgfpathlineto{\pgfqpoint{1.731285in}{1.742660in}}%
\pgfpathlineto{\pgfqpoint{1.982746in}{1.819201in}}%
\pgfpathlineto{\pgfqpoint{2.234207in}{1.867913in}}%
\pgfpathlineto{\pgfqpoint{2.566620in}{1.912924in}}%
\pgfpathlineto{\pgfqpoint{2.818081in}{1.945171in}}%
\pgfusepath{stroke}%
\end{pgfscope}%
\begin{pgfscope}%
\pgfpathrectangle{\pgfqpoint{0.450809in}{0.326389in}}{\pgfqpoint{2.480000in}{1.694000in}}%
\pgfusepath{clip}%
\pgfsetbuttcap%
\pgfsetroundjoin%
\pgfsetlinewidth{1.204500pt}%
\definecolor{currentstroke}{rgb}{0.121569,0.470588,0.705882}%
\pgfsetstrokecolor{currentstroke}%
\pgfsetdash{{1.200000pt}{1.980000pt}}{0.000000pt}%
\pgfpathmoveto{\pgfqpoint{0.563536in}{1.117383in}}%
\pgfpathlineto{\pgfqpoint{0.895949in}{1.447984in}}%
\pgfpathlineto{\pgfqpoint{1.147410in}{1.583620in}}%
\pgfpathlineto{\pgfqpoint{1.398871in}{1.676620in}}%
\pgfpathlineto{\pgfqpoint{1.731285in}{1.767651in}}%
\pgfpathlineto{\pgfqpoint{1.982746in}{1.831220in}}%
\pgfpathlineto{\pgfqpoint{2.234207in}{1.863889in}}%
\pgfpathlineto{\pgfqpoint{2.566620in}{1.914510in}}%
\pgfpathlineto{\pgfqpoint{2.818081in}{1.941832in}}%
\pgfusepath{stroke}%
\end{pgfscope}%
\begin{pgfscope}%
\pgfsetrectcap%
\pgfsetmiterjoin%
\pgfsetlinewidth{1.003750pt}%
\definecolor{currentstroke}{rgb}{0.150000,0.150000,0.150000}%
\pgfsetstrokecolor{currentstroke}%
\pgfsetdash{}{0pt}%
\pgfpathmoveto{\pgfqpoint{0.450809in}{0.326389in}}%
\pgfpathlineto{\pgfqpoint{0.450809in}{2.020389in}}%
\pgfusepath{stroke}%
\end{pgfscope}%
\begin{pgfscope}%
\pgfsetrectcap%
\pgfsetmiterjoin%
\pgfsetlinewidth{1.003750pt}%
\definecolor{currentstroke}{rgb}{0.150000,0.150000,0.150000}%
\pgfsetstrokecolor{currentstroke}%
\pgfsetdash{}{0pt}%
\pgfpathmoveto{\pgfqpoint{2.930809in}{0.326389in}}%
\pgfpathlineto{\pgfqpoint{2.930809in}{2.020389in}}%
\pgfusepath{stroke}%
\end{pgfscope}%
\begin{pgfscope}%
\pgfsetrectcap%
\pgfsetmiterjoin%
\pgfsetlinewidth{1.003750pt}%
\definecolor{currentstroke}{rgb}{0.150000,0.150000,0.150000}%
\pgfsetstrokecolor{currentstroke}%
\pgfsetdash{}{0pt}%
\pgfpathmoveto{\pgfqpoint{0.450809in}{0.326389in}}%
\pgfpathlineto{\pgfqpoint{2.930809in}{0.326389in}}%
\pgfusepath{stroke}%
\end{pgfscope}%
\begin{pgfscope}%
\pgfsetrectcap%
\pgfsetmiterjoin%
\pgfsetlinewidth{1.003750pt}%
\definecolor{currentstroke}{rgb}{0.150000,0.150000,0.150000}%
\pgfsetstrokecolor{currentstroke}%
\pgfsetdash{}{0pt}%
\pgfpathmoveto{\pgfqpoint{0.450809in}{2.020389in}}%
\pgfpathlineto{\pgfqpoint{2.930809in}{2.020389in}}%
\pgfusepath{stroke}%
\end{pgfscope}%
\begin{pgfscope}%
\pgfsetbuttcap%
\pgfsetroundjoin%
\pgfsetlinewidth{1.204500pt}%
\definecolor{currentstroke}{rgb}{1.000000,0.498039,0.000000}%
\pgfsetstrokecolor{currentstroke}%
\pgfsetdash{{7.680000pt}{1.920000pt}{1.200000pt}{1.920000pt}}{0.000000pt}%
\pgfpathmoveto{\pgfqpoint{1.816819in}{0.971250in}}%
\pgfpathlineto{\pgfqpoint{2.061263in}{0.971250in}}%
\pgfusepath{stroke}%
\end{pgfscope}%
\begin{pgfscope}%
\definecolor{textcolor}{rgb}{0.150000,0.150000,0.150000}%
\pgfsetstrokecolor{textcolor}%
\pgfsetfillcolor{textcolor}%
\pgftext[x=2.159041in,y=0.928472in,left,base]{\color{textcolor}\sffamily\fontsize{8.800000}{10.560000}\selectfont Norma}%
\end{pgfscope}%
\begin{pgfscope}%
\pgfsetroundcap%
\pgfsetroundjoin%
\pgfsetlinewidth{1.204500pt}%
\definecolor{currentstroke}{rgb}{0.890196,0.101961,0.109804}%
\pgfsetstrokecolor{currentstroke}%
\pgfsetdash{}{0pt}%
\pgfpathmoveto{\pgfqpoint{1.816819in}{0.799028in}}%
\pgfpathlineto{\pgfqpoint{2.061263in}{0.799028in}}%
\pgfusepath{stroke}%
\end{pgfscope}%
\begin{pgfscope}%
\definecolor{textcolor}{rgb}{0.150000,0.150000,0.150000}%
\pgfsetstrokecolor{textcolor}%
\pgfsetfillcolor{textcolor}%
\pgftext[x=2.159041in,y=0.756250in,left,base]{\color{textcolor}\sffamily\fontsize{8.800000}{10.560000}\selectfont cSMTiser\textsubscript{+LM}}%
\end{pgfscope}%
\begin{pgfscope}%
\pgfsetbuttcap%
\pgfsetroundjoin%
\pgfsetlinewidth{1.204500pt}%
\definecolor{currentstroke}{rgb}{0.200000,0.627451,0.172549}%
\pgfsetstrokecolor{currentstroke}%
\pgfsetdash{{4.440000pt}{1.920000pt}}{0.000000pt}%
\pgfpathmoveto{\pgfqpoint{1.816819in}{0.626806in}}%
\pgfpathlineto{\pgfqpoint{2.061263in}{0.626806in}}%
\pgfusepath{stroke}%
\end{pgfscope}%
\begin{pgfscope}%
\definecolor{textcolor}{rgb}{0.150000,0.150000,0.150000}%
\pgfsetstrokecolor{textcolor}%
\pgfsetfillcolor{textcolor}%
\pgftext[x=2.159041in,y=0.584028in,left,base]{\color{textcolor}\sffamily\fontsize{8.800000}{10.560000}\selectfont NMT-1}%
\end{pgfscope}%
\begin{pgfscope}%
\pgfsetbuttcap%
\pgfsetroundjoin%
\pgfsetlinewidth{1.204500pt}%
\definecolor{currentstroke}{rgb}{0.121569,0.470588,0.705882}%
\pgfsetstrokecolor{currentstroke}%
\pgfsetdash{{1.200000pt}{1.980000pt}}{0.000000pt}%
\pgfpathmoveto{\pgfqpoint{1.816819in}{0.454583in}}%
\pgfpathlineto{\pgfqpoint{2.061263in}{0.454583in}}%
\pgfusepath{stroke}%
\end{pgfscope}%
\begin{pgfscope}%
\definecolor{textcolor}{rgb}{0.150000,0.150000,0.150000}%
\pgfsetstrokecolor{textcolor}%
\pgfsetfillcolor{textcolor}%
\pgftext[x=2.159041in,y=0.411806in,left,base]{\color{textcolor}\sffamily\fontsize{8.800000}{10.560000}\selectfont NMT-2}%
\end{pgfscope}%
\end{pgfpicture}%
\makeatother%
\endgroup%

%% file: curves_portuguese-ps.pgf
\begingroup%
\makeatletter%
\begin{pgfpicture}%
\pgfpathrectangle{\pgfpointorigin}{\pgfqpoint{3.078809in}{2.168389in}}%
\pgfusepath{use as bounding box, clip}%
\begin{pgfscope}%
\pgfsetbuttcap%
\pgfsetmiterjoin%
\definecolor{currentfill}{rgb}{1.000000,1.000000,1.000000}%
\pgfsetfillcolor{currentfill}%
\pgfsetlinewidth{0.000000pt}%
\definecolor{currentstroke}{rgb}{1.000000,1.000000,1.000000}%
\pgfsetstrokecolor{currentstroke}%
\pgfsetdash{}{0pt}%
\pgfpathmoveto{\pgfqpoint{0.000000in}{0.000000in}}%
\pgfpathlineto{\pgfqpoint{3.078809in}{0.000000in}}%
\pgfpathlineto{\pgfqpoint{3.078809in}{2.168389in}}%
\pgfpathlineto{\pgfqpoint{0.000000in}{2.168389in}}%
\pgfpathclose%
\pgfusepath{fill}%
\end{pgfscope}%
\begin{pgfscope}%
\pgfsetbuttcap%
\pgfsetmiterjoin%
\definecolor{currentfill}{rgb}{1.000000,1.000000,1.000000}%
\pgfsetfillcolor{currentfill}%
\pgfsetlinewidth{0.000000pt}%
\definecolor{currentstroke}{rgb}{0.000000,0.000000,0.000000}%
\pgfsetstrokecolor{currentstroke}%
\pgfsetstrokeopacity{0.000000}%
\pgfsetdash{}{0pt}%
\pgfpathmoveto{\pgfqpoint{0.450809in}{0.326389in}}%
\pgfpathlineto{\pgfqpoint{2.930809in}{0.326389in}}%
\pgfpathlineto{\pgfqpoint{2.930809in}{2.020389in}}%
\pgfpathlineto{\pgfqpoint{0.450809in}{2.020389in}}%
\pgfpathclose%
\pgfusepath{fill}%
\end{pgfscope}%
\begin{pgfscope}%
\pgfsetbuttcap%
\pgfsetroundjoin%
\definecolor{currentfill}{rgb}{0.150000,0.150000,0.150000}%
\pgfsetfillcolor{currentfill}%
\pgfsetlinewidth{1.003750pt}%
\definecolor{currentstroke}{rgb}{0.150000,0.150000,0.150000}%
\pgfsetstrokecolor{currentstroke}%
\pgfsetdash{}{0pt}%
\pgfsys@defobject{currentmarker}{\pgfqpoint{0.000000in}{-0.066667in}}{\pgfqpoint{0.000000in}{0.000000in}}{%
\pgfpathmoveto{\pgfqpoint{0.000000in}{0.000000in}}%
\pgfpathlineto{\pgfqpoint{0.000000in}{-0.066667in}}%
\pgfusepath{stroke,fill}%
}%
\begin{pgfscope}%
\pgfsys@transformshift{0.563536in}{0.326389in}%
\pgfsys@useobject{currentmarker}{}%
\end{pgfscope}%
\end{pgfscope}%
\begin{pgfscope}%
\definecolor{textcolor}{rgb}{0.150000,0.150000,0.150000}%
\pgfsetstrokecolor{textcolor}%
\pgfsetfillcolor{textcolor}%
\pgftext[x=0.563536in,y=0.211111in,,top]{\color{textcolor}\sffamily\fontsize{8.800000}{10.560000}\selectfont 100}%
\end{pgfscope}%
\begin{pgfscope}%
\pgfsetbuttcap%
\pgfsetroundjoin%
\definecolor{currentfill}{rgb}{0.150000,0.150000,0.150000}%
\pgfsetfillcolor{currentfill}%
\pgfsetlinewidth{1.003750pt}%
\definecolor{currentstroke}{rgb}{0.150000,0.150000,0.150000}%
\pgfsetstrokecolor{currentstroke}%
\pgfsetdash{}{0pt}%
\pgfsys@defobject{currentmarker}{\pgfqpoint{0.000000in}{-0.066667in}}{\pgfqpoint{0.000000in}{0.000000in}}{%
\pgfpathmoveto{\pgfqpoint{0.000000in}{0.000000in}}%
\pgfpathlineto{\pgfqpoint{0.000000in}{-0.066667in}}%
\pgfusepath{stroke,fill}%
}%
\begin{pgfscope}%
\pgfsys@transformshift{0.895949in}{0.326389in}%
\pgfsys@useobject{currentmarker}{}%
\end{pgfscope}%
\end{pgfscope}%
\begin{pgfscope}%
\definecolor{textcolor}{rgb}{0.150000,0.150000,0.150000}%
\pgfsetstrokecolor{textcolor}%
\pgfsetfillcolor{textcolor}%
\pgftext[x=0.895949in,y=0.211111in,,top]{\color{textcolor}\sffamily\fontsize{8.800000}{10.560000}\selectfont 250}%
\end{pgfscope}%
\begin{pgfscope}%
\pgfsetbuttcap%
\pgfsetroundjoin%
\definecolor{currentfill}{rgb}{0.150000,0.150000,0.150000}%
\pgfsetfillcolor{currentfill}%
\pgfsetlinewidth{1.003750pt}%
\definecolor{currentstroke}{rgb}{0.150000,0.150000,0.150000}%
\pgfsetstrokecolor{currentstroke}%
\pgfsetdash{}{0pt}%
\pgfsys@defobject{currentmarker}{\pgfqpoint{0.000000in}{-0.066667in}}{\pgfqpoint{0.000000in}{0.000000in}}{%
\pgfpathmoveto{\pgfqpoint{0.000000in}{0.000000in}}%
\pgfpathlineto{\pgfqpoint{0.000000in}{-0.066667in}}%
\pgfusepath{stroke,fill}%
}%
\begin{pgfscope}%
\pgfsys@transformshift{1.147410in}{0.326389in}%
\pgfsys@useobject{currentmarker}{}%
\end{pgfscope}%
\end{pgfscope}%
\begin{pgfscope}%
\definecolor{textcolor}{rgb}{0.150000,0.150000,0.150000}%
\pgfsetstrokecolor{textcolor}%
\pgfsetfillcolor{textcolor}%
\pgftext[x=1.147410in,y=0.211111in,,top]{\color{textcolor}\sffamily\fontsize{8.800000}{10.560000}\selectfont 500}%
\end{pgfscope}%
\begin{pgfscope}%
\pgfsetbuttcap%
\pgfsetroundjoin%
\definecolor{currentfill}{rgb}{0.150000,0.150000,0.150000}%
\pgfsetfillcolor{currentfill}%
\pgfsetlinewidth{1.003750pt}%
\definecolor{currentstroke}{rgb}{0.150000,0.150000,0.150000}%
\pgfsetstrokecolor{currentstroke}%
\pgfsetdash{}{0pt}%
\pgfsys@defobject{currentmarker}{\pgfqpoint{0.000000in}{-0.066667in}}{\pgfqpoint{0.000000in}{0.000000in}}{%
\pgfpathmoveto{\pgfqpoint{0.000000in}{0.000000in}}%
\pgfpathlineto{\pgfqpoint{0.000000in}{-0.066667in}}%
\pgfusepath{stroke,fill}%
}%
\begin{pgfscope}%
\pgfsys@transformshift{1.398871in}{0.326389in}%
\pgfsys@useobject{currentmarker}{}%
\end{pgfscope}%
\end{pgfscope}%
\begin{pgfscope}%
\definecolor{textcolor}{rgb}{0.150000,0.150000,0.150000}%
\pgfsetstrokecolor{textcolor}%
\pgfsetfillcolor{textcolor}%
\pgftext[x=1.398871in,y=0.211111in,,top]{\color{textcolor}\sffamily\fontsize{8.800000}{10.560000}\selectfont 1k}%
\end{pgfscope}%
\begin{pgfscope}%
\pgfsetbuttcap%
\pgfsetroundjoin%
\definecolor{currentfill}{rgb}{0.150000,0.150000,0.150000}%
\pgfsetfillcolor{currentfill}%
\pgfsetlinewidth{1.003750pt}%
\definecolor{currentstroke}{rgb}{0.150000,0.150000,0.150000}%
\pgfsetstrokecolor{currentstroke}%
\pgfsetdash{}{0pt}%
\pgfsys@defobject{currentmarker}{\pgfqpoint{0.000000in}{-0.066667in}}{\pgfqpoint{0.000000in}{0.000000in}}{%
\pgfpathmoveto{\pgfqpoint{0.000000in}{0.000000in}}%
\pgfpathlineto{\pgfqpoint{0.000000in}{-0.066667in}}%
\pgfusepath{stroke,fill}%
}%
\begin{pgfscope}%
\pgfsys@transformshift{1.731285in}{0.326389in}%
\pgfsys@useobject{currentmarker}{}%
\end{pgfscope}%
\end{pgfscope}%
\begin{pgfscope}%
\definecolor{textcolor}{rgb}{0.150000,0.150000,0.150000}%
\pgfsetstrokecolor{textcolor}%
\pgfsetfillcolor{textcolor}%
\pgftext[x=1.731285in,y=0.211111in,,top]{\color{textcolor}\sffamily\fontsize{8.800000}{10.560000}\selectfont 2.5k}%
\end{pgfscope}%
\begin{pgfscope}%
\pgfsetbuttcap%
\pgfsetroundjoin%
\definecolor{currentfill}{rgb}{0.150000,0.150000,0.150000}%
\pgfsetfillcolor{currentfill}%
\pgfsetlinewidth{1.003750pt}%
\definecolor{currentstroke}{rgb}{0.150000,0.150000,0.150000}%
\pgfsetstrokecolor{currentstroke}%
\pgfsetdash{}{0pt}%
\pgfsys@defobject{currentmarker}{\pgfqpoint{0.000000in}{-0.066667in}}{\pgfqpoint{0.000000in}{0.000000in}}{%
\pgfpathmoveto{\pgfqpoint{0.000000in}{0.000000in}}%
\pgfpathlineto{\pgfqpoint{0.000000in}{-0.066667in}}%
\pgfusepath{stroke,fill}%
}%
\begin{pgfscope}%
\pgfsys@transformshift{1.982746in}{0.326389in}%
\pgfsys@useobject{currentmarker}{}%
\end{pgfscope}%
\end{pgfscope}%
\begin{pgfscope}%
\definecolor{textcolor}{rgb}{0.150000,0.150000,0.150000}%
\pgfsetstrokecolor{textcolor}%
\pgfsetfillcolor{textcolor}%
\pgftext[x=1.982746in,y=0.211111in,,top]{\color{textcolor}\sffamily\fontsize{8.800000}{10.560000}\selectfont 5k}%
\end{pgfscope}%
\begin{pgfscope}%
\pgfsetbuttcap%
\pgfsetroundjoin%
\definecolor{currentfill}{rgb}{0.150000,0.150000,0.150000}%
\pgfsetfillcolor{currentfill}%
\pgfsetlinewidth{1.003750pt}%
\definecolor{currentstroke}{rgb}{0.150000,0.150000,0.150000}%
\pgfsetstrokecolor{currentstroke}%
\pgfsetdash{}{0pt}%
\pgfsys@defobject{currentmarker}{\pgfqpoint{0.000000in}{-0.066667in}}{\pgfqpoint{0.000000in}{0.000000in}}{%
\pgfpathmoveto{\pgfqpoint{0.000000in}{0.000000in}}%
\pgfpathlineto{\pgfqpoint{0.000000in}{-0.066667in}}%
\pgfusepath{stroke,fill}%
}%
\begin{pgfscope}%
\pgfsys@transformshift{2.234207in}{0.326389in}%
\pgfsys@useobject{currentmarker}{}%
\end{pgfscope}%
\end{pgfscope}%
\begin{pgfscope}%
\definecolor{textcolor}{rgb}{0.150000,0.150000,0.150000}%
\pgfsetstrokecolor{textcolor}%
\pgfsetfillcolor{textcolor}%
\pgftext[x=2.234207in,y=0.211111in,,top]{\color{textcolor}\sffamily\fontsize{8.800000}{10.560000}\selectfont 10k}%
\end{pgfscope}%
\begin{pgfscope}%
\pgfsetbuttcap%
\pgfsetroundjoin%
\definecolor{currentfill}{rgb}{0.150000,0.150000,0.150000}%
\pgfsetfillcolor{currentfill}%
\pgfsetlinewidth{1.003750pt}%
\definecolor{currentstroke}{rgb}{0.150000,0.150000,0.150000}%
\pgfsetstrokecolor{currentstroke}%
\pgfsetdash{}{0pt}%
\pgfsys@defobject{currentmarker}{\pgfqpoint{0.000000in}{-0.066667in}}{\pgfqpoint{0.000000in}{0.000000in}}{%
\pgfpathmoveto{\pgfqpoint{0.000000in}{0.000000in}}%
\pgfpathlineto{\pgfqpoint{0.000000in}{-0.066667in}}%
\pgfusepath{stroke,fill}%
}%
\begin{pgfscope}%
\pgfsys@transformshift{2.566620in}{0.326389in}%
\pgfsys@useobject{currentmarker}{}%
\end{pgfscope}%
\end{pgfscope}%
\begin{pgfscope}%
\definecolor{textcolor}{rgb}{0.150000,0.150000,0.150000}%
\pgfsetstrokecolor{textcolor}%
\pgfsetfillcolor{textcolor}%
\pgftext[x=2.566620in,y=0.211111in,,top]{\color{textcolor}\sffamily\fontsize{8.800000}{10.560000}\selectfont 25k}%
\end{pgfscope}%
\begin{pgfscope}%
\pgfsetbuttcap%
\pgfsetroundjoin%
\definecolor{currentfill}{rgb}{0.150000,0.150000,0.150000}%
\pgfsetfillcolor{currentfill}%
\pgfsetlinewidth{1.003750pt}%
\definecolor{currentstroke}{rgb}{0.150000,0.150000,0.150000}%
\pgfsetstrokecolor{currentstroke}%
\pgfsetdash{}{0pt}%
\pgfsys@defobject{currentmarker}{\pgfqpoint{0.000000in}{-0.066667in}}{\pgfqpoint{0.000000in}{0.000000in}}{%
\pgfpathmoveto{\pgfqpoint{0.000000in}{0.000000in}}%
\pgfpathlineto{\pgfqpoint{0.000000in}{-0.066667in}}%
\pgfusepath{stroke,fill}%
}%
\begin{pgfscope}%
\pgfsys@transformshift{2.818081in}{0.326389in}%
\pgfsys@useobject{currentmarker}{}%
\end{pgfscope}%
\end{pgfscope}%
\begin{pgfscope}%
\definecolor{textcolor}{rgb}{0.150000,0.150000,0.150000}%
\pgfsetstrokecolor{textcolor}%
\pgfsetfillcolor{textcolor}%
\pgftext[x=2.818081in,y=0.211111in,,top]{\color{textcolor}\sffamily\fontsize{8.800000}{10.560000}\selectfont 50k}%
\end{pgfscope}%
\begin{pgfscope}%
\pgfsetbuttcap%
\pgfsetroundjoin%
\definecolor{currentfill}{rgb}{0.150000,0.150000,0.150000}%
\pgfsetfillcolor{currentfill}%
\pgfsetlinewidth{1.003750pt}%
\definecolor{currentstroke}{rgb}{0.150000,0.150000,0.150000}%
\pgfsetstrokecolor{currentstroke}%
\pgfsetdash{}{0pt}%
\pgfsys@defobject{currentmarker}{\pgfqpoint{-0.066667in}{0.000000in}}{\pgfqpoint{0.000000in}{0.000000in}}{%
\pgfpathmoveto{\pgfqpoint{0.000000in}{0.000000in}}%
\pgfpathlineto{\pgfqpoint{-0.066667in}{0.000000in}}%
\pgfusepath{stroke,fill}%
}%
\begin{pgfscope}%
\pgfsys@transformshift{0.450809in}{0.522789in}%
\pgfsys@useobject{currentmarker}{}%
\end{pgfscope}%
\end{pgfscope}%
\begin{pgfscope}%
\definecolor{textcolor}{rgb}{0.150000,0.150000,0.150000}%
\pgfsetstrokecolor{textcolor}%
\pgfsetfillcolor{textcolor}%
\pgftext[x=0.100000in,y=0.479386in,left,base]{\color{textcolor}\sffamily\fontsize{8.800000}{10.560000}\selectfont 20\%}%
\end{pgfscope}%
\begin{pgfscope}%
\pgfsetbuttcap%
\pgfsetroundjoin%
\definecolor{currentfill}{rgb}{0.150000,0.150000,0.150000}%
\pgfsetfillcolor{currentfill}%
\pgfsetlinewidth{1.003750pt}%
\definecolor{currentstroke}{rgb}{0.150000,0.150000,0.150000}%
\pgfsetstrokecolor{currentstroke}%
\pgfsetdash{}{0pt}%
\pgfsys@defobject{currentmarker}{\pgfqpoint{-0.066667in}{0.000000in}}{\pgfqpoint{0.000000in}{0.000000in}}{%
\pgfpathmoveto{\pgfqpoint{0.000000in}{0.000000in}}%
\pgfpathlineto{\pgfqpoint{-0.066667in}{0.000000in}}%
\pgfusepath{stroke,fill}%
}%
\begin{pgfscope}%
\pgfsys@transformshift{0.450809in}{0.915590in}%
\pgfsys@useobject{currentmarker}{}%
\end{pgfscope}%
\end{pgfscope}%
\begin{pgfscope}%
\definecolor{textcolor}{rgb}{0.150000,0.150000,0.150000}%
\pgfsetstrokecolor{textcolor}%
\pgfsetfillcolor{textcolor}%
\pgftext[x=0.100000in,y=0.872187in,left,base]{\color{textcolor}\sffamily\fontsize{8.800000}{10.560000}\selectfont 40\%}%
\end{pgfscope}%
\begin{pgfscope}%
\pgfsetbuttcap%
\pgfsetroundjoin%
\definecolor{currentfill}{rgb}{0.150000,0.150000,0.150000}%
\pgfsetfillcolor{currentfill}%
\pgfsetlinewidth{1.003750pt}%
\definecolor{currentstroke}{rgb}{0.150000,0.150000,0.150000}%
\pgfsetstrokecolor{currentstroke}%
\pgfsetdash{}{0pt}%
\pgfsys@defobject{currentmarker}{\pgfqpoint{-0.066667in}{0.000000in}}{\pgfqpoint{0.000000in}{0.000000in}}{%
\pgfpathmoveto{\pgfqpoint{0.000000in}{0.000000in}}%
\pgfpathlineto{\pgfqpoint{-0.066667in}{0.000000in}}%
\pgfusepath{stroke,fill}%
}%
\begin{pgfscope}%
\pgfsys@transformshift{0.450809in}{1.308390in}%
\pgfsys@useobject{currentmarker}{}%
\end{pgfscope}%
\end{pgfscope}%
\begin{pgfscope}%
\definecolor{textcolor}{rgb}{0.150000,0.150000,0.150000}%
\pgfsetstrokecolor{textcolor}%
\pgfsetfillcolor{textcolor}%
\pgftext[x=0.100000in,y=1.264987in,left,base]{\color{textcolor}\sffamily\fontsize{8.800000}{10.560000}\selectfont 60\%}%
\end{pgfscope}%
\begin{pgfscope}%
\pgfsetbuttcap%
\pgfsetroundjoin%
\definecolor{currentfill}{rgb}{0.150000,0.150000,0.150000}%
\pgfsetfillcolor{currentfill}%
\pgfsetlinewidth{1.003750pt}%
\definecolor{currentstroke}{rgb}{0.150000,0.150000,0.150000}%
\pgfsetstrokecolor{currentstroke}%
\pgfsetdash{}{0pt}%
\pgfsys@defobject{currentmarker}{\pgfqpoint{-0.066667in}{0.000000in}}{\pgfqpoint{0.000000in}{0.000000in}}{%
\pgfpathmoveto{\pgfqpoint{0.000000in}{0.000000in}}%
\pgfpathlineto{\pgfqpoint{-0.066667in}{0.000000in}}%
\pgfusepath{stroke,fill}%
}%
\begin{pgfscope}%
\pgfsys@transformshift{0.450809in}{1.701191in}%
\pgfsys@useobject{currentmarker}{}%
\end{pgfscope}%
\end{pgfscope}%
\begin{pgfscope}%
\definecolor{textcolor}{rgb}{0.150000,0.150000,0.150000}%
\pgfsetstrokecolor{textcolor}%
\pgfsetfillcolor{textcolor}%
\pgftext[x=0.100000in,y=1.657788in,left,base]{\color{textcolor}\sffamily\fontsize{8.800000}{10.560000}\selectfont 80\%}%
\end{pgfscope}%
\begin{pgfscope}%
\pgfsetbuttcap%
\pgfsetroundjoin%
\definecolor{currentfill}{rgb}{0.150000,0.150000,0.150000}%
\pgfsetfillcolor{currentfill}%
\pgfsetlinewidth{0.803000pt}%
\definecolor{currentstroke}{rgb}{0.150000,0.150000,0.150000}%
\pgfsetstrokecolor{currentstroke}%
\pgfsetdash{}{0pt}%
\pgfsys@defobject{currentmarker}{\pgfqpoint{-0.044444in}{0.000000in}}{\pgfqpoint{0.000000in}{0.000000in}}{%
\pgfpathmoveto{\pgfqpoint{0.000000in}{0.000000in}}%
\pgfpathlineto{\pgfqpoint{-0.044444in}{0.000000in}}%
\pgfusepath{stroke,fill}%
}%
\begin{pgfscope}%
\pgfsys@transformshift{0.450809in}{0.326389in}%
\pgfsys@useobject{currentmarker}{}%
\end{pgfscope}%
\end{pgfscope}%
\begin{pgfscope}%
\pgfsetbuttcap%
\pgfsetroundjoin%
\definecolor{currentfill}{rgb}{0.150000,0.150000,0.150000}%
\pgfsetfillcolor{currentfill}%
\pgfsetlinewidth{0.803000pt}%
\definecolor{currentstroke}{rgb}{0.150000,0.150000,0.150000}%
\pgfsetstrokecolor{currentstroke}%
\pgfsetdash{}{0pt}%
\pgfsys@defobject{currentmarker}{\pgfqpoint{-0.044444in}{0.000000in}}{\pgfqpoint{0.000000in}{0.000000in}}{%
\pgfpathmoveto{\pgfqpoint{0.000000in}{0.000000in}}%
\pgfpathlineto{\pgfqpoint{-0.044444in}{0.000000in}}%
\pgfusepath{stroke,fill}%
}%
\begin{pgfscope}%
\pgfsys@transformshift{0.450809in}{0.719189in}%
\pgfsys@useobject{currentmarker}{}%
\end{pgfscope}%
\end{pgfscope}%
\begin{pgfscope}%
\pgfsetbuttcap%
\pgfsetroundjoin%
\definecolor{currentfill}{rgb}{0.150000,0.150000,0.150000}%
\pgfsetfillcolor{currentfill}%
\pgfsetlinewidth{0.803000pt}%
\definecolor{currentstroke}{rgb}{0.150000,0.150000,0.150000}%
\pgfsetstrokecolor{currentstroke}%
\pgfsetdash{}{0pt}%
\pgfsys@defobject{currentmarker}{\pgfqpoint{-0.044444in}{0.000000in}}{\pgfqpoint{0.000000in}{0.000000in}}{%
\pgfpathmoveto{\pgfqpoint{0.000000in}{0.000000in}}%
\pgfpathlineto{\pgfqpoint{-0.044444in}{0.000000in}}%
\pgfusepath{stroke,fill}%
}%
\begin{pgfscope}%
\pgfsys@transformshift{0.450809in}{1.111990in}%
\pgfsys@useobject{currentmarker}{}%
\end{pgfscope}%
\end{pgfscope}%
\begin{pgfscope}%
\pgfsetbuttcap%
\pgfsetroundjoin%
\definecolor{currentfill}{rgb}{0.150000,0.150000,0.150000}%
\pgfsetfillcolor{currentfill}%
\pgfsetlinewidth{0.803000pt}%
\definecolor{currentstroke}{rgb}{0.150000,0.150000,0.150000}%
\pgfsetstrokecolor{currentstroke}%
\pgfsetdash{}{0pt}%
\pgfsys@defobject{currentmarker}{\pgfqpoint{-0.044444in}{0.000000in}}{\pgfqpoint{0.000000in}{0.000000in}}{%
\pgfpathmoveto{\pgfqpoint{0.000000in}{0.000000in}}%
\pgfpathlineto{\pgfqpoint{-0.044444in}{0.000000in}}%
\pgfusepath{stroke,fill}%
}%
\begin{pgfscope}%
\pgfsys@transformshift{0.450809in}{1.504790in}%
\pgfsys@useobject{currentmarker}{}%
\end{pgfscope}%
\end{pgfscope}%
\begin{pgfscope}%
\pgfsetbuttcap%
\pgfsetroundjoin%
\definecolor{currentfill}{rgb}{0.150000,0.150000,0.150000}%
\pgfsetfillcolor{currentfill}%
\pgfsetlinewidth{0.803000pt}%
\definecolor{currentstroke}{rgb}{0.150000,0.150000,0.150000}%
\pgfsetstrokecolor{currentstroke}%
\pgfsetdash{}{0pt}%
\pgfsys@defobject{currentmarker}{\pgfqpoint{-0.044444in}{0.000000in}}{\pgfqpoint{0.000000in}{0.000000in}}{%
\pgfpathmoveto{\pgfqpoint{0.000000in}{0.000000in}}%
\pgfpathlineto{\pgfqpoint{-0.044444in}{0.000000in}}%
\pgfusepath{stroke,fill}%
}%
\begin{pgfscope}%
\pgfsys@transformshift{0.450809in}{1.897591in}%
\pgfsys@useobject{currentmarker}{}%
\end{pgfscope}%
\end{pgfscope}%
\begin{pgfscope}%
\pgfpathrectangle{\pgfqpoint{0.450809in}{0.326389in}}{\pgfqpoint{2.480000in}{1.694000in}}%
\pgfusepath{clip}%
\pgfsetbuttcap%
\pgfsetroundjoin%
\pgfsetlinewidth{1.204500pt}%
\definecolor{currentstroke}{rgb}{1.000000,0.498039,0.000000}%
\pgfsetstrokecolor{currentstroke}%
\pgfsetdash{{7.680000pt}{1.920000pt}{1.200000pt}{1.920000pt}}{0.000000pt}%
\pgfpathmoveto{\pgfqpoint{0.563536in}{1.612319in}}%
\pgfpathlineto{\pgfqpoint{0.895949in}{1.681410in}}%
\pgfpathlineto{\pgfqpoint{1.147410in}{1.723482in}}%
\pgfpathlineto{\pgfqpoint{1.398871in}{1.759012in}}%
\pgfpathlineto{\pgfqpoint{1.731285in}{1.807251in}}%
\pgfpathlineto{\pgfqpoint{1.982746in}{1.843940in}}%
\pgfpathlineto{\pgfqpoint{2.234207in}{1.875336in}}%
\pgfpathlineto{\pgfqpoint{2.566620in}{1.905616in}}%
\pgfpathlineto{\pgfqpoint{2.818081in}{1.929061in}}%
\pgfusepath{stroke}%
\end{pgfscope}%
\begin{pgfscope}%
\pgfpathrectangle{\pgfqpoint{0.450809in}{0.326389in}}{\pgfqpoint{2.480000in}{1.694000in}}%
\pgfusepath{clip}%
\pgfsetroundcap%
\pgfsetroundjoin%
\pgfsetlinewidth{1.204500pt}%
\definecolor{currentstroke}{rgb}{0.890196,0.101961,0.109804}%
\pgfsetstrokecolor{currentstroke}%
\pgfsetdash{}{0pt}%
\pgfpathmoveto{\pgfqpoint{0.563536in}{1.518954in}}%
\pgfpathlineto{\pgfqpoint{0.895949in}{1.593368in}}%
\pgfpathlineto{\pgfqpoint{1.147410in}{1.679964in}}%
\pgfpathlineto{\pgfqpoint{1.398871in}{1.740560in}}%
\pgfpathlineto{\pgfqpoint{1.731285in}{1.817317in}}%
\pgfpathlineto{\pgfqpoint{1.982746in}{1.859234in}}%
\pgfpathlineto{\pgfqpoint{2.234207in}{1.903457in}}%
\pgfpathlineto{\pgfqpoint{2.566620in}{1.933502in}}%
\pgfpathlineto{\pgfqpoint{2.818081in}{1.957791in}}%
\pgfusepath{stroke}%
\end{pgfscope}%
\begin{pgfscope}%
\pgfpathrectangle{\pgfqpoint{0.450809in}{0.326389in}}{\pgfqpoint{2.480000in}{1.694000in}}%
\pgfusepath{clip}%
\pgfsetbuttcap%
\pgfsetroundjoin%
\pgfsetlinewidth{1.204500pt}%
\definecolor{currentstroke}{rgb}{0.200000,0.627451,0.172549}%
\pgfsetstrokecolor{currentstroke}%
\pgfsetdash{{4.440000pt}{1.920000pt}}{0.000000pt}%
\pgfpathmoveto{\pgfqpoint{0.563536in}{0.705826in}}%
\pgfpathlineto{\pgfqpoint{0.895949in}{1.132093in}}%
\pgfpathlineto{\pgfqpoint{1.147410in}{1.368436in}}%
\pgfpathlineto{\pgfqpoint{1.398871in}{1.532875in}}%
\pgfpathlineto{\pgfqpoint{1.731285in}{1.672879in}}%
\pgfpathlineto{\pgfqpoint{1.982746in}{1.755164in}}%
\pgfpathlineto{\pgfqpoint{2.234207in}{1.819586in}}%
\pgfpathlineto{\pgfqpoint{2.566620in}{1.869822in}}%
\pgfpathlineto{\pgfqpoint{2.818081in}{1.909908in}}%
\pgfusepath{stroke}%
\end{pgfscope}%
\begin{pgfscope}%
\pgfpathrectangle{\pgfqpoint{0.450809in}{0.326389in}}{\pgfqpoint{2.480000in}{1.694000in}}%
\pgfusepath{clip}%
\pgfsetbuttcap%
\pgfsetroundjoin%
\pgfsetlinewidth{1.204500pt}%
\definecolor{currentstroke}{rgb}{0.121569,0.470588,0.705882}%
\pgfsetstrokecolor{currentstroke}%
\pgfsetdash{{1.200000pt}{1.980000pt}}{0.000000pt}%
\pgfpathmoveto{\pgfqpoint{0.563536in}{1.015321in}}%
\pgfpathlineto{\pgfqpoint{0.895949in}{1.350542in}}%
\pgfpathlineto{\pgfqpoint{1.147410in}{1.511186in}}%
\pgfpathlineto{\pgfqpoint{1.398871in}{1.605872in}}%
\pgfpathlineto{\pgfqpoint{1.731285in}{1.703379in}}%
\pgfpathlineto{\pgfqpoint{1.982746in}{1.773770in}}%
\pgfpathlineto{\pgfqpoint{2.234207in}{1.836730in}}%
\pgfpathlineto{\pgfqpoint{2.566620in}{1.880574in}}%
\pgfpathlineto{\pgfqpoint{2.818081in}{1.924666in}}%
\pgfusepath{stroke}%
\end{pgfscope}%
\begin{pgfscope}%
\pgfsetrectcap%
\pgfsetmiterjoin%
\pgfsetlinewidth{1.003750pt}%
\definecolor{currentstroke}{rgb}{0.150000,0.150000,0.150000}%
\pgfsetstrokecolor{currentstroke}%
\pgfsetdash{}{0pt}%
\pgfpathmoveto{\pgfqpoint{0.450809in}{0.326389in}}%
\pgfpathlineto{\pgfqpoint{0.450809in}{2.020389in}}%
\pgfusepath{stroke}%
\end{pgfscope}%
\begin{pgfscope}%
\pgfsetrectcap%
\pgfsetmiterjoin%
\pgfsetlinewidth{1.003750pt}%
\definecolor{currentstroke}{rgb}{0.150000,0.150000,0.150000}%
\pgfsetstrokecolor{currentstroke}%
\pgfsetdash{}{0pt}%
\pgfpathmoveto{\pgfqpoint{2.930809in}{0.326389in}}%
\pgfpathlineto{\pgfqpoint{2.930809in}{2.020389in}}%
\pgfusepath{stroke}%
\end{pgfscope}%
\begin{pgfscope}%
\pgfsetrectcap%
\pgfsetmiterjoin%
\pgfsetlinewidth{1.003750pt}%
\definecolor{currentstroke}{rgb}{0.150000,0.150000,0.150000}%
\pgfsetstrokecolor{currentstroke}%
\pgfsetdash{}{0pt}%
\pgfpathmoveto{\pgfqpoint{0.450809in}{0.326389in}}%
\pgfpathlineto{\pgfqpoint{2.930809in}{0.326389in}}%
\pgfusepath{stroke}%
\end{pgfscope}%
\begin{pgfscope}%
\pgfsetrectcap%
\pgfsetmiterjoin%
\pgfsetlinewidth{1.003750pt}%
\definecolor{currentstroke}{rgb}{0.150000,0.150000,0.150000}%
\pgfsetstrokecolor{currentstroke}%
\pgfsetdash{}{0pt}%
\pgfpathmoveto{\pgfqpoint{0.450809in}{2.020389in}}%
\pgfpathlineto{\pgfqpoint{2.930809in}{2.020389in}}%
\pgfusepath{stroke}%
\end{pgfscope}%
\begin{pgfscope}%
\pgfsetbuttcap%
\pgfsetroundjoin%
\pgfsetlinewidth{1.204500pt}%
\definecolor{currentstroke}{rgb}{1.000000,0.498039,0.000000}%
\pgfsetstrokecolor{currentstroke}%
\pgfsetdash{{7.680000pt}{1.920000pt}{1.200000pt}{1.920000pt}}{0.000000pt}%
\pgfpathmoveto{\pgfqpoint{1.816819in}{0.971250in}}%
\pgfpathlineto{\pgfqpoint{2.061263in}{0.971250in}}%
\pgfusepath{stroke}%
\end{pgfscope}%
\begin{pgfscope}%
\definecolor{textcolor}{rgb}{0.150000,0.150000,0.150000}%
\pgfsetstrokecolor{textcolor}%
\pgfsetfillcolor{textcolor}%
\pgftext[x=2.159041in,y=0.928472in,left,base]{\color{textcolor}\sffamily\fontsize{8.800000}{10.560000}\selectfont Norma}%
\end{pgfscope}%
\begin{pgfscope}%
\pgfsetroundcap%
\pgfsetroundjoin%
\pgfsetlinewidth{1.204500pt}%
\definecolor{currentstroke}{rgb}{0.890196,0.101961,0.109804}%
\pgfsetstrokecolor{currentstroke}%
\pgfsetdash{}{0pt}%
\pgfpathmoveto{\pgfqpoint{1.816819in}{0.799028in}}%
\pgfpathlineto{\pgfqpoint{2.061263in}{0.799028in}}%
\pgfusepath{stroke}%
\end{pgfscope}%
\begin{pgfscope}%
\definecolor{textcolor}{rgb}{0.150000,0.150000,0.150000}%
\pgfsetstrokecolor{textcolor}%
\pgfsetfillcolor{textcolor}%
\pgftext[x=2.159041in,y=0.756250in,left,base]{\color{textcolor}\sffamily\fontsize{8.800000}{10.560000}\selectfont cSMTiser\textsubscript{+LM}}%
\end{pgfscope}%
\begin{pgfscope}%
\pgfsetbuttcap%
\pgfsetroundjoin%
\pgfsetlinewidth{1.204500pt}%
\definecolor{currentstroke}{rgb}{0.200000,0.627451,0.172549}%
\pgfsetstrokecolor{currentstroke}%
\pgfsetdash{{4.440000pt}{1.920000pt}}{0.000000pt}%
\pgfpathmoveto{\pgfqpoint{1.816819in}{0.626806in}}%
\pgfpathlineto{\pgfqpoint{2.061263in}{0.626806in}}%
\pgfusepath{stroke}%
\end{pgfscope}%
\begin{pgfscope}%
\definecolor{textcolor}{rgb}{0.150000,0.150000,0.150000}%
\pgfsetstrokecolor{textcolor}%
\pgfsetfillcolor{textcolor}%
\pgftext[x=2.159041in,y=0.584028in,left,base]{\color{textcolor}\sffamily\fontsize{8.800000}{10.560000}\selectfont NMT-1}%
\end{pgfscope}%
\begin{pgfscope}%
\pgfsetbuttcap%
\pgfsetroundjoin%
\pgfsetlinewidth{1.204500pt}%
\definecolor{currentstroke}{rgb}{0.121569,0.470588,0.705882}%
\pgfsetstrokecolor{currentstroke}%
\pgfsetdash{{1.200000pt}{1.980000pt}}{0.000000pt}%
\pgfpathmoveto{\pgfqpoint{1.816819in}{0.454583in}}%
\pgfpathlineto{\pgfqpoint{2.061263in}{0.454583in}}%
\pgfusepath{stroke}%
\end{pgfscope}%
\begin{pgfscope}%
\definecolor{textcolor}{rgb}{0.150000,0.150000,0.150000}%
\pgfsetstrokecolor{textcolor}%
\pgfsetfillcolor{textcolor}%
\pgftext[x=2.159041in,y=0.411806in,left,base]{\color{textcolor}\sffamily\fontsize{8.800000}{10.560000}\selectfont NMT-2}%
\end{pgfscope}%
\end{pgfpicture}%
\makeatother%
\endgroup%

%% file: curves_icelandic-icepahc.pgf
\begingroup%
\makeatletter%
\begin{pgfpicture}%
\pgfpathrectangle{\pgfpointorigin}{\pgfqpoint{3.078809in}{2.168389in}}%
\pgfusepath{use as bounding box, clip}%
\begin{pgfscope}%
\pgfsetbuttcap%
\pgfsetmiterjoin%
\definecolor{currentfill}{rgb}{1.000000,1.000000,1.000000}%
\pgfsetfillcolor{currentfill}%
\pgfsetlinewidth{0.000000pt}%
\definecolor{currentstroke}{rgb}{1.000000,1.000000,1.000000}%
\pgfsetstrokecolor{currentstroke}%
\pgfsetdash{}{0pt}%
\pgfpathmoveto{\pgfqpoint{0.000000in}{0.000000in}}%
\pgfpathlineto{\pgfqpoint{3.078809in}{0.000000in}}%
\pgfpathlineto{\pgfqpoint{3.078809in}{2.168389in}}%
\pgfpathlineto{\pgfqpoint{0.000000in}{2.168389in}}%
\pgfpathclose%
\pgfusepath{fill}%
\end{pgfscope}%
\begin{pgfscope}%
\pgfsetbuttcap%
\pgfsetmiterjoin%
\definecolor{currentfill}{rgb}{1.000000,1.000000,1.000000}%
\pgfsetfillcolor{currentfill}%
\pgfsetlinewidth{0.000000pt}%
\definecolor{currentstroke}{rgb}{0.000000,0.000000,0.000000}%
\pgfsetstrokecolor{currentstroke}%
\pgfsetstrokeopacity{0.000000}%
\pgfsetdash{}{0pt}%
\pgfpathmoveto{\pgfqpoint{0.450809in}{0.326389in}}%
\pgfpathlineto{\pgfqpoint{2.930809in}{0.326389in}}%
\pgfpathlineto{\pgfqpoint{2.930809in}{2.020389in}}%
\pgfpathlineto{\pgfqpoint{0.450809in}{2.020389in}}%
\pgfpathclose%
\pgfusepath{fill}%
\end{pgfscope}%
\begin{pgfscope}%
\pgfsetbuttcap%
\pgfsetroundjoin%
\definecolor{currentfill}{rgb}{0.150000,0.150000,0.150000}%
\pgfsetfillcolor{currentfill}%
\pgfsetlinewidth{1.003750pt}%
\definecolor{currentstroke}{rgb}{0.150000,0.150000,0.150000}%
\pgfsetstrokecolor{currentstroke}%
\pgfsetdash{}{0pt}%
\pgfsys@defobject{currentmarker}{\pgfqpoint{0.000000in}{-0.066667in}}{\pgfqpoint{0.000000in}{0.000000in}}{%
\pgfpathmoveto{\pgfqpoint{0.000000in}{0.000000in}}%
\pgfpathlineto{\pgfqpoint{0.000000in}{-0.066667in}}%
\pgfusepath{stroke,fill}%
}%
\begin{pgfscope}%
\pgfsys@transformshift{0.563536in}{0.326389in}%
\pgfsys@useobject{currentmarker}{}%
\end{pgfscope}%
\end{pgfscope}%
\begin{pgfscope}%
\definecolor{textcolor}{rgb}{0.150000,0.150000,0.150000}%
\pgfsetstrokecolor{textcolor}%
\pgfsetfillcolor{textcolor}%
\pgftext[x=0.563536in,y=0.211111in,,top]{\color{textcolor}\sffamily\fontsize{8.800000}{10.560000}\selectfont 100}%
\end{pgfscope}%
\begin{pgfscope}%
\pgfsetbuttcap%
\pgfsetroundjoin%
\definecolor{currentfill}{rgb}{0.150000,0.150000,0.150000}%
\pgfsetfillcolor{currentfill}%
\pgfsetlinewidth{1.003750pt}%
\definecolor{currentstroke}{rgb}{0.150000,0.150000,0.150000}%
\pgfsetstrokecolor{currentstroke}%
\pgfsetdash{}{0pt}%
\pgfsys@defobject{currentmarker}{\pgfqpoint{0.000000in}{-0.066667in}}{\pgfqpoint{0.000000in}{0.000000in}}{%
\pgfpathmoveto{\pgfqpoint{0.000000in}{0.000000in}}%
\pgfpathlineto{\pgfqpoint{0.000000in}{-0.066667in}}%
\pgfusepath{stroke,fill}%
}%
\begin{pgfscope}%
\pgfsys@transformshift{0.937679in}{0.326389in}%
\pgfsys@useobject{currentmarker}{}%
\end{pgfscope}%
\end{pgfscope}%
\begin{pgfscope}%
\definecolor{textcolor}{rgb}{0.150000,0.150000,0.150000}%
\pgfsetstrokecolor{textcolor}%
\pgfsetfillcolor{textcolor}%
\pgftext[x=0.937679in,y=0.211111in,,top]{\color{textcolor}\sffamily\fontsize{8.800000}{10.560000}\selectfont 250}%
\end{pgfscope}%
\begin{pgfscope}%
\pgfsetbuttcap%
\pgfsetroundjoin%
\definecolor{currentfill}{rgb}{0.150000,0.150000,0.150000}%
\pgfsetfillcolor{currentfill}%
\pgfsetlinewidth{1.003750pt}%
\definecolor{currentstroke}{rgb}{0.150000,0.150000,0.150000}%
\pgfsetstrokecolor{currentstroke}%
\pgfsetdash{}{0pt}%
\pgfsys@defobject{currentmarker}{\pgfqpoint{0.000000in}{-0.066667in}}{\pgfqpoint{0.000000in}{0.000000in}}{%
\pgfpathmoveto{\pgfqpoint{0.000000in}{0.000000in}}%
\pgfpathlineto{\pgfqpoint{0.000000in}{-0.066667in}}%
\pgfusepath{stroke,fill}%
}%
\begin{pgfscope}%
\pgfsys@transformshift{1.220708in}{0.326389in}%
\pgfsys@useobject{currentmarker}{}%
\end{pgfscope}%
\end{pgfscope}%
\begin{pgfscope}%
\definecolor{textcolor}{rgb}{0.150000,0.150000,0.150000}%
\pgfsetstrokecolor{textcolor}%
\pgfsetfillcolor{textcolor}%
\pgftext[x=1.220708in,y=0.211111in,,top]{\color{textcolor}\sffamily\fontsize{8.800000}{10.560000}\selectfont 500}%
\end{pgfscope}%
\begin{pgfscope}%
\pgfsetbuttcap%
\pgfsetroundjoin%
\definecolor{currentfill}{rgb}{0.150000,0.150000,0.150000}%
\pgfsetfillcolor{currentfill}%
\pgfsetlinewidth{1.003750pt}%
\definecolor{currentstroke}{rgb}{0.150000,0.150000,0.150000}%
\pgfsetstrokecolor{currentstroke}%
\pgfsetdash{}{0pt}%
\pgfsys@defobject{currentmarker}{\pgfqpoint{0.000000in}{-0.066667in}}{\pgfqpoint{0.000000in}{0.000000in}}{%
\pgfpathmoveto{\pgfqpoint{0.000000in}{0.000000in}}%
\pgfpathlineto{\pgfqpoint{0.000000in}{-0.066667in}}%
\pgfusepath{stroke,fill}%
}%
\begin{pgfscope}%
\pgfsys@transformshift{1.503737in}{0.326389in}%
\pgfsys@useobject{currentmarker}{}%
\end{pgfscope}%
\end{pgfscope}%
\begin{pgfscope}%
\definecolor{textcolor}{rgb}{0.150000,0.150000,0.150000}%
\pgfsetstrokecolor{textcolor}%
\pgfsetfillcolor{textcolor}%
\pgftext[x=1.503737in,y=0.211111in,,top]{\color{textcolor}\sffamily\fontsize{8.800000}{10.560000}\selectfont 1k}%
\end{pgfscope}%
\begin{pgfscope}%
\pgfsetbuttcap%
\pgfsetroundjoin%
\definecolor{currentfill}{rgb}{0.150000,0.150000,0.150000}%
\pgfsetfillcolor{currentfill}%
\pgfsetlinewidth{1.003750pt}%
\definecolor{currentstroke}{rgb}{0.150000,0.150000,0.150000}%
\pgfsetstrokecolor{currentstroke}%
\pgfsetdash{}{0pt}%
\pgfsys@defobject{currentmarker}{\pgfqpoint{0.000000in}{-0.066667in}}{\pgfqpoint{0.000000in}{0.000000in}}{%
\pgfpathmoveto{\pgfqpoint{0.000000in}{0.000000in}}%
\pgfpathlineto{\pgfqpoint{0.000000in}{-0.066667in}}%
\pgfusepath{stroke,fill}%
}%
\begin{pgfscope}%
\pgfsys@transformshift{1.877880in}{0.326389in}%
\pgfsys@useobject{currentmarker}{}%
\end{pgfscope}%
\end{pgfscope}%
\begin{pgfscope}%
\definecolor{textcolor}{rgb}{0.150000,0.150000,0.150000}%
\pgfsetstrokecolor{textcolor}%
\pgfsetfillcolor{textcolor}%
\pgftext[x=1.877880in,y=0.211111in,,top]{\color{textcolor}\sffamily\fontsize{8.800000}{10.560000}\selectfont 2.5k}%
\end{pgfscope}%
\begin{pgfscope}%
\pgfsetbuttcap%
\pgfsetroundjoin%
\definecolor{currentfill}{rgb}{0.150000,0.150000,0.150000}%
\pgfsetfillcolor{currentfill}%
\pgfsetlinewidth{1.003750pt}%
\definecolor{currentstroke}{rgb}{0.150000,0.150000,0.150000}%
\pgfsetstrokecolor{currentstroke}%
\pgfsetdash{}{0pt}%
\pgfsys@defobject{currentmarker}{\pgfqpoint{0.000000in}{-0.066667in}}{\pgfqpoint{0.000000in}{0.000000in}}{%
\pgfpathmoveto{\pgfqpoint{0.000000in}{0.000000in}}%
\pgfpathlineto{\pgfqpoint{0.000000in}{-0.066667in}}%
\pgfusepath{stroke,fill}%
}%
\begin{pgfscope}%
\pgfsys@transformshift{2.160909in}{0.326389in}%
\pgfsys@useobject{currentmarker}{}%
\end{pgfscope}%
\end{pgfscope}%
\begin{pgfscope}%
\definecolor{textcolor}{rgb}{0.150000,0.150000,0.150000}%
\pgfsetstrokecolor{textcolor}%
\pgfsetfillcolor{textcolor}%
\pgftext[x=2.160909in,y=0.211111in,,top]{\color{textcolor}\sffamily\fontsize{8.800000}{10.560000}\selectfont 5k}%
\end{pgfscope}%
\begin{pgfscope}%
\pgfsetbuttcap%
\pgfsetroundjoin%
\definecolor{currentfill}{rgb}{0.150000,0.150000,0.150000}%
\pgfsetfillcolor{currentfill}%
\pgfsetlinewidth{1.003750pt}%
\definecolor{currentstroke}{rgb}{0.150000,0.150000,0.150000}%
\pgfsetstrokecolor{currentstroke}%
\pgfsetdash{}{0pt}%
\pgfsys@defobject{currentmarker}{\pgfqpoint{0.000000in}{-0.066667in}}{\pgfqpoint{0.000000in}{0.000000in}}{%
\pgfpathmoveto{\pgfqpoint{0.000000in}{0.000000in}}%
\pgfpathlineto{\pgfqpoint{0.000000in}{-0.066667in}}%
\pgfusepath{stroke,fill}%
}%
\begin{pgfscope}%
\pgfsys@transformshift{2.443938in}{0.326389in}%
\pgfsys@useobject{currentmarker}{}%
\end{pgfscope}%
\end{pgfscope}%
\begin{pgfscope}%
\definecolor{textcolor}{rgb}{0.150000,0.150000,0.150000}%
\pgfsetstrokecolor{textcolor}%
\pgfsetfillcolor{textcolor}%
\pgftext[x=2.443938in,y=0.211111in,,top]{\color{textcolor}\sffamily\fontsize{8.800000}{10.560000}\selectfont 10k}%
\end{pgfscope}%
\begin{pgfscope}%
\pgfsetbuttcap%
\pgfsetroundjoin%
\definecolor{currentfill}{rgb}{0.150000,0.150000,0.150000}%
\pgfsetfillcolor{currentfill}%
\pgfsetlinewidth{1.003750pt}%
\definecolor{currentstroke}{rgb}{0.150000,0.150000,0.150000}%
\pgfsetstrokecolor{currentstroke}%
\pgfsetdash{}{0pt}%
\pgfsys@defobject{currentmarker}{\pgfqpoint{0.000000in}{-0.066667in}}{\pgfqpoint{0.000000in}{0.000000in}}{%
\pgfpathmoveto{\pgfqpoint{0.000000in}{0.000000in}}%
\pgfpathlineto{\pgfqpoint{0.000000in}{-0.066667in}}%
\pgfusepath{stroke,fill}%
}%
\begin{pgfscope}%
\pgfsys@transformshift{2.818081in}{0.326389in}%
\pgfsys@useobject{currentmarker}{}%
\end{pgfscope}%
\end{pgfscope}%
\begin{pgfscope}%
\definecolor{textcolor}{rgb}{0.150000,0.150000,0.150000}%
\pgfsetstrokecolor{textcolor}%
\pgfsetfillcolor{textcolor}%
\pgftext[x=2.818081in,y=0.211111in,,top]{\color{textcolor}\sffamily\fontsize{8.800000}{10.560000}\selectfont 25k}%
\end{pgfscope}%
\begin{pgfscope}%
\pgfsetbuttcap%
\pgfsetroundjoin%
\definecolor{currentfill}{rgb}{0.150000,0.150000,0.150000}%
\pgfsetfillcolor{currentfill}%
\pgfsetlinewidth{1.003750pt}%
\definecolor{currentstroke}{rgb}{0.150000,0.150000,0.150000}%
\pgfsetstrokecolor{currentstroke}%
\pgfsetdash{}{0pt}%
\pgfsys@defobject{currentmarker}{\pgfqpoint{-0.066667in}{0.000000in}}{\pgfqpoint{0.000000in}{0.000000in}}{%
\pgfpathmoveto{\pgfqpoint{0.000000in}{0.000000in}}%
\pgfpathlineto{\pgfqpoint{-0.066667in}{0.000000in}}%
\pgfusepath{stroke,fill}%
}%
\begin{pgfscope}%
\pgfsys@transformshift{0.450809in}{0.538139in}%
\pgfsys@useobject{currentmarker}{}%
\end{pgfscope}%
\end{pgfscope}%
\begin{pgfscope}%
\definecolor{textcolor}{rgb}{0.150000,0.150000,0.150000}%
\pgfsetstrokecolor{textcolor}%
\pgfsetfillcolor{textcolor}%
\pgftext[x=0.100000in,y=0.494736in,left,base]{\color{textcolor}\sffamily\fontsize{8.800000}{10.560000}\selectfont 20\%}%
\end{pgfscope}%
\begin{pgfscope}%
\pgfsetbuttcap%
\pgfsetroundjoin%
\definecolor{currentfill}{rgb}{0.150000,0.150000,0.150000}%
\pgfsetfillcolor{currentfill}%
\pgfsetlinewidth{1.003750pt}%
\definecolor{currentstroke}{rgb}{0.150000,0.150000,0.150000}%
\pgfsetstrokecolor{currentstroke}%
\pgfsetdash{}{0pt}%
\pgfsys@defobject{currentmarker}{\pgfqpoint{-0.066667in}{0.000000in}}{\pgfqpoint{0.000000in}{0.000000in}}{%
\pgfpathmoveto{\pgfqpoint{0.000000in}{0.000000in}}%
\pgfpathlineto{\pgfqpoint{-0.066667in}{0.000000in}}%
\pgfusepath{stroke,fill}%
}%
\begin{pgfscope}%
\pgfsys@transformshift{0.450809in}{0.961639in}%
\pgfsys@useobject{currentmarker}{}%
\end{pgfscope}%
\end{pgfscope}%
\begin{pgfscope}%
\definecolor{textcolor}{rgb}{0.150000,0.150000,0.150000}%
\pgfsetstrokecolor{textcolor}%
\pgfsetfillcolor{textcolor}%
\pgftext[x=0.100000in,y=0.918236in,left,base]{\color{textcolor}\sffamily\fontsize{8.800000}{10.560000}\selectfont 40\%}%
\end{pgfscope}%
\begin{pgfscope}%
\pgfsetbuttcap%
\pgfsetroundjoin%
\definecolor{currentfill}{rgb}{0.150000,0.150000,0.150000}%
\pgfsetfillcolor{currentfill}%
\pgfsetlinewidth{1.003750pt}%
\definecolor{currentstroke}{rgb}{0.150000,0.150000,0.150000}%
\pgfsetstrokecolor{currentstroke}%
\pgfsetdash{}{0pt}%
\pgfsys@defobject{currentmarker}{\pgfqpoint{-0.066667in}{0.000000in}}{\pgfqpoint{0.000000in}{0.000000in}}{%
\pgfpathmoveto{\pgfqpoint{0.000000in}{0.000000in}}%
\pgfpathlineto{\pgfqpoint{-0.066667in}{0.000000in}}%
\pgfusepath{stroke,fill}%
}%
\begin{pgfscope}%
\pgfsys@transformshift{0.450809in}{1.385139in}%
\pgfsys@useobject{currentmarker}{}%
\end{pgfscope}%
\end{pgfscope}%
\begin{pgfscope}%
\definecolor{textcolor}{rgb}{0.150000,0.150000,0.150000}%
\pgfsetstrokecolor{textcolor}%
\pgfsetfillcolor{textcolor}%
\pgftext[x=0.100000in,y=1.341736in,left,base]{\color{textcolor}\sffamily\fontsize{8.800000}{10.560000}\selectfont 60\%}%
\end{pgfscope}%
\begin{pgfscope}%
\pgfsetbuttcap%
\pgfsetroundjoin%
\definecolor{currentfill}{rgb}{0.150000,0.150000,0.150000}%
\pgfsetfillcolor{currentfill}%
\pgfsetlinewidth{1.003750pt}%
\definecolor{currentstroke}{rgb}{0.150000,0.150000,0.150000}%
\pgfsetstrokecolor{currentstroke}%
\pgfsetdash{}{0pt}%
\pgfsys@defobject{currentmarker}{\pgfqpoint{-0.066667in}{0.000000in}}{\pgfqpoint{0.000000in}{0.000000in}}{%
\pgfpathmoveto{\pgfqpoint{0.000000in}{0.000000in}}%
\pgfpathlineto{\pgfqpoint{-0.066667in}{0.000000in}}%
\pgfusepath{stroke,fill}%
}%
\begin{pgfscope}%
\pgfsys@transformshift{0.450809in}{1.808639in}%
\pgfsys@useobject{currentmarker}{}%
\end{pgfscope}%
\end{pgfscope}%
\begin{pgfscope}%
\definecolor{textcolor}{rgb}{0.150000,0.150000,0.150000}%
\pgfsetstrokecolor{textcolor}%
\pgfsetfillcolor{textcolor}%
\pgftext[x=0.100000in,y=1.765236in,left,base]{\color{textcolor}\sffamily\fontsize{8.800000}{10.560000}\selectfont 80\%}%
\end{pgfscope}%
\begin{pgfscope}%
\pgfsetbuttcap%
\pgfsetroundjoin%
\definecolor{currentfill}{rgb}{0.150000,0.150000,0.150000}%
\pgfsetfillcolor{currentfill}%
\pgfsetlinewidth{0.803000pt}%
\definecolor{currentstroke}{rgb}{0.150000,0.150000,0.150000}%
\pgfsetstrokecolor{currentstroke}%
\pgfsetdash{}{0pt}%
\pgfsys@defobject{currentmarker}{\pgfqpoint{-0.044444in}{0.000000in}}{\pgfqpoint{0.000000in}{0.000000in}}{%
\pgfpathmoveto{\pgfqpoint{0.000000in}{0.000000in}}%
\pgfpathlineto{\pgfqpoint{-0.044444in}{0.000000in}}%
\pgfusepath{stroke,fill}%
}%
\begin{pgfscope}%
\pgfsys@transformshift{0.450809in}{0.326389in}%
\pgfsys@useobject{currentmarker}{}%
\end{pgfscope}%
\end{pgfscope}%
\begin{pgfscope}%
\pgfsetbuttcap%
\pgfsetroundjoin%
\definecolor{currentfill}{rgb}{0.150000,0.150000,0.150000}%
\pgfsetfillcolor{currentfill}%
\pgfsetlinewidth{0.803000pt}%
\definecolor{currentstroke}{rgb}{0.150000,0.150000,0.150000}%
\pgfsetstrokecolor{currentstroke}%
\pgfsetdash{}{0pt}%
\pgfsys@defobject{currentmarker}{\pgfqpoint{-0.044444in}{0.000000in}}{\pgfqpoint{0.000000in}{0.000000in}}{%
\pgfpathmoveto{\pgfqpoint{0.000000in}{0.000000in}}%
\pgfpathlineto{\pgfqpoint{-0.044444in}{0.000000in}}%
\pgfusepath{stroke,fill}%
}%
\begin{pgfscope}%
\pgfsys@transformshift{0.450809in}{0.749889in}%
\pgfsys@useobject{currentmarker}{}%
\end{pgfscope}%
\end{pgfscope}%
\begin{pgfscope}%
\pgfsetbuttcap%
\pgfsetroundjoin%
\definecolor{currentfill}{rgb}{0.150000,0.150000,0.150000}%
\pgfsetfillcolor{currentfill}%
\pgfsetlinewidth{0.803000pt}%
\definecolor{currentstroke}{rgb}{0.150000,0.150000,0.150000}%
\pgfsetstrokecolor{currentstroke}%
\pgfsetdash{}{0pt}%
\pgfsys@defobject{currentmarker}{\pgfqpoint{-0.044444in}{0.000000in}}{\pgfqpoint{0.000000in}{0.000000in}}{%
\pgfpathmoveto{\pgfqpoint{0.000000in}{0.000000in}}%
\pgfpathlineto{\pgfqpoint{-0.044444in}{0.000000in}}%
\pgfusepath{stroke,fill}%
}%
\begin{pgfscope}%
\pgfsys@transformshift{0.450809in}{1.173389in}%
\pgfsys@useobject{currentmarker}{}%
\end{pgfscope}%
\end{pgfscope}%
\begin{pgfscope}%
\pgfsetbuttcap%
\pgfsetroundjoin%
\definecolor{currentfill}{rgb}{0.150000,0.150000,0.150000}%
\pgfsetfillcolor{currentfill}%
\pgfsetlinewidth{0.803000pt}%
\definecolor{currentstroke}{rgb}{0.150000,0.150000,0.150000}%
\pgfsetstrokecolor{currentstroke}%
\pgfsetdash{}{0pt}%
\pgfsys@defobject{currentmarker}{\pgfqpoint{-0.044444in}{0.000000in}}{\pgfqpoint{0.000000in}{0.000000in}}{%
\pgfpathmoveto{\pgfqpoint{0.000000in}{0.000000in}}%
\pgfpathlineto{\pgfqpoint{-0.044444in}{0.000000in}}%
\pgfusepath{stroke,fill}%
}%
\begin{pgfscope}%
\pgfsys@transformshift{0.450809in}{1.596889in}%
\pgfsys@useobject{currentmarker}{}%
\end{pgfscope}%
\end{pgfscope}%
\begin{pgfscope}%
\pgfsetbuttcap%
\pgfsetroundjoin%
\definecolor{currentfill}{rgb}{0.150000,0.150000,0.150000}%
\pgfsetfillcolor{currentfill}%
\pgfsetlinewidth{0.803000pt}%
\definecolor{currentstroke}{rgb}{0.150000,0.150000,0.150000}%
\pgfsetstrokecolor{currentstroke}%
\pgfsetdash{}{0pt}%
\pgfsys@defobject{currentmarker}{\pgfqpoint{-0.044444in}{0.000000in}}{\pgfqpoint{0.000000in}{0.000000in}}{%
\pgfpathmoveto{\pgfqpoint{0.000000in}{0.000000in}}%
\pgfpathlineto{\pgfqpoint{-0.044444in}{0.000000in}}%
\pgfusepath{stroke,fill}%
}%
\begin{pgfscope}%
\pgfsys@transformshift{0.450809in}{2.020389in}%
\pgfsys@useobject{currentmarker}{}%
\end{pgfscope}%
\end{pgfscope}%
\begin{pgfscope}%
\pgfpathrectangle{\pgfqpoint{0.450809in}{0.326389in}}{\pgfqpoint{2.480000in}{1.694000in}}%
\pgfusepath{clip}%
\pgfsetbuttcap%
\pgfsetroundjoin%
\pgfsetlinewidth{1.204500pt}%
\definecolor{currentstroke}{rgb}{1.000000,0.498039,0.000000}%
\pgfsetstrokecolor{currentstroke}%
\pgfsetdash{{7.680000pt}{1.920000pt}{1.200000pt}{1.920000pt}}{0.000000pt}%
\pgfpathmoveto{\pgfqpoint{0.563536in}{1.430858in}}%
\pgfpathlineto{\pgfqpoint{0.937679in}{1.457028in}}%
\pgfpathlineto{\pgfqpoint{1.220708in}{1.516750in}}%
\pgfpathlineto{\pgfqpoint{1.503737in}{1.615052in}}%
\pgfpathlineto{\pgfqpoint{1.877880in}{1.702157in}}%
\pgfpathlineto{\pgfqpoint{2.160909in}{1.695572in}}%
\pgfpathlineto{\pgfqpoint{2.443938in}{1.757876in}}%
\pgfpathlineto{\pgfqpoint{2.818081in}{1.825901in}}%
\pgfusepath{stroke}%
\end{pgfscope}%
\begin{pgfscope}%
\pgfpathrectangle{\pgfqpoint{0.450809in}{0.326389in}}{\pgfqpoint{2.480000in}{1.694000in}}%
\pgfusepath{clip}%
\pgfsetroundcap%
\pgfsetroundjoin%
\pgfsetlinewidth{1.204500pt}%
\definecolor{currentstroke}{rgb}{0.890196,0.101961,0.109804}%
\pgfsetstrokecolor{currentstroke}%
\pgfsetdash{}{0pt}%
\pgfpathmoveto{\pgfqpoint{0.563536in}{1.439489in}}%
\pgfpathlineto{\pgfqpoint{0.937679in}{1.450095in}}%
\pgfpathlineto{\pgfqpoint{1.220708in}{1.545208in}}%
\pgfpathlineto{\pgfqpoint{1.503737in}{1.645554in}}%
\pgfpathlineto{\pgfqpoint{1.877880in}{1.773838in}}%
\pgfpathlineto{\pgfqpoint{2.160909in}{1.732244in}}%
\pgfpathlineto{\pgfqpoint{2.443938in}{1.779627in}}%
\pgfpathlineto{\pgfqpoint{2.818081in}{1.846004in}}%
\pgfusepath{stroke}%
\end{pgfscope}%
\begin{pgfscope}%
\pgfpathrectangle{\pgfqpoint{0.450809in}{0.326389in}}{\pgfqpoint{2.480000in}{1.694000in}}%
\pgfusepath{clip}%
\pgfsetbuttcap%
\pgfsetroundjoin%
\pgfsetlinewidth{1.204500pt}%
\definecolor{currentstroke}{rgb}{0.200000,0.627451,0.172549}%
\pgfsetstrokecolor{currentstroke}%
\pgfsetdash{{4.440000pt}{1.920000pt}}{0.000000pt}%
\pgfpathmoveto{\pgfqpoint{0.563536in}{0.566666in}}%
\pgfpathlineto{\pgfqpoint{0.937679in}{0.994706in}}%
\pgfpathlineto{\pgfqpoint{1.220708in}{1.260078in}}%
\pgfpathlineto{\pgfqpoint{1.503737in}{1.444376in}}%
\pgfpathlineto{\pgfqpoint{1.877880in}{1.612625in}}%
\pgfpathlineto{\pgfqpoint{2.160909in}{1.657339in}}%
\pgfpathlineto{\pgfqpoint{2.443938in}{1.716282in}}%
\pgfpathlineto{\pgfqpoint{2.818081in}{1.799904in}}%
\pgfusepath{stroke}%
\end{pgfscope}%
\begin{pgfscope}%
\pgfpathrectangle{\pgfqpoint{0.450809in}{0.326389in}}{\pgfqpoint{2.480000in}{1.694000in}}%
\pgfusepath{clip}%
\pgfsetbuttcap%
\pgfsetroundjoin%
\pgfsetlinewidth{1.204500pt}%
\definecolor{currentstroke}{rgb}{0.121569,0.470588,0.705882}%
\pgfsetstrokecolor{currentstroke}%
\pgfsetdash{{1.200000pt}{1.980000pt}}{0.000000pt}%
\pgfpathmoveto{\pgfqpoint{0.563536in}{0.866422in}}%
\pgfpathlineto{\pgfqpoint{0.937679in}{1.170512in}}%
\pgfpathlineto{\pgfqpoint{1.220708in}{1.353978in}}%
\pgfpathlineto{\pgfqpoint{1.503737in}{1.511759in}}%
\pgfpathlineto{\pgfqpoint{1.877880in}{1.622781in}}%
\pgfpathlineto{\pgfqpoint{2.160909in}{1.627322in}}%
\pgfpathlineto{\pgfqpoint{2.443938in}{1.699341in}}%
\pgfpathlineto{\pgfqpoint{2.818081in}{1.758136in}}%
\pgfusepath{stroke}%
\end{pgfscope}%
\begin{pgfscope}%
\pgfsetrectcap%
\pgfsetmiterjoin%
\pgfsetlinewidth{1.003750pt}%
\definecolor{currentstroke}{rgb}{0.150000,0.150000,0.150000}%
\pgfsetstrokecolor{currentstroke}%
\pgfsetdash{}{0pt}%
\pgfpathmoveto{\pgfqpoint{0.450809in}{0.326389in}}%
\pgfpathlineto{\pgfqpoint{0.450809in}{2.020389in}}%
\pgfusepath{stroke}%
\end{pgfscope}%
\begin{pgfscope}%
\pgfsetrectcap%
\pgfsetmiterjoin%
\pgfsetlinewidth{1.003750pt}%
\definecolor{currentstroke}{rgb}{0.150000,0.150000,0.150000}%
\pgfsetstrokecolor{currentstroke}%
\pgfsetdash{}{0pt}%
\pgfpathmoveto{\pgfqpoint{2.930809in}{0.326389in}}%
\pgfpathlineto{\pgfqpoint{2.930809in}{2.020389in}}%
\pgfusepath{stroke}%
\end{pgfscope}%
\begin{pgfscope}%
\pgfsetrectcap%
\pgfsetmiterjoin%
\pgfsetlinewidth{1.003750pt}%
\definecolor{currentstroke}{rgb}{0.150000,0.150000,0.150000}%
\pgfsetstrokecolor{currentstroke}%
\pgfsetdash{}{0pt}%
\pgfpathmoveto{\pgfqpoint{0.450809in}{0.326389in}}%
\pgfpathlineto{\pgfqpoint{2.930809in}{0.326389in}}%
\pgfusepath{stroke}%
\end{pgfscope}%
\begin{pgfscope}%
\pgfsetrectcap%
\pgfsetmiterjoin%
\pgfsetlinewidth{1.003750pt}%
\definecolor{currentstroke}{rgb}{0.150000,0.150000,0.150000}%
\pgfsetstrokecolor{currentstroke}%
\pgfsetdash{}{0pt}%
\pgfpathmoveto{\pgfqpoint{0.450809in}{2.020389in}}%
\pgfpathlineto{\pgfqpoint{2.930809in}{2.020389in}}%
\pgfusepath{stroke}%
\end{pgfscope}%
\begin{pgfscope}%
\pgfsetbuttcap%
\pgfsetroundjoin%
\pgfsetlinewidth{1.204500pt}%
\definecolor{currentstroke}{rgb}{1.000000,0.498039,0.000000}%
\pgfsetstrokecolor{currentstroke}%
\pgfsetdash{{7.680000pt}{1.920000pt}{1.200000pt}{1.920000pt}}{0.000000pt}%
\pgfpathmoveto{\pgfqpoint{1.816819in}{0.971250in}}%
\pgfpathlineto{\pgfqpoint{2.061263in}{0.971250in}}%
\pgfusepath{stroke}%
\end{pgfscope}%
\begin{pgfscope}%
\definecolor{textcolor}{rgb}{0.150000,0.150000,0.150000}%
\pgfsetstrokecolor{textcolor}%
\pgfsetfillcolor{textcolor}%
\pgftext[x=2.159041in,y=0.928472in,left,base]{\color{textcolor}\sffamily\fontsize{8.800000}{10.560000}\selectfont Norma}%
\end{pgfscope}%
\begin{pgfscope}%
\pgfsetroundcap%
\pgfsetroundjoin%
\pgfsetlinewidth{1.204500pt}%
\definecolor{currentstroke}{rgb}{0.890196,0.101961,0.109804}%
\pgfsetstrokecolor{currentstroke}%
\pgfsetdash{}{0pt}%
\pgfpathmoveto{\pgfqpoint{1.816819in}{0.799028in}}%
\pgfpathlineto{\pgfqpoint{2.061263in}{0.799028in}}%
\pgfusepath{stroke}%
\end{pgfscope}%
\begin{pgfscope}%
\definecolor{textcolor}{rgb}{0.150000,0.150000,0.150000}%
\pgfsetstrokecolor{textcolor}%
\pgfsetfillcolor{textcolor}%
\pgftext[x=2.159041in,y=0.756250in,left,base]{\color{textcolor}\sffamily\fontsize{8.800000}{10.560000}\selectfont cSMTiser\textsubscript{+LM}}%
\end{pgfscope}%
\begin{pgfscope}%
\pgfsetbuttcap%
\pgfsetroundjoin%
\pgfsetlinewidth{1.204500pt}%
\definecolor{currentstroke}{rgb}{0.200000,0.627451,0.172549}%
\pgfsetstrokecolor{currentstroke}%
\pgfsetdash{{4.440000pt}{1.920000pt}}{0.000000pt}%
\pgfpathmoveto{\pgfqpoint{1.816819in}{0.626806in}}%
\pgfpathlineto{\pgfqpoint{2.061263in}{0.626806in}}%
\pgfusepath{stroke}%
\end{pgfscope}%
\begin{pgfscope}%
\definecolor{textcolor}{rgb}{0.150000,0.150000,0.150000}%
\pgfsetstrokecolor{textcolor}%
\pgfsetfillcolor{textcolor}%
\pgftext[x=2.159041in,y=0.584028in,left,base]{\color{textcolor}\sffamily\fontsize{8.800000}{10.560000}\selectfont NMT-1}%
\end{pgfscope}%
\begin{pgfscope}%
\pgfsetbuttcap%
\pgfsetroundjoin%
\pgfsetlinewidth{1.204500pt}%
\definecolor{currentstroke}{rgb}{0.121569,0.470588,0.705882}%
\pgfsetstrokecolor{currentstroke}%
\pgfsetdash{{1.200000pt}{1.980000pt}}{0.000000pt}%
\pgfpathmoveto{\pgfqpoint{1.816819in}{0.454583in}}%
\pgfpathlineto{\pgfqpoint{2.061263in}{0.454583in}}%
\pgfusepath{stroke}%
\end{pgfscope}%
\begin{pgfscope}%
\definecolor{textcolor}{rgb}{0.150000,0.150000,0.150000}%
\pgfsetstrokecolor{textcolor}%
\pgfsetfillcolor{textcolor}%
\pgftext[x=2.159041in,y=0.411806in,left,base]{\color{textcolor}\sffamily\fontsize{8.800000}{10.560000}\selectfont NMT-2}%
\end{pgfscope}%
\end{pgfpicture}%
\makeatother%
\endgroup%

%% file: curves_swedish-gaw.pgf
\begingroup%
\makeatletter%
\begin{pgfpicture}%
\pgfpathrectangle{\pgfpointorigin}{\pgfqpoint{3.078809in}{2.168389in}}%
\pgfusepath{use as bounding box, clip}%
\begin{pgfscope}%
\pgfsetbuttcap%
\pgfsetmiterjoin%
\definecolor{currentfill}{rgb}{1.000000,1.000000,1.000000}%
\pgfsetfillcolor{currentfill}%
\pgfsetlinewidth{0.000000pt}%
\definecolor{currentstroke}{rgb}{1.000000,1.000000,1.000000}%
\pgfsetstrokecolor{currentstroke}%
\pgfsetdash{}{0pt}%
\pgfpathmoveto{\pgfqpoint{0.000000in}{0.000000in}}%
\pgfpathlineto{\pgfqpoint{3.078809in}{0.000000in}}%
\pgfpathlineto{\pgfqpoint{3.078809in}{2.168389in}}%
\pgfpathlineto{\pgfqpoint{0.000000in}{2.168389in}}%
\pgfpathclose%
\pgfusepath{fill}%
\end{pgfscope}%
\begin{pgfscope}%
\pgfsetbuttcap%
\pgfsetmiterjoin%
\definecolor{currentfill}{rgb}{1.000000,1.000000,1.000000}%
\pgfsetfillcolor{currentfill}%
\pgfsetlinewidth{0.000000pt}%
\definecolor{currentstroke}{rgb}{0.000000,0.000000,0.000000}%
\pgfsetstrokecolor{currentstroke}%
\pgfsetstrokeopacity{0.000000}%
\pgfsetdash{}{0pt}%
\pgfpathmoveto{\pgfqpoint{0.450809in}{0.326389in}}%
\pgfpathlineto{\pgfqpoint{2.930809in}{0.326389in}}%
\pgfpathlineto{\pgfqpoint{2.930809in}{2.020389in}}%
\pgfpathlineto{\pgfqpoint{0.450809in}{2.020389in}}%
\pgfpathclose%
\pgfusepath{fill}%
\end{pgfscope}%
\begin{pgfscope}%
\pgfsetbuttcap%
\pgfsetroundjoin%
\definecolor{currentfill}{rgb}{0.150000,0.150000,0.150000}%
\pgfsetfillcolor{currentfill}%
\pgfsetlinewidth{1.003750pt}%
\definecolor{currentstroke}{rgb}{0.150000,0.150000,0.150000}%
\pgfsetstrokecolor{currentstroke}%
\pgfsetdash{}{0pt}%
\pgfsys@defobject{currentmarker}{\pgfqpoint{0.000000in}{-0.066667in}}{\pgfqpoint{0.000000in}{0.000000in}}{%
\pgfpathmoveto{\pgfqpoint{0.000000in}{0.000000in}}%
\pgfpathlineto{\pgfqpoint{0.000000in}{-0.066667in}}%
\pgfusepath{stroke,fill}%
}%
\begin{pgfscope}%
\pgfsys@transformshift{0.563536in}{0.326389in}%
\pgfsys@useobject{currentmarker}{}%
\end{pgfscope}%
\end{pgfscope}%
\begin{pgfscope}%
\definecolor{textcolor}{rgb}{0.150000,0.150000,0.150000}%
\pgfsetstrokecolor{textcolor}%
\pgfsetfillcolor{textcolor}%
\pgftext[x=0.563536in,y=0.211111in,,top]{\color{textcolor}\sffamily\fontsize{8.800000}{10.560000}\selectfont 100}%
\end{pgfscope}%
\begin{pgfscope}%
\pgfsetbuttcap%
\pgfsetroundjoin%
\definecolor{currentfill}{rgb}{0.150000,0.150000,0.150000}%
\pgfsetfillcolor{currentfill}%
\pgfsetlinewidth{1.003750pt}%
\definecolor{currentstroke}{rgb}{0.150000,0.150000,0.150000}%
\pgfsetstrokecolor{currentstroke}%
\pgfsetdash{}{0pt}%
\pgfsys@defobject{currentmarker}{\pgfqpoint{0.000000in}{-0.066667in}}{\pgfqpoint{0.000000in}{0.000000in}}{%
\pgfpathmoveto{\pgfqpoint{0.000000in}{0.000000in}}%
\pgfpathlineto{\pgfqpoint{0.000000in}{-0.066667in}}%
\pgfusepath{stroke,fill}%
}%
\begin{pgfscope}%
\pgfsys@transformshift{1.012123in}{0.326389in}%
\pgfsys@useobject{currentmarker}{}%
\end{pgfscope}%
\end{pgfscope}%
\begin{pgfscope}%
\definecolor{textcolor}{rgb}{0.150000,0.150000,0.150000}%
\pgfsetstrokecolor{textcolor}%
\pgfsetfillcolor{textcolor}%
\pgftext[x=1.012123in,y=0.211111in,,top]{\color{textcolor}\sffamily\fontsize{8.800000}{10.560000}\selectfont 250}%
\end{pgfscope}%
\begin{pgfscope}%
\pgfsetbuttcap%
\pgfsetroundjoin%
\definecolor{currentfill}{rgb}{0.150000,0.150000,0.150000}%
\pgfsetfillcolor{currentfill}%
\pgfsetlinewidth{1.003750pt}%
\definecolor{currentstroke}{rgb}{0.150000,0.150000,0.150000}%
\pgfsetstrokecolor{currentstroke}%
\pgfsetdash{}{0pt}%
\pgfsys@defobject{currentmarker}{\pgfqpoint{0.000000in}{-0.066667in}}{\pgfqpoint{0.000000in}{0.000000in}}{%
\pgfpathmoveto{\pgfqpoint{0.000000in}{0.000000in}}%
\pgfpathlineto{\pgfqpoint{0.000000in}{-0.066667in}}%
\pgfusepath{stroke,fill}%
}%
\begin{pgfscope}%
\pgfsys@transformshift{1.351466in}{0.326389in}%
\pgfsys@useobject{currentmarker}{}%
\end{pgfscope}%
\end{pgfscope}%
\begin{pgfscope}%
\definecolor{textcolor}{rgb}{0.150000,0.150000,0.150000}%
\pgfsetstrokecolor{textcolor}%
\pgfsetfillcolor{textcolor}%
\pgftext[x=1.351466in,y=0.211111in,,top]{\color{textcolor}\sffamily\fontsize{8.800000}{10.560000}\selectfont 500}%
\end{pgfscope}%
\begin{pgfscope}%
\pgfsetbuttcap%
\pgfsetroundjoin%
\definecolor{currentfill}{rgb}{0.150000,0.150000,0.150000}%
\pgfsetfillcolor{currentfill}%
\pgfsetlinewidth{1.003750pt}%
\definecolor{currentstroke}{rgb}{0.150000,0.150000,0.150000}%
\pgfsetstrokecolor{currentstroke}%
\pgfsetdash{}{0pt}%
\pgfsys@defobject{currentmarker}{\pgfqpoint{0.000000in}{-0.066667in}}{\pgfqpoint{0.000000in}{0.000000in}}{%
\pgfpathmoveto{\pgfqpoint{0.000000in}{0.000000in}}%
\pgfpathlineto{\pgfqpoint{0.000000in}{-0.066667in}}%
\pgfusepath{stroke,fill}%
}%
\begin{pgfscope}%
\pgfsys@transformshift{1.690809in}{0.326389in}%
\pgfsys@useobject{currentmarker}{}%
\end{pgfscope}%
\end{pgfscope}%
\begin{pgfscope}%
\definecolor{textcolor}{rgb}{0.150000,0.150000,0.150000}%
\pgfsetstrokecolor{textcolor}%
\pgfsetfillcolor{textcolor}%
\pgftext[x=1.690809in,y=0.211111in,,top]{\color{textcolor}\sffamily\fontsize{8.800000}{10.560000}\selectfont 1k}%
\end{pgfscope}%
\begin{pgfscope}%
\pgfsetbuttcap%
\pgfsetroundjoin%
\definecolor{currentfill}{rgb}{0.150000,0.150000,0.150000}%
\pgfsetfillcolor{currentfill}%
\pgfsetlinewidth{1.003750pt}%
\definecolor{currentstroke}{rgb}{0.150000,0.150000,0.150000}%
\pgfsetstrokecolor{currentstroke}%
\pgfsetdash{}{0pt}%
\pgfsys@defobject{currentmarker}{\pgfqpoint{0.000000in}{-0.066667in}}{\pgfqpoint{0.000000in}{0.000000in}}{%
\pgfpathmoveto{\pgfqpoint{0.000000in}{0.000000in}}%
\pgfpathlineto{\pgfqpoint{0.000000in}{-0.066667in}}%
\pgfusepath{stroke,fill}%
}%
\begin{pgfscope}%
\pgfsys@transformshift{2.139396in}{0.326389in}%
\pgfsys@useobject{currentmarker}{}%
\end{pgfscope}%
\end{pgfscope}%
\begin{pgfscope}%
\definecolor{textcolor}{rgb}{0.150000,0.150000,0.150000}%
\pgfsetstrokecolor{textcolor}%
\pgfsetfillcolor{textcolor}%
\pgftext[x=2.139396in,y=0.211111in,,top]{\color{textcolor}\sffamily\fontsize{8.800000}{10.560000}\selectfont 2.5k}%
\end{pgfscope}%
\begin{pgfscope}%
\pgfsetbuttcap%
\pgfsetroundjoin%
\definecolor{currentfill}{rgb}{0.150000,0.150000,0.150000}%
\pgfsetfillcolor{currentfill}%
\pgfsetlinewidth{1.003750pt}%
\definecolor{currentstroke}{rgb}{0.150000,0.150000,0.150000}%
\pgfsetstrokecolor{currentstroke}%
\pgfsetdash{}{0pt}%
\pgfsys@defobject{currentmarker}{\pgfqpoint{0.000000in}{-0.066667in}}{\pgfqpoint{0.000000in}{0.000000in}}{%
\pgfpathmoveto{\pgfqpoint{0.000000in}{0.000000in}}%
\pgfpathlineto{\pgfqpoint{0.000000in}{-0.066667in}}%
\pgfusepath{stroke,fill}%
}%
\begin{pgfscope}%
\pgfsys@transformshift{2.478738in}{0.326389in}%
\pgfsys@useobject{currentmarker}{}%
\end{pgfscope}%
\end{pgfscope}%
\begin{pgfscope}%
\definecolor{textcolor}{rgb}{0.150000,0.150000,0.150000}%
\pgfsetstrokecolor{textcolor}%
\pgfsetfillcolor{textcolor}%
\pgftext[x=2.478738in,y=0.211111in,,top]{\color{textcolor}\sffamily\fontsize{8.800000}{10.560000}\selectfont 5k}%
\end{pgfscope}%
\begin{pgfscope}%
\pgfsetbuttcap%
\pgfsetroundjoin%
\definecolor{currentfill}{rgb}{0.150000,0.150000,0.150000}%
\pgfsetfillcolor{currentfill}%
\pgfsetlinewidth{1.003750pt}%
\definecolor{currentstroke}{rgb}{0.150000,0.150000,0.150000}%
\pgfsetstrokecolor{currentstroke}%
\pgfsetdash{}{0pt}%
\pgfsys@defobject{currentmarker}{\pgfqpoint{0.000000in}{-0.066667in}}{\pgfqpoint{0.000000in}{0.000000in}}{%
\pgfpathmoveto{\pgfqpoint{0.000000in}{0.000000in}}%
\pgfpathlineto{\pgfqpoint{0.000000in}{-0.066667in}}%
\pgfusepath{stroke,fill}%
}%
\begin{pgfscope}%
\pgfsys@transformshift{2.818081in}{0.326389in}%
\pgfsys@useobject{currentmarker}{}%
\end{pgfscope}%
\end{pgfscope}%
\begin{pgfscope}%
\definecolor{textcolor}{rgb}{0.150000,0.150000,0.150000}%
\pgfsetstrokecolor{textcolor}%
\pgfsetfillcolor{textcolor}%
\pgftext[x=2.818081in,y=0.211111in,,top]{\color{textcolor}\sffamily\fontsize{8.800000}{10.560000}\selectfont 10k}%
\end{pgfscope}%
\begin{pgfscope}%
\pgfsetbuttcap%
\pgfsetroundjoin%
\definecolor{currentfill}{rgb}{0.150000,0.150000,0.150000}%
\pgfsetfillcolor{currentfill}%
\pgfsetlinewidth{1.003750pt}%
\definecolor{currentstroke}{rgb}{0.150000,0.150000,0.150000}%
\pgfsetstrokecolor{currentstroke}%
\pgfsetdash{}{0pt}%
\pgfsys@defobject{currentmarker}{\pgfqpoint{-0.066667in}{0.000000in}}{\pgfqpoint{0.000000in}{0.000000in}}{%
\pgfpathmoveto{\pgfqpoint{0.000000in}{0.000000in}}%
\pgfpathlineto{\pgfqpoint{-0.066667in}{0.000000in}}%
\pgfusepath{stroke,fill}%
}%
\begin{pgfscope}%
\pgfsys@transformshift{0.450809in}{0.533242in}%
\pgfsys@useobject{currentmarker}{}%
\end{pgfscope}%
\end{pgfscope}%
\begin{pgfscope}%
\definecolor{textcolor}{rgb}{0.150000,0.150000,0.150000}%
\pgfsetstrokecolor{textcolor}%
\pgfsetfillcolor{textcolor}%
\pgftext[x=0.100000in,y=0.489839in,left,base]{\color{textcolor}\sffamily\fontsize{8.800000}{10.560000}\selectfont 20\%}%
\end{pgfscope}%
\begin{pgfscope}%
\pgfsetbuttcap%
\pgfsetroundjoin%
\definecolor{currentfill}{rgb}{0.150000,0.150000,0.150000}%
\pgfsetfillcolor{currentfill}%
\pgfsetlinewidth{1.003750pt}%
\definecolor{currentstroke}{rgb}{0.150000,0.150000,0.150000}%
\pgfsetstrokecolor{currentstroke}%
\pgfsetdash{}{0pt}%
\pgfsys@defobject{currentmarker}{\pgfqpoint{-0.066667in}{0.000000in}}{\pgfqpoint{0.000000in}{0.000000in}}{%
\pgfpathmoveto{\pgfqpoint{0.000000in}{0.000000in}}%
\pgfpathlineto{\pgfqpoint{-0.066667in}{0.000000in}}%
\pgfusepath{stroke,fill}%
}%
\begin{pgfscope}%
\pgfsys@transformshift{0.450809in}{0.946947in}%
\pgfsys@useobject{currentmarker}{}%
\end{pgfscope}%
\end{pgfscope}%
\begin{pgfscope}%
\definecolor{textcolor}{rgb}{0.150000,0.150000,0.150000}%
\pgfsetstrokecolor{textcolor}%
\pgfsetfillcolor{textcolor}%
\pgftext[x=0.100000in,y=0.903544in,left,base]{\color{textcolor}\sffamily\fontsize{8.800000}{10.560000}\selectfont 40\%}%
\end{pgfscope}%
\begin{pgfscope}%
\pgfsetbuttcap%
\pgfsetroundjoin%
\definecolor{currentfill}{rgb}{0.150000,0.150000,0.150000}%
\pgfsetfillcolor{currentfill}%
\pgfsetlinewidth{1.003750pt}%
\definecolor{currentstroke}{rgb}{0.150000,0.150000,0.150000}%
\pgfsetstrokecolor{currentstroke}%
\pgfsetdash{}{0pt}%
\pgfsys@defobject{currentmarker}{\pgfqpoint{-0.066667in}{0.000000in}}{\pgfqpoint{0.000000in}{0.000000in}}{%
\pgfpathmoveto{\pgfqpoint{0.000000in}{0.000000in}}%
\pgfpathlineto{\pgfqpoint{-0.066667in}{0.000000in}}%
\pgfusepath{stroke,fill}%
}%
\begin{pgfscope}%
\pgfsys@transformshift{0.450809in}{1.360653in}%
\pgfsys@useobject{currentmarker}{}%
\end{pgfscope}%
\end{pgfscope}%
\begin{pgfscope}%
\definecolor{textcolor}{rgb}{0.150000,0.150000,0.150000}%
\pgfsetstrokecolor{textcolor}%
\pgfsetfillcolor{textcolor}%
\pgftext[x=0.100000in,y=1.317250in,left,base]{\color{textcolor}\sffamily\fontsize{8.800000}{10.560000}\selectfont 60\%}%
\end{pgfscope}%
\begin{pgfscope}%
\pgfsetbuttcap%
\pgfsetroundjoin%
\definecolor{currentfill}{rgb}{0.150000,0.150000,0.150000}%
\pgfsetfillcolor{currentfill}%
\pgfsetlinewidth{1.003750pt}%
\definecolor{currentstroke}{rgb}{0.150000,0.150000,0.150000}%
\pgfsetstrokecolor{currentstroke}%
\pgfsetdash{}{0pt}%
\pgfsys@defobject{currentmarker}{\pgfqpoint{-0.066667in}{0.000000in}}{\pgfqpoint{0.000000in}{0.000000in}}{%
\pgfpathmoveto{\pgfqpoint{0.000000in}{0.000000in}}%
\pgfpathlineto{\pgfqpoint{-0.066667in}{0.000000in}}%
\pgfusepath{stroke,fill}%
}%
\begin{pgfscope}%
\pgfsys@transformshift{0.450809in}{1.774358in}%
\pgfsys@useobject{currentmarker}{}%
\end{pgfscope}%
\end{pgfscope}%
\begin{pgfscope}%
\definecolor{textcolor}{rgb}{0.150000,0.150000,0.150000}%
\pgfsetstrokecolor{textcolor}%
\pgfsetfillcolor{textcolor}%
\pgftext[x=0.100000in,y=1.730956in,left,base]{\color{textcolor}\sffamily\fontsize{8.800000}{10.560000}\selectfont 80\%}%
\end{pgfscope}%
\begin{pgfscope}%
\pgfsetbuttcap%
\pgfsetroundjoin%
\definecolor{currentfill}{rgb}{0.150000,0.150000,0.150000}%
\pgfsetfillcolor{currentfill}%
\pgfsetlinewidth{0.803000pt}%
\definecolor{currentstroke}{rgb}{0.150000,0.150000,0.150000}%
\pgfsetstrokecolor{currentstroke}%
\pgfsetdash{}{0pt}%
\pgfsys@defobject{currentmarker}{\pgfqpoint{-0.044444in}{0.000000in}}{\pgfqpoint{0.000000in}{0.000000in}}{%
\pgfpathmoveto{\pgfqpoint{0.000000in}{0.000000in}}%
\pgfpathlineto{\pgfqpoint{-0.044444in}{0.000000in}}%
\pgfusepath{stroke,fill}%
}%
\begin{pgfscope}%
\pgfsys@transformshift{0.450809in}{0.326389in}%
\pgfsys@useobject{currentmarker}{}%
\end{pgfscope}%
\end{pgfscope}%
\begin{pgfscope}%
\pgfsetbuttcap%
\pgfsetroundjoin%
\definecolor{currentfill}{rgb}{0.150000,0.150000,0.150000}%
\pgfsetfillcolor{currentfill}%
\pgfsetlinewidth{0.803000pt}%
\definecolor{currentstroke}{rgb}{0.150000,0.150000,0.150000}%
\pgfsetstrokecolor{currentstroke}%
\pgfsetdash{}{0pt}%
\pgfsys@defobject{currentmarker}{\pgfqpoint{-0.044444in}{0.000000in}}{\pgfqpoint{0.000000in}{0.000000in}}{%
\pgfpathmoveto{\pgfqpoint{0.000000in}{0.000000in}}%
\pgfpathlineto{\pgfqpoint{-0.044444in}{0.000000in}}%
\pgfusepath{stroke,fill}%
}%
\begin{pgfscope}%
\pgfsys@transformshift{0.450809in}{0.740094in}%
\pgfsys@useobject{currentmarker}{}%
\end{pgfscope}%
\end{pgfscope}%
\begin{pgfscope}%
\pgfsetbuttcap%
\pgfsetroundjoin%
\definecolor{currentfill}{rgb}{0.150000,0.150000,0.150000}%
\pgfsetfillcolor{currentfill}%
\pgfsetlinewidth{0.803000pt}%
\definecolor{currentstroke}{rgb}{0.150000,0.150000,0.150000}%
\pgfsetstrokecolor{currentstroke}%
\pgfsetdash{}{0pt}%
\pgfsys@defobject{currentmarker}{\pgfqpoint{-0.044444in}{0.000000in}}{\pgfqpoint{0.000000in}{0.000000in}}{%
\pgfpathmoveto{\pgfqpoint{0.000000in}{0.000000in}}%
\pgfpathlineto{\pgfqpoint{-0.044444in}{0.000000in}}%
\pgfusepath{stroke,fill}%
}%
\begin{pgfscope}%
\pgfsys@transformshift{0.450809in}{1.153800in}%
\pgfsys@useobject{currentmarker}{}%
\end{pgfscope}%
\end{pgfscope}%
\begin{pgfscope}%
\pgfsetbuttcap%
\pgfsetroundjoin%
\definecolor{currentfill}{rgb}{0.150000,0.150000,0.150000}%
\pgfsetfillcolor{currentfill}%
\pgfsetlinewidth{0.803000pt}%
\definecolor{currentstroke}{rgb}{0.150000,0.150000,0.150000}%
\pgfsetstrokecolor{currentstroke}%
\pgfsetdash{}{0pt}%
\pgfsys@defobject{currentmarker}{\pgfqpoint{-0.044444in}{0.000000in}}{\pgfqpoint{0.000000in}{0.000000in}}{%
\pgfpathmoveto{\pgfqpoint{0.000000in}{0.000000in}}%
\pgfpathlineto{\pgfqpoint{-0.044444in}{0.000000in}}%
\pgfusepath{stroke,fill}%
}%
\begin{pgfscope}%
\pgfsys@transformshift{0.450809in}{1.567506in}%
\pgfsys@useobject{currentmarker}{}%
\end{pgfscope}%
\end{pgfscope}%
\begin{pgfscope}%
\pgfsetbuttcap%
\pgfsetroundjoin%
\definecolor{currentfill}{rgb}{0.150000,0.150000,0.150000}%
\pgfsetfillcolor{currentfill}%
\pgfsetlinewidth{0.803000pt}%
\definecolor{currentstroke}{rgb}{0.150000,0.150000,0.150000}%
\pgfsetstrokecolor{currentstroke}%
\pgfsetdash{}{0pt}%
\pgfsys@defobject{currentmarker}{\pgfqpoint{-0.044444in}{0.000000in}}{\pgfqpoint{0.000000in}{0.000000in}}{%
\pgfpathmoveto{\pgfqpoint{0.000000in}{0.000000in}}%
\pgfpathlineto{\pgfqpoint{-0.044444in}{0.000000in}}%
\pgfusepath{stroke,fill}%
}%
\begin{pgfscope}%
\pgfsys@transformshift{0.450809in}{1.981211in}%
\pgfsys@useobject{currentmarker}{}%
\end{pgfscope}%
\end{pgfscope}%
\begin{pgfscope}%
\pgfpathrectangle{\pgfqpoint{0.450809in}{0.326389in}}{\pgfqpoint{2.480000in}{1.694000in}}%
\pgfusepath{clip}%
\pgfsetbuttcap%
\pgfsetroundjoin%
\pgfsetlinewidth{1.204500pt}%
\definecolor{currentstroke}{rgb}{1.000000,0.498039,0.000000}%
\pgfsetstrokecolor{currentstroke}%
\pgfsetdash{{7.680000pt}{1.920000pt}{1.200000pt}{1.920000pt}}{0.000000pt}%
\pgfpathmoveto{\pgfqpoint{0.563536in}{1.521528in}}%
\pgfpathlineto{\pgfqpoint{1.012123in}{1.601874in}}%
\pgfpathlineto{\pgfqpoint{1.351466in}{1.631727in}}%
\pgfpathlineto{\pgfqpoint{1.690809in}{1.661396in}}%
\pgfpathlineto{\pgfqpoint{2.139396in}{1.720549in}}%
\pgfpathlineto{\pgfqpoint{2.478738in}{1.781499in}}%
\pgfpathlineto{\pgfqpoint{2.818081in}{1.844077in}}%
\pgfusepath{stroke}%
\end{pgfscope}%
\begin{pgfscope}%
\pgfpathrectangle{\pgfqpoint{0.450809in}{0.326389in}}{\pgfqpoint{2.480000in}{1.694000in}}%
\pgfusepath{clip}%
\pgfsetroundcap%
\pgfsetroundjoin%
\pgfsetlinewidth{1.204500pt}%
\definecolor{currentstroke}{rgb}{0.890196,0.101961,0.109804}%
\pgfsetstrokecolor{currentstroke}%
\pgfsetdash{}{0pt}%
\pgfpathmoveto{\pgfqpoint{0.563536in}{1.487437in}}%
\pgfpathlineto{\pgfqpoint{1.012123in}{1.579207in}}%
\pgfpathlineto{\pgfqpoint{1.351466in}{1.670978in}}%
\pgfpathlineto{\pgfqpoint{1.690809in}{1.743492in}}%
\pgfpathlineto{\pgfqpoint{2.139396in}{1.807344in}}%
\pgfpathlineto{\pgfqpoint{2.478738in}{1.883083in}}%
\pgfpathlineto{\pgfqpoint{2.818081in}{1.951573in}}%
\pgfusepath{stroke}%
\end{pgfscope}%
\begin{pgfscope}%
\pgfpathrectangle{\pgfqpoint{0.450809in}{0.326389in}}{\pgfqpoint{2.480000in}{1.694000in}}%
\pgfusepath{clip}%
\pgfsetbuttcap%
\pgfsetroundjoin%
\pgfsetlinewidth{1.204500pt}%
\definecolor{currentstroke}{rgb}{0.200000,0.627451,0.172549}%
\pgfsetstrokecolor{currentstroke}%
\pgfsetdash{{4.440000pt}{1.920000pt}}{0.000000pt}%
\pgfpathmoveto{\pgfqpoint{0.563536in}{0.575257in}}%
\pgfpathlineto{\pgfqpoint{1.012123in}{1.072994in}}%
\pgfpathlineto{\pgfqpoint{1.351466in}{1.285099in}}%
\pgfpathlineto{\pgfqpoint{1.690809in}{1.493702in}}%
\pgfpathlineto{\pgfqpoint{2.139396in}{1.650339in}}%
\pgfpathlineto{\pgfqpoint{2.478738in}{1.774474in}}%
\pgfpathlineto{\pgfqpoint{2.818081in}{1.865576in}}%
\pgfusepath{stroke}%
\end{pgfscope}%
\begin{pgfscope}%
\pgfpathrectangle{\pgfqpoint{0.450809in}{0.326389in}}{\pgfqpoint{2.480000in}{1.694000in}}%
\pgfusepath{clip}%
\pgfsetbuttcap%
\pgfsetroundjoin%
\pgfsetlinewidth{1.204500pt}%
\definecolor{currentstroke}{rgb}{0.121569,0.470588,0.705882}%
\pgfsetstrokecolor{currentstroke}%
\pgfsetdash{{1.200000pt}{1.980000pt}}{0.000000pt}%
\pgfpathmoveto{\pgfqpoint{0.563536in}{0.924834in}}%
\pgfpathlineto{\pgfqpoint{1.012123in}{1.317347in}}%
\pgfpathlineto{\pgfqpoint{1.351466in}{1.456570in}}%
\pgfpathlineto{\pgfqpoint{1.690809in}{1.610719in}}%
\pgfpathlineto{\pgfqpoint{2.139396in}{1.691064in}}%
\pgfpathlineto{\pgfqpoint{2.478738in}{1.802346in}}%
\pgfpathlineto{\pgfqpoint{2.818081in}{1.882776in}}%
\pgfusepath{stroke}%
\end{pgfscope}%
\begin{pgfscope}%
\pgfsetrectcap%
\pgfsetmiterjoin%
\pgfsetlinewidth{1.003750pt}%
\definecolor{currentstroke}{rgb}{0.150000,0.150000,0.150000}%
\pgfsetstrokecolor{currentstroke}%
\pgfsetdash{}{0pt}%
\pgfpathmoveto{\pgfqpoint{0.450809in}{0.326389in}}%
\pgfpathlineto{\pgfqpoint{0.450809in}{2.020389in}}%
\pgfusepath{stroke}%
\end{pgfscope}%
\begin{pgfscope}%
\pgfsetrectcap%
\pgfsetmiterjoin%
\pgfsetlinewidth{1.003750pt}%
\definecolor{currentstroke}{rgb}{0.150000,0.150000,0.150000}%
\pgfsetstrokecolor{currentstroke}%
\pgfsetdash{}{0pt}%
\pgfpathmoveto{\pgfqpoint{2.930809in}{0.326389in}}%
\pgfpathlineto{\pgfqpoint{2.930809in}{2.020389in}}%
\pgfusepath{stroke}%
\end{pgfscope}%
\begin{pgfscope}%
\pgfsetrectcap%
\pgfsetmiterjoin%
\pgfsetlinewidth{1.003750pt}%
\definecolor{currentstroke}{rgb}{0.150000,0.150000,0.150000}%
\pgfsetstrokecolor{currentstroke}%
\pgfsetdash{}{0pt}%
\pgfpathmoveto{\pgfqpoint{0.450809in}{0.326389in}}%
\pgfpathlineto{\pgfqpoint{2.930809in}{0.326389in}}%
\pgfusepath{stroke}%
\end{pgfscope}%
\begin{pgfscope}%
\pgfsetrectcap%
\pgfsetmiterjoin%
\pgfsetlinewidth{1.003750pt}%
\definecolor{currentstroke}{rgb}{0.150000,0.150000,0.150000}%
\pgfsetstrokecolor{currentstroke}%
\pgfsetdash{}{0pt}%
\pgfpathmoveto{\pgfqpoint{0.450809in}{2.020389in}}%
\pgfpathlineto{\pgfqpoint{2.930809in}{2.020389in}}%
\pgfusepath{stroke}%
\end{pgfscope}%
\begin{pgfscope}%
\pgfsetbuttcap%
\pgfsetroundjoin%
\pgfsetlinewidth{1.204500pt}%
\definecolor{currentstroke}{rgb}{1.000000,0.498039,0.000000}%
\pgfsetstrokecolor{currentstroke}%
\pgfsetdash{{7.680000pt}{1.920000pt}{1.200000pt}{1.920000pt}}{0.000000pt}%
\pgfpathmoveto{\pgfqpoint{1.816819in}{0.971250in}}%
\pgfpathlineto{\pgfqpoint{2.061263in}{0.971250in}}%
\pgfusepath{stroke}%
\end{pgfscope}%
\begin{pgfscope}%
\definecolor{textcolor}{rgb}{0.150000,0.150000,0.150000}%
\pgfsetstrokecolor{textcolor}%
\pgfsetfillcolor{textcolor}%
\pgftext[x=2.159041in,y=0.928472in,left,base]{\color{textcolor}\sffamily\fontsize{8.800000}{10.560000}\selectfont Norma}%
\end{pgfscope}%
\begin{pgfscope}%
\pgfsetroundcap%
\pgfsetroundjoin%
\pgfsetlinewidth{1.204500pt}%
\definecolor{currentstroke}{rgb}{0.890196,0.101961,0.109804}%
\pgfsetstrokecolor{currentstroke}%
\pgfsetdash{}{0pt}%
\pgfpathmoveto{\pgfqpoint{1.816819in}{0.799028in}}%
\pgfpathlineto{\pgfqpoint{2.061263in}{0.799028in}}%
\pgfusepath{stroke}%
\end{pgfscope}%
\begin{pgfscope}%
\definecolor{textcolor}{rgb}{0.150000,0.150000,0.150000}%
\pgfsetstrokecolor{textcolor}%
\pgfsetfillcolor{textcolor}%
\pgftext[x=2.159041in,y=0.756250in,left,base]{\color{textcolor}\sffamily\fontsize{8.800000}{10.560000}\selectfont cSMTiser\textsubscript{+LM}}%
\end{pgfscope}%
\begin{pgfscope}%
\pgfsetbuttcap%
\pgfsetroundjoin%
\pgfsetlinewidth{1.204500pt}%
\definecolor{currentstroke}{rgb}{0.200000,0.627451,0.172549}%
\pgfsetstrokecolor{currentstroke}%
\pgfsetdash{{4.440000pt}{1.920000pt}}{0.000000pt}%
\pgfpathmoveto{\pgfqpoint{1.816819in}{0.626806in}}%
\pgfpathlineto{\pgfqpoint{2.061263in}{0.626806in}}%
\pgfusepath{stroke}%
\end{pgfscope}%
\begin{pgfscope}%
\definecolor{textcolor}{rgb}{0.150000,0.150000,0.150000}%
\pgfsetstrokecolor{textcolor}%
\pgfsetfillcolor{textcolor}%
\pgftext[x=2.159041in,y=0.584028in,left,base]{\color{textcolor}\sffamily\fontsize{8.800000}{10.560000}\selectfont NMT-1}%
\end{pgfscope}%
\begin{pgfscope}%
\pgfsetbuttcap%
\pgfsetroundjoin%
\pgfsetlinewidth{1.204500pt}%
\definecolor{currentstroke}{rgb}{0.121569,0.470588,0.705882}%
\pgfsetstrokecolor{currentstroke}%
\pgfsetdash{{1.200000pt}{1.980000pt}}{0.000000pt}%
\pgfpathmoveto{\pgfqpoint{1.816819in}{0.454583in}}%
\pgfpathlineto{\pgfqpoint{2.061263in}{0.454583in}}%
\pgfusepath{stroke}%
\end{pgfscope}%
\begin{pgfscope}%
\definecolor{textcolor}{rgb}{0.150000,0.150000,0.150000}%
\pgfsetstrokecolor{textcolor}%
\pgfsetfillcolor{textcolor}%
\pgftext[x=2.159041in,y=0.411806in,left,base]{\color{textcolor}\sffamily\fontsize{8.800000}{10.560000}\selectfont NMT-2}%
\end{pgfscope}%
\end{pgfpicture}%
\makeatother%
\endgroup%

%% file: curves_slovene-goo300k-bohoric.pgf
\begingroup%
\makeatletter%
\begin{pgfpicture}%
\pgfpathrectangle{\pgfpointorigin}{\pgfqpoint{3.078809in}{2.168389in}}%
\pgfusepath{use as bounding box, clip}%
\begin{pgfscope}%
\pgfsetbuttcap%
\pgfsetmiterjoin%
\definecolor{currentfill}{rgb}{1.000000,1.000000,1.000000}%
\pgfsetfillcolor{currentfill}%
\pgfsetlinewidth{0.000000pt}%
\definecolor{currentstroke}{rgb}{1.000000,1.000000,1.000000}%
\pgfsetstrokecolor{currentstroke}%
\pgfsetdash{}{0pt}%
\pgfpathmoveto{\pgfqpoint{0.000000in}{0.000000in}}%
\pgfpathlineto{\pgfqpoint{3.078809in}{0.000000in}}%
\pgfpathlineto{\pgfqpoint{3.078809in}{2.168389in}}%
\pgfpathlineto{\pgfqpoint{0.000000in}{2.168389in}}%
\pgfpathclose%
\pgfusepath{fill}%
\end{pgfscope}%
\begin{pgfscope}%
\pgfsetbuttcap%
\pgfsetmiterjoin%
\definecolor{currentfill}{rgb}{1.000000,1.000000,1.000000}%
\pgfsetfillcolor{currentfill}%
\pgfsetlinewidth{0.000000pt}%
\definecolor{currentstroke}{rgb}{0.000000,0.000000,0.000000}%
\pgfsetstrokecolor{currentstroke}%
\pgfsetstrokeopacity{0.000000}%
\pgfsetdash{}{0pt}%
\pgfpathmoveto{\pgfqpoint{0.450809in}{0.326389in}}%
\pgfpathlineto{\pgfqpoint{2.930809in}{0.326389in}}%
\pgfpathlineto{\pgfqpoint{2.930809in}{2.020389in}}%
\pgfpathlineto{\pgfqpoint{0.450809in}{2.020389in}}%
\pgfpathclose%
\pgfusepath{fill}%
\end{pgfscope}%
\begin{pgfscope}%
\pgfsetbuttcap%
\pgfsetroundjoin%
\definecolor{currentfill}{rgb}{0.150000,0.150000,0.150000}%
\pgfsetfillcolor{currentfill}%
\pgfsetlinewidth{1.003750pt}%
\definecolor{currentstroke}{rgb}{0.150000,0.150000,0.150000}%
\pgfsetstrokecolor{currentstroke}%
\pgfsetdash{}{0pt}%
\pgfsys@defobject{currentmarker}{\pgfqpoint{0.000000in}{-0.066667in}}{\pgfqpoint{0.000000in}{0.000000in}}{%
\pgfpathmoveto{\pgfqpoint{0.000000in}{0.000000in}}%
\pgfpathlineto{\pgfqpoint{0.000000in}{-0.066667in}}%
\pgfusepath{stroke,fill}%
}%
\begin{pgfscope}%
\pgfsys@transformshift{0.563536in}{0.326389in}%
\pgfsys@useobject{currentmarker}{}%
\end{pgfscope}%
\end{pgfscope}%
\begin{pgfscope}%
\definecolor{textcolor}{rgb}{0.150000,0.150000,0.150000}%
\pgfsetstrokecolor{textcolor}%
\pgfsetfillcolor{textcolor}%
\pgftext[x=0.563536in,y=0.211111in,,top]{\color{textcolor}\sffamily\fontsize{8.800000}{10.560000}\selectfont 100}%
\end{pgfscope}%
\begin{pgfscope}%
\pgfsetbuttcap%
\pgfsetroundjoin%
\definecolor{currentfill}{rgb}{0.150000,0.150000,0.150000}%
\pgfsetfillcolor{currentfill}%
\pgfsetlinewidth{1.003750pt}%
\definecolor{currentstroke}{rgb}{0.150000,0.150000,0.150000}%
\pgfsetstrokecolor{currentstroke}%
\pgfsetdash{}{0pt}%
\pgfsys@defobject{currentmarker}{\pgfqpoint{0.000000in}{-0.066667in}}{\pgfqpoint{0.000000in}{0.000000in}}{%
\pgfpathmoveto{\pgfqpoint{0.000000in}{0.000000in}}%
\pgfpathlineto{\pgfqpoint{0.000000in}{-0.066667in}}%
\pgfusepath{stroke,fill}%
}%
\begin{pgfscope}%
\pgfsys@transformshift{0.937679in}{0.326389in}%
\pgfsys@useobject{currentmarker}{}%
\end{pgfscope}%
\end{pgfscope}%
\begin{pgfscope}%
\definecolor{textcolor}{rgb}{0.150000,0.150000,0.150000}%
\pgfsetstrokecolor{textcolor}%
\pgfsetfillcolor{textcolor}%
\pgftext[x=0.937679in,y=0.211111in,,top]{\color{textcolor}\sffamily\fontsize{8.800000}{10.560000}\selectfont 250}%
\end{pgfscope}%
\begin{pgfscope}%
\pgfsetbuttcap%
\pgfsetroundjoin%
\definecolor{currentfill}{rgb}{0.150000,0.150000,0.150000}%
\pgfsetfillcolor{currentfill}%
\pgfsetlinewidth{1.003750pt}%
\definecolor{currentstroke}{rgb}{0.150000,0.150000,0.150000}%
\pgfsetstrokecolor{currentstroke}%
\pgfsetdash{}{0pt}%
\pgfsys@defobject{currentmarker}{\pgfqpoint{0.000000in}{-0.066667in}}{\pgfqpoint{0.000000in}{0.000000in}}{%
\pgfpathmoveto{\pgfqpoint{0.000000in}{0.000000in}}%
\pgfpathlineto{\pgfqpoint{0.000000in}{-0.066667in}}%
\pgfusepath{stroke,fill}%
}%
\begin{pgfscope}%
\pgfsys@transformshift{1.220708in}{0.326389in}%
\pgfsys@useobject{currentmarker}{}%
\end{pgfscope}%
\end{pgfscope}%
\begin{pgfscope}%
\definecolor{textcolor}{rgb}{0.150000,0.150000,0.150000}%
\pgfsetstrokecolor{textcolor}%
\pgfsetfillcolor{textcolor}%
\pgftext[x=1.220708in,y=0.211111in,,top]{\color{textcolor}\sffamily\fontsize{8.800000}{10.560000}\selectfont 500}%
\end{pgfscope}%
\begin{pgfscope}%
\pgfsetbuttcap%
\pgfsetroundjoin%
\definecolor{currentfill}{rgb}{0.150000,0.150000,0.150000}%
\pgfsetfillcolor{currentfill}%
\pgfsetlinewidth{1.003750pt}%
\definecolor{currentstroke}{rgb}{0.150000,0.150000,0.150000}%
\pgfsetstrokecolor{currentstroke}%
\pgfsetdash{}{0pt}%
\pgfsys@defobject{currentmarker}{\pgfqpoint{0.000000in}{-0.066667in}}{\pgfqpoint{0.000000in}{0.000000in}}{%
\pgfpathmoveto{\pgfqpoint{0.000000in}{0.000000in}}%
\pgfpathlineto{\pgfqpoint{0.000000in}{-0.066667in}}%
\pgfusepath{stroke,fill}%
}%
\begin{pgfscope}%
\pgfsys@transformshift{1.503737in}{0.326389in}%
\pgfsys@useobject{currentmarker}{}%
\end{pgfscope}%
\end{pgfscope}%
\begin{pgfscope}%
\definecolor{textcolor}{rgb}{0.150000,0.150000,0.150000}%
\pgfsetstrokecolor{textcolor}%
\pgfsetfillcolor{textcolor}%
\pgftext[x=1.503737in,y=0.211111in,,top]{\color{textcolor}\sffamily\fontsize{8.800000}{10.560000}\selectfont 1k}%
\end{pgfscope}%
\begin{pgfscope}%
\pgfsetbuttcap%
\pgfsetroundjoin%
\definecolor{currentfill}{rgb}{0.150000,0.150000,0.150000}%
\pgfsetfillcolor{currentfill}%
\pgfsetlinewidth{1.003750pt}%
\definecolor{currentstroke}{rgb}{0.150000,0.150000,0.150000}%
\pgfsetstrokecolor{currentstroke}%
\pgfsetdash{}{0pt}%
\pgfsys@defobject{currentmarker}{\pgfqpoint{0.000000in}{-0.066667in}}{\pgfqpoint{0.000000in}{0.000000in}}{%
\pgfpathmoveto{\pgfqpoint{0.000000in}{0.000000in}}%
\pgfpathlineto{\pgfqpoint{0.000000in}{-0.066667in}}%
\pgfusepath{stroke,fill}%
}%
\begin{pgfscope}%
\pgfsys@transformshift{1.877880in}{0.326389in}%
\pgfsys@useobject{currentmarker}{}%
\end{pgfscope}%
\end{pgfscope}%
\begin{pgfscope}%
\definecolor{textcolor}{rgb}{0.150000,0.150000,0.150000}%
\pgfsetstrokecolor{textcolor}%
\pgfsetfillcolor{textcolor}%
\pgftext[x=1.877880in,y=0.211111in,,top]{\color{textcolor}\sffamily\fontsize{8.800000}{10.560000}\selectfont 2.5k}%
\end{pgfscope}%
\begin{pgfscope}%
\pgfsetbuttcap%
\pgfsetroundjoin%
\definecolor{currentfill}{rgb}{0.150000,0.150000,0.150000}%
\pgfsetfillcolor{currentfill}%
\pgfsetlinewidth{1.003750pt}%
\definecolor{currentstroke}{rgb}{0.150000,0.150000,0.150000}%
\pgfsetstrokecolor{currentstroke}%
\pgfsetdash{}{0pt}%
\pgfsys@defobject{currentmarker}{\pgfqpoint{0.000000in}{-0.066667in}}{\pgfqpoint{0.000000in}{0.000000in}}{%
\pgfpathmoveto{\pgfqpoint{0.000000in}{0.000000in}}%
\pgfpathlineto{\pgfqpoint{0.000000in}{-0.066667in}}%
\pgfusepath{stroke,fill}%
}%
\begin{pgfscope}%
\pgfsys@transformshift{2.160909in}{0.326389in}%
\pgfsys@useobject{currentmarker}{}%
\end{pgfscope}%
\end{pgfscope}%
\begin{pgfscope}%
\definecolor{textcolor}{rgb}{0.150000,0.150000,0.150000}%
\pgfsetstrokecolor{textcolor}%
\pgfsetfillcolor{textcolor}%
\pgftext[x=2.160909in,y=0.211111in,,top]{\color{textcolor}\sffamily\fontsize{8.800000}{10.560000}\selectfont 5k}%
\end{pgfscope}%
\begin{pgfscope}%
\pgfsetbuttcap%
\pgfsetroundjoin%
\definecolor{currentfill}{rgb}{0.150000,0.150000,0.150000}%
\pgfsetfillcolor{currentfill}%
\pgfsetlinewidth{1.003750pt}%
\definecolor{currentstroke}{rgb}{0.150000,0.150000,0.150000}%
\pgfsetstrokecolor{currentstroke}%
\pgfsetdash{}{0pt}%
\pgfsys@defobject{currentmarker}{\pgfqpoint{0.000000in}{-0.066667in}}{\pgfqpoint{0.000000in}{0.000000in}}{%
\pgfpathmoveto{\pgfqpoint{0.000000in}{0.000000in}}%
\pgfpathlineto{\pgfqpoint{0.000000in}{-0.066667in}}%
\pgfusepath{stroke,fill}%
}%
\begin{pgfscope}%
\pgfsys@transformshift{2.443938in}{0.326389in}%
\pgfsys@useobject{currentmarker}{}%
\end{pgfscope}%
\end{pgfscope}%
\begin{pgfscope}%
\definecolor{textcolor}{rgb}{0.150000,0.150000,0.150000}%
\pgfsetstrokecolor{textcolor}%
\pgfsetfillcolor{textcolor}%
\pgftext[x=2.443938in,y=0.211111in,,top]{\color{textcolor}\sffamily\fontsize{8.800000}{10.560000}\selectfont 10k}%
\end{pgfscope}%
\begin{pgfscope}%
\pgfsetbuttcap%
\pgfsetroundjoin%
\definecolor{currentfill}{rgb}{0.150000,0.150000,0.150000}%
\pgfsetfillcolor{currentfill}%
\pgfsetlinewidth{1.003750pt}%
\definecolor{currentstroke}{rgb}{0.150000,0.150000,0.150000}%
\pgfsetstrokecolor{currentstroke}%
\pgfsetdash{}{0pt}%
\pgfsys@defobject{currentmarker}{\pgfqpoint{0.000000in}{-0.066667in}}{\pgfqpoint{0.000000in}{0.000000in}}{%
\pgfpathmoveto{\pgfqpoint{0.000000in}{0.000000in}}%
\pgfpathlineto{\pgfqpoint{0.000000in}{-0.066667in}}%
\pgfusepath{stroke,fill}%
}%
\begin{pgfscope}%
\pgfsys@transformshift{2.818081in}{0.326389in}%
\pgfsys@useobject{currentmarker}{}%
\end{pgfscope}%
\end{pgfscope}%
\begin{pgfscope}%
\definecolor{textcolor}{rgb}{0.150000,0.150000,0.150000}%
\pgfsetstrokecolor{textcolor}%
\pgfsetfillcolor{textcolor}%
\pgftext[x=2.818081in,y=0.211111in,,top]{\color{textcolor}\sffamily\fontsize{8.800000}{10.560000}\selectfont 25k}%
\end{pgfscope}%
\begin{pgfscope}%
\pgfsetbuttcap%
\pgfsetroundjoin%
\definecolor{currentfill}{rgb}{0.150000,0.150000,0.150000}%
\pgfsetfillcolor{currentfill}%
\pgfsetlinewidth{1.003750pt}%
\definecolor{currentstroke}{rgb}{0.150000,0.150000,0.150000}%
\pgfsetstrokecolor{currentstroke}%
\pgfsetdash{}{0pt}%
\pgfsys@defobject{currentmarker}{\pgfqpoint{-0.066667in}{0.000000in}}{\pgfqpoint{0.000000in}{0.000000in}}{%
\pgfpathmoveto{\pgfqpoint{0.000000in}{0.000000in}}%
\pgfpathlineto{\pgfqpoint{-0.066667in}{0.000000in}}%
\pgfusepath{stroke,fill}%
}%
\begin{pgfscope}%
\pgfsys@transformshift{0.450809in}{0.526996in}%
\pgfsys@useobject{currentmarker}{}%
\end{pgfscope}%
\end{pgfscope}%
\begin{pgfscope}%
\definecolor{textcolor}{rgb}{0.150000,0.150000,0.150000}%
\pgfsetstrokecolor{textcolor}%
\pgfsetfillcolor{textcolor}%
\pgftext[x=0.100000in,y=0.483593in,left,base]{\color{textcolor}\sffamily\fontsize{8.800000}{10.560000}\selectfont 20\%}%
\end{pgfscope}%
\begin{pgfscope}%
\pgfsetbuttcap%
\pgfsetroundjoin%
\definecolor{currentfill}{rgb}{0.150000,0.150000,0.150000}%
\pgfsetfillcolor{currentfill}%
\pgfsetlinewidth{1.003750pt}%
\definecolor{currentstroke}{rgb}{0.150000,0.150000,0.150000}%
\pgfsetstrokecolor{currentstroke}%
\pgfsetdash{}{0pt}%
\pgfsys@defobject{currentmarker}{\pgfqpoint{-0.066667in}{0.000000in}}{\pgfqpoint{0.000000in}{0.000000in}}{%
\pgfpathmoveto{\pgfqpoint{0.000000in}{0.000000in}}%
\pgfpathlineto{\pgfqpoint{-0.066667in}{0.000000in}}%
\pgfusepath{stroke,fill}%
}%
\begin{pgfscope}%
\pgfsys@transformshift{0.450809in}{0.928210in}%
\pgfsys@useobject{currentmarker}{}%
\end{pgfscope}%
\end{pgfscope}%
\begin{pgfscope}%
\definecolor{textcolor}{rgb}{0.150000,0.150000,0.150000}%
\pgfsetstrokecolor{textcolor}%
\pgfsetfillcolor{textcolor}%
\pgftext[x=0.100000in,y=0.884807in,left,base]{\color{textcolor}\sffamily\fontsize{8.800000}{10.560000}\selectfont 40\%}%
\end{pgfscope}%
\begin{pgfscope}%
\pgfsetbuttcap%
\pgfsetroundjoin%
\definecolor{currentfill}{rgb}{0.150000,0.150000,0.150000}%
\pgfsetfillcolor{currentfill}%
\pgfsetlinewidth{1.003750pt}%
\definecolor{currentstroke}{rgb}{0.150000,0.150000,0.150000}%
\pgfsetstrokecolor{currentstroke}%
\pgfsetdash{}{0pt}%
\pgfsys@defobject{currentmarker}{\pgfqpoint{-0.066667in}{0.000000in}}{\pgfqpoint{0.000000in}{0.000000in}}{%
\pgfpathmoveto{\pgfqpoint{0.000000in}{0.000000in}}%
\pgfpathlineto{\pgfqpoint{-0.066667in}{0.000000in}}%
\pgfusepath{stroke,fill}%
}%
\begin{pgfscope}%
\pgfsys@transformshift{0.450809in}{1.329424in}%
\pgfsys@useobject{currentmarker}{}%
\end{pgfscope}%
\end{pgfscope}%
\begin{pgfscope}%
\definecolor{textcolor}{rgb}{0.150000,0.150000,0.150000}%
\pgfsetstrokecolor{textcolor}%
\pgfsetfillcolor{textcolor}%
\pgftext[x=0.100000in,y=1.286022in,left,base]{\color{textcolor}\sffamily\fontsize{8.800000}{10.560000}\selectfont 60\%}%
\end{pgfscope}%
\begin{pgfscope}%
\pgfsetbuttcap%
\pgfsetroundjoin%
\definecolor{currentfill}{rgb}{0.150000,0.150000,0.150000}%
\pgfsetfillcolor{currentfill}%
\pgfsetlinewidth{1.003750pt}%
\definecolor{currentstroke}{rgb}{0.150000,0.150000,0.150000}%
\pgfsetstrokecolor{currentstroke}%
\pgfsetdash{}{0pt}%
\pgfsys@defobject{currentmarker}{\pgfqpoint{-0.066667in}{0.000000in}}{\pgfqpoint{0.000000in}{0.000000in}}{%
\pgfpathmoveto{\pgfqpoint{0.000000in}{0.000000in}}%
\pgfpathlineto{\pgfqpoint{-0.066667in}{0.000000in}}%
\pgfusepath{stroke,fill}%
}%
\begin{pgfscope}%
\pgfsys@transformshift{0.450809in}{1.730639in}%
\pgfsys@useobject{currentmarker}{}%
\end{pgfscope}%
\end{pgfscope}%
\begin{pgfscope}%
\definecolor{textcolor}{rgb}{0.150000,0.150000,0.150000}%
\pgfsetstrokecolor{textcolor}%
\pgfsetfillcolor{textcolor}%
\pgftext[x=0.100000in,y=1.687236in,left,base]{\color{textcolor}\sffamily\fontsize{8.800000}{10.560000}\selectfont 80\%}%
\end{pgfscope}%
\begin{pgfscope}%
\pgfsetbuttcap%
\pgfsetroundjoin%
\definecolor{currentfill}{rgb}{0.150000,0.150000,0.150000}%
\pgfsetfillcolor{currentfill}%
\pgfsetlinewidth{0.803000pt}%
\definecolor{currentstroke}{rgb}{0.150000,0.150000,0.150000}%
\pgfsetstrokecolor{currentstroke}%
\pgfsetdash{}{0pt}%
\pgfsys@defobject{currentmarker}{\pgfqpoint{-0.044444in}{0.000000in}}{\pgfqpoint{0.000000in}{0.000000in}}{%
\pgfpathmoveto{\pgfqpoint{0.000000in}{0.000000in}}%
\pgfpathlineto{\pgfqpoint{-0.044444in}{0.000000in}}%
\pgfusepath{stroke,fill}%
}%
\begin{pgfscope}%
\pgfsys@transformshift{0.450809in}{0.326389in}%
\pgfsys@useobject{currentmarker}{}%
\end{pgfscope}%
\end{pgfscope}%
\begin{pgfscope}%
\pgfsetbuttcap%
\pgfsetroundjoin%
\definecolor{currentfill}{rgb}{0.150000,0.150000,0.150000}%
\pgfsetfillcolor{currentfill}%
\pgfsetlinewidth{0.803000pt}%
\definecolor{currentstroke}{rgb}{0.150000,0.150000,0.150000}%
\pgfsetstrokecolor{currentstroke}%
\pgfsetdash{}{0pt}%
\pgfsys@defobject{currentmarker}{\pgfqpoint{-0.044444in}{0.000000in}}{\pgfqpoint{0.000000in}{0.000000in}}{%
\pgfpathmoveto{\pgfqpoint{0.000000in}{0.000000in}}%
\pgfpathlineto{\pgfqpoint{-0.044444in}{0.000000in}}%
\pgfusepath{stroke,fill}%
}%
\begin{pgfscope}%
\pgfsys@transformshift{0.450809in}{0.727603in}%
\pgfsys@useobject{currentmarker}{}%
\end{pgfscope}%
\end{pgfscope}%
\begin{pgfscope}%
\pgfsetbuttcap%
\pgfsetroundjoin%
\definecolor{currentfill}{rgb}{0.150000,0.150000,0.150000}%
\pgfsetfillcolor{currentfill}%
\pgfsetlinewidth{0.803000pt}%
\definecolor{currentstroke}{rgb}{0.150000,0.150000,0.150000}%
\pgfsetstrokecolor{currentstroke}%
\pgfsetdash{}{0pt}%
\pgfsys@defobject{currentmarker}{\pgfqpoint{-0.044444in}{0.000000in}}{\pgfqpoint{0.000000in}{0.000000in}}{%
\pgfpathmoveto{\pgfqpoint{0.000000in}{0.000000in}}%
\pgfpathlineto{\pgfqpoint{-0.044444in}{0.000000in}}%
\pgfusepath{stroke,fill}%
}%
\begin{pgfscope}%
\pgfsys@transformshift{0.450809in}{1.128817in}%
\pgfsys@useobject{currentmarker}{}%
\end{pgfscope}%
\end{pgfscope}%
\begin{pgfscope}%
\pgfsetbuttcap%
\pgfsetroundjoin%
\definecolor{currentfill}{rgb}{0.150000,0.150000,0.150000}%
\pgfsetfillcolor{currentfill}%
\pgfsetlinewidth{0.803000pt}%
\definecolor{currentstroke}{rgb}{0.150000,0.150000,0.150000}%
\pgfsetstrokecolor{currentstroke}%
\pgfsetdash{}{0pt}%
\pgfsys@defobject{currentmarker}{\pgfqpoint{-0.044444in}{0.000000in}}{\pgfqpoint{0.000000in}{0.000000in}}{%
\pgfpathmoveto{\pgfqpoint{0.000000in}{0.000000in}}%
\pgfpathlineto{\pgfqpoint{-0.044444in}{0.000000in}}%
\pgfusepath{stroke,fill}%
}%
\begin{pgfscope}%
\pgfsys@transformshift{0.450809in}{1.530031in}%
\pgfsys@useobject{currentmarker}{}%
\end{pgfscope}%
\end{pgfscope}%
\begin{pgfscope}%
\pgfsetbuttcap%
\pgfsetroundjoin%
\definecolor{currentfill}{rgb}{0.150000,0.150000,0.150000}%
\pgfsetfillcolor{currentfill}%
\pgfsetlinewidth{0.803000pt}%
\definecolor{currentstroke}{rgb}{0.150000,0.150000,0.150000}%
\pgfsetstrokecolor{currentstroke}%
\pgfsetdash{}{0pt}%
\pgfsys@defobject{currentmarker}{\pgfqpoint{-0.044444in}{0.000000in}}{\pgfqpoint{0.000000in}{0.000000in}}{%
\pgfpathmoveto{\pgfqpoint{0.000000in}{0.000000in}}%
\pgfpathlineto{\pgfqpoint{-0.044444in}{0.000000in}}%
\pgfusepath{stroke,fill}%
}%
\begin{pgfscope}%
\pgfsys@transformshift{0.450809in}{1.931246in}%
\pgfsys@useobject{currentmarker}{}%
\end{pgfscope}%
\end{pgfscope}%
\begin{pgfscope}%
\pgfpathrectangle{\pgfqpoint{0.450809in}{0.326389in}}{\pgfqpoint{2.480000in}{1.694000in}}%
\pgfusepath{clip}%
\pgfsetbuttcap%
\pgfsetroundjoin%
\pgfsetlinewidth{1.204500pt}%
\definecolor{currentstroke}{rgb}{1.000000,0.498039,0.000000}%
\pgfsetstrokecolor{currentstroke}%
\pgfsetdash{{7.680000pt}{1.920000pt}{1.200000pt}{1.920000pt}}{0.000000pt}%
\pgfpathmoveto{\pgfqpoint{0.563536in}{1.562006in}}%
\pgfpathlineto{\pgfqpoint{0.937679in}{1.615927in}}%
\pgfpathlineto{\pgfqpoint{1.220708in}{1.670501in}}%
\pgfpathlineto{\pgfqpoint{1.503737in}{1.709963in}}%
\pgfpathlineto{\pgfqpoint{1.877880in}{1.758698in}}%
\pgfpathlineto{\pgfqpoint{2.160909in}{1.781846in}}%
\pgfpathlineto{\pgfqpoint{2.443938in}{1.817485in}}%
\pgfpathlineto{\pgfqpoint{2.818081in}{1.867971in}}%
\pgfusepath{stroke}%
\end{pgfscope}%
\begin{pgfscope}%
\pgfpathrectangle{\pgfqpoint{0.450809in}{0.326389in}}{\pgfqpoint{2.480000in}{1.694000in}}%
\pgfusepath{clip}%
\pgfsetroundcap%
\pgfsetroundjoin%
\pgfsetlinewidth{1.204500pt}%
\definecolor{currentstroke}{rgb}{0.890196,0.101961,0.109804}%
\pgfsetstrokecolor{currentstroke}%
\pgfsetdash{}{0pt}%
\pgfpathmoveto{\pgfqpoint{0.563536in}{1.484353in}}%
\pgfpathlineto{\pgfqpoint{0.937679in}{1.645464in}}%
\pgfpathlineto{\pgfqpoint{1.220708in}{1.723598in}}%
\pgfpathlineto{\pgfqpoint{1.503737in}{1.788818in}}%
\pgfpathlineto{\pgfqpoint{1.877880in}{1.863415in}}%
\pgfpathlineto{\pgfqpoint{2.160909in}{1.892986in}}%
\pgfpathlineto{\pgfqpoint{2.443938in}{1.923075in}}%
\pgfpathlineto{\pgfqpoint{2.818081in}{1.952230in}}%
\pgfusepath{stroke}%
\end{pgfscope}%
\begin{pgfscope}%
\pgfpathrectangle{\pgfqpoint{0.450809in}{0.326389in}}{\pgfqpoint{2.480000in}{1.694000in}}%
\pgfusepath{clip}%
\pgfsetbuttcap%
\pgfsetroundjoin%
\pgfsetlinewidth{1.204500pt}%
\definecolor{currentstroke}{rgb}{0.200000,0.627451,0.172549}%
\pgfsetstrokecolor{currentstroke}%
\pgfsetdash{{4.440000pt}{1.920000pt}}{0.000000pt}%
\pgfpathmoveto{\pgfqpoint{0.563536in}{0.589057in}}%
\pgfpathlineto{\pgfqpoint{0.937679in}{1.209424in}}%
\pgfpathlineto{\pgfqpoint{1.220708in}{1.446814in}}%
\pgfpathlineto{\pgfqpoint{1.503737in}{1.606792in}}%
\pgfpathlineto{\pgfqpoint{1.877880in}{1.713707in}}%
\pgfpathlineto{\pgfqpoint{2.160909in}{1.806368in}}%
\pgfpathlineto{\pgfqpoint{2.443938in}{1.859271in}}%
\pgfpathlineto{\pgfqpoint{2.818081in}{1.920175in}}%
\pgfusepath{stroke}%
\end{pgfscope}%
\begin{pgfscope}%
\pgfpathrectangle{\pgfqpoint{0.450809in}{0.326389in}}{\pgfqpoint{2.480000in}{1.694000in}}%
\pgfusepath{clip}%
\pgfsetbuttcap%
\pgfsetroundjoin%
\pgfsetlinewidth{1.204500pt}%
\definecolor{currentstroke}{rgb}{0.121569,0.470588,0.705882}%
\pgfsetstrokecolor{currentstroke}%
\pgfsetdash{{1.200000pt}{1.980000pt}}{0.000000pt}%
\pgfpathmoveto{\pgfqpoint{0.563536in}{0.951942in}}%
\pgfpathlineto{\pgfqpoint{0.937679in}{1.408623in}}%
\pgfpathlineto{\pgfqpoint{1.220708in}{1.585223in}}%
\pgfpathlineto{\pgfqpoint{1.503737in}{1.693272in}}%
\pgfpathlineto{\pgfqpoint{1.877880in}{1.757599in}}%
\pgfpathlineto{\pgfqpoint{2.160909in}{1.817221in}}%
\pgfpathlineto{\pgfqpoint{2.443938in}{1.871673in}}%
\pgfpathlineto{\pgfqpoint{2.818081in}{1.926586in}}%
\pgfusepath{stroke}%
\end{pgfscope}%
\begin{pgfscope}%
\pgfsetrectcap%
\pgfsetmiterjoin%
\pgfsetlinewidth{1.003750pt}%
\definecolor{currentstroke}{rgb}{0.150000,0.150000,0.150000}%
\pgfsetstrokecolor{currentstroke}%
\pgfsetdash{}{0pt}%
\pgfpathmoveto{\pgfqpoint{0.450809in}{0.326389in}}%
\pgfpathlineto{\pgfqpoint{0.450809in}{2.020389in}}%
\pgfusepath{stroke}%
\end{pgfscope}%
\begin{pgfscope}%
\pgfsetrectcap%
\pgfsetmiterjoin%
\pgfsetlinewidth{1.003750pt}%
\definecolor{currentstroke}{rgb}{0.150000,0.150000,0.150000}%
\pgfsetstrokecolor{currentstroke}%
\pgfsetdash{}{0pt}%
\pgfpathmoveto{\pgfqpoint{2.930809in}{0.326389in}}%
\pgfpathlineto{\pgfqpoint{2.930809in}{2.020389in}}%
\pgfusepath{stroke}%
\end{pgfscope}%
\begin{pgfscope}%
\pgfsetrectcap%
\pgfsetmiterjoin%
\pgfsetlinewidth{1.003750pt}%
\definecolor{currentstroke}{rgb}{0.150000,0.150000,0.150000}%
\pgfsetstrokecolor{currentstroke}%
\pgfsetdash{}{0pt}%
\pgfpathmoveto{\pgfqpoint{0.450809in}{0.326389in}}%
\pgfpathlineto{\pgfqpoint{2.930809in}{0.326389in}}%
\pgfusepath{stroke}%
\end{pgfscope}%
\begin{pgfscope}%
\pgfsetrectcap%
\pgfsetmiterjoin%
\pgfsetlinewidth{1.003750pt}%
\definecolor{currentstroke}{rgb}{0.150000,0.150000,0.150000}%
\pgfsetstrokecolor{currentstroke}%
\pgfsetdash{}{0pt}%
\pgfpathmoveto{\pgfqpoint{0.450809in}{2.020389in}}%
\pgfpathlineto{\pgfqpoint{2.930809in}{2.020389in}}%
\pgfusepath{stroke}%
\end{pgfscope}%
\begin{pgfscope}%
\pgfsetbuttcap%
\pgfsetroundjoin%
\pgfsetlinewidth{1.204500pt}%
\definecolor{currentstroke}{rgb}{1.000000,0.498039,0.000000}%
\pgfsetstrokecolor{currentstroke}%
\pgfsetdash{{7.680000pt}{1.920000pt}{1.200000pt}{1.920000pt}}{0.000000pt}%
\pgfpathmoveto{\pgfqpoint{1.816819in}{0.971250in}}%
\pgfpathlineto{\pgfqpoint{2.061263in}{0.971250in}}%
\pgfusepath{stroke}%
\end{pgfscope}%
\begin{pgfscope}%
\definecolor{textcolor}{rgb}{0.150000,0.150000,0.150000}%
\pgfsetstrokecolor{textcolor}%
\pgfsetfillcolor{textcolor}%
\pgftext[x=2.159041in,y=0.928472in,left,base]{\color{textcolor}\sffamily\fontsize{8.800000}{10.560000}\selectfont Norma}%
\end{pgfscope}%
\begin{pgfscope}%
\pgfsetroundcap%
\pgfsetroundjoin%
\pgfsetlinewidth{1.204500pt}%
\definecolor{currentstroke}{rgb}{0.890196,0.101961,0.109804}%
\pgfsetstrokecolor{currentstroke}%
\pgfsetdash{}{0pt}%
\pgfpathmoveto{\pgfqpoint{1.816819in}{0.799028in}}%
\pgfpathlineto{\pgfqpoint{2.061263in}{0.799028in}}%
\pgfusepath{stroke}%
\end{pgfscope}%
\begin{pgfscope}%
\definecolor{textcolor}{rgb}{0.150000,0.150000,0.150000}%
\pgfsetstrokecolor{textcolor}%
\pgfsetfillcolor{textcolor}%
\pgftext[x=2.159041in,y=0.756250in,left,base]{\color{textcolor}\sffamily\fontsize{8.800000}{10.560000}\selectfont cSMTiser\textsubscript{+LM}}%
\end{pgfscope}%
\begin{pgfscope}%
\pgfsetbuttcap%
\pgfsetroundjoin%
\pgfsetlinewidth{1.204500pt}%
\definecolor{currentstroke}{rgb}{0.200000,0.627451,0.172549}%
\pgfsetstrokecolor{currentstroke}%
\pgfsetdash{{4.440000pt}{1.920000pt}}{0.000000pt}%
\pgfpathmoveto{\pgfqpoint{1.816819in}{0.626806in}}%
\pgfpathlineto{\pgfqpoint{2.061263in}{0.626806in}}%
\pgfusepath{stroke}%
\end{pgfscope}%
\begin{pgfscope}%
\definecolor{textcolor}{rgb}{0.150000,0.150000,0.150000}%
\pgfsetstrokecolor{textcolor}%
\pgfsetfillcolor{textcolor}%
\pgftext[x=2.159041in,y=0.584028in,left,base]{\color{textcolor}\sffamily\fontsize{8.800000}{10.560000}\selectfont NMT-1}%
\end{pgfscope}%
\begin{pgfscope}%
\pgfsetbuttcap%
\pgfsetroundjoin%
\pgfsetlinewidth{1.204500pt}%
\definecolor{currentstroke}{rgb}{0.121569,0.470588,0.705882}%
\pgfsetstrokecolor{currentstroke}%
\pgfsetdash{{1.200000pt}{1.980000pt}}{0.000000pt}%
\pgfpathmoveto{\pgfqpoint{1.816819in}{0.454583in}}%
\pgfpathlineto{\pgfqpoint{2.061263in}{0.454583in}}%
\pgfusepath{stroke}%
\end{pgfscope}%
\begin{pgfscope}%
\definecolor{textcolor}{rgb}{0.150000,0.150000,0.150000}%
\pgfsetstrokecolor{textcolor}%
\pgfsetfillcolor{textcolor}%
\pgftext[x=2.159041in,y=0.411806in,left,base]{\color{textcolor}\sffamily\fontsize{8.800000}{10.560000}\selectfont NMT-2}%
\end{pgfscope}%
\end{pgfpicture}%
\makeatother%
\endgroup%

%% file: curves_slovene-goo300k-gaj.pgf
\begingroup%
\makeatletter%
\begin{pgfpicture}%
\pgfpathrectangle{\pgfpointorigin}{\pgfqpoint{3.078809in}{2.168389in}}%
\pgfusepath{use as bounding box, clip}%
\begin{pgfscope}%
\pgfsetbuttcap%
\pgfsetmiterjoin%
\definecolor{currentfill}{rgb}{1.000000,1.000000,1.000000}%
\pgfsetfillcolor{currentfill}%
\pgfsetlinewidth{0.000000pt}%
\definecolor{currentstroke}{rgb}{1.000000,1.000000,1.000000}%
\pgfsetstrokecolor{currentstroke}%
\pgfsetdash{}{0pt}%
\pgfpathmoveto{\pgfqpoint{0.000000in}{0.000000in}}%
\pgfpathlineto{\pgfqpoint{3.078809in}{0.000000in}}%
\pgfpathlineto{\pgfqpoint{3.078809in}{2.168389in}}%
\pgfpathlineto{\pgfqpoint{0.000000in}{2.168389in}}%
\pgfpathclose%
\pgfusepath{fill}%
\end{pgfscope}%
\begin{pgfscope}%
\pgfsetbuttcap%
\pgfsetmiterjoin%
\definecolor{currentfill}{rgb}{1.000000,1.000000,1.000000}%
\pgfsetfillcolor{currentfill}%
\pgfsetlinewidth{0.000000pt}%
\definecolor{currentstroke}{rgb}{0.000000,0.000000,0.000000}%
\pgfsetstrokecolor{currentstroke}%
\pgfsetstrokeopacity{0.000000}%
\pgfsetdash{}{0pt}%
\pgfpathmoveto{\pgfqpoint{0.450809in}{0.326389in}}%
\pgfpathlineto{\pgfqpoint{2.930809in}{0.326389in}}%
\pgfpathlineto{\pgfqpoint{2.930809in}{2.020389in}}%
\pgfpathlineto{\pgfqpoint{0.450809in}{2.020389in}}%
\pgfpathclose%
\pgfusepath{fill}%
\end{pgfscope}%
\begin{pgfscope}%
\pgfsetbuttcap%
\pgfsetroundjoin%
\definecolor{currentfill}{rgb}{0.150000,0.150000,0.150000}%
\pgfsetfillcolor{currentfill}%
\pgfsetlinewidth{1.003750pt}%
\definecolor{currentstroke}{rgb}{0.150000,0.150000,0.150000}%
\pgfsetstrokecolor{currentstroke}%
\pgfsetdash{}{0pt}%
\pgfsys@defobject{currentmarker}{\pgfqpoint{0.000000in}{-0.066667in}}{\pgfqpoint{0.000000in}{0.000000in}}{%
\pgfpathmoveto{\pgfqpoint{0.000000in}{0.000000in}}%
\pgfpathlineto{\pgfqpoint{0.000000in}{-0.066667in}}%
\pgfusepath{stroke,fill}%
}%
\begin{pgfscope}%
\pgfsys@transformshift{0.563536in}{0.326389in}%
\pgfsys@useobject{currentmarker}{}%
\end{pgfscope}%
\end{pgfscope}%
\begin{pgfscope}%
\definecolor{textcolor}{rgb}{0.150000,0.150000,0.150000}%
\pgfsetstrokecolor{textcolor}%
\pgfsetfillcolor{textcolor}%
\pgftext[x=0.563536in,y=0.211111in,,top]{\color{textcolor}\sffamily\fontsize{8.800000}{10.560000}\selectfont 100}%
\end{pgfscope}%
\begin{pgfscope}%
\pgfsetbuttcap%
\pgfsetroundjoin%
\definecolor{currentfill}{rgb}{0.150000,0.150000,0.150000}%
\pgfsetfillcolor{currentfill}%
\pgfsetlinewidth{1.003750pt}%
\definecolor{currentstroke}{rgb}{0.150000,0.150000,0.150000}%
\pgfsetstrokecolor{currentstroke}%
\pgfsetdash{}{0pt}%
\pgfsys@defobject{currentmarker}{\pgfqpoint{0.000000in}{-0.066667in}}{\pgfqpoint{0.000000in}{0.000000in}}{%
\pgfpathmoveto{\pgfqpoint{0.000000in}{0.000000in}}%
\pgfpathlineto{\pgfqpoint{0.000000in}{-0.066667in}}%
\pgfusepath{stroke,fill}%
}%
\begin{pgfscope}%
\pgfsys@transformshift{0.895949in}{0.326389in}%
\pgfsys@useobject{currentmarker}{}%
\end{pgfscope}%
\end{pgfscope}%
\begin{pgfscope}%
\definecolor{textcolor}{rgb}{0.150000,0.150000,0.150000}%
\pgfsetstrokecolor{textcolor}%
\pgfsetfillcolor{textcolor}%
\pgftext[x=0.895949in,y=0.211111in,,top]{\color{textcolor}\sffamily\fontsize{8.800000}{10.560000}\selectfont 250}%
\end{pgfscope}%
\begin{pgfscope}%
\pgfsetbuttcap%
\pgfsetroundjoin%
\definecolor{currentfill}{rgb}{0.150000,0.150000,0.150000}%
\pgfsetfillcolor{currentfill}%
\pgfsetlinewidth{1.003750pt}%
\definecolor{currentstroke}{rgb}{0.150000,0.150000,0.150000}%
\pgfsetstrokecolor{currentstroke}%
\pgfsetdash{}{0pt}%
\pgfsys@defobject{currentmarker}{\pgfqpoint{0.000000in}{-0.066667in}}{\pgfqpoint{0.000000in}{0.000000in}}{%
\pgfpathmoveto{\pgfqpoint{0.000000in}{0.000000in}}%
\pgfpathlineto{\pgfqpoint{0.000000in}{-0.066667in}}%
\pgfusepath{stroke,fill}%
}%
\begin{pgfscope}%
\pgfsys@transformshift{1.147410in}{0.326389in}%
\pgfsys@useobject{currentmarker}{}%
\end{pgfscope}%
\end{pgfscope}%
\begin{pgfscope}%
\definecolor{textcolor}{rgb}{0.150000,0.150000,0.150000}%
\pgfsetstrokecolor{textcolor}%
\pgfsetfillcolor{textcolor}%
\pgftext[x=1.147410in,y=0.211111in,,top]{\color{textcolor}\sffamily\fontsize{8.800000}{10.560000}\selectfont 500}%
\end{pgfscope}%
\begin{pgfscope}%
\pgfsetbuttcap%
\pgfsetroundjoin%
\definecolor{currentfill}{rgb}{0.150000,0.150000,0.150000}%
\pgfsetfillcolor{currentfill}%
\pgfsetlinewidth{1.003750pt}%
\definecolor{currentstroke}{rgb}{0.150000,0.150000,0.150000}%
\pgfsetstrokecolor{currentstroke}%
\pgfsetdash{}{0pt}%
\pgfsys@defobject{currentmarker}{\pgfqpoint{0.000000in}{-0.066667in}}{\pgfqpoint{0.000000in}{0.000000in}}{%
\pgfpathmoveto{\pgfqpoint{0.000000in}{0.000000in}}%
\pgfpathlineto{\pgfqpoint{0.000000in}{-0.066667in}}%
\pgfusepath{stroke,fill}%
}%
\begin{pgfscope}%
\pgfsys@transformshift{1.398871in}{0.326389in}%
\pgfsys@useobject{currentmarker}{}%
\end{pgfscope}%
\end{pgfscope}%
\begin{pgfscope}%
\definecolor{textcolor}{rgb}{0.150000,0.150000,0.150000}%
\pgfsetstrokecolor{textcolor}%
\pgfsetfillcolor{textcolor}%
\pgftext[x=1.398871in,y=0.211111in,,top]{\color{textcolor}\sffamily\fontsize{8.800000}{10.560000}\selectfont 1k}%
\end{pgfscope}%
\begin{pgfscope}%
\pgfsetbuttcap%
\pgfsetroundjoin%
\definecolor{currentfill}{rgb}{0.150000,0.150000,0.150000}%
\pgfsetfillcolor{currentfill}%
\pgfsetlinewidth{1.003750pt}%
\definecolor{currentstroke}{rgb}{0.150000,0.150000,0.150000}%
\pgfsetstrokecolor{currentstroke}%
\pgfsetdash{}{0pt}%
\pgfsys@defobject{currentmarker}{\pgfqpoint{0.000000in}{-0.066667in}}{\pgfqpoint{0.000000in}{0.000000in}}{%
\pgfpathmoveto{\pgfqpoint{0.000000in}{0.000000in}}%
\pgfpathlineto{\pgfqpoint{0.000000in}{-0.066667in}}%
\pgfusepath{stroke,fill}%
}%
\begin{pgfscope}%
\pgfsys@transformshift{1.731285in}{0.326389in}%
\pgfsys@useobject{currentmarker}{}%
\end{pgfscope}%
\end{pgfscope}%
\begin{pgfscope}%
\definecolor{textcolor}{rgb}{0.150000,0.150000,0.150000}%
\pgfsetstrokecolor{textcolor}%
\pgfsetfillcolor{textcolor}%
\pgftext[x=1.731285in,y=0.211111in,,top]{\color{textcolor}\sffamily\fontsize{8.800000}{10.560000}\selectfont 2.5k}%
\end{pgfscope}%
\begin{pgfscope}%
\pgfsetbuttcap%
\pgfsetroundjoin%
\definecolor{currentfill}{rgb}{0.150000,0.150000,0.150000}%
\pgfsetfillcolor{currentfill}%
\pgfsetlinewidth{1.003750pt}%
\definecolor{currentstroke}{rgb}{0.150000,0.150000,0.150000}%
\pgfsetstrokecolor{currentstroke}%
\pgfsetdash{}{0pt}%
\pgfsys@defobject{currentmarker}{\pgfqpoint{0.000000in}{-0.066667in}}{\pgfqpoint{0.000000in}{0.000000in}}{%
\pgfpathmoveto{\pgfqpoint{0.000000in}{0.000000in}}%
\pgfpathlineto{\pgfqpoint{0.000000in}{-0.066667in}}%
\pgfusepath{stroke,fill}%
}%
\begin{pgfscope}%
\pgfsys@transformshift{1.982746in}{0.326389in}%
\pgfsys@useobject{currentmarker}{}%
\end{pgfscope}%
\end{pgfscope}%
\begin{pgfscope}%
\definecolor{textcolor}{rgb}{0.150000,0.150000,0.150000}%
\pgfsetstrokecolor{textcolor}%
\pgfsetfillcolor{textcolor}%
\pgftext[x=1.982746in,y=0.211111in,,top]{\color{textcolor}\sffamily\fontsize{8.800000}{10.560000}\selectfont 5k}%
\end{pgfscope}%
\begin{pgfscope}%
\pgfsetbuttcap%
\pgfsetroundjoin%
\definecolor{currentfill}{rgb}{0.150000,0.150000,0.150000}%
\pgfsetfillcolor{currentfill}%
\pgfsetlinewidth{1.003750pt}%
\definecolor{currentstroke}{rgb}{0.150000,0.150000,0.150000}%
\pgfsetstrokecolor{currentstroke}%
\pgfsetdash{}{0pt}%
\pgfsys@defobject{currentmarker}{\pgfqpoint{0.000000in}{-0.066667in}}{\pgfqpoint{0.000000in}{0.000000in}}{%
\pgfpathmoveto{\pgfqpoint{0.000000in}{0.000000in}}%
\pgfpathlineto{\pgfqpoint{0.000000in}{-0.066667in}}%
\pgfusepath{stroke,fill}%
}%
\begin{pgfscope}%
\pgfsys@transformshift{2.234207in}{0.326389in}%
\pgfsys@useobject{currentmarker}{}%
\end{pgfscope}%
\end{pgfscope}%
\begin{pgfscope}%
\definecolor{textcolor}{rgb}{0.150000,0.150000,0.150000}%
\pgfsetstrokecolor{textcolor}%
\pgfsetfillcolor{textcolor}%
\pgftext[x=2.234207in,y=0.211111in,,top]{\color{textcolor}\sffamily\fontsize{8.800000}{10.560000}\selectfont 10k}%
\end{pgfscope}%
\begin{pgfscope}%
\pgfsetbuttcap%
\pgfsetroundjoin%
\definecolor{currentfill}{rgb}{0.150000,0.150000,0.150000}%
\pgfsetfillcolor{currentfill}%
\pgfsetlinewidth{1.003750pt}%
\definecolor{currentstroke}{rgb}{0.150000,0.150000,0.150000}%
\pgfsetstrokecolor{currentstroke}%
\pgfsetdash{}{0pt}%
\pgfsys@defobject{currentmarker}{\pgfqpoint{0.000000in}{-0.066667in}}{\pgfqpoint{0.000000in}{0.000000in}}{%
\pgfpathmoveto{\pgfqpoint{0.000000in}{0.000000in}}%
\pgfpathlineto{\pgfqpoint{0.000000in}{-0.066667in}}%
\pgfusepath{stroke,fill}%
}%
\begin{pgfscope}%
\pgfsys@transformshift{2.566620in}{0.326389in}%
\pgfsys@useobject{currentmarker}{}%
\end{pgfscope}%
\end{pgfscope}%
\begin{pgfscope}%
\definecolor{textcolor}{rgb}{0.150000,0.150000,0.150000}%
\pgfsetstrokecolor{textcolor}%
\pgfsetfillcolor{textcolor}%
\pgftext[x=2.566620in,y=0.211111in,,top]{\color{textcolor}\sffamily\fontsize{8.800000}{10.560000}\selectfont 25k}%
\end{pgfscope}%
\begin{pgfscope}%
\pgfsetbuttcap%
\pgfsetroundjoin%
\definecolor{currentfill}{rgb}{0.150000,0.150000,0.150000}%
\pgfsetfillcolor{currentfill}%
\pgfsetlinewidth{1.003750pt}%
\definecolor{currentstroke}{rgb}{0.150000,0.150000,0.150000}%
\pgfsetstrokecolor{currentstroke}%
\pgfsetdash{}{0pt}%
\pgfsys@defobject{currentmarker}{\pgfqpoint{0.000000in}{-0.066667in}}{\pgfqpoint{0.000000in}{0.000000in}}{%
\pgfpathmoveto{\pgfqpoint{0.000000in}{0.000000in}}%
\pgfpathlineto{\pgfqpoint{0.000000in}{-0.066667in}}%
\pgfusepath{stroke,fill}%
}%
\begin{pgfscope}%
\pgfsys@transformshift{2.818081in}{0.326389in}%
\pgfsys@useobject{currentmarker}{}%
\end{pgfscope}%
\end{pgfscope}%
\begin{pgfscope}%
\definecolor{textcolor}{rgb}{0.150000,0.150000,0.150000}%
\pgfsetstrokecolor{textcolor}%
\pgfsetfillcolor{textcolor}%
\pgftext[x=2.818081in,y=0.211111in,,top]{\color{textcolor}\sffamily\fontsize{8.800000}{10.560000}\selectfont 50k}%
\end{pgfscope}%
\begin{pgfscope}%
\pgfsetbuttcap%
\pgfsetroundjoin%
\definecolor{currentfill}{rgb}{0.150000,0.150000,0.150000}%
\pgfsetfillcolor{currentfill}%
\pgfsetlinewidth{1.003750pt}%
\definecolor{currentstroke}{rgb}{0.150000,0.150000,0.150000}%
\pgfsetstrokecolor{currentstroke}%
\pgfsetdash{}{0pt}%
\pgfsys@defobject{currentmarker}{\pgfqpoint{-0.066667in}{0.000000in}}{\pgfqpoint{0.000000in}{0.000000in}}{%
\pgfpathmoveto{\pgfqpoint{0.000000in}{0.000000in}}%
\pgfpathlineto{\pgfqpoint{-0.066667in}{0.000000in}}%
\pgfusepath{stroke,fill}%
}%
\begin{pgfscope}%
\pgfsys@transformshift{0.450809in}{0.517208in}%
\pgfsys@useobject{currentmarker}{}%
\end{pgfscope}%
\end{pgfscope}%
\begin{pgfscope}%
\definecolor{textcolor}{rgb}{0.150000,0.150000,0.150000}%
\pgfsetstrokecolor{textcolor}%
\pgfsetfillcolor{textcolor}%
\pgftext[x=0.100000in,y=0.473805in,left,base]{\color{textcolor}\sffamily\fontsize{8.800000}{10.560000}\selectfont 20\%}%
\end{pgfscope}%
\begin{pgfscope}%
\pgfsetbuttcap%
\pgfsetroundjoin%
\definecolor{currentfill}{rgb}{0.150000,0.150000,0.150000}%
\pgfsetfillcolor{currentfill}%
\pgfsetlinewidth{1.003750pt}%
\definecolor{currentstroke}{rgb}{0.150000,0.150000,0.150000}%
\pgfsetstrokecolor{currentstroke}%
\pgfsetdash{}{0pt}%
\pgfsys@defobject{currentmarker}{\pgfqpoint{-0.066667in}{0.000000in}}{\pgfqpoint{0.000000in}{0.000000in}}{%
\pgfpathmoveto{\pgfqpoint{0.000000in}{0.000000in}}%
\pgfpathlineto{\pgfqpoint{-0.066667in}{0.000000in}}%
\pgfusepath{stroke,fill}%
}%
\begin{pgfscope}%
\pgfsys@transformshift{0.450809in}{0.898846in}%
\pgfsys@useobject{currentmarker}{}%
\end{pgfscope}%
\end{pgfscope}%
\begin{pgfscope}%
\definecolor{textcolor}{rgb}{0.150000,0.150000,0.150000}%
\pgfsetstrokecolor{textcolor}%
\pgfsetfillcolor{textcolor}%
\pgftext[x=0.100000in,y=0.855443in,left,base]{\color{textcolor}\sffamily\fontsize{8.800000}{10.560000}\selectfont 40\%}%
\end{pgfscope}%
\begin{pgfscope}%
\pgfsetbuttcap%
\pgfsetroundjoin%
\definecolor{currentfill}{rgb}{0.150000,0.150000,0.150000}%
\pgfsetfillcolor{currentfill}%
\pgfsetlinewidth{1.003750pt}%
\definecolor{currentstroke}{rgb}{0.150000,0.150000,0.150000}%
\pgfsetstrokecolor{currentstroke}%
\pgfsetdash{}{0pt}%
\pgfsys@defobject{currentmarker}{\pgfqpoint{-0.066667in}{0.000000in}}{\pgfqpoint{0.000000in}{0.000000in}}{%
\pgfpathmoveto{\pgfqpoint{0.000000in}{0.000000in}}%
\pgfpathlineto{\pgfqpoint{-0.066667in}{0.000000in}}%
\pgfusepath{stroke,fill}%
}%
\begin{pgfscope}%
\pgfsys@transformshift{0.450809in}{1.280484in}%
\pgfsys@useobject{currentmarker}{}%
\end{pgfscope}%
\end{pgfscope}%
\begin{pgfscope}%
\definecolor{textcolor}{rgb}{0.150000,0.150000,0.150000}%
\pgfsetstrokecolor{textcolor}%
\pgfsetfillcolor{textcolor}%
\pgftext[x=0.100000in,y=1.237081in,left,base]{\color{textcolor}\sffamily\fontsize{8.800000}{10.560000}\selectfont 60\%}%
\end{pgfscope}%
\begin{pgfscope}%
\pgfsetbuttcap%
\pgfsetroundjoin%
\definecolor{currentfill}{rgb}{0.150000,0.150000,0.150000}%
\pgfsetfillcolor{currentfill}%
\pgfsetlinewidth{1.003750pt}%
\definecolor{currentstroke}{rgb}{0.150000,0.150000,0.150000}%
\pgfsetstrokecolor{currentstroke}%
\pgfsetdash{}{0pt}%
\pgfsys@defobject{currentmarker}{\pgfqpoint{-0.066667in}{0.000000in}}{\pgfqpoint{0.000000in}{0.000000in}}{%
\pgfpathmoveto{\pgfqpoint{0.000000in}{0.000000in}}%
\pgfpathlineto{\pgfqpoint{-0.066667in}{0.000000in}}%
\pgfusepath{stroke,fill}%
}%
\begin{pgfscope}%
\pgfsys@transformshift{0.450809in}{1.662122in}%
\pgfsys@useobject{currentmarker}{}%
\end{pgfscope}%
\end{pgfscope}%
\begin{pgfscope}%
\definecolor{textcolor}{rgb}{0.150000,0.150000,0.150000}%
\pgfsetstrokecolor{textcolor}%
\pgfsetfillcolor{textcolor}%
\pgftext[x=0.100000in,y=1.618720in,left,base]{\color{textcolor}\sffamily\fontsize{8.800000}{10.560000}\selectfont 80\%}%
\end{pgfscope}%
\begin{pgfscope}%
\pgfsetbuttcap%
\pgfsetroundjoin%
\definecolor{currentfill}{rgb}{0.150000,0.150000,0.150000}%
\pgfsetfillcolor{currentfill}%
\pgfsetlinewidth{0.803000pt}%
\definecolor{currentstroke}{rgb}{0.150000,0.150000,0.150000}%
\pgfsetstrokecolor{currentstroke}%
\pgfsetdash{}{0pt}%
\pgfsys@defobject{currentmarker}{\pgfqpoint{-0.044444in}{0.000000in}}{\pgfqpoint{0.000000in}{0.000000in}}{%
\pgfpathmoveto{\pgfqpoint{0.000000in}{0.000000in}}%
\pgfpathlineto{\pgfqpoint{-0.044444in}{0.000000in}}%
\pgfusepath{stroke,fill}%
}%
\begin{pgfscope}%
\pgfsys@transformshift{0.450809in}{0.326389in}%
\pgfsys@useobject{currentmarker}{}%
\end{pgfscope}%
\end{pgfscope}%
\begin{pgfscope}%
\pgfsetbuttcap%
\pgfsetroundjoin%
\definecolor{currentfill}{rgb}{0.150000,0.150000,0.150000}%
\pgfsetfillcolor{currentfill}%
\pgfsetlinewidth{0.803000pt}%
\definecolor{currentstroke}{rgb}{0.150000,0.150000,0.150000}%
\pgfsetstrokecolor{currentstroke}%
\pgfsetdash{}{0pt}%
\pgfsys@defobject{currentmarker}{\pgfqpoint{-0.044444in}{0.000000in}}{\pgfqpoint{0.000000in}{0.000000in}}{%
\pgfpathmoveto{\pgfqpoint{0.000000in}{0.000000in}}%
\pgfpathlineto{\pgfqpoint{-0.044444in}{0.000000in}}%
\pgfusepath{stroke,fill}%
}%
\begin{pgfscope}%
\pgfsys@transformshift{0.450809in}{0.708027in}%
\pgfsys@useobject{currentmarker}{}%
\end{pgfscope}%
\end{pgfscope}%
\begin{pgfscope}%
\pgfsetbuttcap%
\pgfsetroundjoin%
\definecolor{currentfill}{rgb}{0.150000,0.150000,0.150000}%
\pgfsetfillcolor{currentfill}%
\pgfsetlinewidth{0.803000pt}%
\definecolor{currentstroke}{rgb}{0.150000,0.150000,0.150000}%
\pgfsetstrokecolor{currentstroke}%
\pgfsetdash{}{0pt}%
\pgfsys@defobject{currentmarker}{\pgfqpoint{-0.044444in}{0.000000in}}{\pgfqpoint{0.000000in}{0.000000in}}{%
\pgfpathmoveto{\pgfqpoint{0.000000in}{0.000000in}}%
\pgfpathlineto{\pgfqpoint{-0.044444in}{0.000000in}}%
\pgfusepath{stroke,fill}%
}%
\begin{pgfscope}%
\pgfsys@transformshift{0.450809in}{1.089665in}%
\pgfsys@useobject{currentmarker}{}%
\end{pgfscope}%
\end{pgfscope}%
\begin{pgfscope}%
\pgfsetbuttcap%
\pgfsetroundjoin%
\definecolor{currentfill}{rgb}{0.150000,0.150000,0.150000}%
\pgfsetfillcolor{currentfill}%
\pgfsetlinewidth{0.803000pt}%
\definecolor{currentstroke}{rgb}{0.150000,0.150000,0.150000}%
\pgfsetstrokecolor{currentstroke}%
\pgfsetdash{}{0pt}%
\pgfsys@defobject{currentmarker}{\pgfqpoint{-0.044444in}{0.000000in}}{\pgfqpoint{0.000000in}{0.000000in}}{%
\pgfpathmoveto{\pgfqpoint{0.000000in}{0.000000in}}%
\pgfpathlineto{\pgfqpoint{-0.044444in}{0.000000in}}%
\pgfusepath{stroke,fill}%
}%
\begin{pgfscope}%
\pgfsys@transformshift{0.450809in}{1.471303in}%
\pgfsys@useobject{currentmarker}{}%
\end{pgfscope}%
\end{pgfscope}%
\begin{pgfscope}%
\pgfsetbuttcap%
\pgfsetroundjoin%
\definecolor{currentfill}{rgb}{0.150000,0.150000,0.150000}%
\pgfsetfillcolor{currentfill}%
\pgfsetlinewidth{0.803000pt}%
\definecolor{currentstroke}{rgb}{0.150000,0.150000,0.150000}%
\pgfsetstrokecolor{currentstroke}%
\pgfsetdash{}{0pt}%
\pgfsys@defobject{currentmarker}{\pgfqpoint{-0.044444in}{0.000000in}}{\pgfqpoint{0.000000in}{0.000000in}}{%
\pgfpathmoveto{\pgfqpoint{0.000000in}{0.000000in}}%
\pgfpathlineto{\pgfqpoint{-0.044444in}{0.000000in}}%
\pgfusepath{stroke,fill}%
}%
\begin{pgfscope}%
\pgfsys@transformshift{0.450809in}{1.852941in}%
\pgfsys@useobject{currentmarker}{}%
\end{pgfscope}%
\end{pgfscope}%
\begin{pgfscope}%
\pgfpathrectangle{\pgfqpoint{0.450809in}{0.326389in}}{\pgfqpoint{2.480000in}{1.694000in}}%
\pgfusepath{clip}%
\pgfsetbuttcap%
\pgfsetroundjoin%
\pgfsetlinewidth{1.204500pt}%
\definecolor{currentstroke}{rgb}{1.000000,0.498039,0.000000}%
\pgfsetstrokecolor{currentstroke}%
\pgfsetdash{{7.680000pt}{1.920000pt}{1.200000pt}{1.920000pt}}{0.000000pt}%
\pgfpathmoveto{\pgfqpoint{0.563536in}{1.711111in}}%
\pgfpathlineto{\pgfqpoint{0.895949in}{1.732014in}}%
\pgfpathlineto{\pgfqpoint{1.147410in}{1.737918in}}%
\pgfpathlineto{\pgfqpoint{1.398871in}{1.745385in}}%
\pgfpathlineto{\pgfqpoint{1.731285in}{1.762294in}}%
\pgfpathlineto{\pgfqpoint{1.982746in}{1.777447in}}%
\pgfpathlineto{\pgfqpoint{2.234207in}{1.794210in}}%
\pgfpathlineto{\pgfqpoint{2.566620in}{1.823164in}}%
\pgfpathlineto{\pgfqpoint{2.818081in}{1.846617in}}%
\pgfusepath{stroke}%
\end{pgfscope}%
\begin{pgfscope}%
\pgfpathrectangle{\pgfqpoint{0.450809in}{0.326389in}}{\pgfqpoint{2.480000in}{1.694000in}}%
\pgfusepath{clip}%
\pgfsetroundcap%
\pgfsetroundjoin%
\pgfsetlinewidth{1.204500pt}%
\definecolor{currentstroke}{rgb}{0.890196,0.101961,0.109804}%
\pgfsetstrokecolor{currentstroke}%
\pgfsetdash{}{0pt}%
\pgfpathmoveto{\pgfqpoint{0.563536in}{1.791221in}}%
\pgfpathlineto{\pgfqpoint{0.895949in}{1.794904in}}%
\pgfpathlineto{\pgfqpoint{1.147410in}{1.813961in}}%
\pgfpathlineto{\pgfqpoint{1.398871in}{1.853024in}}%
\pgfpathlineto{\pgfqpoint{1.731285in}{1.889592in}}%
\pgfpathlineto{\pgfqpoint{1.982746in}{1.905979in}}%
\pgfpathlineto{\pgfqpoint{2.234207in}{1.920255in}}%
\pgfpathlineto{\pgfqpoint{2.566620in}{1.942447in}}%
\pgfpathlineto{\pgfqpoint{2.818081in}{1.955197in}}%
\pgfusepath{stroke}%
\end{pgfscope}%
\begin{pgfscope}%
\pgfpathrectangle{\pgfqpoint{0.450809in}{0.326389in}}{\pgfqpoint{2.480000in}{1.694000in}}%
\pgfusepath{clip}%
\pgfsetbuttcap%
\pgfsetroundjoin%
\pgfsetlinewidth{1.204500pt}%
\definecolor{currentstroke}{rgb}{0.200000,0.627451,0.172549}%
\pgfsetstrokecolor{currentstroke}%
\pgfsetdash{{4.440000pt}{1.920000pt}}{0.000000pt}%
\pgfpathmoveto{\pgfqpoint{0.563536in}{0.651352in}}%
\pgfpathlineto{\pgfqpoint{0.895949in}{1.416410in}}%
\pgfpathlineto{\pgfqpoint{1.147410in}{1.697648in}}%
\pgfpathlineto{\pgfqpoint{1.398871in}{1.782538in}}%
\pgfpathlineto{\pgfqpoint{1.731285in}{1.830202in}}%
\pgfpathlineto{\pgfqpoint{1.982746in}{1.856168in}}%
\pgfpathlineto{\pgfqpoint{2.234207in}{1.890652in}}%
\pgfpathlineto{\pgfqpoint{2.566620in}{1.917504in}}%
\pgfpathlineto{\pgfqpoint{2.818081in}{1.938617in}}%
\pgfusepath{stroke}%
\end{pgfscope}%
\begin{pgfscope}%
\pgfpathrectangle{\pgfqpoint{0.450809in}{0.326389in}}{\pgfqpoint{2.480000in}{1.694000in}}%
\pgfusepath{clip}%
\pgfsetbuttcap%
\pgfsetroundjoin%
\pgfsetlinewidth{1.204500pt}%
\definecolor{currentstroke}{rgb}{0.121569,0.470588,0.705882}%
\pgfsetstrokecolor{currentstroke}%
\pgfsetdash{{1.200000pt}{1.980000pt}}{0.000000pt}%
\pgfpathmoveto{\pgfqpoint{0.563536in}{1.222182in}}%
\pgfpathlineto{\pgfqpoint{0.895949in}{1.645278in}}%
\pgfpathlineto{\pgfqpoint{1.147410in}{1.651922in}}%
\pgfpathlineto{\pgfqpoint{1.398871in}{1.816766in}}%
\pgfpathlineto{\pgfqpoint{1.731285in}{1.843921in}}%
\pgfpathlineto{\pgfqpoint{1.982746in}{1.860335in}}%
\pgfpathlineto{\pgfqpoint{2.234207in}{1.884163in}}%
\pgfpathlineto{\pgfqpoint{2.566620in}{1.923939in}}%
\pgfpathlineto{\pgfqpoint{2.818081in}{1.947062in}}%
\pgfusepath{stroke}%
\end{pgfscope}%
\begin{pgfscope}%
\pgfsetrectcap%
\pgfsetmiterjoin%
\pgfsetlinewidth{1.003750pt}%
\definecolor{currentstroke}{rgb}{0.150000,0.150000,0.150000}%
\pgfsetstrokecolor{currentstroke}%
\pgfsetdash{}{0pt}%
\pgfpathmoveto{\pgfqpoint{0.450809in}{0.326389in}}%
\pgfpathlineto{\pgfqpoint{0.450809in}{2.020389in}}%
\pgfusepath{stroke}%
\end{pgfscope}%
\begin{pgfscope}%
\pgfsetrectcap%
\pgfsetmiterjoin%
\pgfsetlinewidth{1.003750pt}%
\definecolor{currentstroke}{rgb}{0.150000,0.150000,0.150000}%
\pgfsetstrokecolor{currentstroke}%
\pgfsetdash{}{0pt}%
\pgfpathmoveto{\pgfqpoint{2.930809in}{0.326389in}}%
\pgfpathlineto{\pgfqpoint{2.930809in}{2.020389in}}%
\pgfusepath{stroke}%
\end{pgfscope}%
\begin{pgfscope}%
\pgfsetrectcap%
\pgfsetmiterjoin%
\pgfsetlinewidth{1.003750pt}%
\definecolor{currentstroke}{rgb}{0.150000,0.150000,0.150000}%
\pgfsetstrokecolor{currentstroke}%
\pgfsetdash{}{0pt}%
\pgfpathmoveto{\pgfqpoint{0.450809in}{0.326389in}}%
\pgfpathlineto{\pgfqpoint{2.930809in}{0.326389in}}%
\pgfusepath{stroke}%
\end{pgfscope}%
\begin{pgfscope}%
\pgfsetrectcap%
\pgfsetmiterjoin%
\pgfsetlinewidth{1.003750pt}%
\definecolor{currentstroke}{rgb}{0.150000,0.150000,0.150000}%
\pgfsetstrokecolor{currentstroke}%
\pgfsetdash{}{0pt}%
\pgfpathmoveto{\pgfqpoint{0.450809in}{2.020389in}}%
\pgfpathlineto{\pgfqpoint{2.930809in}{2.020389in}}%
\pgfusepath{stroke}%
\end{pgfscope}%
\begin{pgfscope}%
\pgfsetbuttcap%
\pgfsetroundjoin%
\pgfsetlinewidth{1.204500pt}%
\definecolor{currentstroke}{rgb}{1.000000,0.498039,0.000000}%
\pgfsetstrokecolor{currentstroke}%
\pgfsetdash{{7.680000pt}{1.920000pt}{1.200000pt}{1.920000pt}}{0.000000pt}%
\pgfpathmoveto{\pgfqpoint{1.816819in}{0.971250in}}%
\pgfpathlineto{\pgfqpoint{2.061263in}{0.971250in}}%
\pgfusepath{stroke}%
\end{pgfscope}%
\begin{pgfscope}%
\definecolor{textcolor}{rgb}{0.150000,0.150000,0.150000}%
\pgfsetstrokecolor{textcolor}%
\pgfsetfillcolor{textcolor}%
\pgftext[x=2.159041in,y=0.928472in,left,base]{\color{textcolor}\sffamily\fontsize{8.800000}{10.560000}\selectfont Norma}%
\end{pgfscope}%
\begin{pgfscope}%
\pgfsetroundcap%
\pgfsetroundjoin%
\pgfsetlinewidth{1.204500pt}%
\definecolor{currentstroke}{rgb}{0.890196,0.101961,0.109804}%
\pgfsetstrokecolor{currentstroke}%
\pgfsetdash{}{0pt}%
\pgfpathmoveto{\pgfqpoint{1.816819in}{0.799028in}}%
\pgfpathlineto{\pgfqpoint{2.061263in}{0.799028in}}%
\pgfusepath{stroke}%
\end{pgfscope}%
\begin{pgfscope}%
\definecolor{textcolor}{rgb}{0.150000,0.150000,0.150000}%
\pgfsetstrokecolor{textcolor}%
\pgfsetfillcolor{textcolor}%
\pgftext[x=2.159041in,y=0.756250in,left,base]{\color{textcolor}\sffamily\fontsize{8.800000}{10.560000}\selectfont cSMTiser\textsubscript{+LM}}%
\end{pgfscope}%
\begin{pgfscope}%
\pgfsetbuttcap%
\pgfsetroundjoin%
\pgfsetlinewidth{1.204500pt}%
\definecolor{currentstroke}{rgb}{0.200000,0.627451,0.172549}%
\pgfsetstrokecolor{currentstroke}%
\pgfsetdash{{4.440000pt}{1.920000pt}}{0.000000pt}%
\pgfpathmoveto{\pgfqpoint{1.816819in}{0.626806in}}%
\pgfpathlineto{\pgfqpoint{2.061263in}{0.626806in}}%
\pgfusepath{stroke}%
\end{pgfscope}%
\begin{pgfscope}%
\definecolor{textcolor}{rgb}{0.150000,0.150000,0.150000}%
\pgfsetstrokecolor{textcolor}%
\pgfsetfillcolor{textcolor}%
\pgftext[x=2.159041in,y=0.584028in,left,base]{\color{textcolor}\sffamily\fontsize{8.800000}{10.560000}\selectfont NMT-1}%
\end{pgfscope}%
\begin{pgfscope}%
\pgfsetbuttcap%
\pgfsetroundjoin%
\pgfsetlinewidth{1.204500pt}%
\definecolor{currentstroke}{rgb}{0.121569,0.470588,0.705882}%
\pgfsetstrokecolor{currentstroke}%
\pgfsetdash{{1.200000pt}{1.980000pt}}{0.000000pt}%
\pgfpathmoveto{\pgfqpoint{1.816819in}{0.454583in}}%
\pgfpathlineto{\pgfqpoint{2.061263in}{0.454583in}}%
\pgfusepath{stroke}%
\end{pgfscope}%
\begin{pgfscope}%
\definecolor{textcolor}{rgb}{0.150000,0.150000,0.150000}%
\pgfsetstrokecolor{textcolor}%
\pgfsetfillcolor{textcolor}%
\pgftext[x=2.159041in,y=0.411806in,left,base]{\color{textcolor}\sffamily\fontsize{8.800000}{10.560000}\selectfont NMT-2}%
\end{pgfscope}%
\end{pgfpicture}%
\makeatother%
\endgroup%

%% file: master.bbl
\begin{thebibliography}{54}
\expandafter\ifx\csname natexlab\endcsname\relax\def\natexlab#1{#1}\fi

\bibitem[{Adesam et~al.(2012)Adesam, Ahlberg, and Bouma}]{Adesam-etal2012}
Yvonne Adesam, Malin Ahlberg, and Gerlof Bouma. 2012.
\newblock \href {http://www.oegai.at/konvens2012/proceedings/54_adesam12w/}
  {{\emph{bokstaffua, bokstaffwa, bokstafwa, bokstaua, bokstawa\ldots{}}
  Towards lexical link-up for a corpus of Old Swedish}}.
\newblock In \emph{{Proceedings of the 11th Conference on Natural Language
  Processing (KONVENS~2012), LThist~2012 workshop}}, pages 365--369, Vienna,
  Austria.

\bibitem[{Al~Azawi et~al.(2013)Al~Azawi, Afzal, and Breuel}]{AlAzawi-etal2013}
Mayce Al~Azawi, Muhammad~Zeshan Afzal, and Thomas~M. Breuel. 2013.
\newblock \href {https://doi.org/10.1145/2501115.2501131} {Normalizing
  historical orthography for {OCR} historical documents using {LSTM}}.
\newblock In \emph{Proceedings of the 2nd International Workshop on Historical
  Document Imaging and Processing (HIP '13)}, pages 80--85, Washington, DC.

\bibitem[{Amoia and Mart\'{i}nez(2013)}]{Amoia-Martinez2013}
Marilisa Amoia and Jos\'{e}~Manuel Mart\'{i}nez. 2013.
\newblock \href {http://www.aclweb.org/anthology/W13-2711} {Using comparable
  collections of historical texts for building a diachronic dictionary for
  spelling normalization}.
\newblock In \emph{Proceedings of the 7th Workshop on Language Technology for
  Cultural Heritage, Social Sciences, and Humanities (LaTeCH)}, pages 84--89.
  Association for Computational Linguistics.

\bibitem[{Baron and Rayson(2008)}]{Baron-Rayson2008}
Alistair Baron and Paul Rayson. 2008.
\newblock {VARD 2}: A tool for dealing with spelling variation in historical
  corpora.
\newblock In \emph{{Proceedings of the Postgraduate Conference in Corpus
  Linguistics}}.

\bibitem[{Barteld et~al.(2015)Barteld, Schr\"{o}der, and
  Zinsmeister}]{Barteld-etal2015}
Fabian Barteld, Ingrid Schr\"{o}der, and Heike Zinsmeister. 2015.
\newblock Unsupervised regularization of historical texts for {POS} tagging.
\newblock In \emph{Proceedings of the Workshop on Corpus-Based Research in the
  Humanities (CRH)}, pages 3--12, Warsaw, Poland.

\bibitem[{Bjarnad\'{o}ttir(2012)}]{Bjarnadottir2012}
Krist\'{i}n Bjarnad\'{o}ttir. 2012.
\newblock The database of modern {I}celandic inflection ({B}eygingarlýsing
  \'{i}slensks n\'{u}t\'{i}mam\'{a}ls).
\newblock In \emph{Proceedings of the Workshop on Language Technology for
  Normalisation of Less-Resourced Languages}, pages 13--18.

\bibitem[{Bollmann(2012)}]{Bollmann2012}
Marcel Bollmann. 2012.
\newblock ({S}emi-)automatic normalization of historical texts using distance
  measures and the {Norma} tool.
\newblock In \emph{{Proceedings of the Second Workshop on Annotation of Corpora
  for Research in the Humanities (ACRH-2)}}, pages 3--12, Lisbon, Portugal.

\bibitem[{Bollmann(2018)}]{BollmannPhD}
Marcel Bollmann. 2018.
\newblock \href
  {http://www.linguistics.rub.de/forschung/arbeitsberichte/22.pdf}
  {Normalization of historical texts with neural network models}.
\newblock \emph{Bochumer Linguistische Arbeitsberichte}, 22.

\bibitem[{Bollmann et~al.(2017)Bollmann, Bingel, and
  S{\o}gaard}]{Bollmann-etal2017}
Marcel Bollmann, Joachim Bingel, and Anders S{\o}gaard. 2017.
\newblock \href {https://doi.org/10.18653/v1/P17-1031} {Learning attention for
  historical text normalization by learning to pronounce}.
\newblock In \emph{Proceedings of the 55th Annual Meeting of the Association
  for Computational Linguistics (Volume 1: Long Papers)}, pages 332--344.
  Association for Computational Linguistics.

\bibitem[{Bollmann et~al.(2011)Bollmann, Petran, and
  Dipper}]{Bollmann-etal2011}
Marcel Bollmann, Florian Petran, and Stefanie Dipper. 2011.
\newblock Rule-based normalization of historical texts.
\newblock In \emph{{Proceedings of the International Workshop on Language
  Technologies for Digital Humanities and Cultural Heritage}}, pages 34--42,
  Hissar, Bulgaria.

\bibitem[{Bollmann and S{\o}gaard(2016)}]{Bollmann-Sogaard2016}
Marcel Bollmann and Anders S{\o}gaard. 2016.
\newblock \href {http://www.aclweb.org/anthology/C16-1013} {Improving
  historical spelling normalization with bi-directional {LSTMs} and multi-task
  learning}.
\newblock In \emph{Proceedings of COLING 2016, the 26th International
  Conference on Computational Linguistics: Technical Papers}, pages 131--139.
  The COLING 2016 Organizing Committee.

\bibitem[{Bollmann et~al.(2018)Bollmann, S{\o}gaard, and
  Bingel}]{Bollmann-etal2018}
Marcel Bollmann, Anders S{\o}gaard, and Joachim Bingel. 2018.
\newblock \href {http://aclweb.org/anthology/W18-3403} {Multi-task learning for
  historical text normalization: Size matters}.
\newblock In \emph{Proceedings of the Workshop on Deep Learning Approaches for
  Low-Resource NLP}, pages 19--24. Association for Computational Linguistics.

\bibitem[{Christodouloupoulos and Steedman(2015)}]{Christo-Steedman2015}
Christos Christodouloupoulos and Mark Steedman. 2015.
\newblock \href {https://doi.org/10.1007/s10579-014-9287-y} {A massively
  parallel corpus: the {B}ible in 100 languages}.
\newblock \emph{Language Resources and Evaluation}, 49(2):375--395.

\bibitem[{Domingo and Casacuberta(2018)}]{Domingo-Casacuberta2018}
Miguel Domingo and Francisco Casacuberta. 2018.
\newblock \href {http://hdl.handle.net/10045/76035} {Spelling normalization of
  historical documents by using a machine translation approach}.
\newblock In \emph{Proceedings of the 21st Annual Conference of the European
  Association for Machine Translation}, pages 129--137.

\bibitem[{Ernst-Gerlach and Fuhr(2006)}]{ErnstGerlach-Fuhr2006}
Andrea Ernst-Gerlach and Norbert Fuhr. 2006.
\newblock \href {https://doi.org/10.1007/11735106} {Generating search term
  variants for text collections with historic spellings}.
\newblock In \emph{Proceedings of the 28th European Conference on Information
  Retrieval Research (ECIR 2006)}, Lecture Notes in Computer Science, pages
  49--60, Berlin. Springer.

\bibitem[{Etxeberria et~al.(2016)Etxeberria, {n}aki Alegria, Uria, and
  Hulden}]{Etxeberria-etal2016}
Izaskun Etxeberria, I\~{n}aki Alegria, Larraitz Uria, and Mans Hulden. 2016.
\newblock \href
  {http://www.lrec-conf.org/proceedings/lrec2016/pdf/147_Paper.pdf} {Evaluating
  the noisy channel model for the normalization of historical texts: {Basque},
  {Spanish} and {Slovene}}.
\newblock In \emph{Proceedings of the Tenth International Conference on
  Language Resources and Evaluation (LREC 2016)}, pages 1064--1069, Paris,
  France. European Language Resources Association (ELRA).

\bibitem[{Fix(1980)}]{Fix1980}
Hans Fix. 1980.
\newblock {A}utomatische {N}ormalisierung -- {V}orarbeit zur {L}emmatisierung
  eines diplomatischen altisl\"{a}ndischen {T}extes.
\newblock In Paul Sappler and Erich Stra{\ss}ner, editors, \emph{{M}aschinelle
  {V}erarbeitung altdeutscher {T}exte. {B}eitr\"{a}ge zum dritten {S}ymposion,
  {T}\"{u}bingen 17.--19. {F}ebruar 1977}, pages 92--100. Niemeyer,
  T\"{u}bingen.

\bibitem[{Giusti et~al.(2007)Giusti, Candido~Jr, Muniz, Cucatto, and
  Alu\'{i}sio}]{Giusti-etal2007}
Rafael Giusti, Arnaldo Candido~Jr, Marcelo Muniz, L\'{i}via Cucatto, and Sandra
  Alu\'{i}sio. 2007.
\newblock \href
  {http://ucrel.lancs.ac.uk/publications/cl2007/paper/238_Paper.pdf} {Automatic
  detection of spelling variation in historical corpus: An application to build
  a {Brazilian} {Portuguese} spelling variants dictionary}.
\newblock In \emph{Proceedings of the Corpus Linguistics Conference (CL2007)},
  Birmingham, UK.

\bibitem[{van Halteren and Rem(2013)}]{vanHalteren-Rem2013}
Hans van Halteren and Margit Rem. 2013.
\newblock \href {https://doi.org/10.1007/s10579-013-9236-1} {Dealing with
  orthographic variation in a tagger-lemmatizer for fourteenth century {Dutch}
  charters}.
\newblock \emph{Language Resources and Evaluation}, 47(4):1233--1259.

\bibitem[{H{\"a}m{\"a}l{\"a}inen et~al.(2018)H{\"a}m{\"a}l{\"a}inen, S{\"a}ily,
  Rueter, Tiedemann, and M{\"a}kel{\"a}}]{Hamalainen-etal2018}
Mika H{\"a}m{\"a}l{\"a}inen, Tanja S{\"a}ily, Jack Rueter, J{\"o}rg Tiedemann,
  and Eetu M{\"a}kel{\"a}. 2018.
\newblock \href {http://aclweb.org/anthology/W18-4510} {Normalizing {E}arly
  {E}nglish letters to present-day {E}nglish spelling}.
\newblock In \emph{Proceedings of the Second Joint SIGHUM Workshop on
  Computational Linguistics for Cultural Heritage, Social Sciences, Humanities
  and Literature}, pages 87--96. Association for Computational Linguistics.

\bibitem[{Hauser and Schulz(2007)}]{Hauser-Schulz2007}
Andreas~W. Hauser and Klaus~U. Schulz. 2007.
\newblock Unsupervised learning of edit distance weights for retrieving
  historical spelling variations.
\newblock In \emph{{Proceedings of the First Workshop on Finite-State
  Techniques and Approximate Search (FSTAS 2007)}}, pages 1--6, Borovets,
  Bulgaria.

\bibitem[{Helgad\'{o}ttir et~al.(2012)Helgad\'{o}ttir, \'{A}sta
  Svavarsd\'{o}ttir, R\"{o}gnvaldsson, Bjarnad\'{o}ttir, and
  Loftsson}]{Helgadottir-etal2012}
Sigr\'{u}n Helgad\'{o}ttir, \'{A}sta Svavarsd\'{o}ttir, Eir\'{i}kur
  R\"{o}gnvaldsson, Krist\'{i}n Bjarnad\'{o}ttir, and Hrafn Loftsson. 2012.
\newblock The tagged {I}celandic corpus {(M\'{i}M)}.
\newblock In \emph{Proceedings of the Workshop on Language Technology for
  Normalisation of Less-Resourced Languages}, pages 67--72.

\bibitem[{Junczys-Dowmunt et~al.(2018)Junczys-Dowmunt, Grundkiewicz, Dwojak,
  Hoang, Heafield, Neckermann, Seide, Germann, Fikri~Aji, Bogoychev, Martins,
  and Birch}]{mariannmt}
Marcin Junczys-Dowmunt, Roman Grundkiewicz, Tomasz Dwojak, Hieu Hoang, Kenneth
  Heafield, Tom Neckermann, Frank Seide, Ulrich Germann, Alham Fikri~Aji,
  Nikolay Bogoychev, Andr\'{e} F.~T. Martins, and Alexandra Birch. 2018.
\newblock \href {http://www.aclweb.org/anthology/P18-4020} {Marian: Fast neural
  machine translation in {C++}}.
\newblock In \emph{Proceedings of ACL 2018, System Demonstrations}, pages
  116--121, Melbourne, Australia. Association for Computational Linguistics.

\bibitem[{Jurish(2010{\natexlab{a}})}]{Jurish2010b}
Bryan Jurish. 2010{\natexlab{a}}.
\newblock \href {http://www.aclweb.org/anthology/W10-2209} {Comparing
  canonicalizations of historical {German} text}.
\newblock In \emph{Proceedings of the 11th Meeting of the ACL Special Interest
  Group on Computational Morphology and Phonology}, pages 72--77, Uppsala,
  Sweden. Association for Computational Linguistics.

\bibitem[{Jurish(2010{\natexlab{b}})}]{Jurish2010}
Bryan Jurish. 2010{\natexlab{b}}.
\newblock \href {http://www.jlcl.org/2010_Heft1/bryan_jurish.pdf} {{More than
  words: using token context to improve canonicalization of historical
  {G}erman}}.
\newblock \emph{{Journal for Language Technology and Computational
  Linguistics}}, 25(1):23--39.

\bibitem[{Kempken et~al.(2006)Kempken, Luther, and Pilz}]{Kempken-etal2006}
Sebastian Kempken, Wolfram Luther, and Thomas Pilz. 2006.
\newblock \href {https://doi.org/10.1007/978-0-387-34747-9_31} {Comparison of
  distance measures for historical spelling variants}.
\newblock In Max Bramer, editor, \emph{Artificial Intelligence in Theory and
  Practice}, pages 295--304. Springer, Boston, MA.

\bibitem[{Kestemont et~al.(2010)Kestemont, Daelemans, and
  De~Pauw}]{Kestemont-etal2010}
Mike Kestemont, Walter Daelemans, and Guy De~Pauw. 2010.
\newblock \href {https://doi.org/10.1093/llc/fqq011} {Weigh your
  words---memory-based lemmatization for {Middle Dutch}}.
\newblock \emph{Literary and Linguistic Computing}, 25(3):287--301.

\bibitem[{Kestemont et~al.(2016)Kestemont, de~Pauw, van Nie, and
  Daelemans}]{Kestemont-etal2016}
Mike Kestemont, Guy de~Pauw, Renske van Nie, and Walter Daelemans. 2016.
\newblock \href {https://doi.org/10.1093/llc/fqw034} {Lemmatization for
  variation-rich languages using deep learning}.
\newblock \emph{Digital Scholarship in the Humanities}.

\bibitem[{Koehn(2005)}]{Koehn2005}
Philipp Koehn. 2005.
\newblock \href
  {http://www.iccs.inf.ed.ac.uk/~pkoehn/publications/europarl-mtsummit05.pdf}
  {Europarl: A parallel corpus for statistical machine translation}.
\newblock In \emph{Proceedings of MT Summit}.

\bibitem[{Koehn et~al.(2007)Koehn, Hoang, Birch, Callison-Burch, Federico,
  Bertoldi, Cowan, Shen, Moran, Zens, Dyer, Bojar, Constantin, and
  Herbst}]{Moses}
Philipp Koehn, Hieu Hoang, Alexandra Birch, Chris Callison-Burch, Mercello
  Federico, Nicola Bertoldi, Brooke Cowan, Wade Shen, Christine Moran, Richard
  Zens, Chris Dyer, Ond\v{r}ej Bojar, Alexandra Constantin, and Evan Herbst.
  2007.
\newblock Moses: Open source toolkit for statistical machine translation.
\newblock In \emph{{Proceedings of the ACL 2007 Demo and Poster Sessions}},
  pages 177--180, Prague, Czech Republic.

\bibitem[{Koller(1983)}]{Koller1983}
Gerhard Koller. 1983.
\newblock Ein maschinelles {V}erfahren zur {N}ormalisierung altdeutscher
  {T}exte.
\newblock In Dietmar Peschel, editor, \emph{{G}ermanistik in {E}rlangen},
  volume~31 of \emph{{E}rlanger {F}orschungen}, pages 611--620.
  Universit\"{a}tsbund Erlangen-N\"{u}rnberg, Erlangen.

\bibitem[{Koolen et~al.(2006)Koolen, Adriaans, Kamps, and
  de~Rijke}]{Koolen-etal2006}
Marijn Koolen, Frans Adriaans, Jaap Kamps, and Maarten de~Rijke. 2006.
\newblock \href {https://doi.org/10.1007/11735106} {A cross-language approach
  to historic document retrieval}.
\newblock In \emph{Proceedings of the 28th European Conference on Information
  Retrieval Research (ECIR 2006)}, Lecture Notes in Computer Science, pages
  407--419, Berlin. Springer.

\bibitem[{Korchagina(2017)}]{Korchagina2017}
Natalia Korchagina. 2017.
\newblock \href {http://www.aclweb.org/anthology/W17-0504} {Normalizing
  medieval {G}erman texts: from rules to deep learning}.
\newblock In \emph{Proceedings of the NoDaLiDa 2017 Workshop on Processing
  Historical Language}, pages 12--17. Link{\"o}ping University Electronic
  Press.

\bibitem[{Levenshtein(1966)}]{Levenshtein1966}
Vladimir~I. Levenshtein. 1966.
\newblock Binary codes capable of correcting deletions, insertions, and
  reversals.
\newblock \emph{Soviet Physics Doklady}, 10(8):707--710.

\bibitem[{Ljube\v{s}i\'{c} et~al.(2016)Ljube\v{s}i\'{c}, Zupan, Fi\v{s}er, and
  Erjavec}]{Ljubesic-etal2016}
Nikola Ljube\v{s}i\'{c}, Katja Zupan, Darja Fi\v{s}er, and Toma\v{z} Erjavec.
  2016.
\newblock \href
  {https://www.linguistics.rub.de/konvens16/pub/19_konvensproc.pdf}
  {Normalising {Slovene} data: historical texts vs. user-generated content}.
\newblock In \emph{Proceedings of the 13th Conference on Natural Language
  Processing (KONVENS 2016)}, volume~16 of \emph{Bochumer Linguistische
  Arbeitsberichte}, pages 146--155, Bochum, Germany.

\bibitem[{Mitankin et~al.(2014)Mitankin, Gerdjikov, and
  Mihov}]{Mitankin-etal2014}
Petar Mitankin, Stefan Gerdjikov, and Stoyan Mihov. 2014.
\newblock \href {https://doi.org/10.1145/2595188.2595191} {An approach to
  unsupervised historical text normalisation}.
\newblock In \emph{Proceedings of the First International Conference on Digital
  Access to Textual Cultural Heritage (DATeCH '14)}, Madrid, Spain.

\bibitem[{Neubig et~al.(2018)Neubig, Sperber, Wang, Felix, Matthews,
  Padmanabhan, Qi, Sachan, Arthur, Godard, Hewitt, Riad, and
  Wang}]{Neubig-etal2018}
Graham Neubig, Matthias Sperber, Xinyi Wang, Matthieu Felix, Austin Matthews,
  Sarguna Padmanabhan, Ye~Qi, Devendra~Singh Sachan, Philip Arthur, Pierre
  Godard, John Hewitt, Rachid Riad, and Liming Wang. 2018.
\newblock {XNMT}: The extensible neural machine translation toolkit.
\newblock In \emph{Conference of the Association for Machine Translation in the
  Americas (AMTA) Open Source Software Showcase}, Boston.

\bibitem[{Oravecz et~al.(2010)Oravecz, Sass, and Simon}]{Oravecz-etal2010}
Csaba Oravecz, B\'alint Sass, and Eszter Simon. 2010.
\newblock \href {https://ilk.uvt.nl/LaTeCH2010/paperlist.html} {Semi-automatic
  normalization of {Old Hungarian} codices}.
\newblock In \emph{Proceedings of the ECAI 2010 Workshop on Language Technology
  for Cultural Heritage}, pages 55--59.

\bibitem[{Pettersson(2016)}]{Pettersson2016}
Eva Pettersson. 2016.
\newblock \href {http://urn.kb.se/resolve?urn=urn:nbn:se:uu:diva-269753}
  {\emph{Spelling Normalisation and Linguistic Analysis of Historical Text for
  Information Extraction}}.
\newblock Doctoral dissertation, Uppsala University, Department of Linguistics
  and Philology, Uppsala.

\bibitem[{Pettersson et~al.(2013{\natexlab{a}})Pettersson, Megyesi, and
  Nivre}]{Pettersson-Megyesi-Nivre2013}
Eva Pettersson, Be\'{a}ta Megyesi, and Joakim Nivre. 2013{\natexlab{a}}.
\newblock \href {http://www.aclweb.org/anthology/W13-5617} {Normalisation of
  historical text using context-sensitive weighted {L}evenshtein distance and
  compound splitting}.
\newblock In \emph{Proceedings of the 19th Nordic Conference of Computational
  Linguistics (NODALIDA 2013)}, pages 163--179. Link\"{o}ping University
  Electronic Press.

\bibitem[{Pettersson et~al.(2014)Pettersson, Megyesi, and
  Nivre}]{Pettersson-etal2014}
Eva Pettersson, Be\'{a}ta Megyesi, and Joakim Nivre. 2014.
\newblock \href {https://doi.org/10.3115/v1/W14-0605} {A multilingual
  evaluation of three spelling normalisation methods for historical text}.
\newblock In \emph{Proceedings of the 8th Workshop on Language Technology for
  Cultural Heritage, Social Sciences, and Humanities (LaTeCH)}, pages 32--41.
  Association for Computational Linguistics.

\bibitem[{Pettersson et~al.(2013{\natexlab{b}})Pettersson, Megyesi, and
  Tiedemann}]{Pettersson-etal2013}
Eva Pettersson, Be\'{a}ta Megyesi, and J\"{o}rg Tiedemann. 2013{\natexlab{b}}.
\newblock \href {http://www.ep.liu.se/ecp/article.asp?issue=087&article=005}
  {An {SMT} approach to automatic annotation of historical text}.
\newblock In \emph{Proceedings of the Workshop on Computational Historical
  Linguistics at {NODALIDA} 2013}, NEALT Proceedings Series 18/Link\"{o}ping
  Electronic Conference Proceedings 87, pages 54--69. Link\"{o}ping University
  Electronic Press.

\bibitem[{Porta et~al.(2013)Porta, Sancho, and G\'{o}mez}]{Porta-etal2013}
Jordi Porta, Jos\'{e}-Luis Sancho, and Javier G\'{o}mez. 2013.
\newblock \href {http://www.ep.liu.se/ecp/article.asp?issue=087&article=006}
  {Edit transducers for spelling variation in {Old Spanish}}.
\newblock In \emph{Proceedings of the Workshop on Computational Historical
  Linguistics at {NODALIDA} 2013}, NEALT Proceedings Series 18/Link\"{o}ping
  Electronic Conference Proceedings 87, pages 70--79. Link\"{o}ping University
  Electronic Press.

\bibitem[{Porter(2001)}]{Snowball}
Martin Porter. 2001.
\newblock \href {http://snowballstem.org/texts/introduction.html} {Snowball:
  {A} language for stemming algorithms}.

\bibitem[{Rayson et~al.(2005)Rayson, Archer, and Smith}]{Rayson-etal2005}
Paul Rayson, Dawn Archer, and Nicholas Smith. 2005.
\newblock {VARD} versus {Word}: A comparison of the {UCREL} variant detector
  and modern spell checkers on {English} historical corpora.
\newblock In \emph{Proceedings of the Corpus Linguistics Conference CL2005},
  Birmingham, UK. University of Birmingham.

\bibitem[{Robertson and Goldwater(2018)}]{Robertson-Goldwater2018}
Alexander Robertson and Sharon Goldwater. 2018.
\newblock \href {http://aclweb.org/anthology/N18-2113} {Evaluating historical
  text normalization systems: How well do they generalize?}
\newblock In \emph{Proceedings of the 2018 Conference of the North American
  Chapter of the Association for Computational Linguistics: Human Language
  Technologies, Volume 2 (Short Papers)}, pages 720--725. Association for
  Computational Linguistics.

\bibitem[{Robertson and Willett(1993)}]{Robertson-Willett1993}
Alexander~M. Robertson and Peter Willett. 1993.
\newblock \href {https://doi.org/10.1093/llc/8.3.143} {A comparison of
  spelling-correction methods for the identification of word forms in
  historical text databases}.
\newblock \emph{Literary and Linguistic Computing}, 8(3):143--152.

\bibitem[{S\'{a}nchez-Mart\'{i}nez et~al.(2013)S\'{a}nchez-Mart\'{i}nez,
  Mart\'{i}nez-Sempere, Ivars-Ribes, and Carrasco}]{SanchezMartinez-etal2013}
Felipe S\'{a}nchez-Mart\'{i}nez, Isabel Mart\'{i}nez-Sempere, Xavier
  Ivars-Ribes, and Rafael~C. Carrasco. 2013.
\newblock \href {http://arxiv.org/abs/1306.3692} {An open diachronic corpus of
  historical {Spanish}: annotation criteria and automatic modernisation of
  spelling}.
\newblock \emph{CoRR}, abs/1306.3692.

\bibitem[{Scherrer and Erjavec(2013)}]{Scherrer-Erjavec2013}
Yves Scherrer and Toma\v{z} Erjavec. 2013.
\newblock \href {http://www.aclweb.org/anthology/W13-2409} {Modernizing
  historical slovene words with character-based {SMT}}.
\newblock In \emph{Proceedings of the 4th Biennial International Workshop on
  Balto-Slavic Natural Language Processing}, pages 58--62. Association for
  Computational Linguistics.

\bibitem[{Scherrer and Erjavec(2016)}]{Scherrer-Erjavec2016}
Yves Scherrer and Toma\v{z} Erjavec. 2016.
\newblock \href {https://doi.org/10.1017/S1351324915000236} {Modernising
  historical {Slovene} words}.
\newblock \emph{Natural Language Engineering}, 22(6):881--905.

\bibitem[{Scherrer and Ljube\v{s}i\'{c}(2016)}]{Scherrer-Ljubesic2016}
Yves Scherrer and Nikola Ljube\v{s}i\'{c}. 2016.
\newblock \href
  {https://www.linguistics.rub.de/konvens16/pub/32_konvensproc.pdf} {Automatic
  normalisation of the {Swiss} {German} {ArchiMob} corpus using character-level
  machine translation}.
\newblock In \emph{Proceedings of the 13th Conference on Natural Language
  Processing (KONVENS 2016)}, volume~16 of \emph{Bochumer Linguistische
  Arbeitsberichte}, pages 248--255, Bochum, Germany.

\bibitem[{Schneider et~al.(2017)Schneider, Pettersson, and
  Percillier}]{Schneider-etal2017}
Gerold Schneider, Eva Pettersson, and Michael Percillier. 2017.
\newblock \href {http://www.aclweb.org/anthology/W17-0508} {Comparing
  rule-based and {SMT}-based spelling normalisation for {E}nglish historical
  texts}.
\newblock In \emph{Proceedings of the NoDaLiDa 2017 Workshop on Processing
  Historical Language}, pages 40--46. Link{\"o}ping University Electronic
  Press.

\bibitem[{Sennrich et~al.(2017)Sennrich, Birch, Currey, Germann, Haddow,
  Heafield, Miceli~Barone, and Williams}]{Sennrich-etal2017}
Rico Sennrich, Alexandra Birch, Anna Currey, Ulrich Germann, Barry Haddow,
  Kenneth Heafield, Antonio~Valerio Miceli~Barone, and Philip Williams. 2017.
\newblock \href {https://doi.org/10.18653/v1/W17-4739} {The {U}niversity of
  {E}dinburgh's neural {MT} systems for {WMT17}}.
\newblock In \emph{Proceedings of the Second Conference on Machine
  Translation}, pages 389--399. Association for Computational Linguistics.

\bibitem[{Tang et~al.(2018)Tang, Cap, Pettersson, and Nivre}]{Tang-etal2018}
Gongbo Tang, Fabienne Cap, Eva Pettersson, and Joakim Nivre. 2018.
\newblock \href {http://aclweb.org/anthology/C18-1112} {An evaluation of neural
  machine translation models on historical spelling normalization}.
\newblock In \emph{Proceedings of the 27th International Conference on
  Computational Linguistics}, pages 1320--1331. Association for Computational
  Linguistics.

\end{thebibliography}
